\documentclass[sigconf]{aamas} 

\usepackage{microtype}
\usepackage{graphicx}
\usepackage{subcaption}
\usepackage{booktabs} 
\usepackage{enumitem}
\newenvironment{faq}{\begin{description}[style=nextline]}{\end{description}}

\usepackage{balance} 
\usepackage{hyperref}
\usepackage{algorithm}
\usepackage{algorithmic}

\usepackage{amsmath}
\usepackage{mathtools}
\usepackage{amsthm}
\usepackage[capitalize,noabbrev]{cleveref}
\usepackage{pifont}
\theoremstyle{plain}

\theoremstyle{definition}

\theoremstyle{remark}

\newcommand{\expectation}{\mathbb{E}_{\epsilon\sim\mathcal{N}(\mathbf{0},I_{d})}}

\usepackage{balance} 

\setcopyright{ifaamas}
\acmConference[AAMAS '24]{Proc.\@ of the 23rd International Conference
on Autonomous Agents and Multiagent Systems (AAMAS 2024)}{May 6 -- 10, 2024}
{Auckland, New Zealand}{N.~Alechina, V.~Dignum, M.~Dastani, J.S.~Sichman (eds.)}
\copyrightyear{2024}
\acmYear{2024}
\acmDOI{}
\acmPrice{}
\acmISBN{}

\acmSubmissionID{437}

\title[AAMAS-2024 Formatting Instructions]{Scaling Opponent Shaping to High Dimensional Games}

\author{Akbir Khan}
\authornote{Equal Contribution.}
\affiliation{
  \institution{University College London}
  \country{}
  }
\email{akbir.khan.13@ucl.ac.uk}

\author{Timon Willi\footnotemark[1]}
\affiliation{
  \institution{University of Oxford}
  \country{}
  }
\email{timon.willi@eng.ox.ac.uk}

\author{Newton Kwan\footnotemark[1]}
\affiliation{
  \institution{University College London}
  \country{}
  }

\author{Andrea Tacchetti}
\affiliation{
  \institution{Deepmind}
  \country{}
  }

\author{Chris Lu}
\affiliation{
  \institution{University of Oxford}
  \country{}
  }

\author{Edward Grefenstette}
\affiliation{
  \institution{University College London}
  \country{}
  }

\author{Tim Rocktäschel}
\affiliation{
  \institution{University College London}
  \country{}
  }

\author{Jakob Foerster}
\affiliation{
  \institution{University of Oxford}
  \country{}
  }

\newcommand{\method}[0]{\textsc{Shaper}}
\begin{abstract}
In multi-agent settings with mixed incentives, methods developed for zero-sum games have been shown to lead to detrimental outcomes. To address this issue,  \textit{opponent shaping} (OS) methods explicitly learn to influence the learning dynamics of co-players and empirically lead to improved individual \textit{and} collective outcomes. However, OS methods have only been evaluated in low-dimensional environments due to the challenges associated with estimating higher-order derivatives or scaling model-free meta-learning. Alternative methods that scale to more complex settings either converge to undesirable solutions or rely on unrealistic assumptions about the environment or co-players. In this paper, we successfully scale an OS-based approach to general-sum games with temporally-extended actions and long-time horizons for the first time. After analysing the representations of the meta-state and history used by previous algorithms, we propose a simplified version called \textsc{Shaper}. We show empirically that \method{} leads to improved individual and collective outcomes in a range of challenging settings from literature. We further formalize a technique previously implicit in the literature, and analyse its contribution to opponent shaping. We show empirically that this technique is helpful for the functioning of prior methods in certain environments. Lastly, we show that previous environments, such as the CoinGame, are inadequate for analysing temporally-extended general-sum interactions\footnote{Blogpost available at \href{https://sites.google.com/view/scale-os/}{sites.google.com/view/scale-os/}}.
\end{abstract}


\newcommand{\BibTeX}{\rm B\kern-.05em{\sc i\kern-.025em b}\kern-.08em\TeX}

\begin{document}


\pagestyle{fancy}
\fancyhead{}


\maketitle 


\section{Introduction}
\label{introduction}
From personal assistants and chat-bots to self-driving cars and recommendation systems, the world of software is becoming increasingly \textit{multi-agent} as these systems are continuously learning and interacting with each other in fully cooperative, fully competitive, and general-sum settings. 

In this paper we investigate interacting, learning agents in general-sum settings. In such settings, commonly-used RL methods developed for zero-sum games can lead to disastrous outcomes \citep{dafoe2020open}. For example, in real-world scenarios like pollution and international arms races \citep{snyder1971prisoner, dawes1980social}, such agents would fail to realise that they're better off cooperating, even if it means they're potentially worse off than their co-players. Poor performance could also lead to being extorted in social-dilemma-like scenarios \citep{press_iterated_2012, lu2022model}.

While multi-agent learning research has shown great success in strictly competitive~\citep{silver2016mastering, brown2017libratus, vinyals2019alphastar,jarderberg_2019_hideandseek} and fully cooperative settings~\citep{rashid2018qmix,foerster2019bayesian}, this success does not transfer to general-sum settings ~\citep{foerster_learning_2018,letcher_stable_2019}: In competitive games, agents can learn Nash equilibrium strategies by iteratively best-responding to suitable mixtures of past opponents. Similarly, best-responding to rational co-players leads to the desirable equilibria in cooperative games (assuming joint training). In contrast, many Nash equilibria coincide with globally worst welfare outcomes in general-sum settings, rendering the above learning paradigms ineffective. For example, in the iterated prisoner's dilemma ~\citep[IPD,][]{axelrod1981evolution, Harper_2017}, naive best-response dynamics converge to unconditional mutual defection~\citep{foerster_learning_2018} rather than Nash equilibria with higher social welfare.

It is important that general-sum learning methods scale to \textit{high-dimensional} settings, such as those with longer-time horizons, larger state spaces and temporally-extended actions, as these environments are more akin to the real world. In matrix games, cooperation and defection are clearly defined atomic actions, whilst in more complex environments such as autonomous driving, cooperation and defection are defined over sequences of actions (e.g. a path towards a cooperative/defective location). Scalable methods~\citep{lupu2020gifting, yuan2022adherence, jaques2019social, koster2020model} that manage to avoid \textit{unconditional defection} in these settings rely heavily on \textit{reward shaping}, which blurs the line between the problem setting (``social dilemma'') and the method. 

As an alternative approach, \textit{opponent shaping} (OS) methods recognise that the actions of any one agent influence their co-player's policy and seek to use this mechanism to their advantage \citep{foerster_learning_2018, letcher_differentiable_2019, kim2021meta-mapg, willi2022cola}. However, many past OS methods require privileged information to shape their opponents and are myopic since anticipating many steps is intractable. Model-Free Opponent Shaping~\citep[M-FOS,][]{lu2022model} and The Good Shepherd~\citep[GS,][]{balaguer2022good} solve the issues above by framing opponent shaping as a meta-learning problem, which our method inherits and builds upon. However, M-FOS presents only preliminary results on the higher-dimensional CoinGame~\citep{lerer2017coingame} benchmark, and GS none at all.

To scale OS agents to higher-dimensional benchmarks, we systematically evaluate the architectural components of the \textsc{M-FOS} and GS algorithms. We identify two forms of memory---\textit{history} and \textit{context}. \textit{History} captures intra-episode information and \textit{context} inter-episode information. We find empirically that both are necessary to achieve shaping. \textsc{M-FOS} captures both types of memory (though not completely), whereas GS does not. However, we identify a bottleneck in the M-FOS method, as M-FOS requires two separate policies to capture \textit{context} and \textit{history}, where only one is necessary. Using this finding, we propose a new method, called \method{}, removing the unnecessary bottleneck from \textsc{M-FOS}. 

Beyond these memory components, we uncover another element used implicitly in prior work but never formally introduced or analysed: averaging across the batch of trajectories at each meta step. This ensures that the hidden states of the opponent shaping algorithm carry information from the entire batch, rather than just a single batch dimension. We formalise this technique and empirically investigate its importance. Our analysis shows that while this technique improves previous methods like \textsc{M-FOS} in certain environments, it is not essential for our proposed method, \method{}, in typical environment settings. This highlights the value of our formalisation and empirical analysis in understanding and improving upon existing methods.

\method{} outperforms previous OS methods in general-sum games with long-time horizons and temporally-extended actions. We showcase our performance on the ``IPD in the Matrix'' and ``IMP in the Matrix`` games, introduced by the \textit{melting pot} suite~\cite{leibo2021meltingpot}. These have a 30x state-space than environments previously used to evaluate OS and contain more complex interaction dynamics. Additionally, we consider shaping on sequential matrix games with 100x longer horizons than their previously used counterparts. We demonstrate empirically that our simplification of M-FOS helps scalability, that only \textit{evolutionary-based} meta-learning is effective in these long-horizon games, and that previous evaluation environments, such as the CoinGame \citep{lerer2017coingame}, are inadequate for analysing OS in temporally-extended, general-sum interactions.

\section{Background}
\label{Background}

\textbf{What is a Game?} We formalise our environments as Partially Observable Stochastic Games \citep[POSG]{shapley_stochastic_1953}. A POSG is given by the tuple $\mathcal{M} = \langle \mathcal{N}, \mathcal{A}, \mathcal{O}, S, \mathcal{T}, \mathcal{I}, \mathcal{R}, \gamma \rangle$, 
where $\mathcal{A}$, $\mathcal{O}$, and $S$ denote the action, observation, and state space, respectively. These parameters can be distinct at every time step and also incorporated into the transition function $\mathcal{T}: S \times \mathbf{A} \rightarrow \Delta S$, where $\mathbf{A} \equiv \mathcal{A}^{n}$ is the joint action of all agents.
Each agent draws individual observations according to the observation function $\mathcal{I}: S \times N \rightarrow \mathcal{O}$  and obtains a reward according to their reward function $\mathcal{R}: S \times \mathbf{A} \times  N \rightarrow \mathbb{R}$ where $N=\{ 1, \dots, n\}$. POSGs represent general-sum games. The single-player case, $N=\{1\}$, of POSGs are Partially Observable Markov Decision Processes (POMDPs).

\textbf{What is Shaping?}
Shaping is acting to manipulate the co-player's learning dynamics (and subsequent behaviour) \citep{foerster_learning_2018}, where co-players are any other participants in the game. Newer shaping methods frame shaping as a meta-learning problem \citep{kim2021meta-mapg, lu2022model, balaguer2022good}. We next present the meta-learning problem setting presented by \textsc{M-FOS} since our work is a simplified case of the \textsc{M-FOS} framework.

\textbf{What is M-FOS?} Conceptually, \textsc{M-FOS} separates the task of shaping (the meta-game) from the task of playing the game. Specifically, the meta-game is formulated as a POMDP $\langle  \overline{\mathcal{A}},\overline{\mathcal{O}},\overline{\mathcal{S}}, \overline{\mathcal{T}}, \overline{\mathcal{I}}, \overline{\mathcal{R}}, \bar{\gamma}\rangle$ over an underlying general-sum game, represented by a POSG $\mathcal{M}$, where the overline indicates the single-agent version of the elements defined for POSGs. In the ``shaping'' POMDP, the meta-state $\overline{\mathcal{S}}$ contains the policies of all players in the underlying POSG: $\bar{s}_e=\left(\phi^{e-1}_i, \phi^{e-1}_{-i}\right) \in \overline{\mathcal{S}}$, where $e$ indexes the episodes and $(i, -i)$ index all agents. The meta-observation is all observations of the previous episode in the underlying game, i.e., $\bar{o}_{e}=(o^{e-1}_{0}, o^{e-1}_{1}, ..., o^{e-1}_{K})$, where $K$ is the length of an episode. The meta-action space $\overline{\mathcal{A}}$ consists of the policy parameterisation of the inner agent $i$ (in practice a vector conditioning the policy), i.e., $\overline{a}_{e}=\phi^{e}_i$. 

\textsc{M-FOS} training works as follows. The meta-agent trains over a sequence of $T$ \textit{trials} (denoted ``meta-episodes'' in the original paper). Each trial contains $E$ episodes. At the end of each episode $e$ within a trial $t$, conditioned on both agent's policies, the co-players update their parameters with respect to the expected episodic return $J^{e}_{-i}=\mathbb{E}\left[\sum_{k=0}^K\gamma^{k} r^{k}_{-i}(\phi^{e}_{i}, \phi^{e}_{-i})\right]$, where $K$ is the length of an episode. For example, if the co-players were Naive Learners, i.e., agents not accounting for the learning dynamics of the co-player, with learning rate $\alpha$, the update is:
$   \phi^{e+1}_{j} = \phi^{e}_{j} + \alpha\nabla_{\phi^{e}_{j}}J^{e}_{j}, \text{for } j \in -i$.


\begin{algorithm}[t] 
\caption{\method{} Update: Given a POSG $\mathcal{M}$, policies $\pi_{\phi_{i}}, \pi_{\phi_{-i}}$ and their respective initial hidden states $h_{i}, h_{-i}$ and a distribution of initial co-players $\rho_{\phi}$, this algorithm updates a meta-agent policy $\phi_{i}$ over $T$ trials consisting of $E$ episodes.}\label{alg:cap}. 
\begin{algorithmic}[1]
\REQUIRE $\mathcal{M},\phi_{i}, \rho_{\phi}, E, T$
        \FOR{$t=0$ \textbf{to} $T$}
            \STATE Initialise trial reward $\bar{J}=0$
            \STATE Initialise meta-agent hidden state $h_i=\textbf{0}$
            \STATE Sample co-players $\phi_{-i}\sim \rho_{\phi}$
            \FOR{$e=0$ \textbf{to} $E$}
            \STATE Initialise co-players' $h_{-i}=\textbf{0}$
            \STATE $J_{i}, J_{-i}, h'_{i}, h'_{-i}$ = $\mathcal{M}(\phi_{i}, \phi_{-i}, h_{i}, h_{-i})$
            \STATE Update $\phi_{-i}$ according to co-players' update rule.
            \STATE  $h_{i} \leftarrow{} h'_{i}$
            \STATE  $\bar{J} \leftarrow \bar{J} + J_{i}$ 
            \ENDFOR{} \\
            \STATE{Update $\phi_{i}$ with respect to $\bar{J}$}
            \ENDFOR{} 
\end{algorithmic}
\end{algorithm}

In contrast to the co-player's update, the meta-agent learns an update function for the parameters of their inner agent, i.e., $\bar{a}_e=\phi_i^e \sim \pi_\theta\left(\cdot \mid \bar{o}_e\right)$, where $\theta$ is the parameters of the meta-agent. The meta-agent optimises the meta-return $\bar{J}=\sum_{e=0}^E J^{e}_{i}$ (summed over all episodes) at the end of a trial $t$ using any model-free optimisation technique, e.g., PPO \citep{schulman2017ppo} or Evolution Strategies~\citep[ES,][]{salimans2017evolutionstrategies}. The meta-agent and the inner agent are usually represented as recurrent neural networks, such as LSTMs \citep{hochreiter1997lstm}. The meta-game setup allows the meta-agent to observe the results of the co-player's learning dynamics, enabling it \textit{to learn to shape}. Though not formalized or discussed in detail in the paper, the original \textsc{M-FOS} averages across a batch of trajectories at each update step to ensure access to all information necessary for shaping. In Section \ref{sec:method}, we introduce a formal definition of averaging across the batch and investigate its role for shaping. While published concurrently, the Good Shepherd~\citep{balaguer2022good} is a simplified version of \textsc{M-FOS}, in which the meta-agent and underlying agent are collapsed into a single agent without memory \textit{that only updates after each trial}. The agent is represented by a feedforward neural network and has no memory. However, as GS was evaluated on infinitely iterated matrix games, where the state usually represents a one-step history, we consider GS to have one-step history.

\textbf{Where have current shaping algorithms been evaluated?} Both \textsc{M-FOS} and GS evaluate their shaping on infinitely-iterated matrix games. While this is a fruitful playground to discover complex strategies, such as tit-for-tat, infinitely-iterated matrix games do not contain temporally-extended actions or high dimensional state spaces. For example, in matrix games, cooperation simply consists of playing the ``cooperation'' action. However, in the real world, cooperation requires a repeated commitment to a cooperative strategy (where it is often unclear whether a given atomic action is cooperative). The CoinGame \citep{lerer2017coingame} supposedly addresses this shortcoming by incorporating IPD-like game dynamics into a gridworld. \textsc{M-FOS} presents very preliminary results on the CoinGame with no detailed analysis of the emerging strategies. However, as we show in Section~\ref{sec:experiments} the CoinGame suffers from pathologies that enable shaping with simple strategies.

\section{\method{}: A Scalable OS Method}
\label{sec:method}

To introduce our method, we first analyse the role of memory in meta-learning-based OS. 
Memory is important because it enables a meta-agent to adapt its meta-policy within a trial since it only updates \textit{parameters} after a trial. 

If the meta-agent cannot adapt their policy within a trial, a co-player could simply learn the best-response to the meta-agent's policy. For example, in Rock-Paper-Scissors, the meta-agent would be forced to play the fully mixed strategy, as any deviation from it will be taken advantage of by the co-player. Instead a meta-agent with memory can adapt to the co-player's best response within a trial and potentially achieve a better meta-return, which we show in Section \ref{sec:results}. Thus, memory is important if a meta-agent is to perform well in all general-sum games. 

We define two forms of memory: \textit{context} and \textit{history}. Let us define \textit{history} as a trajectory of a (partial) episode $e$, $\tau_{e} = \left(o^{0}_{e}, a^{0}_{e}, r^{0}_{e},..., r^{K}_{e}\right)$, and \textit{context} as a trajectory of a (partial) trial t, $\bar{\tau}_{t} = \left(\tau_{0}, ..., \tau_{E}\right)$. \textit{History} captures the dynamics within an episode and is crucial for implementing policies such as TFT that reward/punish the co-player based on past actions. In contrast, \textit{context} captures the learning dynamics of the co-player as it contains the co-player's policy changes over many parameter updates. \textit{Context} is important for shaping when the co-players \textit{update dynamics} are non-stationary across a trial or need to be inferred from the changing policy itself across different episodes. It allows the shaper to adapt its inner policy to, e.g., a change of the co-player's learning rate or implicitely infer their objective function. 

Using the above definitions, we express the policies as the following: $\textsc{M-FOS}: a \sim \pi_{\phi}(\cdot\mid\tau_{e}, \pi_{\theta}(\bar{o}_{e}))$ and $\textsc{GS}: a \sim \pi_{\phi}(\cdot\mid o_{e}^{t})$. 
M-FOS captures one-step \textit{context} via the memory of the meta-agent, and \textit{history} via the memory of the inner agent but requires two agents to do so. In contrast, GS captures one-step history (if given by environment) but does not require a separate inner agent. 

We propose \method{}, an algorithm requiring only one agent to capture \textit{context and history}. This is accomplished via an RNN that retains its hidden state over episodes and only resets \textit{after each trial} 
\begin{equation}
    \textrm{\method{}:}~a \sim \pi_{\phi}(\cdot\mid \tau_{e}, \bar{\tau}_{t})
\end{equation}

Compared to GS, \method{} has access to history and context by \textit{adding memory to the architecture and retaining the hidden state over the episodes}. Compared to M-FOS, \method{} only requires sampling from one action space. To contrast \method{} to \textsc{M-FOS} in more detail,\ we refer to Appendix \ref{app:faq}.

\method{} is trained as follows. Given a POSG $\mathcal{M}$, at the start of a trial, co-players $\phi_{-i}\sim \rho_{\phi}$ are sampled, where $\rho_{\phi}$ is the respective sampling distribution. 
\method{}'s parameters $\phi_i$ and hidden state $h_i$ are randomly initialised.  During an episode of length $K$, agents take their actions, $a_{i}^k \sim \pi_{\phi_{i}}(\cdot\mid o^{k}_{i}, h^{k}_{i})$. At each time step in the episode, the hidden state of the meta-agent is updated: $h_{i}^{k+1} = f(o_{i}^{k}, h_{i}^{k})$. On receiving actions, the POSG returns rewards $r_{i}^{k}$, new observations $o_{i}^{k+1}$ and a done flag $d$, indicating if an episode has ended.

When an inner episode terminates, the updated co-player $\phi^{e+1}_{-i}$ \textit{and the meta-agent's hidden state} $h_{i}^{K}$ are passed to the next episode. This process is repeated over $E$ episodes in a trial. When a trial terminates, the meta-agent's policy is updated, maximising total \textit{trial} reward, $\bar{J}=\sum_{e}^E J^{e}_{i}$. This leads to the following objective,
\begin{equation}
\max_{\phi_{i}}\mathbb{E}_{\small \rho(\phi), \rho(\mathcal{M})}  \left[\bar{J}\right].
\end{equation}
\begin{table*}[ht]
\caption{Converged reward per episode (meta-agent, co-player) for agents trained with Naive Learners on the CoinGame, IPDitM and IMPitM. We report reward per episode for better interpretability. The mean is reported across 100 seeds with standard deviations.
\label{tab:result_gridworlds}}
    \centering
    \begin{tabular}{l rl rl rl}
     & \multicolumn{2}{c}{CoinGame} & \multicolumn{2}{c}{IPD in the Matrix} & \multicolumn{2}{c}{IMP in the Matrix} \\ 
    \midrule
     \method{} 
     & $\mathbf{4.63 \pm 0.66}$,&$ \mathbf{-3.35 \pm0.67}$
     & $\mathbf{22.44 \pm 1.12}$, & $\mathbf{21.49 \pm 0.67}$
     & $\mathbf{0.14 \pm 0.06}$, & $\mathbf{-0.14 \pm 0.06}$\\
     \textsc{M-FOS} (ES) & $3.13 \pm 0.40$,&$ 2.27 \pm 0.38$
      & $15.49 \pm 1.28$,&$ 23.88 \pm 0.93$
      & $\mathbf{0.11 \pm 0.07}$,&$\mathbf{-0.11 \pm 0.07}$
      \\
     \textsc{M-FOS} (RL) 
     & $0.94 \pm 0.68$,&$ -0.23 \pm 0.52$
     & $7.42 \pm 0.21$,&$ 7.28 \pm 0.15 $
     & $0.04 \pm 0.00$,&$ -0.04 \pm 0.00$ \\
     GS 
     & $\mathbf{5.44 \pm 0.61}$, & $\mathbf{-4.17 \pm 0.48}$
      & $16.16 \pm 1.33$,&$ 24.35 \pm 0.83$ 
      & $0.00 \pm 0.00$,&$ 0.00 \pm 0.00$ \\
     PT-NL 
     & $0.70 \pm 0.58$,&$ -0.3 \pm 0.47$
     & $6.33 \pm 0.33$,&$ 6.96 \pm 0.31 $
     & $-0.17 \pm 0.10$,&$ 0.17 \pm 0.10 $\\ 
     CT-NL 
     & $0.47 \pm 0.83$,&$ 0.26 \pm 0.30$
     & $5.56 \pm 0.02$,&$ 5.56 \pm 0.02$
     & $-0.10 \pm 0.06$,&$ 0.10 \pm 0.06$ \\ 
    \bottomrule
    \end{tabular}
\end{table*}
In practice, the co-players optimise their parameters using some form of gradient descent, which typically involves batching the episodal trajectories.
Assume $\phi^{e}_{-i} = G(\phi^{e-1}_{-i}, \boldsymbol{\tau}_{e-1})$ is some co-player's update function $G: \mathbb{R}^{P}\times\mathbb{R}^{B\times T} \rightarrow \mathbb{R}^{P}$, where P is the number of parameters of $\phi_{-i}$, $B$ is the batch size, i.e., number of environments for parallel training, and $T$ is the length of the trajectory. Shaper then interacts with a co-player over a batch of environments, i.e., $\textbf{a}_{i}^k \sim \pi_{\phi_{i}}(\cdot\mid \textbf{o}^{k}_{i}, \textbf{h}^{k}_{i})$, where $\textbf{a}_{i}^k \in \mathbb{R}^{B\times A}$, $\textbf{o}^{k}_{i} \in \mathbb{R}^{B\times O}$, and $\textbf{h}^{k}_{i} \in \mathbb{R}^{B\times H}$, and $A$, $O$, and $H$ are the action-, observation-, and hidden-state-size respectively.


Shaper needs to account for the batched updates of the co-player because opponent shaping requires all the information determining the learning update of the co-player. For example, imagine Shaper plays with a co-player across a batch of environments with different reward functions. While the co-player updates its parameters based on a diverse set of trajectories from many reward functions, each of Shaper's hidden states only observes the trajectory of its respective reward function. Intuitively, if the reward functions are very diverse, the update derived from the whole batch would significantly differ from the update estimated from a single environment trajectory, as observed by the hidden state. We thus define the \textit{batched context} as a trajectory of a batched (partial) trial $\boldsymbol{\bar{\tau}}_{t} = \left(\boldsymbol{\tau}_{0}, ..., \boldsymbol{\tau}_{E}\right)$. 
$$\textrm{Shaper:}~a \sim \pi_{\phi}(\cdot\mid \boldsymbol{\tau_{e}}, \boldsymbol{\bar{\tau}}_{t})$$ 
This insight leads to the consequence that Shaper needs to consolidate information across its batch of hidden states, at least at every co-player update. To address this issue, Shaper averages over its hidden states across the batch at each step, combined with a skip connection to ensure ``situational'' awareness of the hidden state's respective environment (see Figure \ref{fig:shaper_batching}).

\begin{align}
    \textbf{h}_{i}^{k+1} &= \lambda\left(\frac{1}{B}\sum_{l=0}^{B}h_{i,l}^{k}\right) + (1-\lambda)\textbf{h}_{i}^{k}\\
    \textbf{h}_{i}^{k+1} &= f(o_{i}^{k}, \textbf{h}_{i}^{k+1})
\end{align}
This approach ensures that \method{} effectively shapes its co-players while accounting for the diverse set of trajectories that inform their gradient updates, captured by the following meta-return function with the expectation over the batched gradient update of the co-player:


\begin{equation}
\begin{aligned}
    \Bar{J}^{\text{ES}} &= \expectation\Bigg[\mathbb{E}_{\Bar{\tau}\sim\pi_{\phi_{i}+\epsilon\sigma};\phi_{-i}^{e}\sim G(\phi_{-i}^{e-1}, \boldsymbol{\tau}_{e-1})} \\
    &\quad \times \left[\mathbb{E}_{\tau_{e}\sim\pi_{\phi_{-i}^{e}}} \left[\Sigma_{k=0}^{K}\gamma^{k}r^{k}(\phi_{-i}^{e}, \phi_{i}+\epsilon\sigma)\right]\right]\Bigg]
\end{aligned}
\end{equation}


\section{Experiments}
Here we present the test environments and our evaluation protocol for \method{}. We also explain our ablation experiments helping us evaluate the role of memory in OS. 

\label{sec:experiments}
The \textbf{Prisoner's Dilemma} is a well-known and widely studied general-sum game illustrating that two self-interested agents do not cooperate even if it is globally optimal. The players either cooperate (C) or defect (D) and receive a payoff according to Table \ref{tab:payoff_ipd}. In the \textit{iterated} prisoner's dilemma (IPD), the agents repeatedly play the prisoner's dilemma and observe the previous action of both players.Past research used the infinite IPD in their experiments \citep{foerster_learning_2018,letcher_stable_2019, willi2022cola, lu2022model, balaguer2022good}. In the infinite version, the exact value function and gradients thereof are calculated directly from the policy weights~\citep{foerster_learning_2018}; 
In our work, we consider the finitely iterated PD (f-IPD), where we cannot calculate the exact value function and have to rely on sample-based approaches such as RL and ES.

\textbf{Iterated Matching Pennies} (IMP) is an iterated matrix game like the IPD. The players choose heads (H) or tails (T) and receive a payoff according to both players' choices. In contrast to the IPD, a general-sum game, IMP is a zero-sum game. 
In the IMP one player gets +1 for playing the same action as the other player, while the other player is rewarded for playing a \textit{different} action.
Thus, the only equilibrium strategy for each one-memory agent is to play a random policy, resulting in an expected joint payoff of (0,0). Only with intra-episode memory can a meta-agent observe a co-player's current policy and thus shape it.

\textbf{CoinGame} is a multi-agent gridworld environment that simulates social dilemmas (like the IPD) with high-dimensional states and multi-step actions \citep{lerer2017coingame}. Two players, blue and orange,  move around a wrap-around grid and pick up blue and orange coloured coins. When a player picks up a coin of any colour, this player receives a reward of $+1$. When a player picks up a coin of the co-player's colour, the co-player receives a reward of $-2$. Whenever a coin gets picked up, a new coin of the same colour appears in a random location on the grid at the next time step. If both agents reach a coin simultaneously, then both agents pick up that coin (the coin is duplicated). When both players pick up coins without regard to colour, the expected reward is $0$. In contrast to matrix games, the CoinGame requires learning from high-dimensional states with multi-step actions.

\begin{figure*}[h]
    \centering
    \begin{subfigure}[b]{0.24\textwidth}
      \centering
      \includegraphics[width=\textwidth]{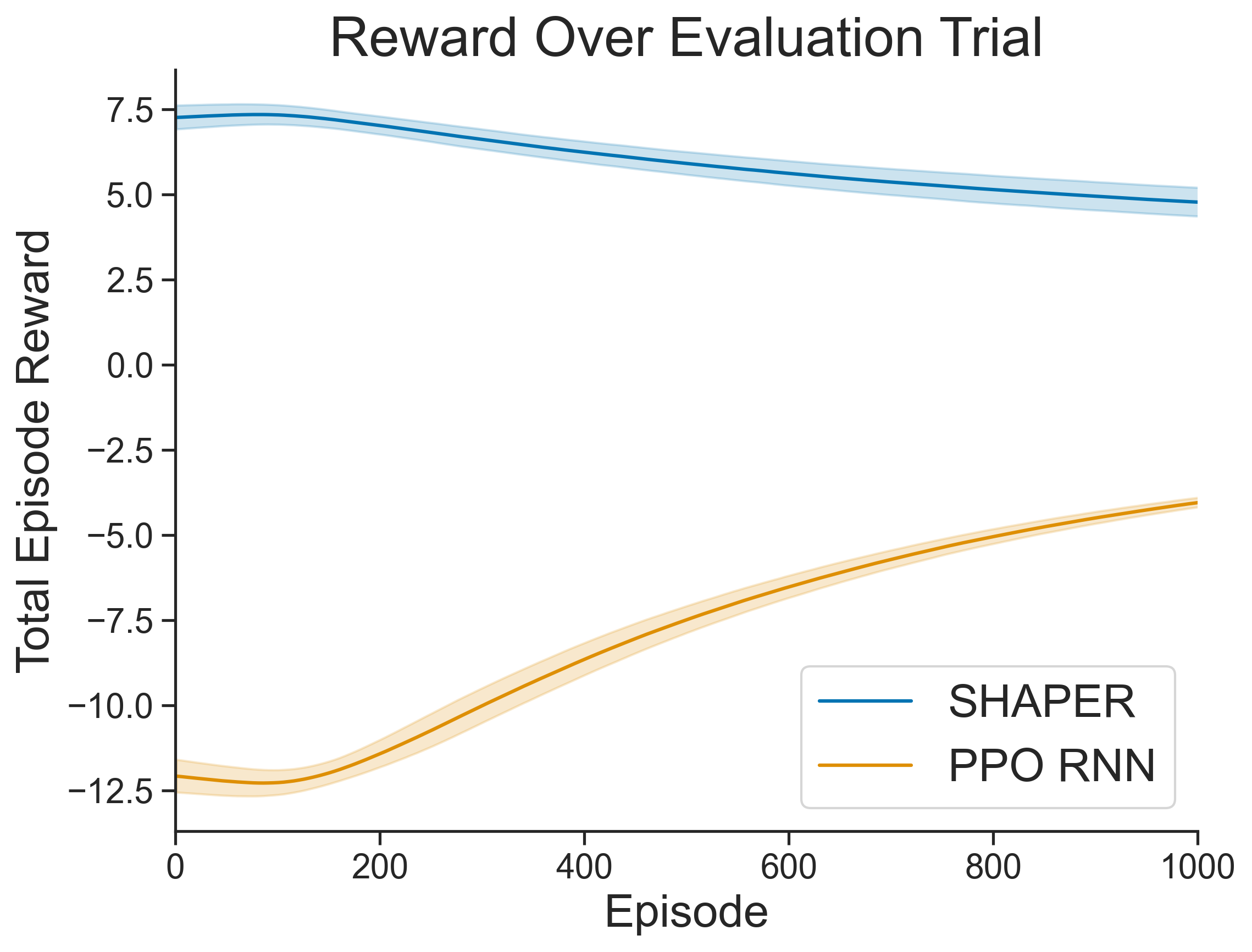}
      \caption{}
      \label{fig:earl_cg_eval_reward}
    \end{subfigure}
    \begin{subfigure}[b]{0.24\textwidth}
      \centering
      \includegraphics[width=\textwidth]{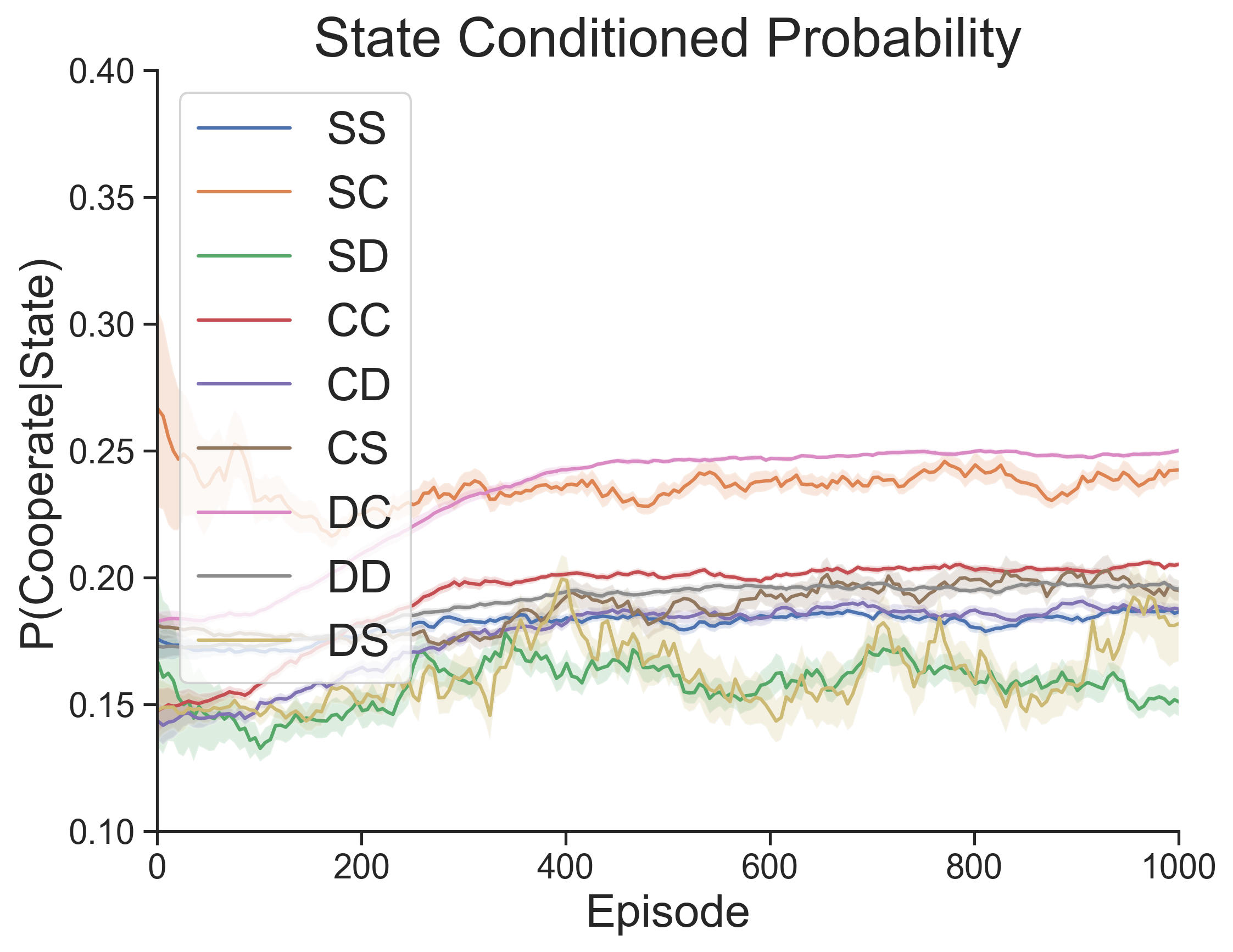}
      \caption{}
      \label{fig:earl_cg_eval_coop}
    \end{subfigure}
    \begin{subfigure}[b]{0.24\textwidth}
      \centering
      \includegraphics[width=\textwidth]{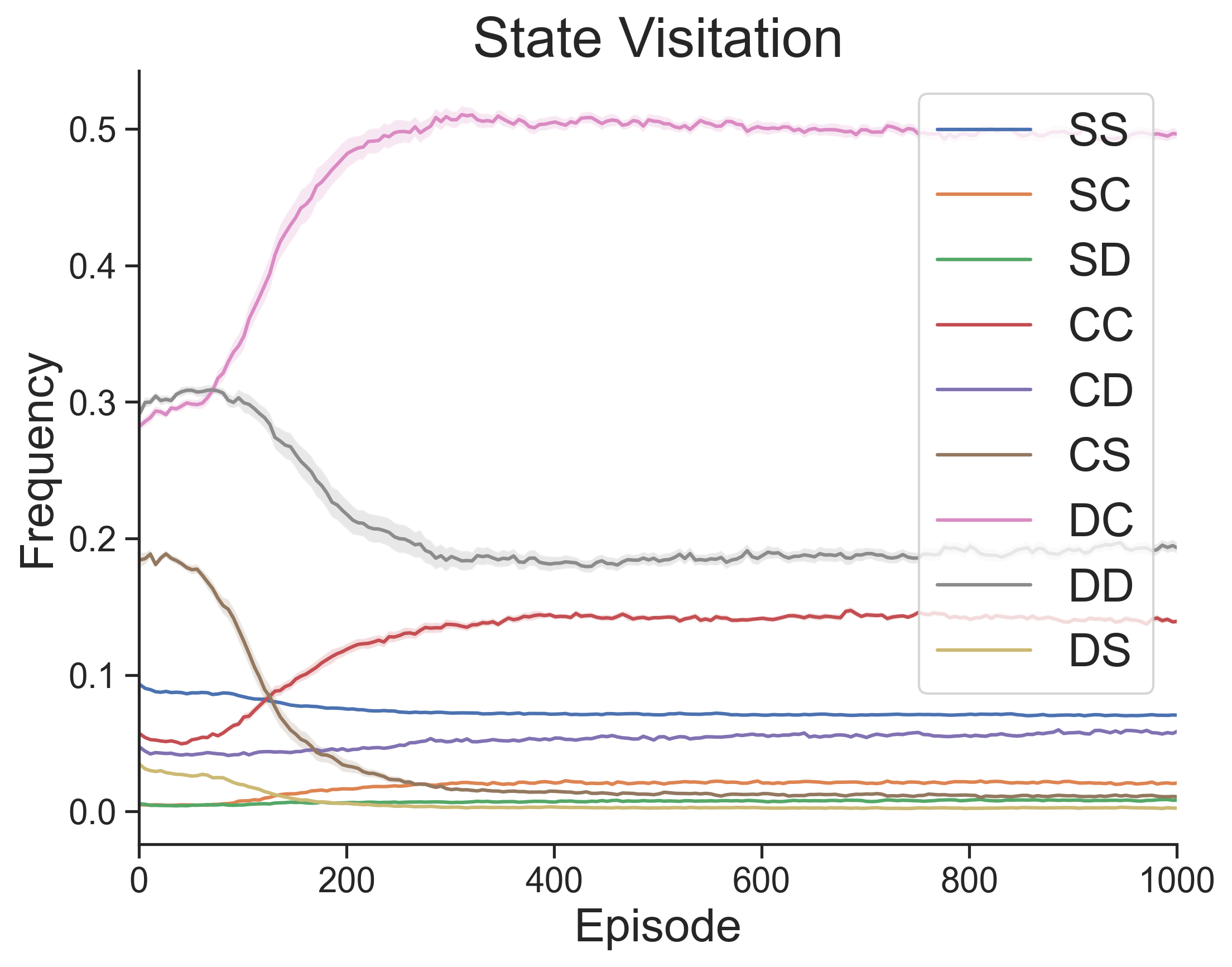}
      \caption{}
      \label{fig:earl_cg_eval_state}
    \end{subfigure}
    \begin{subfigure}[b]{0.24\textwidth}
      \centering
      \includegraphics[width=\textwidth]{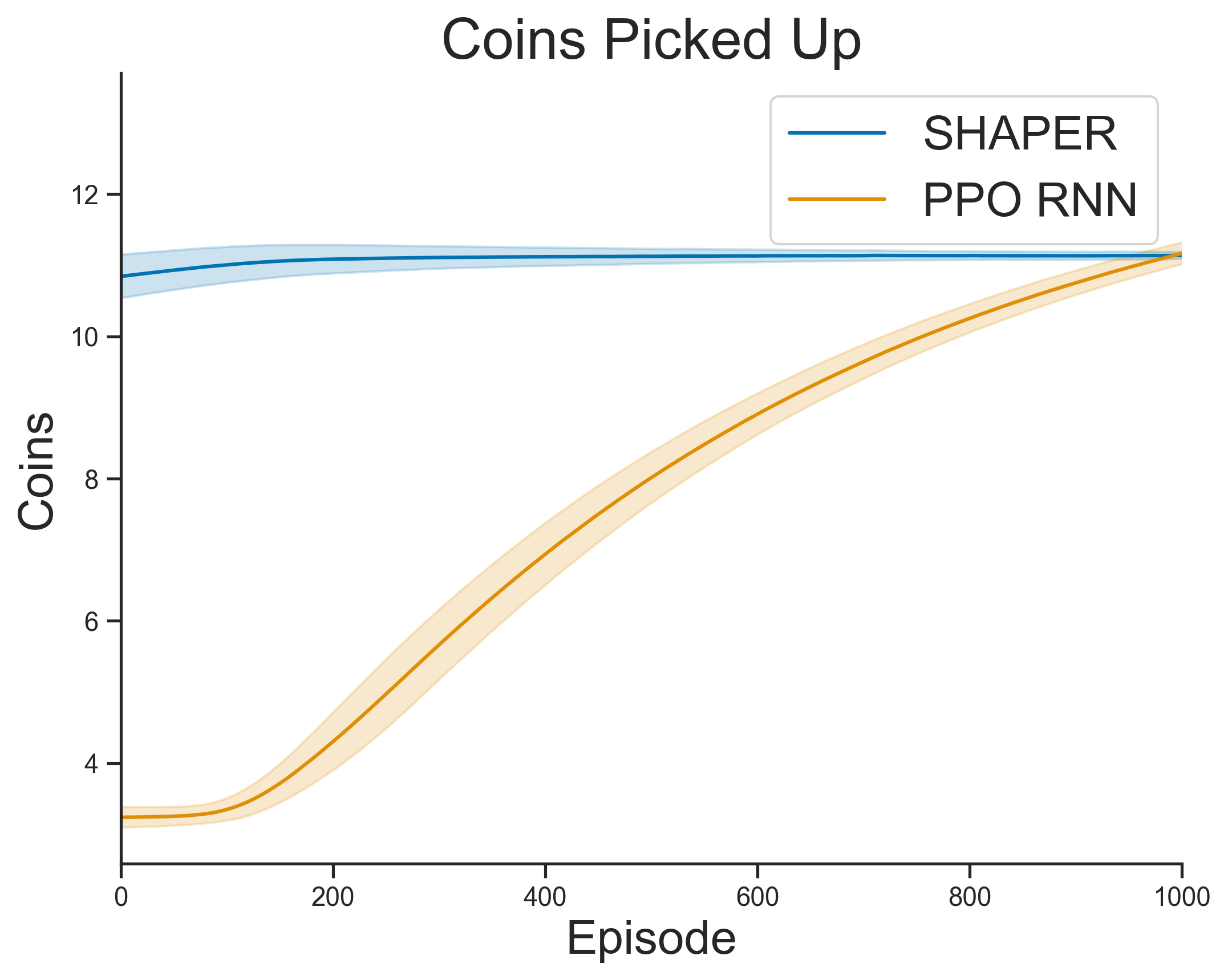}
      \caption{}
      \label{fig:earl_cg_eval_coins}
    \end{subfigure}
    \caption{Evaluation results over a single trial (with co-player) compromising over 100 seeds for the CoinGame. (a) Reward, (b) \method{}'s frequency of picking up its own colour coin, (c) state visitation, and (d) the number of coins picked up per episode. \method{} successfully elicits exploitation with a co-player with a high state visitation for DC and strong competency.}
    \label{fig:earl_ego_cg_viz}
\end{figure*}
\textbf{* in the Matrix} extends matrix games to gridworld environments~\citep{vezhnevets2020options}, where * is any normal-form game. For visual descriptions, see Figures \ref{fig:coin_game_viz}(c,d), \ref{fig:8subfigures}, and\ref{fig:ipditm_annotated}. Agents collect two types of resources into their inventory: \textit{Cooperate} and \textit{Defect} coins. Once an agent has collected any coin, the agent's colour changes, representing that the agent is ``ready'' for interaction. Agents can fire an `interact' beam to an area in front of them. If an agent's interact beam catches a ``ready'' agent, both receive rewards equivalent to playing a matrix game *, where their inventory represents their policy. For example, when agent 1's inventory is 1 \textit{Cooperate} coin and 3 \textit{Defect} coins, agent 1's probability to cooperate is 25\%. For all details see Appendix \ref{app:environment}.

* in the Matrix introduces a series of novel complexities for shaping over the CoinGame and finite matrix games. The environment is substantially more demanding than the previous games---it is partially observable, has complex interactions, and much longer time horizons. For shaping, partial observability makes temporally-extended actions harder to estimate. Shapers are also incentivised to speed up co-players learning, as the environment only allows interactions after both agents have picked up a coin. We explore two specific implementations of the game: ``IPD in the Matrix'' (IPDitM) and ``IMP in the Matrix'' (IMPitM).

\begin{figure}{}
    \centering
    \begin{subfigure}[b]{0.2\textwidth}
      \centering
      \includegraphics[width=\textwidth]{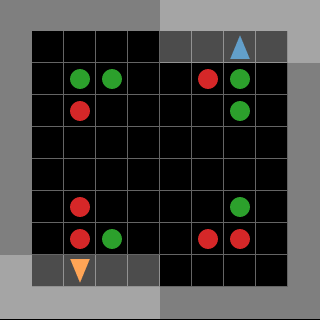}
      \label{fig:ipditm_a}
    \end{subfigure}
    \begin{subfigure}[b]{0.2\textwidth}
      \centering
      \includegraphics[width=\textwidth]{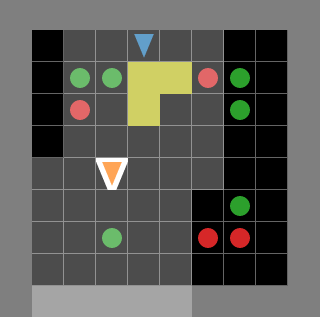}
      \label{fig:ipditm_b}
    \end{subfigure}
    \caption{Render of the IPDitM games, a multi-step, gridworld-based general-sum game. Agents with restricted visibility and orientation traverse a grid picking up either \textit{Defect} or \textit{Cooperate} coins. (left) shows an initial state of the game before either agent has a coin. Once agents pick up a coin, their appearance changes, and they can interact. (right) shows the orange agent having collected a coin and the blue agent firing their interact beam.
    }
\label{fig:ipditm_viz}
\end{figure}

\textbf{For our baseline comparisons}, we compare \method{} against multiple baselines: Naive Learners (NLs), variants of \textsc{M-FOS}, and \textsc{GS}. A NL does not explicitly account for the learning of the co-player across different episodes. In all of our experiments, the co-player is a NL. We train meta-agents until convergence in their respective environments. Then, we evaluate the performance of fixed meta-agents against new co-player (NL) initialisation $\phi_{-i}$.
Additional implementation details and hyperparameters for each game are provided in Appendix~\ref{appendix:cg_hyper}.\footnote{The codebase is open-source \citep{willi2023pax}.} 

In finite matrix games, our NL is parameterised as a tabular policy trained using PPO. In the gridworld environments, the NL is parameterised by a recurrent neural network and trained using PPO. Furthermore, in gridworlds, we compare to both M-FOS optimised with PPO and by ES. For GS, we only use ES, consistent with the original paper. 
We compare the performance of \method{} to two different types of NL pairs:
The first type, co-training NL (CT-NL), two NLs are initialised randomly and trained together using independent learning. This shows that avoiding unconditional defection is a challenge in the first place. 
The second type, pre-trained NL (PT-NL) instead takes an agent from a fully trained CT-NL pair and uses it as a naive shaper baseline, i.e., trains a NL as a best response to the fixed final policy.
This ensures that the performance of \method{} is not simply due to breaking the learning dynamics of the co-player, e.g., because the fully trained NL deprives a randomly initialised agent of all rewards. 
Specific details are provided in Appendix \ref{appendix:ipd_hyper}. Next, for our \textbf{ablations}, we consider three challenges:

\begin{figure*}[ht]
    \centering
    \begin{subfigure}[b]{0.32\textwidth}
      \centering
      \includegraphics[width=\textwidth]{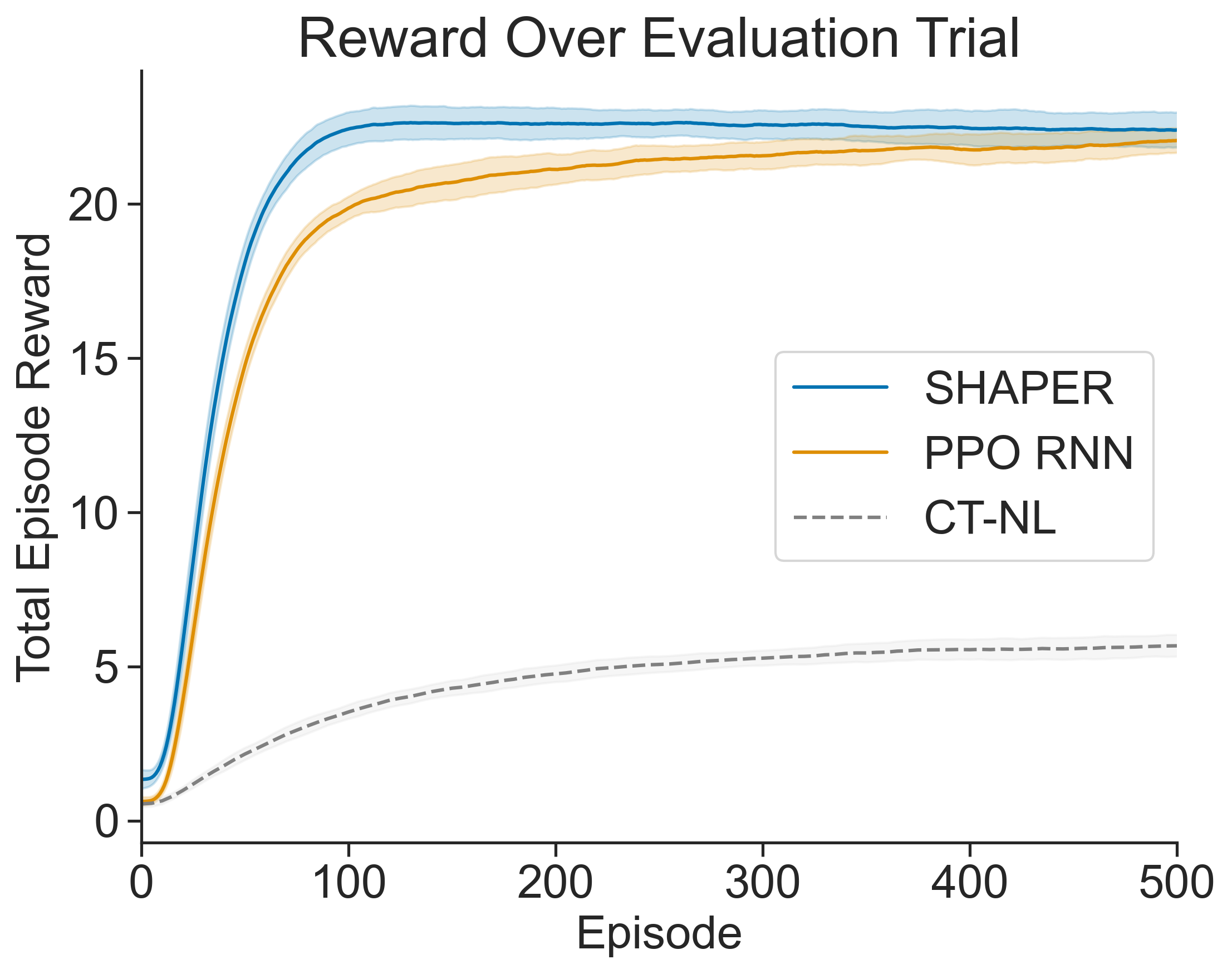}
      \caption{}
      \label{fig:shaper_ipditm_eval_reward}
    \end{subfigure}
    \hfill
    \begin{subfigure}[b]{0.32\textwidth}
      \centering
      \includegraphics[width=\textwidth]{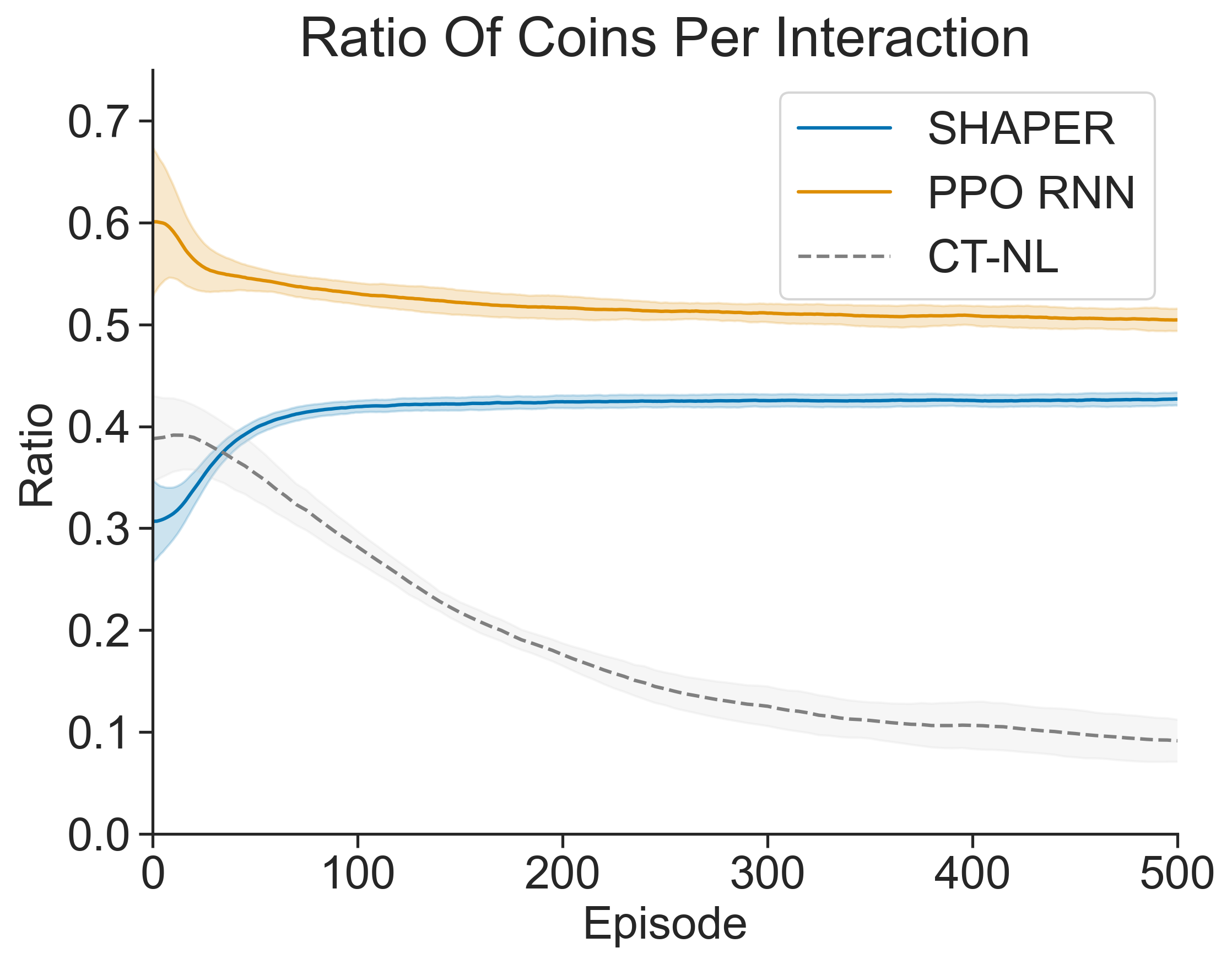}
      \caption{}
      \label{fig:shaper_ipditm_eval_ratio}
    \end{subfigure}
    \hfill
    \begin{subfigure}[b]{0.32\textwidth}
      \centering
      \includegraphics[width=\textwidth]{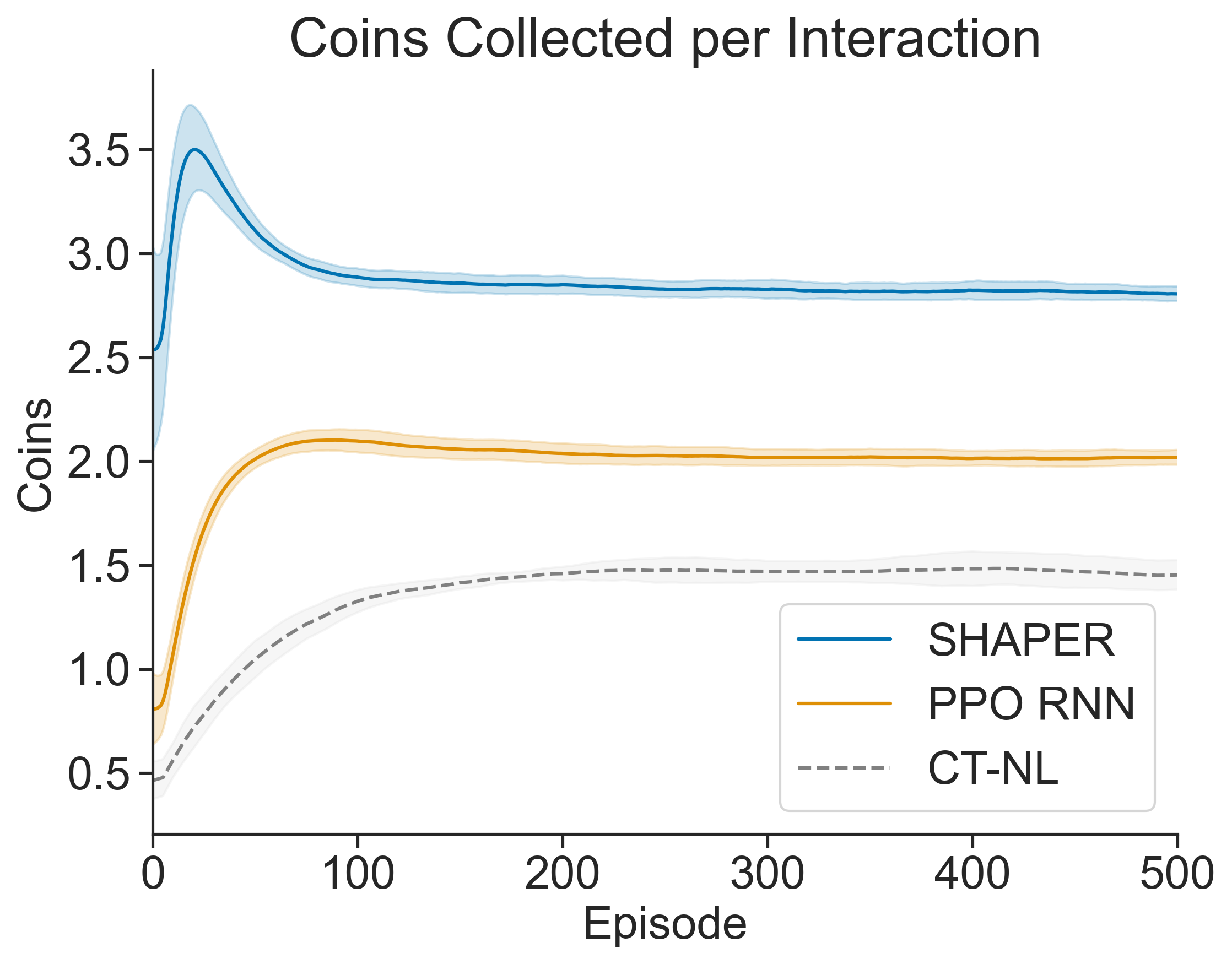}
      \caption{}
      \label{fig:shaper_ipditm_eval_coins}
    \end{subfigure}
    \caption{Evaluation results over a single trial (with co-player) compromising over 100 seeds for the IPDitM. (a) Mean reward per timestep, (b) mean ratio of picking up cooperate coins per soft-reset, (c) total number of coins picked up per soft-reset. The independent learner is shown to contrast what learning without a meta-agent would look like.
    }
    \label{fig:Shaper_ego_ipditm_viz}
\end{figure*}
\textbf{Context Challenge:} During a trial, after $k$ episodes, the co-player stops updating their parameters. When they stop updating, the shaper's optimal behaviour is to exploit the co-player's fixed policy (effectively stop shaping). We evaluate in the IPD and choose $k=2$. This challenge tests if shapers: 1) identify the sudden change in a co-player's learning dynamics, and 2) deploy a more suitable exploitative policy. We hypothesise that shapers without context cannot identify the change. We evaluate \method{} and compare against GS to understand the importance of context for shaping. 

\textbf{History Challenge:} We reset the hidden state of \method{} between episodes, removing its ability to use \textit{context} to shape (\method{} w/o context). We evaluate in IMP, and agents must infer the co-player's current policy using only history. Finally, we evaluate \method{} within IMP environment over short and long episode lengths (2 and 100, respectively) to limit the relative strength of \textit{history}.

\textbf{Average Challenge}: We also analyse the role of averaging across the batch in matrix games by comparing the performance difference of both MFOS and Shaper with and without averaging.

\section{Results}
\label{sec:results}
\textbf{Shaping in Finite Matrix Games:} We evaluate \method{}, \textsc{M-FOS} and GS on finite matrix games, i.e., long-time-horizon variants of the infinite matrix games used in prior work. We recreate previously reported extortion behaviour in a more challenging setting~\citep{lu2022model}.

\textbf{Insight 1: \method{} shapes the best in long-horizon iterated matrix games.} We inspect the converged reward for each shaping algorithm against a PPO agent in the IPD (see Table \ref{tab:matrix_results}). Here, \method{} shapes its co-player more effectively than the baselines, achieving an average return of -0.13 per episode. All shaping baselines reach extortion-like policies.

\textbf{Insight 2: Memory is important for shaping in the IMP.} In the IMP, \method{} exploits its opponent to achieve a score of $(0.9, -0.9)$ (see Table \ref{tab:matrix_results_mean}). As expected, GS cannot shape the opponent, achieving a score close to the Nash equilibrium, $(0.0, 0.0)$. With only a single-step history, it is impossible to shape the opponent because the opponent can switch to a random strategy between episodes to achieve a score of at least 0. Thus memory is required to find shaping strategies. We find that \textsc{M-FOS}, an agent with memory, shapes too. Next, we present our \textbf{CoinGame} results.

\textbf{Insight 3: Knowing how to navigate the gridworld and pick up coins is already enough to suppress co-player's learning.} Towards the end of meta-training, newly initialised co-players have to play against already competent meta-agents who have seen the game many times. We found that in CoinGame, it was sufficient for the meta-agents to pick up all coins before the co-player could reach them to hinder training. Therefore, we suggest checking that the co-player learns against pre-trained Naive Learners. This mitigates behaviours that prohibit the co-players from learning at all. We found that changing from a global to an egocentric observation space in the CoinGame helped co-players learn against pre-trained agents. Examples of sanity tests are found in Appendix \ref{appendix:cg_sanity}.

\begin{table}[tp]
\caption{
Converged reward per step (meta-agent, co-player) for agents against Naive Learners in finite matrix games. \method{} can shape co-players to exploitative equilibria. We report mean and standard deviation over 20 randomised co-players.
\label{tab:matrix_results_mean}}
    \centering
    \begin{tabular}{l cc}
     &  IPD  & IMP \\ 
    \midrule
     \method{} & \textbf{-0.1 $\pm$ 0.02, -2.8 $\pm$ 0.05}  & \textbf{0.9 $\pm$ 0.02, -0.9 $\pm$ 0.02}\\
     M-FOS & -0.6 $\pm$ 0.14, -2.3 $\pm$ 0.14
     & \textbf{0.8 $\pm$ 0.09, -0.8 $\pm$ 0.09} \\
     GS & -1.0 $\pm$ 0.03, -1.3 $\pm$ 0.10 & 
     0.0 $\pm$ 0.01, 0.0 $\pm$ 0.01 \\
     CT-NL & 
     -2.0 $\pm$ 0.00, -2.0 $\pm$ 0.00 & 0.0 $\pm$ 0.00, 0.0 $\pm$ 0.00 \\
    \bottomrule
    \end{tabular}
\end{table}

\begin{figure*}[h]
\centering
    \begin{subfigure}[b]{0.3\textwidth}
      \centering
      \includegraphics[width=\textwidth]{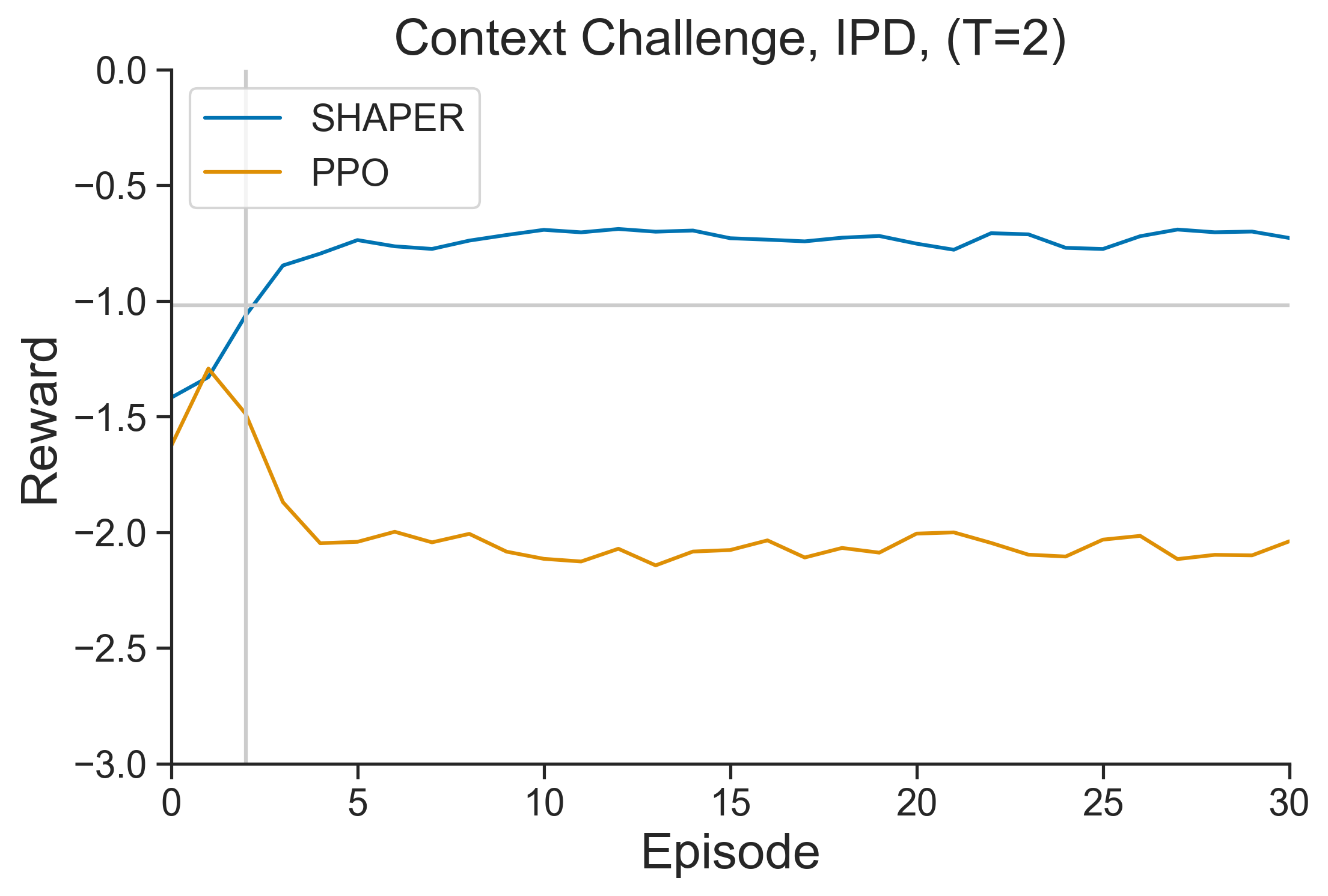}
      \caption{}
    \end{subfigure}
    \begin{subfigure}[b]{0.3\textwidth}
      \centering
      \includegraphics[width=\textwidth]{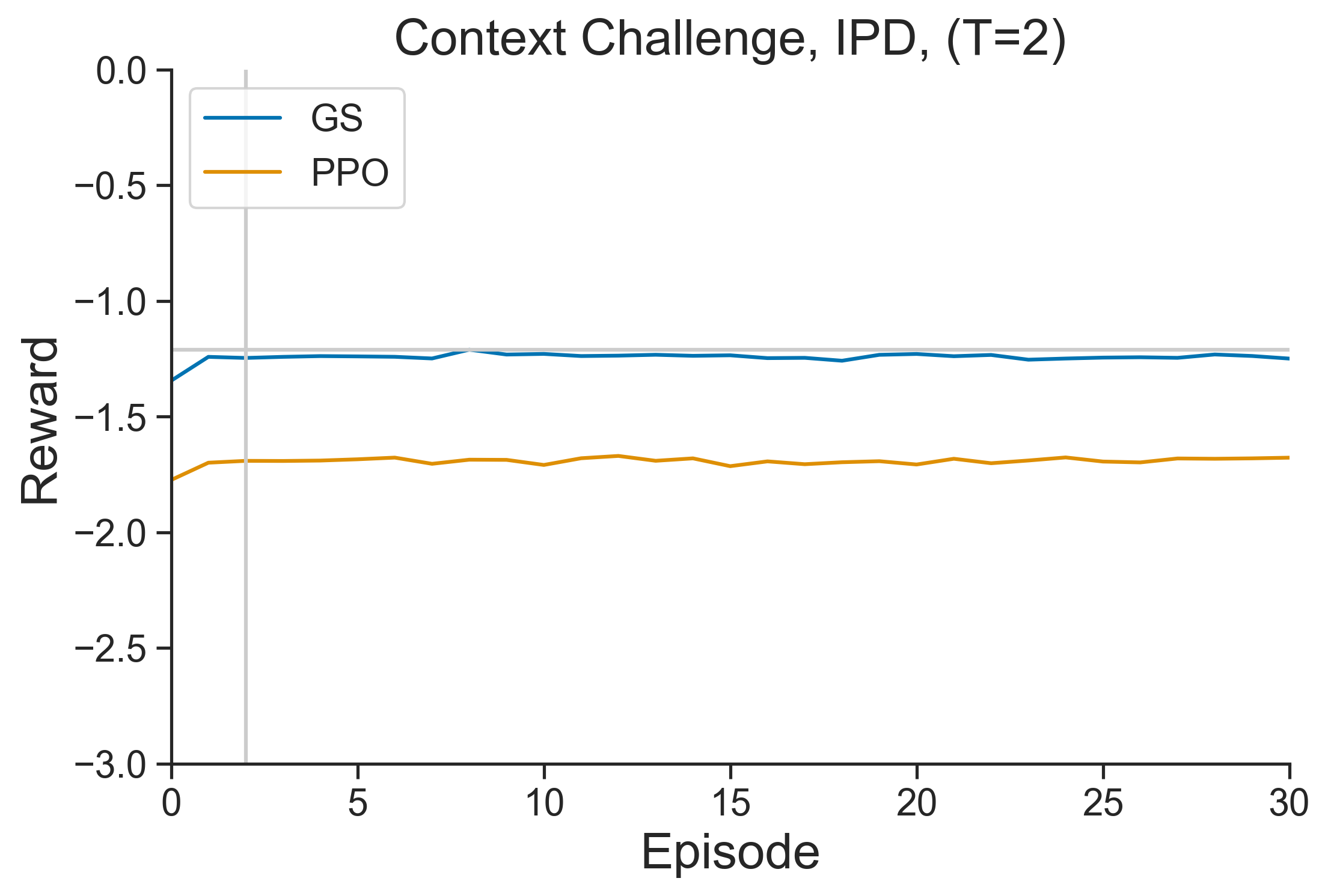}
        \caption{}
    \end{subfigure}
    \begin{subfigure}[b]{0.3\textwidth}
      \centering
      \includegraphics[width=\textwidth]{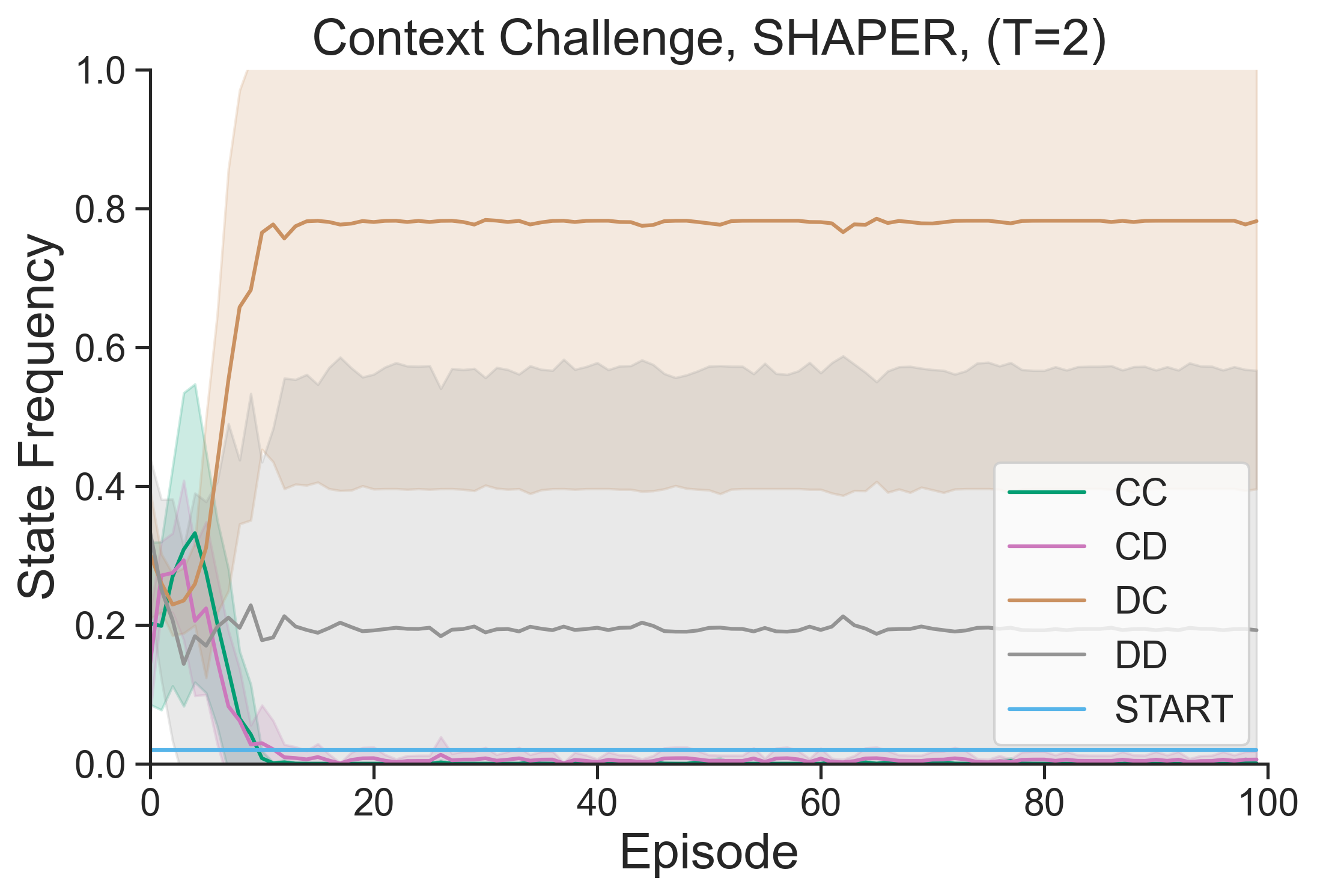}
      \caption{\label{fig:earl_coop_hardstop}}
    \end{subfigure}
\caption{Hardstop Challenge: Average reward per timestep over an evaluation trial for \method{} (a) and GS (b) against a Naive Learner in the IPD. Here GS fails to generalise to a co-player that stops learning after an unknown number of timesteps (unseen during training). (c)  State Visitation through the evaluation shows \method{} responds to co-players frozen policy by moving into either DD (the best response to a defective agent) or DC (the best response to a fully cooperative agent).
}
\label{fig:hardstop}
\end{figure*}

Reiterating \textit{Insight 1}, we find meta-agents find extortion-like policies in the CoinGame. To better understand behaviour in CoinGame, we extend the five states from the IPD (S, CC, CD, DC, DD) to include the start states (SS, SC, SD, CS, DS). At the start of an episode, the state is SS until a player picks up a coin. To understand how \method{} shapes, we inspect the probability of the meta-agent picking up a coin of its own colour at the start, i.e., SC $\rightarrow$ CC. For example, suppose the meta-agent were to cooperate unconditionally in the CoinGame. In that case, it only picks up coins of their own colour no matter the state and would relate to a high probability of cooperating over all states.  

Figure \ref{fig:earl_cg_eval_coop} demonstrates how \method{} shapes its co-players effectively already at the start. The difference between cooperating in SC and SS ($25\%, 15\%$ resp.) highlights how \method{} uses context to evaluate the exploitability of its co-player. In SS, when both agents have not picked up coins, \method{} probes for exploitability by not cooperating. In SC, where the co-player has already shown they are cooperative, \method{} also cooperates. Moreover, Figure \ref{fig:earl_cg_eval_state} shows that CS is visited more often than DS in early episodes ($18\%, 15\%$ resp.), indicating that \method{} is shaping the co-player to form a preference for picking up their own colour. This preference is then exploited by \method{} as indicated by the increasing visitation of DC. The meta-agent's probability of cooperating in DC converges to $25\%$, i.e., occasionally rewarding the co-player, as never cooperating would probably make the co-player learn pure defection.

\textbf{Insight 4: CoinGame is not suitable as a multi-step action environment}. We found GS produces comparable results to \method{}. At first, this is surprising since GS is a feedforward network and does not have access to the history (or, at most, one step). Therefore it should not be able to retaliate against a defecting agent since it has no memory of their past actions. However, a close investigation of the problem setting shows that due to particular environment dynamics, the \textit{current state} is often indicative of \textit{past actions}. For example, seeing two agents and a coin on the same square is a strong signal that one of the agents defected since this situation only could have arisen when either all objects spawn on the same square (occurs with a probability of $0.12\%$ and only at the beginning of an episode) or when both agents went for the same coin and the coin respawned on top of them (see Figure \ref{fig:coin_game_convention_break}). This illustrates that CoinGame allows for simple shaping strategies that do not require \textit{context} or \textit{history}, limiting its utility as a benchmark to measure temporally-extended actions.

We continue with our results for the \textbf{* in the Matrix} environments. Motivated by \textit{Insight 3}, we show that co-players learn against pre-trained agents by the number of coins collected in Table \ref{tab:result_gridworlds_1000}. 

\textbf{Insight 5: \method{} outperforms other shaping methods in the IPDitM by a considerable margin}
\method{} outperforms other shaping methods in the IPDitM by a considerable margin (see Table \ref{tab:result_gridworlds}), e.g., Shaper gets $\sim22.44$ points against NL, where M-FOS gets $\sim15.49$. Furthermore, \method{} finds a \textit{collectively better equilibrium for both players} over any other shaping method, e.g., in comparison with M-FOS, Shaper achieves $\left(\sim22.44, \sim21.49\right)$ and M-FOS gets $\left(\sim15.49, \sim 23.88\right) $.

\textbf{Insight 6: Shaping in \textit{IPDitM} leads to collectively and individually better outcomes.}
Table \ref{tab:result_gridworlds} (second column) shows that shaping (\method{}, \textsc{M-FOS}, and \textsc{GS}) leads to collectively and individually better outcomes in IPDitM compared to PT-NL or CT-NL.

\begin{table}[t]
\caption{
Ablations highlighting the importance of context and history for Shaping. We report converged reward per step (meta-agent, co-player) for agents against Naive Learners.
\label{tab:ablation}}
    \centering
    \begin{tabular}{l c }
    \toprule
     \multicolumn{2}{c}{Context Challenge:  IPD}   \\  
     \method{} & -0.8, -2.0 \\ 
     \method{} w/o Context & -1.25, -1.75\\ 
    \midrule
    \midrule
    \multicolumn{2}{c}{History Challenge: IMP (Length=2)} \\ 
     \method{} & 0.5, -0.5  \\ 
    \method{} w/o History & 
     0.0, 0.0 \\ 
    \midrule
    \midrule
    \multicolumn{2}{c}{History Challenge: IMP (Length=100)} \\ 
     \method{} & 0.5, -0.5 \\ 
    \method{} w/o History & 
     0.5, -0.5 \\ 
    \bottomrule
    \end{tabular}
\end{table}

\textbf{Insight 7: \method{} shapes by picking up almost all coins at the beginning of a trial.}
The meta-agent picks up almost all coins in the grid in the first 20 episodes ($\approx 3.5$, see Figure \ref{fig:shaper_ipditm_eval_coins}), especially \textit{Defect} coins. This leaves only \textit{Cooperate} coins for co-players. Interacting with a more cooperative ratio, the co-player receives some reward, reinforcing the co-player to play a cooperative ratio in the future. Figure \ref{fig:shaper_ipditm_eval_ratio} shows the meta-agent and co-player converge to collecting a large ratio of \textit{Cooperate} coins ($\approx(0.4, 0.6)$), in contrast to independent learners ($\approx 0.1$) (grey dashed line). Interestingly, a (meta-agent, co-player) pair collects more coins ($\approx(3.0, 2.0)$) than a pair of independent agents ($\approx(1.5, 1.5)$) - this is because the independent learners maximise their return under mutual defection only by increasing interactions within an episode.

In the IMPitM, GS does not learn to shape, as expected from \textit{Insight 2}, whereas \textsc{M-FOS} and \method{} does. \method{} and \textsc{M-FOS} achieve similar performances. (see Table \ref{tab:result_gridworlds}).

\begin{figure*}[h]
    \begin{subfigure}[b]{0.23\textwidth}
      \centering
      \includegraphics[width=\textwidth]{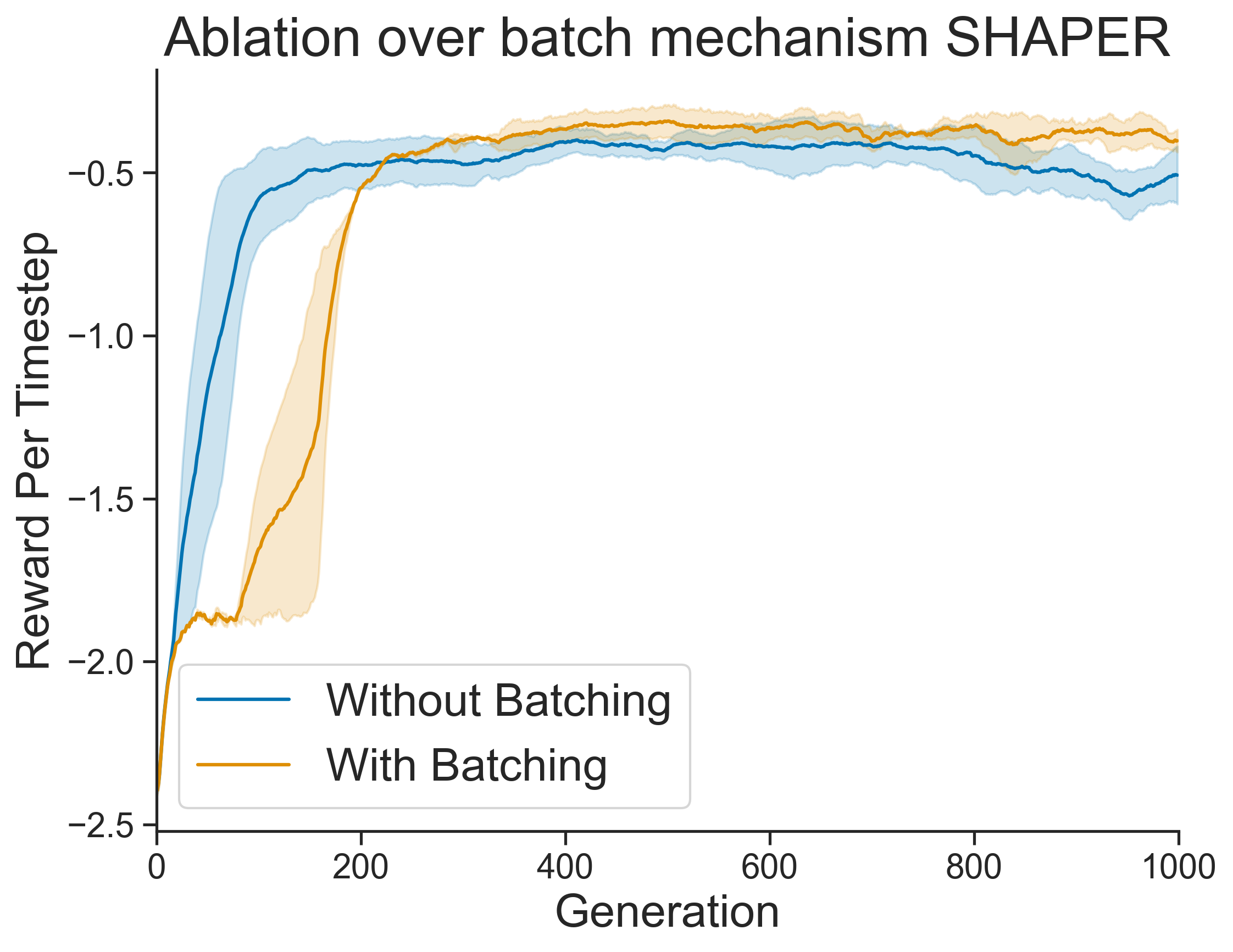}
      \caption{}
      \label{fig:batching_shaper_ipd}
    \end{subfigure}
    \begin{subfigure}[b]{0.23\textwidth}
      \centering
      \includegraphics[width=\textwidth]{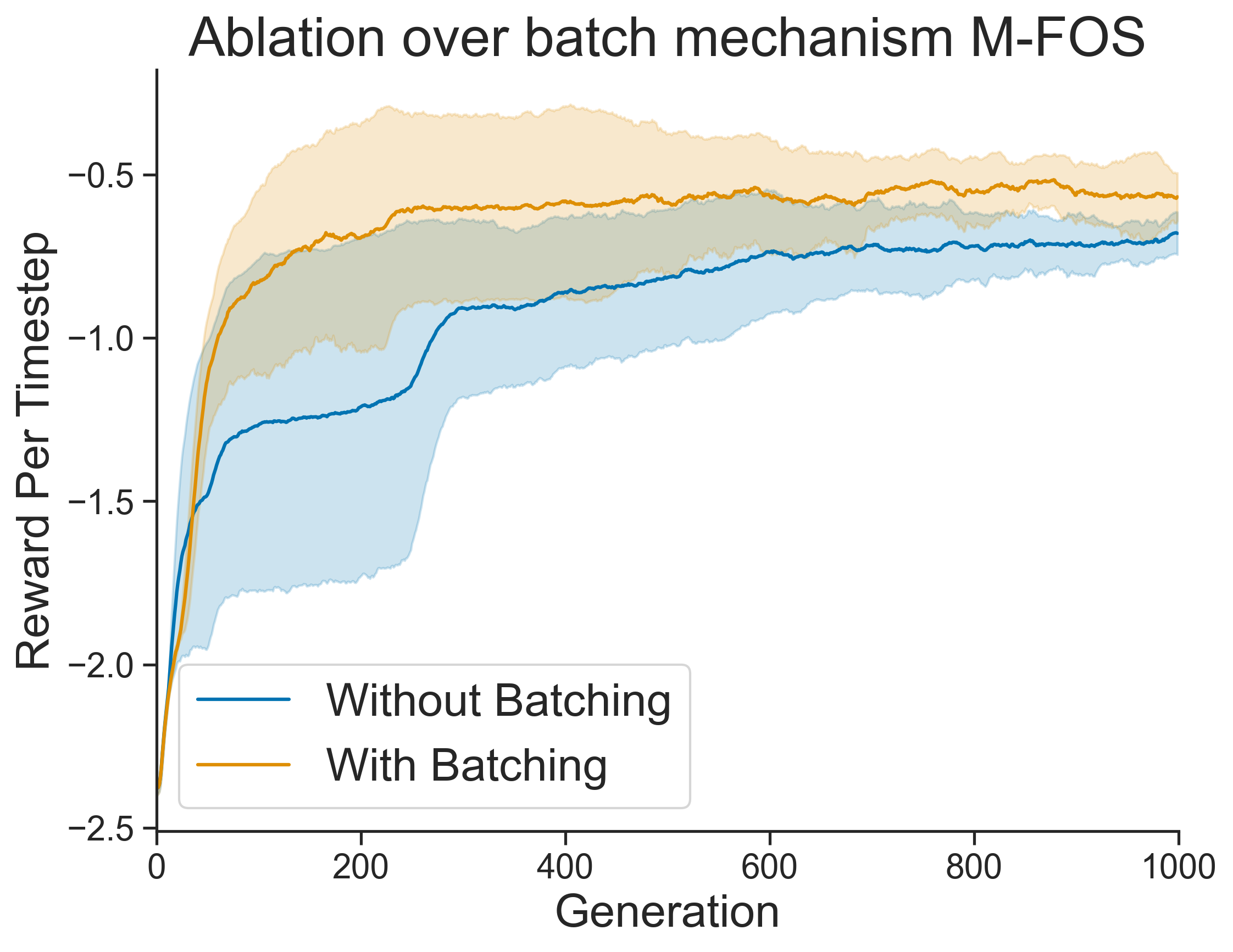}
      \caption{}
      \label{fig:batching_mfos_ipd}
    \end{subfigure}
    \centering
    \begin{subfigure}[b]{0.23\textwidth}
      \centering
      \includegraphics[width=\textwidth]{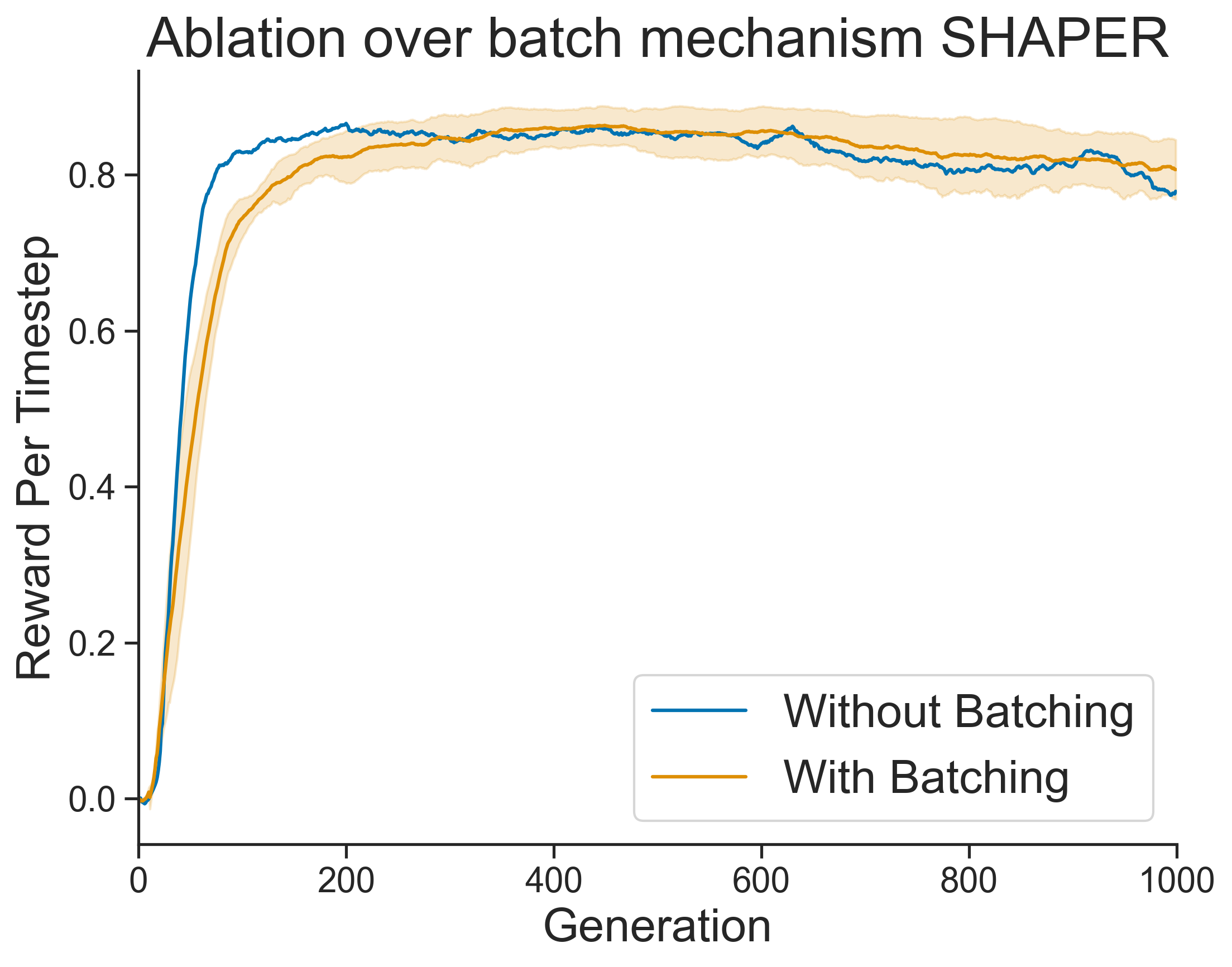}
      \caption{}
      \label{fig:batching_shaper_imp}
    \end{subfigure}
    \begin{subfigure}[b]{0.23\textwidth}
      \centering
      \includegraphics[width=\textwidth]{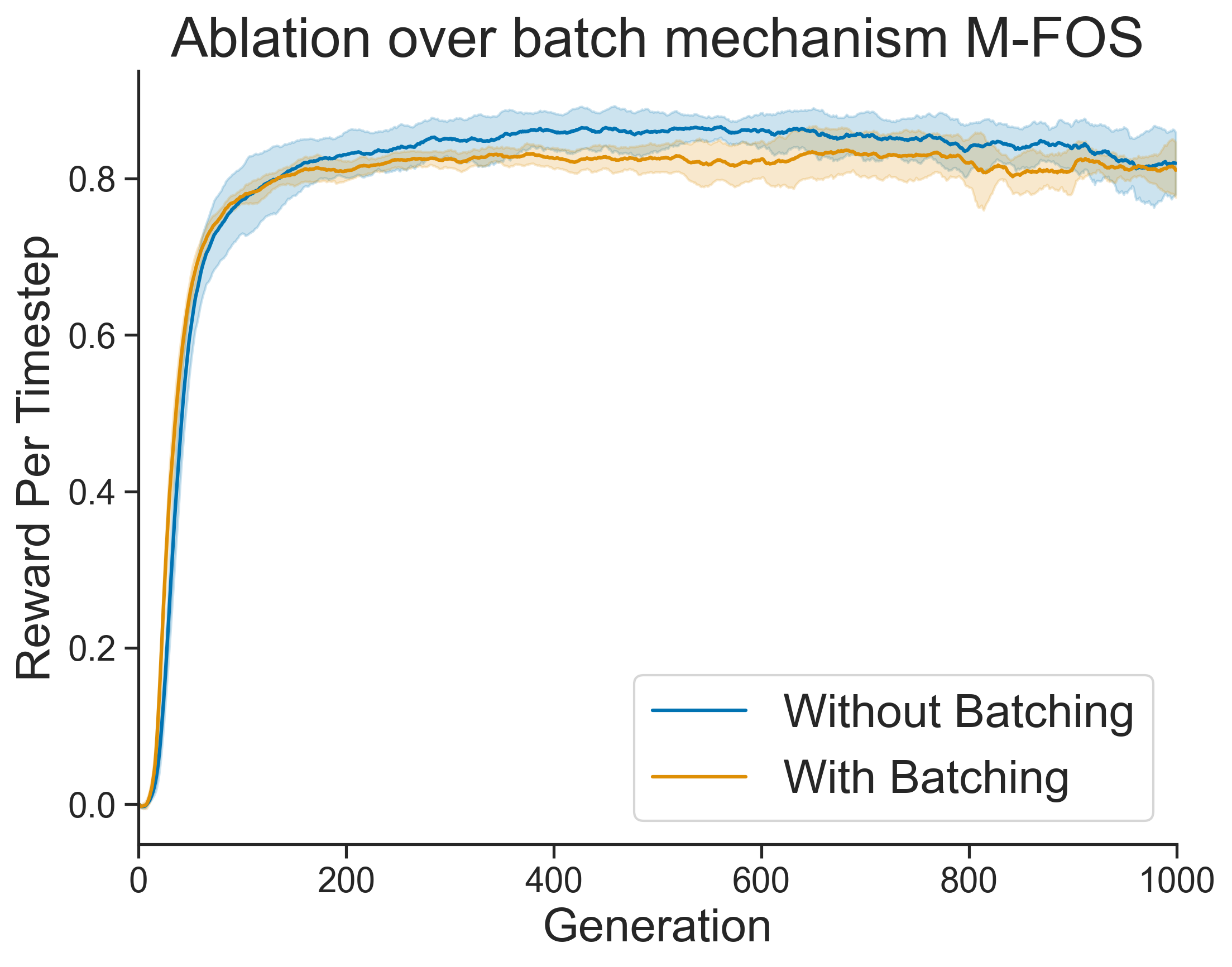}
      \caption{}
      \label{fig:batching_mfos_imp}
    \end{subfigure}
    \caption{Reward per timestep throughout training for The ``Average'' challenge. Results are presented over matrix games for 5 seeds. In a) and b) we evaluate OS methods on IPD and in c) and d) we evaluate on IMP. We note that batching only helps M-FOS in the IPD. This indicates batching is only useful in sufficiently diverse environments, relative to the OS method.} 
\end{figure*}

\textbf{Insight 8: \textsc{Shaper} empirically tends to find better shaping policies than \textsc{M-FOS} in IPDitM.}
\method{} outperforms \textsc{M-FOS} in Table \ref{tab:result_gridworlds}, providing evidence that \method{} scales to more complex policies. \method{} demonstrates shaping, as indicated by the final rewards, which are significantly higher for both agents than M-FOS IPDitM. We postulate that as \textsc{M-FOS} architecture is as expressive as \textsc{Shaper}, its complexities and biases hinder ES' ability to find optimal solutions (for training training curves, see Appendix \ref{appendix:ipditm_training}).

In Table \ref{tab:x_play_results}, we show that \method{} finds policies leading to improved global welfare in cross-play with M-FOS and GS. In cross-play, the shaping algorithms are trained against Naive Learners and evaluated against each other. This experiment motivates that \method{}'s inductive biases leads to finding more robust policies even when evaluated out of distribution. Note that Shaper vs. Shaper achieves similar scores as M-FOS vs M-FOS. However, Shaper achieves better scores against M-FOS (7.32 ± 0.34, 5.08 ± 0.36) and GS (28.61 ± 1.82, 20.23 ± 1.27). Also, note how GS achieves its highest payoff when playing against Shaper. 

In our \textbf{ablations}, we find that context is beneficial for shaping in the IMP. In the ``Context Challenge'', \method{} (-0.8) outperforms \method{} w/o Context (-1.25) (see Table \ref{tab:ablation}). For shaping to occur in this challenge, we expect methods to change their strategy at $e=2$ episodes. \method{} demonstrates dynamic shaping by switching, yet \method{} w/o Context's policy does not adapt and does not exploit the stop (see Fig. \ref{fig:hardstop}). This result provides evidence that context is needed to shape.

In the ``History Challenge'', when playing the IMP with a small number of inner-episodes ($e=2$), we expect meta-agents without context to be unable to identify co-players' current learning and thus cannot shape. We find that \method{} shapes agents, whilst \method{} w/o Context does not shape agents as indicated by better rewards, 0.5 vs 0.0 (Fig. \ref{fig: only_history}). Interestingly, we also found that with a longer inner-episode length ($E=100$), \method{} w/o Context uses \textit{history} to shape its co-player (Fig.\ref{fig:gs_two_step_length_100}). This shows that history can encode co-players' learning dynamics in some environments.

In the ``Average Challenge'', we find that averaging across the batch only helps M-FOS in the IPD, as it improves convergence speed. In all other scenarios, averaging across the batch did not significantly improve performance (see Figure \ref{fig:batching_mfos_ipd}). Shaping agents must approximate, via observations, a co-players update rule. If this update is batched (such as with stochastic gradient descent), the batching mechanism should in theory provide a better estimate. If the batching mechanism is not required, this suggests experience in the update is not diverse. Comparing games, the diversity of co-player behaviours within the IMP is much less than IPD. Within the IPD, \method{} sees no improvement with the batch mechanism compared to M-FOS (see Figure \ref{fig:batching_shaper_ipd} - \ref{fig:batching_mfos_ipd}). Here we postulate that given M-FOS has a limited context (1-step), batching provides M-FOS with greater context such that it can infer co-player learner. Shaper does not require averaging as it captures more context via its hidden than M-FOS does. This suggests that moving forward, OS methods should consider Context, History and Batching, as mechanisms for observing the experience / learning of co-players.

\begin{table}[t]
    \centering
    \caption{Episodic reward for in a single evaluation trial against different OS shaping methods in IPDitM. Neither agent takes gradient updates, but those with memory \textsc{Shaper} and \textsc{M-FOS} use memory to change their policy during the trial. Results are reported for the row player in each match. We report mean and std over 5 seeds.}
    \label{tab:x_play_results}

    \begin{tabular}{@{}lccc@{}}
        \toprule
        & \textsc{Shaper} & \textsc{GS} & \textsc{M-FOS} \\
        \midrule
    \textsc{shaper} & $16.48 \pm 0.88$ & $28.61 \pm 1.82$ & $7.32 \pm 0.34$ \\
    \textsc{gs} & $20.23 \pm 1.27$ & $0 \pm 0$ & $1.91 \pm 0.27$ \\
    \textsc{m-fos} & $5.08 \pm 0.36$ & $1.35 \pm 0.28$ & $16.25 \pm 0.95$ \\
    \bottomrule
			\bottomrule
		\end{tabular}
  \vspace{-1.5\baselineskip}
\end{table}
\section{Related Work}
\label{sec:related-work}

\textbf{Opponent Shaping} methods explicitly account for their opponent's learning. Just like \method{}, these approaches recognise that the actions of any one agent influence their co-players policy and seek to use this mechanism to their advantage~\citep{foerster_learning_2018, letcher_differentiable_2019, kim2021meta-mapg, willi2022cola, zhao2022proximal, fung2023analyzing}. However, in contrast to \method{}, these approaches require privileged information to shape their opponents. These models are also myopic since anticipating many steps is intractable due to the difficulty of estimating higher-order gradients. \citet{balaguer2022good} and  \citet{lu2022model} solve the issues above by framing opponent shaping as a meta reinforcement learning problem, which allows them to account for long-term shaping, where there is no need for higher-order gradients.

\textbf{Algorithms for Social Dilemmas} often achieve desirable outcomes in high-dimensional social dilemmas yet assume access to hand-crafted notions of adherence~\cite{yuan2022adherence}, social influence~\cite{jaques2019social, cicero2022brown}, gifting~\cite{lupu2020gifting} or social conventions~\cite{koster2020model}. While these approaches can achieve desirable outcomes, they change the agent's objectives and alter the dynamics of the underlying game.

\textbf{Multi-Agent Meta-Learning} methods have also shown success in general-sum games with other learners \citep{alshedivat2018mpg, kim2021policy, wu2021l2e}. Similar to \method{}, they take inspiration from meta-RL - their approach is to learn the optimal initial parameterisation for the meta-agent akin to Model-Agnostic Meta Learning \citep{finn2017maml}. In contrast, \method{} uses an approach similar to RL$^2$ \citep{duan2016rl2}, which trains an RNN-based agent to implement efficient learning for its next task. Finally, \method{} is optimised using ES, which empirically performs better with long-time horizons than policy-gradient methods \citep{lu2022dpo,lu2022model,lu2022adversarial}.
\vspace{-0.25\baselineskip}

\section{Conclusion}
When agents interact, the actions of each agent influence the rewards and observations of others and, through their learning, ultimately affect their behaviour. Leveraging this connection is called opponent shaping, and has received considerable attention recently.

This paper introduces \method{}, a shaping method suitable for high-dimensional games. We are the first to scale shaping successfully to long-time horizon general-sum games with temporally-extended actions, and we provide extensive performance analysis in these settings. We formalise the concept of history and context for shaping and analyse their respective roles empirically. Next, we formalise the previously implicit concept of averaging across the batch and show that it's helpful for previous methods to learn. Future work might investigate scenarios where averaging across a batch is also necessary for \method{}. Finally, we identify a fundamental problem in the widely-used CoinGame.
\newpage


\begin{acks}
We’d like to thank members of the UCL DARK lab, members of the Oxford FLAIR, members of Deepmind Gamma, Mika Semyalven and Joel Liebo for fruitful discussions and comments on an earlier draft of this paper. AK was supported by the EPSRC Grant EP/S021566/1, UCL International Scholar Award for Doctoral Training Centres and Cooperative AI Foundation Grant (1201294). This work was supported by an Oracle for Research Cloud Grant (19158657).
\end{acks}

\section{Ethics Statement*}
Shaping can be used for good and bad. Empirically, shaping has lead learning agents to find more prosocial solutions in mixed-incentive settings. However, one can imagine scenarios where shaping is used with a negative impact on society.
Assuming that learning agents will be deployed in the real world, e.g., online learning self-driving cars, it is important we understand how such agents interact. Early opponent shaping research has already shown that two naive agents mutually defect in the iterated prisoner’s dilemma and that opponent shaping leads to the more prosocial tit-for-tat strategy. It is important that we develop these methods further, investigate if they keep leading to more prosocial outcomes even in more difficult environments, and if not, what improvements can we make such that they do. In our work, we show that in grid-worlds with temporally-extended actions and long-time horizons, opponent shaping tends to find more prosocial solutions than Naive Learners.
Investigating shaping is important to prevent misuse of the paradigm. We are at the beginning of fundamental research in shaping and better understanding the necessary components to achieve shaping will help us to better control shaping agents.
Opponent Shaping is still in an early phase of development and practical implications are limited so immediate negative societal influence is unlikely.


\bibliographystyle{ACM-Reference-Format} 
\bibliography{main}


\begin{thebibliography}{47}


\ifx \showCODEN    \undefined \def \showCODEN     #1{\unskip}     \fi
\ifx \showDOI      \undefined \def \showDOI       #1{#1}\fi
\ifx \showISBNx    \undefined \def \showISBNx     #1{\unskip}     \fi
\ifx \showISBNxiii \undefined \def \showISBNxiii  #1{\unskip}     \fi
\ifx \showISSN     \undefined \def \showISSN      #1{\unskip}     \fi
\ifx \showLCCN     \undefined \def \showLCCN      #1{\unskip}     \fi
\ifx \shownote     \undefined \def \shownote      #1{#1}          \fi
\ifx \showarticletitle \undefined \def \showarticletitle #1{#1}   \fi
\ifx \showURL      \undefined \def \showURL       {\relax}        \fi
\providecommand\bibfield[2]{#2}
\providecommand\bibinfo[2]{#2}
\providecommand\natexlab[1]{#1}
\providecommand\showeprint[2][]{arXiv:#2}

\bibitem[\protect\citeauthoryear{Al-Shedivat, Bansal, Burda, Sutskever, Mordatch, and Abbeel}{Al-Shedivat et~al\mbox{.}}{2018}]%
        {alshedivat2018mpg}
\bibfield{author}{\bibinfo{person}{Maruan Al-Shedivat}, \bibinfo{person}{Trapit Bansal}, \bibinfo{person}{Yura Burda}, \bibinfo{person}{Ilya Sutskever}, \bibinfo{person}{Igor Mordatch}, {and} \bibinfo{person}{Pieter Abbeel}.} \bibinfo{year}{2018}\natexlab{}.
\newblock \showarticletitle{Continuous Adaptation via Meta-Learning in Nonstationary and Competitive Environments}. In \bibinfo{booktitle}{\emph{International Conference on Learning Representations}}.
\newblock


\bibitem[\protect\citeauthoryear{Axelrod and Hamilton}{Axelrod and Hamilton}{1981}]%
        {axelrod1981evolution}
\bibfield{author}{\bibinfo{person}{Robert Axelrod} {and} \bibinfo{person}{William~D Hamilton}.} \bibinfo{year}{1981}\natexlab{}.
\newblock \showarticletitle{The evolution of cooperation}.
\newblock \bibinfo{journal}{\emph{science}} \bibinfo{volume}{211}, \bibinfo{number}{4489} (\bibinfo{year}{1981}), \bibinfo{pages}{1390--1396}.
\newblock


\bibitem[\protect\citeauthoryear{Bakhtin, Brown, Dinan, Farina, Flaherty, Fried, Goff, Gray, Hu, Jacob, Komeili, Konath, Kwon, Lerer, Lewis, Miller, Mitts, Renduchintala, Roller, Rowe, Shi, Spisak, Wei, Wu, Zhang, and Zijlstra}{Bakhtin et~al\mbox{.}}{2022}]%
        {cicero2022brown}
\bibfield{author}{\bibinfo{person}{Anton Bakhtin}, \bibinfo{person}{Noam Brown}, \bibinfo{person}{Emily Dinan}, \bibinfo{person}{Gabriele Farina}, \bibinfo{person}{Colin Flaherty}, \bibinfo{person}{Daniel Fried}, \bibinfo{person}{Andrew Goff}, \bibinfo{person}{Jonathan Gray}, \bibinfo{person}{Hengyuan Hu}, \bibinfo{person}{Athul~Paul Jacob}, \bibinfo{person}{Mojtaba Komeili}, \bibinfo{person}{Karthik Konath}, \bibinfo{person}{Minae Kwon}, \bibinfo{person}{Adam Lerer}, \bibinfo{person}{Mike Lewis}, \bibinfo{person}{Alexander~H. Miller}, \bibinfo{person}{Sasha Mitts}, \bibinfo{person}{Adithya Renduchintala}, \bibinfo{person}{Stephen Roller}, \bibinfo{person}{Dirk Rowe}, \bibinfo{person}{Weiyan Shi}, \bibinfo{person}{Joe Spisak}, \bibinfo{person}{Alexander Wei}, \bibinfo{person}{David Wu}, \bibinfo{person}{Hugh Zhang}, {and} \bibinfo{person}{Markus Zijlstra}.} \bibinfo{year}{2022}\natexlab{}.
\newblock \showarticletitle{Human-level play in the game of Diplomacy by combining language models with strategic reasoning}.
\newblock \bibinfo{journal}{\emph{Science}} \bibinfo{volume}{378}, \bibinfo{number}{6624} (\bibinfo{year}{2022}), \bibinfo{pages}{1067--1074}.
\newblock
\urldef\tempurl%
\url{https://doi.org/10.1126/science.ade9097}
\showDOI{\tempurl}


\bibitem[\protect\citeauthoryear{Balaguer, Koster, Summerfield, and Tacchetti}{Balaguer et~al\mbox{.}}{2022}]%
        {balaguer2022good}
\bibfield{author}{\bibinfo{person}{Jan Balaguer}, \bibinfo{person}{Raphael Koster}, \bibinfo{person}{Christopher Summerfield}, {and} \bibinfo{person}{Andrea Tacchetti}.} \bibinfo{year}{2022}\natexlab{}.
\newblock \bibinfo{title}{The Good Shepherd: An Oracle Agent for Mechanism Design}.
\newblock \bibinfo{howpublished}{arXiv preprint arXiv:2202.10135}.
\newblock


\bibitem[\protect\citeauthoryear{Bradbury, Frostig, Hawkins, Johnson, Leary, Maclaurin, Necula, Paszke, Vander{P}las, Wanderman-{M}ilne, and Zhang}{Bradbury et~al\mbox{.}}{2018}]%
        {jax2018github}
\bibfield{author}{\bibinfo{person}{James Bradbury}, \bibinfo{person}{Roy Frostig}, \bibinfo{person}{Peter Hawkins}, \bibinfo{person}{Matthew~James Johnson}, \bibinfo{person}{Chris Leary}, \bibinfo{person}{Dougal Maclaurin}, \bibinfo{person}{George Necula}, \bibinfo{person}{Adam Paszke}, \bibinfo{person}{Jake Vander{P}las}, \bibinfo{person}{Skye Wanderman-{M}ilne}, {and} \bibinfo{person}{Qiao Zhang}.} \bibinfo{year}{2018}\natexlab{}.
\newblock \bibinfo{booktitle}{\emph{{JAX}: composable transformations of {P}ython+{N}um{P}y programs}}.
\newblock
\urldef\tempurl%
\url{http://github.com/google/jax}
\showURL{%
\tempurl}


\bibitem[\protect\citeauthoryear{Brown and Sandholm}{Brown and Sandholm}{2017}]%
        {brown2017libratus}
\bibfield{author}{\bibinfo{person}{Noam Brown} {and} \bibinfo{person}{Tuomas Sandholm}.} \bibinfo{year}{2017}\natexlab{}.
\newblock \showarticletitle{Libratus: the superhuman AI for no-limit poker}. In \bibinfo{booktitle}{\emph{Proceedings of the Twenty-Sixth International Joint Conference on Artificial Intelligence}}.
\newblock


\bibitem[\protect\citeauthoryear{Dafoe, Hughes, Bachrach, Collins, McKee, Leibo, Larson, and Graepel}{Dafoe et~al\mbox{.}}{2021}]%
        {dafoe2020open}
\bibfield{author}{\bibinfo{person}{Allan Dafoe}, \bibinfo{person}{Edward Hughes}, \bibinfo{person}{Yoram Bachrach}, \bibinfo{person}{Tantum Collins}, \bibinfo{person}{Kevin~R McKee}, \bibinfo{person}{Joel~Z Leibo}, \bibinfo{person}{Kate Larson}, {and} \bibinfo{person}{Thore Graepel}.} \bibinfo{year}{2021}\natexlab{}.
\newblock \showarticletitle{Open Problems in Cooperative AI}. In \bibinfo{booktitle}{\emph{Cooperative AI workshop}}.
\newblock


\bibitem[\protect\citeauthoryear{Dawes}{Dawes}{1980}]%
        {dawes1980social}
\bibfield{author}{\bibinfo{person}{Robyn~M Dawes}.} \bibinfo{year}{1980}\natexlab{}.
\newblock \showarticletitle{Social dilemmas}.
\newblock \bibinfo{journal}{\emph{Annual review of psychology}} \bibinfo{volume}{31}, \bibinfo{number}{1} (\bibinfo{year}{1980}), \bibinfo{pages}{169--193}.
\newblock


\bibitem[\protect\citeauthoryear{Duan, Schulman, Chen, Bartlett, Sutskever, and Abbeel}{Duan et~al\mbox{.}}{2016}]%
        {duan2016rl2}
\bibfield{author}{\bibinfo{person}{Yan Duan}, \bibinfo{person}{John Schulman}, \bibinfo{person}{Xi Chen}, \bibinfo{person}{Peter~L. Bartlett}, \bibinfo{person}{Ilya Sutskever}, {and} \bibinfo{person}{Pieter Abbeel}.} \bibinfo{year}{2016}\natexlab{}.
\newblock \bibinfo{title}{RL{\textdollar}{\^{}}2{\textdollar}: Fast Reinforcement Learning via Slow Reinforcement Learning}.
\newblock \bibinfo{howpublished}{arXiv preprint arXiv:1611.02779}.
\newblock


\bibitem[\protect\citeauthoryear{Ellis, Moalla, Samvelyan, Sun, Mahajan, Foerster, and Whiteson}{Ellis et~al\mbox{.}}{2022}]%
        {ellis22022smacv2}
\bibfield{author}{\bibinfo{person}{Benjamin Ellis}, \bibinfo{person}{Skander Moalla}, \bibinfo{person}{Mikayel Samvelyan}, \bibinfo{person}{Mingfei Sun}, \bibinfo{person}{Anuj Mahajan}, \bibinfo{person}{Jakob~N. Foerster}, {and} \bibinfo{person}{Shimon Whiteson}.} \bibinfo{year}{2022}\natexlab{}.
\newblock \showarticletitle{SMACv2: An Improved Benchmark for Cooperative Multi-Agent Reinforcement Learning}.
\newblock \bibinfo{journal}{\emph{arXiv preprint arXiv:2212.07489}} (\bibinfo{year}{2022}).
\newblock


\bibitem[\protect\citeauthoryear{Finn, Abbeel, and Levine}{Finn et~al\mbox{.}}{2017}]%
        {finn2017maml}
\bibfield{author}{\bibinfo{person}{Chelsea Finn}, \bibinfo{person}{Pieter Abbeel}, {and} \bibinfo{person}{Sergey Levine}.} \bibinfo{year}{2017}\natexlab{}.
\newblock \showarticletitle{Model-Agnostic Meta-Learning for Fast Adaptation of Deep Networks}. In \bibinfo{booktitle}{\emph{International Conference on Machine Learning}} \emph{(\bibinfo{series}{Proceedings of Machine Learning Research}, Vol.~\bibinfo{volume}{70})}. \bibinfo{pages}{1126--1135}.
\newblock


\bibitem[\protect\citeauthoryear{Foerster, Chen, Al-Shedivat, Whiteson, Abbeel, and Mordatch}{Foerster et~al\mbox{.}}{2018}]%
        {foerster_learning_2018}
\bibfield{author}{\bibinfo{person}{Jakob Foerster}, \bibinfo{person}{Richard~Y Chen}, \bibinfo{person}{Maruan Al-Shedivat}, \bibinfo{person}{Shimon Whiteson}, \bibinfo{person}{Pieter Abbeel}, {and} \bibinfo{person}{Igor Mordatch}.} \bibinfo{year}{2018}\natexlab{}.
\newblock \showarticletitle{Learning with Opponent-Learning Awareness}. In \bibinfo{booktitle}{\emph{Proceedings of the 17th International Conference on Autonomous Agents and MultiAgent Systems}}. \bibinfo{pages}{122--130}.
\newblock


\bibitem[\protect\citeauthoryear{Foerster, Song, Hughes, Burch, Dunning, Whiteson, Botvinick, and Bowling}{Foerster et~al\mbox{.}}{2019}]%
        {foerster2019bayesian}
\bibfield{author}{\bibinfo{person}{Jakob Foerster}, \bibinfo{person}{Francis Song}, \bibinfo{person}{Edward Hughes}, \bibinfo{person}{Neil Burch}, \bibinfo{person}{Iain Dunning}, \bibinfo{person}{Shimon Whiteson}, \bibinfo{person}{Matthew Botvinick}, {and} \bibinfo{person}{Michael Bowling}.} \bibinfo{year}{2019}\natexlab{}.
\newblock \showarticletitle{Bayesian action decoder for deep multi-agent reinforcement learning}. In \bibinfo{booktitle}{\emph{International Conference on Machine Learning}}. PMLR, \bibinfo{pages}{1942--1951}.
\newblock


\bibitem[\protect\citeauthoryear{Fung, Zhang, Lu, Willi, and Foerster}{Fung et~al\mbox{.}}{2023}]%
        {fung2023analyzing}
\bibfield{author}{\bibinfo{person}{Kitty Fung}, \bibinfo{person}{Qizhen Zhang}, \bibinfo{person}{Chris Lu}, \bibinfo{person}{Timon Willi}, {and} \bibinfo{person}{Jakob~Nicolaus Foerster}.} \bibinfo{year}{2023}\natexlab{}.
\newblock \showarticletitle{Analyzing the Sample Complexity of Model-Free Opponent Shaping}. In \bibinfo{booktitle}{\emph{ICML Workshop on New Frontiers in Learning, Control, and Dynamical Systems}}.
\newblock
\urldef\tempurl%
\url{https://openreview.net/forum?id=Dm2fbPpU6v}
\showURL{%
\tempurl}


\bibitem[\protect\citeauthoryear{Harper, Knight, Jones, Koutsovoulos, Glynatsi, and Campbell}{Harper et~al\mbox{.}}{2017}]%
        {Harper_2017}
\bibfield{author}{\bibinfo{person}{Marc Harper}, \bibinfo{person}{Vincent Knight}, \bibinfo{person}{Martin Jones}, \bibinfo{person}{Georgios Koutsovoulos}, \bibinfo{person}{Nikoleta~E. Glynatsi}, {and} \bibinfo{person}{Owen Campbell}.} \bibinfo{year}{2017}\natexlab{}.
\newblock \showarticletitle{Reinforcement learning produces dominant strategies for the Iterated Prisoner’s Dilemma}.
\newblock \bibinfo{journal}{\emph{PLOS ONE}} \bibinfo{volume}{12}, \bibinfo{number}{12} (\bibinfo{year}{2017}), \bibinfo{pages}{e0188046}.
\newblock


\bibitem[\protect\citeauthoryear{Hennigan, Cai, Norman, and Babuschkin}{Hennigan et~al\mbox{.}}{2020}]%
        {haiku2020github}
\bibfield{author}{\bibinfo{person}{Tom Hennigan}, \bibinfo{person}{Trevor Cai}, \bibinfo{person}{Tamara Norman}, {and} \bibinfo{person}{Igor Babuschkin}.} \bibinfo{year}{2020}\natexlab{}.
\newblock \bibinfo{booktitle}{\emph{{H}aiku: {S}onnet for {JAX}}}.
\newblock
\urldef\tempurl%
\url{http://github.com/deepmind/dm-haiku}
\showURL{%
\tempurl}


\bibitem[\protect\citeauthoryear{Hochreiter and Schmidhuber}{Hochreiter and Schmidhuber}{1997}]%
        {hochreiter1997lstm}
\bibfield{author}{\bibinfo{person}{Sepp Hochreiter} {and} \bibinfo{person}{J{\"{u}}rgen Schmidhuber}.} \bibinfo{year}{1997}\natexlab{}.
\newblock \showarticletitle{Long Short-Term Memory}.
\newblock \bibinfo{journal}{\emph{Neural Comput.}} \bibinfo{volume}{9}, \bibinfo{number}{8} (\bibinfo{year}{1997}), \bibinfo{pages}{1735--1780}.
\newblock
\urldef\tempurl%
\url{https://doi.org/10.1162/neco.1997.9.8.1735}
\showDOI{\tempurl}


\bibitem[\protect\citeauthoryear{Jaderberg, Czarnecki, Dunning, Marris, Lever, Castañeda, Beattie, Rabinowitz, Morcos, Ruderman, Sonnerat, Green, Deason, Leibo, Silver, Hassabis, Kavukcuoglu, and Graepel}{Jaderberg et~al\mbox{.}}{2019}]%
        {jarderberg_2019_hideandseek}
\bibfield{author}{\bibinfo{person}{Max Jaderberg}, \bibinfo{person}{Wojciech~M. Czarnecki}, \bibinfo{person}{Iain Dunning}, \bibinfo{person}{Luke Marris}, \bibinfo{person}{Guy Lever}, \bibinfo{person}{Antonio~Garcia Castañeda}, \bibinfo{person}{Charles Beattie}, \bibinfo{person}{Neil~C. Rabinowitz}, \bibinfo{person}{Ari~S. Morcos}, \bibinfo{person}{Avraham Ruderman}, \bibinfo{person}{Nicolas Sonnerat}, \bibinfo{person}{Tim Green}, \bibinfo{person}{Louise Deason}, \bibinfo{person}{Joel~Z. Leibo}, \bibinfo{person}{David Silver}, \bibinfo{person}{Demis Hassabis}, \bibinfo{person}{Koray Kavukcuoglu}, {and} \bibinfo{person}{Thore Graepel}.} \bibinfo{year}{2019}\natexlab{}.
\newblock \showarticletitle{Human-level performance in 3D multiplayer games with population-based reinforcement learning}.
\newblock \bibinfo{journal}{\emph{Science}} \bibinfo{volume}{364}, \bibinfo{number}{6443} (\bibinfo{year}{2019}), \bibinfo{pages}{859--865}.
\newblock
\urldef\tempurl%
\url{https://doi.org/10.1126/science.aau6249}
\showDOI{\tempurl}


\bibitem[\protect\citeauthoryear{Jaques, Lazaridou, Hughes, Gulcehre, Ortega, Strouse, Leibo, and De~Freitas}{Jaques et~al\mbox{.}}{2019}]%
        {jaques2019social}
\bibfield{author}{\bibinfo{person}{Natasha Jaques}, \bibinfo{person}{Angeliki Lazaridou}, \bibinfo{person}{Edward Hughes}, \bibinfo{person}{Caglar Gulcehre}, \bibinfo{person}{Pedro Ortega}, \bibinfo{person}{DJ Strouse}, \bibinfo{person}{Joel~Z Leibo}, {and} \bibinfo{person}{Nando De~Freitas}.} \bibinfo{year}{2019}\natexlab{}.
\newblock \showarticletitle{Social influence as intrinsic motivation for multi-agent deep reinforcement learning}. In \bibinfo{booktitle}{\emph{International conference on machine learning}}. PMLR, \bibinfo{pages}{3040--3049}.
\newblock


\bibitem[\protect\citeauthoryear{Kim, Liu, Riemer, Sun, Abdulhai, Habibi, Lopez{-}Cot, Tesauro, and How}{Kim et~al\mbox{.}}{2021a}]%
        {kim2021meta-mapg}
\bibfield{author}{\bibinfo{person}{Dong{-}Ki Kim}, \bibinfo{person}{Miao Liu}, \bibinfo{person}{Matthew Riemer}, \bibinfo{person}{Chuangchuang Sun}, \bibinfo{person}{Marwa Abdulhai}, \bibinfo{person}{Golnaz Habibi}, \bibinfo{person}{Sebastian Lopez{-}Cot}, \bibinfo{person}{Gerald Tesauro}, {and} \bibinfo{person}{Jonathan~P. How}.} \bibinfo{year}{2021}\natexlab{a}.
\newblock \showarticletitle{A Policy Gradient Algorithm for Learning to Learn in Multiagent Reinforcement Learning}. In \bibinfo{booktitle}{\emph{International Conference on Machine Learning}} \emph{(\bibinfo{series}{Proceedings of Machine Learning Research}, Vol.~\bibinfo{volume}{139})}. \bibinfo{pages}{5541--5550}.
\newblock


\bibitem[\protect\citeauthoryear{Kim, Liu, Riemer, Sun, Abdulhai, Habibi, Lopez-Cot, Tesauro, and How}{Kim et~al\mbox{.}}{2021b}]%
        {kim2021policy}
\bibfield{author}{\bibinfo{person}{Dong~Ki Kim}, \bibinfo{person}{Miao Liu}, \bibinfo{person}{Matthew~D Riemer}, \bibinfo{person}{Chuangchuang Sun}, \bibinfo{person}{Marwa Abdulhai}, \bibinfo{person}{Golnaz Habibi}, \bibinfo{person}{Sebastian Lopez-Cot}, \bibinfo{person}{Gerald Tesauro}, {and} \bibinfo{person}{Jonathan How}.} \bibinfo{year}{2021}\natexlab{b}.
\newblock \showarticletitle{A policy gradient algorithm for learning to learn in multiagent reinforcement learning}. In \bibinfo{booktitle}{\emph{International Conference on Machine Learning}}. PMLR, \bibinfo{pages}{5541--5550}.
\newblock


\bibitem[\protect\citeauthoryear{K{\"o}ster, McKee, Everett, Weidinger, Isaac, Hughes, Du{\'e}{\~n}ez-Guzm{\'a}n, Graepel, Botvinick, and Leibo}{K{\"o}ster et~al\mbox{.}}{2020}]%
        {koster2020model}
\bibfield{author}{\bibinfo{person}{Raphael K{\"o}ster}, \bibinfo{person}{Kevin~R McKee}, \bibinfo{person}{Richard Everett}, \bibinfo{person}{Laura Weidinger}, \bibinfo{person}{William~S Isaac}, \bibinfo{person}{Edward Hughes}, \bibinfo{person}{Edgar~A Du{\'e}{\~n}ez-Guzm{\'a}n}, \bibinfo{person}{Thore Graepel}, \bibinfo{person}{Matthew Botvinick}, {and} \bibinfo{person}{Joel~Z Leibo}.} \bibinfo{year}{2020}\natexlab{}.
\newblock \showarticletitle{Model-free conventions in multi-agent reinforcement learning with heterogeneous preferences}.
\newblock \bibinfo{journal}{\emph{arXiv preprint arXiv:2010.09054}} (\bibinfo{year}{2020}).
\newblock


\bibitem[\protect\citeauthoryear{Lange}{Lange}{2022a}]%
        {evosax2022github}
\bibfield{author}{\bibinfo{person}{Robert~Tjarko Lange}.} \bibinfo{year}{2022}\natexlab{a}.
\newblock \bibinfo{booktitle}{\emph{{evosax}: JAX-based Evolution Strategies}}.
\newblock
\urldef\tempurl%
\url{http://github.com/RobertTLange/evosax}
\showURL{%
\tempurl}


\bibitem[\protect\citeauthoryear{Lange}{Lange}{2022b}]%
        {gymnax2022github}
\bibfield{author}{\bibinfo{person}{Robert~Tjarko Lange}.} \bibinfo{year}{2022}\natexlab{b}.
\newblock \bibinfo{booktitle}{\emph{{gymnax}: A {JAX}-based Reinforcement Learning Environment Library}}.
\newblock
\urldef\tempurl%
\url{http://github.com/RobertTLange/gymnax}
\showURL{%
\tempurl}


\bibitem[\protect\citeauthoryear{Leibo, Du{\'{e}}{\~{n}}ez{-}Guzm{\'{a}}n, Vezhnevets, Agapiou, Sunehag, Koster, Matyas, Beattie, Mordatch, and Graepel}{Leibo et~al\mbox{.}}{2021}]%
        {leibo2021meltingpot}
\bibfield{author}{\bibinfo{person}{Joel~Z. Leibo}, \bibinfo{person}{Edgar~A. Du{\'{e}}{\~{n}}ez{-}Guzm{\'{a}}n}, \bibinfo{person}{Alexander Vezhnevets}, \bibinfo{person}{John~P. Agapiou}, \bibinfo{person}{Peter Sunehag}, \bibinfo{person}{Raphael Koster}, \bibinfo{person}{Jayd Matyas}, \bibinfo{person}{Charlie Beattie}, \bibinfo{person}{Igor Mordatch}, {and} \bibinfo{person}{Thore Graepel}.} \bibinfo{year}{2021}\natexlab{}.
\newblock \showarticletitle{Scalable Evaluation of Multi-Agent Reinforcement Learning with Melting Pot}. In \bibinfo{booktitle}{\emph{Proceedings of the 38th International Conference on Machine Learning}} \emph{(\bibinfo{series}{Proceedings of Machine Learning Research}, Vol.~\bibinfo{volume}{139})}. \bibinfo{publisher}{{PMLR}}, \bibinfo{pages}{6187--6199}.
\newblock


\bibitem[\protect\citeauthoryear{Lerer and Peysakhovich}{Lerer and Peysakhovich}{2017}]%
        {lerer2017coingame}
\bibfield{author}{\bibinfo{person}{Adam Lerer} {and} \bibinfo{person}{Alexander Peysakhovich}.} \bibinfo{year}{2017}\natexlab{}.
\newblock \showarticletitle{Maintaining cooperation in complex social dilemmas using deep reinforcement learning}.
\newblock \bibinfo{journal}{\emph{CoRR}}  \bibinfo{volume}{abs/1707.01068} (\bibinfo{year}{2017}).
\newblock


\bibitem[\protect\citeauthoryear{Letcher, Balduzzi, Racani{\`{e}}re, Martens, Foerster, Tuyls, and Graepel}{Letcher et~al\mbox{.}}{2019a}]%
        {letcher_differentiable_2019}
\bibfield{author}{\bibinfo{person}{Alistair Letcher}, \bibinfo{person}{David Balduzzi}, \bibinfo{person}{S{\'{e}}bastien Racani{\`{e}}re}, \bibinfo{person}{James Martens}, \bibinfo{person}{Jakob~N. Foerster}, \bibinfo{person}{Karl Tuyls}, {and} \bibinfo{person}{Thore Graepel}.} \bibinfo{year}{2019}\natexlab{a}.
\newblock \showarticletitle{Differentiable Game Mechanics}.
\newblock \bibinfo{journal}{\emph{J. Mach. Learn. Res.}}  \bibinfo{volume}{20} (\bibinfo{year}{2019}), \bibinfo{pages}{84:1--84:40}.
\newblock


\bibitem[\protect\citeauthoryear{Letcher, Foerster, Balduzzi, Rockt{\"{a}}schel, and Whiteson}{Letcher et~al\mbox{.}}{2019b}]%
        {letcher_stable_2019}
\bibfield{author}{\bibinfo{person}{Alistair Letcher}, \bibinfo{person}{Jakob~N. Foerster}, \bibinfo{person}{David Balduzzi}, \bibinfo{person}{Tim Rockt{\"{a}}schel}, {and} \bibinfo{person}{Shimon Whiteson}.} \bibinfo{year}{2019}\natexlab{b}.
\newblock \showarticletitle{Stable Opponent Shaping in Differentiable Games}. In \bibinfo{booktitle}{\emph{7th International Conference on Learning Representations}}.
\newblock


\bibitem[\protect\citeauthoryear{Lu, Kuba, Letcher, Metz, de~Witt, and Foerster}{Lu et~al\mbox{.}}{2022a}]%
        {lu2022dpo}
\bibfield{author}{\bibinfo{person}{Chris Lu}, \bibinfo{person}{Jakub~Grudzien Kuba}, \bibinfo{person}{Alistair Letcher}, \bibinfo{person}{Luke Metz}, \bibinfo{person}{Christian~Schr{\"{o}}der de Witt}, {and} \bibinfo{person}{Jakob~N. Foerster}.} \bibinfo{year}{2022}\natexlab{a}.
\newblock \showarticletitle{Discovered Policy Optimisation}.
\newblock \bibinfo{journal}{\emph{CoRR}}  \bibinfo{volume}{abs/2210.05639} (\bibinfo{year}{2022}).
\newblock
\urldef\tempurl%
\url{https://doi.org/10.48550/arXiv.2210.05639}
\showDOI{\tempurl}
\showeprint[arXiv]{2210.05639}


\bibitem[\protect\citeauthoryear{Lu, Willi, De~Witt, and Foerster}{Lu et~al\mbox{.}}{2022b}]%
        {lu2022model}
\bibfield{author}{\bibinfo{person}{Christopher Lu}, \bibinfo{person}{Timon Willi}, \bibinfo{person}{Christian A~Schroeder De~Witt}, {and} \bibinfo{person}{Jakob Foerster}.} \bibinfo{year}{2022}\natexlab{b}.
\newblock \showarticletitle{Model-Free Opponent Shaping}. In \bibinfo{booktitle}{\emph{International Conference on Machine Learning}}. PMLR, \bibinfo{pages}{14398--14411}.
\newblock


\bibitem[\protect\citeauthoryear{Lu, Willi, Letcher, and Foerster}{Lu et~al\mbox{.}}{2022c}]%
        {lu2022adversarial}
\bibfield{author}{\bibinfo{person}{Chris Lu}, \bibinfo{person}{Timon Willi}, \bibinfo{person}{Alistair Letcher}, {and} \bibinfo{person}{Jakob~Nicolaus Foerster}.} \bibinfo{year}{2022}\natexlab{c}.
\newblock \showarticletitle{Adversarial Cheap Talk}. In \bibinfo{booktitle}{\emph{Decision Awareness in Reinforcement Learning Workshop at ICML 2022}}.
\newblock


\bibitem[\protect\citeauthoryear{Lupu and Precup}{Lupu and Precup}{2020}]%
        {lupu2020gifting}
\bibfield{author}{\bibinfo{person}{Andrei Lupu} {and} \bibinfo{person}{Doina Precup}.} \bibinfo{year}{2020}\natexlab{}.
\newblock \showarticletitle{Gifting in Multi-Agent Reinforcement Learning}. In \bibinfo{booktitle}{\emph{Proceedings of the 19th International Conference on Autonomous Agents and MultiAgent Systems}} (Auckland, New Zealand) \emph{(\bibinfo{series}{AAMAS '20})}. \bibinfo{publisher}{International Foundation for Autonomous Agents and Multiagent Systems}, \bibinfo{address}{Richland, SC}, \bibinfo{pages}{789–797}.
\newblock
\showISBNx{9781450375184}


\bibitem[\protect\citeauthoryear{Press and Dyson}{Press and Dyson}{2012}]%
        {press_iterated_2012}
\bibfield{author}{\bibinfo{person}{William~H. Press} {and} \bibinfo{person}{Freeman~J. Dyson}.} \bibinfo{year}{2012}\natexlab{}.
\newblock \showarticletitle{Iterated {Prisoner}’s {Dilemma} contains strategies that dominate any evolutionary opponent}.
\newblock \bibinfo{journal}{\emph{Proceedings of the National Academy of Sciences}} \bibinfo{volume}{109}, \bibinfo{number}{26} (\bibinfo{year}{2012}), \bibinfo{pages}{10409--10413}.
\newblock
\showISSN{0027-8424}


\bibitem[\protect\citeauthoryear{Rashid, Samvelyan, Schroeder, Farquhar, Foerster, and Whiteson}{Rashid et~al\mbox{.}}{2018}]%
        {rashid2018qmix}
\bibfield{author}{\bibinfo{person}{Tabish Rashid}, \bibinfo{person}{Mikayel Samvelyan}, \bibinfo{person}{Christian Schroeder}, \bibinfo{person}{Gregory Farquhar}, \bibinfo{person}{Jakob Foerster}, {and} \bibinfo{person}{Shimon Whiteson}.} \bibinfo{year}{2018}\natexlab{}.
\newblock \showarticletitle{Qmix: Monotonic value function factorisation for deep multi-agent reinforcement learning}. In \bibinfo{booktitle}{\emph{International conference on machine learning}}. PMLR, \bibinfo{pages}{4295--4304}.
\newblock


\bibitem[\protect\citeauthoryear{Salimans, Ho, Chen, and Sutskever}{Salimans et~al\mbox{.}}{2017}]%
        {salimans2017evolutionstrategies}
\bibfield{author}{\bibinfo{person}{Tim Salimans}, \bibinfo{person}{Jonathan Ho}, \bibinfo{person}{Xi Chen}, {and} \bibinfo{person}{Ilya Sutskever}.} \bibinfo{year}{2017}\natexlab{}.
\newblock \bibinfo{title}{Evolution Strategies as a Scalable Alternative to Reinforcement Learning}.
\newblock \bibinfo{howpublished}{arXiv preprint arXiv:1703.03864}.
\newblock


\bibitem[\protect\citeauthoryear{Samvelyan, Rashid, de~Witt, Farquhar, Nardelli, Rudner, Hung, Torr, Foerster, and Whiteson}{Samvelyan et~al\mbox{.}}{2019}]%
        {samvelyan19smac}
\bibfield{author}{\bibinfo{person}{Mikayel Samvelyan}, \bibinfo{person}{Tabish Rashid}, \bibinfo{person}{Christian~Schroeder de Witt}, \bibinfo{person}{Gregory Farquhar}, \bibinfo{person}{Nantas Nardelli}, \bibinfo{person}{Tim G.~J. Rudner}, \bibinfo{person}{Chia-Man Hung}, \bibinfo{person}{Philiph H.~S. Torr}, \bibinfo{person}{Jakob Foerster}, {and} \bibinfo{person}{Shimon Whiteson}.} \bibinfo{year}{2019}\natexlab{}.
\newblock \showarticletitle{{The} {StarCraft} {Multi}-{Agent} {Challenge}}.
\newblock \bibinfo{journal}{\emph{CoRR}}  \bibinfo{volume}{abs/1902.04043} (\bibinfo{year}{2019}).
\newblock


\bibitem[\protect\citeauthoryear{Schulman, Wolski, Dhariwal, Radford, and Klimov}{Schulman et~al\mbox{.}}{2017}]%
        {schulman2017ppo}
\bibfield{author}{\bibinfo{person}{John Schulman}, \bibinfo{person}{Filip Wolski}, \bibinfo{person}{Prafulla Dhariwal}, \bibinfo{person}{Alec Radford}, {and} \bibinfo{person}{Oleg Klimov}.} \bibinfo{year}{2017}\natexlab{}.
\newblock \bibinfo{title}{Proximal Policy Optimization Algorithms}.
\newblock \bibinfo{howpublished}{arXiv preprint arXiv:1707.06347}.
\newblock


\bibitem[\protect\citeauthoryear{Shapley}{Shapley}{1953}]%
        {shapley_stochastic_1953}
\bibfield{author}{\bibinfo{person}{L.~S. Shapley}.} \bibinfo{year}{1953}\natexlab{}.
\newblock \showarticletitle{Stochastic {Games}}.
\newblock \bibinfo{journal}{\emph{Proceedings of the National Academy of Sciences}} \bibinfo{volume}{39}, \bibinfo{number}{10} (\bibinfo{year}{1953}), \bibinfo{pages}{1095--1100}.
\newblock
\showISSN{0027-8424}


\bibitem[\protect\citeauthoryear{Silver, Huang, Maddison, Guez, Sifre, van~den Driessche, Schrittwieser, Antonoglou, Panneershelvam, Lanctot, Dieleman, Grewe, Nham, Kalchbrenner, Sutskever, Lillicrap, Leach, Kavukcuoglu, Graepel, and Hassabis}{Silver et~al\mbox{.}}{2016}]%
        {silver2016mastering}
\bibfield{author}{\bibinfo{person}{David Silver}, \bibinfo{person}{Aja Huang}, \bibinfo{person}{Chris~J. Maddison}, \bibinfo{person}{Arthur Guez}, \bibinfo{person}{Laurent Sifre}, \bibinfo{person}{George van~den Driessche}, \bibinfo{person}{Julian Schrittwieser}, \bibinfo{person}{Ioannis Antonoglou}, \bibinfo{person}{Vedavyas Panneershelvam}, \bibinfo{person}{Marc Lanctot}, \bibinfo{person}{Sander Dieleman}, \bibinfo{person}{Dominik Grewe}, \bibinfo{person}{John Nham}, \bibinfo{person}{Nal Kalchbrenner}, \bibinfo{person}{Ilya Sutskever}, \bibinfo{person}{Timothy~P. Lillicrap}, \bibinfo{person}{Madeleine Leach}, \bibinfo{person}{Koray Kavukcuoglu}, \bibinfo{person}{Thore Graepel}, {and} \bibinfo{person}{Demis Hassabis}.} \bibinfo{year}{2016}\natexlab{}.
\newblock \showarticletitle{Mastering the game of Go with deep neural networks and tree search}.
\newblock \bibinfo{journal}{\emph{Nat.}} \bibinfo{volume}{529}, \bibinfo{number}{7587} (\bibinfo{year}{2016}), \bibinfo{pages}{484--489}.
\newblock


\bibitem[\protect\citeauthoryear{Snyder}{Snyder}{1971}]%
        {snyder1971prisoner}
\bibfield{author}{\bibinfo{person}{Glenn~H Snyder}.} \bibinfo{year}{1971}\natexlab{}.
\newblock \showarticletitle{" Prisoner's Dilemma" and" Chicken" Models in International Politics}.
\newblock \bibinfo{journal}{\emph{International Studies Quarterly}} \bibinfo{volume}{15}, \bibinfo{number}{1} (\bibinfo{year}{1971}), \bibinfo{pages}{66--103}.
\newblock


\bibitem[\protect\citeauthoryear{Vezhnevets, Wu, Eckstein, Leblond, and Leibo}{Vezhnevets et~al\mbox{.}}{2020}]%
        {vezhnevets2020options}
\bibfield{author}{\bibinfo{person}{Alexander Vezhnevets}, \bibinfo{person}{Yuhuai Wu}, \bibinfo{person}{Maria Eckstein}, \bibinfo{person}{R{\'e}mi Leblond}, {and} \bibinfo{person}{Joel~Z Leibo}.} \bibinfo{year}{2020}\natexlab{}.
\newblock \showarticletitle{Options as responses: Grounding behavioural hierarchies in multi-agent reinforcement learning}. In \bibinfo{booktitle}{\emph{International Conference on Machine Learning}}. PMLR.
\newblock


\bibitem[\protect\citeauthoryear{Vinyals, Babuschkin, Czarnecki, Mathieu, Dudzik, Chung, Choi, Powell, Ewalds, Georgiev, Oh, Horgan, Kroiss, Danihelka, Huang, Sifre, Cai, Agapiou, Jaderberg, Vezhnevets, Leblond, Pohlen, Dalibard, Budden, Sulsky, Molloy, Paine, G{\"{u}}l{\c{c}}ehre, Wang, Pfaff, Wu, Ring, Yogatama, W{\"{u}}nsch, McKinney, Smith, Schaul, Lillicrap, Kavukcuoglu, Hassabis, Apps, and Silver}{Vinyals et~al\mbox{.}}{2019}]%
        {vinyals2019alphastar}
\bibfield{author}{\bibinfo{person}{Oriol Vinyals}, \bibinfo{person}{Igor Babuschkin}, \bibinfo{person}{Wojciech~M. Czarnecki}, \bibinfo{person}{Micha{\"{e}}l Mathieu}, \bibinfo{person}{Andrew Dudzik}, \bibinfo{person}{Junyoung Chung}, \bibinfo{person}{David~H. Choi}, \bibinfo{person}{Richard Powell}, \bibinfo{person}{Timo Ewalds}, \bibinfo{person}{Petko Georgiev}, \bibinfo{person}{Junhyuk Oh}, \bibinfo{person}{Dan Horgan}, \bibinfo{person}{Manuel Kroiss}, \bibinfo{person}{Ivo Danihelka}, \bibinfo{person}{Aja Huang}, \bibinfo{person}{Laurent Sifre}, \bibinfo{person}{Trevor Cai}, \bibinfo{person}{John~P. Agapiou}, \bibinfo{person}{Max Jaderberg}, \bibinfo{person}{Alexander~Sasha Vezhnevets}, \bibinfo{person}{R{\'{e}}mi Leblond}, \bibinfo{person}{Tobias Pohlen}, \bibinfo{person}{Valentin Dalibard}, \bibinfo{person}{David Budden}, \bibinfo{person}{Yury Sulsky}, \bibinfo{person}{James Molloy}, \bibinfo{person}{Tom~Le Paine}, \bibinfo{person}{{\c{C}}aglar G{\"{u}}l{\c{c}}ehre}, \bibinfo{person}{Ziyu Wang},
  \bibinfo{person}{Tobias Pfaff}, \bibinfo{person}{Yuhuai Wu}, \bibinfo{person}{Roman Ring}, \bibinfo{person}{Dani Yogatama}, \bibinfo{person}{Dario W{\"{u}}nsch}, \bibinfo{person}{Katrina McKinney}, \bibinfo{person}{Oliver Smith}, \bibinfo{person}{Tom Schaul}, \bibinfo{person}{Timothy~P. Lillicrap}, \bibinfo{person}{Koray Kavukcuoglu}, \bibinfo{person}{Demis Hassabis}, \bibinfo{person}{Chris Apps}, {and} \bibinfo{person}{David Silver}.} \bibinfo{year}{2019}\natexlab{}.
\newblock \showarticletitle{Grandmaster level in StarCraft {II} using multi-agent reinforcement learning}.
\newblock \bibinfo{journal}{\emph{Nat.}} \bibinfo{volume}{575}, \bibinfo{number}{7782} (\bibinfo{year}{2019}), \bibinfo{pages}{350--354}.
\newblock


\bibitem[\protect\citeauthoryear{Willi, Khan, Kwan, Samvelyan, Lu, and Foerster}{Willi et~al\mbox{.}}{2023}]%
        {willi2023pax}
\bibfield{author}{\bibinfo{person}{Timon Willi}, \bibinfo{person}{Akbir Khan}, \bibinfo{person}{Newton Kwan}, \bibinfo{person}{Mikayel Samvelyan}, \bibinfo{person}{Chris Lu}, {and} \bibinfo{person}{Jakob Foerster}.} \bibinfo{year}{2023}\natexlab{}.
\newblock \bibinfo{title}{Pax: Multi-Agent Learning in JAX}.
\newblock \bibinfo{howpublished}{\url{https://github.com/ucl-dark/pax}}.
\newblock


\bibitem[\protect\citeauthoryear{Willi, Letcher, Treutlein, and Foerster}{Willi et~al\mbox{.}}{2022}]%
        {willi2022cola}
\bibfield{author}{\bibinfo{person}{Timon Willi}, \bibinfo{person}{Alistair Letcher}, \bibinfo{person}{Johannes Treutlein}, {and} \bibinfo{person}{Jakob~N. Foerster}.} \bibinfo{year}{2022}\natexlab{}.
\newblock \showarticletitle{{COLA:} Consistent Learning with Opponent-Learning Awareness}. In \bibinfo{booktitle}{\emph{International Conference on Machine Learning}} \emph{(\bibinfo{series}{Proceedings of Machine Learning Research}, Vol.~\bibinfo{volume}{162})}. \bibinfo{pages}{23804--23831}.
\newblock


\bibitem[\protect\citeauthoryear{Wu, Li, Zhao, Xu, Zhang, Fu, An, and Xing}{Wu et~al\mbox{.}}{2021}]%
        {wu2021l2e}
\bibfield{author}{\bibinfo{person}{Zhe Wu}, \bibinfo{person}{Kai Li}, \bibinfo{person}{Enmin Zhao}, \bibinfo{person}{Hang Xu}, \bibinfo{person}{Meng Zhang}, \bibinfo{person}{Haobo Fu}, \bibinfo{person}{Bo An}, {and} \bibinfo{person}{Junliang Xing}.} \bibinfo{year}{2021}\natexlab{}.
\newblock \bibinfo{title}{{L2E:} Learning to Exploit Your Opponent}.
\newblock \bibinfo{howpublished}{arXiv preprint arXiv:2102.09381}.
\newblock


\bibitem[\protect\citeauthoryear{Yuan, Guo, Zhao, and Jiang}{Yuan et~al\mbox{.}}{2022}]%
        {yuan2022adherence}
\bibfield{author}{\bibinfo{person}{Yuyu Yuan}, \bibinfo{person}{Ting Guo}, \bibinfo{person}{Pengqian Zhao}, {and} \bibinfo{person}{Hongpu Jiang}.} \bibinfo{year}{2022}\natexlab{}.
\newblock \showarticletitle{Adherence Improves Cooperation in Sequential Social Dilemmas}.
\newblock \bibinfo{journal}{\emph{Applied Sciences}} \bibinfo{volume}{12}, \bibinfo{number}{16} (\bibinfo{year}{2022}), \bibinfo{pages}{8004}.
\newblock


\bibitem[\protect\citeauthoryear{Zhao, Lu, Grosse, and Foerster}{Zhao et~al\mbox{.}}{2022}]%
        {zhao2022proximal}
\bibfield{author}{\bibinfo{person}{Stephen Zhao}, \bibinfo{person}{Chris Lu}, \bibinfo{person}{Roger~Baker Grosse}, {and} \bibinfo{person}{Jakob~Nicolaus Foerster}.} \bibinfo{year}{2022}\natexlab{}.
\newblock \showarticletitle{Proximal Learning With Opponent-Learning Awareness}.
\newblock \bibinfo{journal}{\emph{arXiv preprint arXiv:2210.10125}} (\bibinfo{year}{2022}).
\newblock


\end{thebibliography}


\newpage
\appendix
\onecolumn

\newpage

\section{\textsc{Shaper} details}
Below we list the \textsc{SHAPER} algorithm for both batching and un-batched version.

\begin{minipage}{\linewidth}
\begin{algorithm}[H]
\caption{\method{} Update: Given a POSG $\mathcal{M}$, policies $\pi_{\phi_{i}}, \pi_{\phi_{-i}}$ and their respective initial hidden states $h_{i}, h_{-i}$ and a distribution of initial co-players $\rho_{\phi}$, this algorithm updates a meta-agent policy $\phi_{i}$ over $T$ trials consisting of $E$ episodes.}\label{alg:cap}. 
\begin{algorithmic}[1]
\REQUIRE $\mathcal{M},\phi_{i}, \rho_{\phi}, E, T, f$
        \FOR{$t=0$ \textbf{to} $T$}
            \STATE Initialise trial reward $\bar{J}=0$
            \STATE Initialise meta-agent hidden states $h_i=\textbf{0}$
            \STATE Sample co-players $\phi_{-i}\sim \rho_{\phi}$
            \FOR{$e=0$ \textbf{to} $E$}
            \STATE Initialise co-players' $h_{-i}=\textbf{0}$
            \STATE $J_{i}, J_{-i}, h'_{i}, h'_{-i}$ = $\mathcal{M}(\phi_{i}, \phi_{-i}, h_{i}, h_{-i})$
            \STATE Update $\phi_{-i}$ according to co-players' update rule.
            \STATE  $h_{i} \leftarrow{} f(h'_{i})$    
            \STATE  $\bar{J} \leftarrow \bar{J} + J_{i}$ 
            \ENDFOR{} \\
            \STATE{Update $\phi_{i}$ with respect to $\bar{J}$}
            \ENDFOR{} 
\end{algorithmic}
\end{algorithm}
\end{minipage}

Here we also provide a diagram of the batching method.

\begin{figure}[h!]
    \centering
    \includegraphics[width=\textwidth]{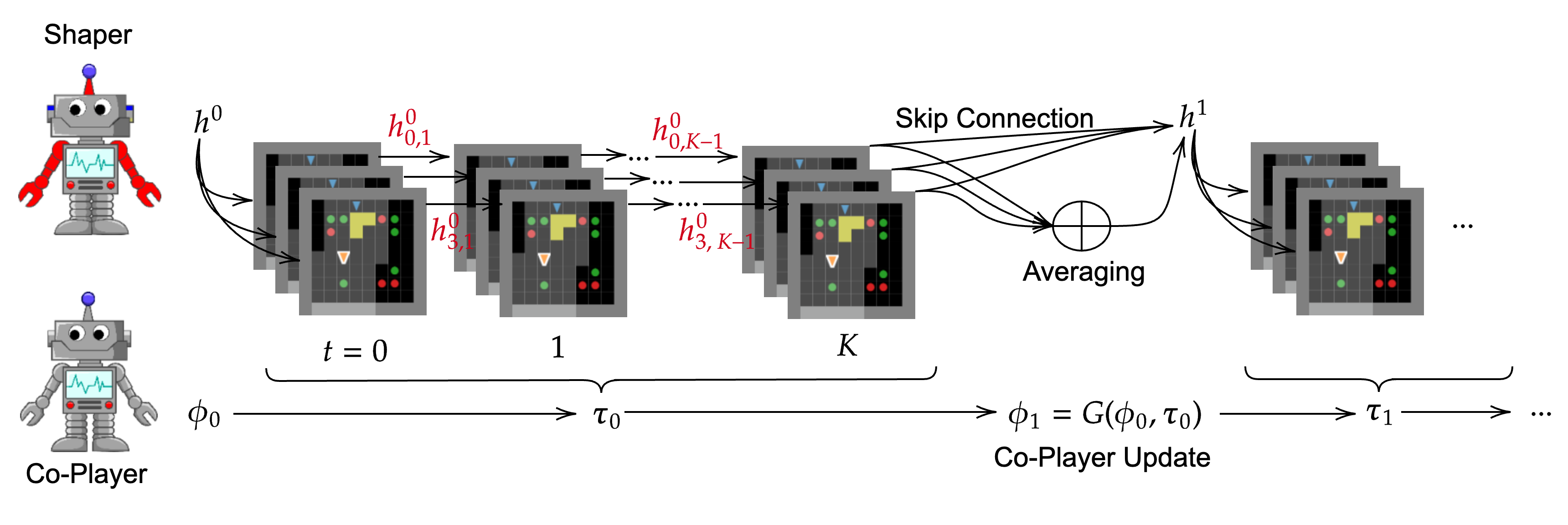}
    \caption{Typically, co-players are trained over vectorized environments, depicted as superimposed in-game frames. The co-players update their parameters after an episode of K steps. The co-players parameter update depends on the trajectories from the whole batch of vectorized environments. However, the shaping method deploys one hidden state per environment. Without averaging, the respective hidden states miss context information, which could be important to ensure proper shaping.}
    \label{fig:shaper_batching}
\end{figure}

\pagebreak
\newpage

\section{Matrix Game Details}
Here we present details of training of shaping agents in Iterated Matrix Games.
\subsection{Payoff Matrices}
\begin{table}[h]
    \centering
    \caption{Payoff Matrices}
    \begin{subtable}[h]{0.45\textwidth}
    \centering
    \begin{tabular}{ c | c | c } 
         \hline
         & C & D \\ 
         \hline
         C & (-1,-1) & (0, -3)  \\ 
         \hline
         D & (-3, 0) & (-2, -2)  \\
         \hline
    \end{tabular}
    \caption{Iterated Prisoners Dilemma (IPD)}
    \label{tab:payoff_ipd}
    \end{subtable}
    \begin{subtable}[h]{0.45\textwidth}
    \centering
    \begin{tabular}{ c | c | c } 
         \hline
         & H & T \\ 
         \hline
         H & (1,-1) & (-1, 1)  \\ 
         \hline
         T & (-1, 1) & (1, 1)  \\
         \hline
    \end{tabular}
    \caption{Iterated Matching Pennies (IMP)}
    \label{tab:payoff_imp}
    \end{subtable}
\end{table}

\subsection{Training Details}
We present training curves for both IPD and IMP below.
\begin{figure}[H]
    \begin{subfigure}[b]{0.32\textwidth}
      \centering
      \includegraphics[width=\textwidth]{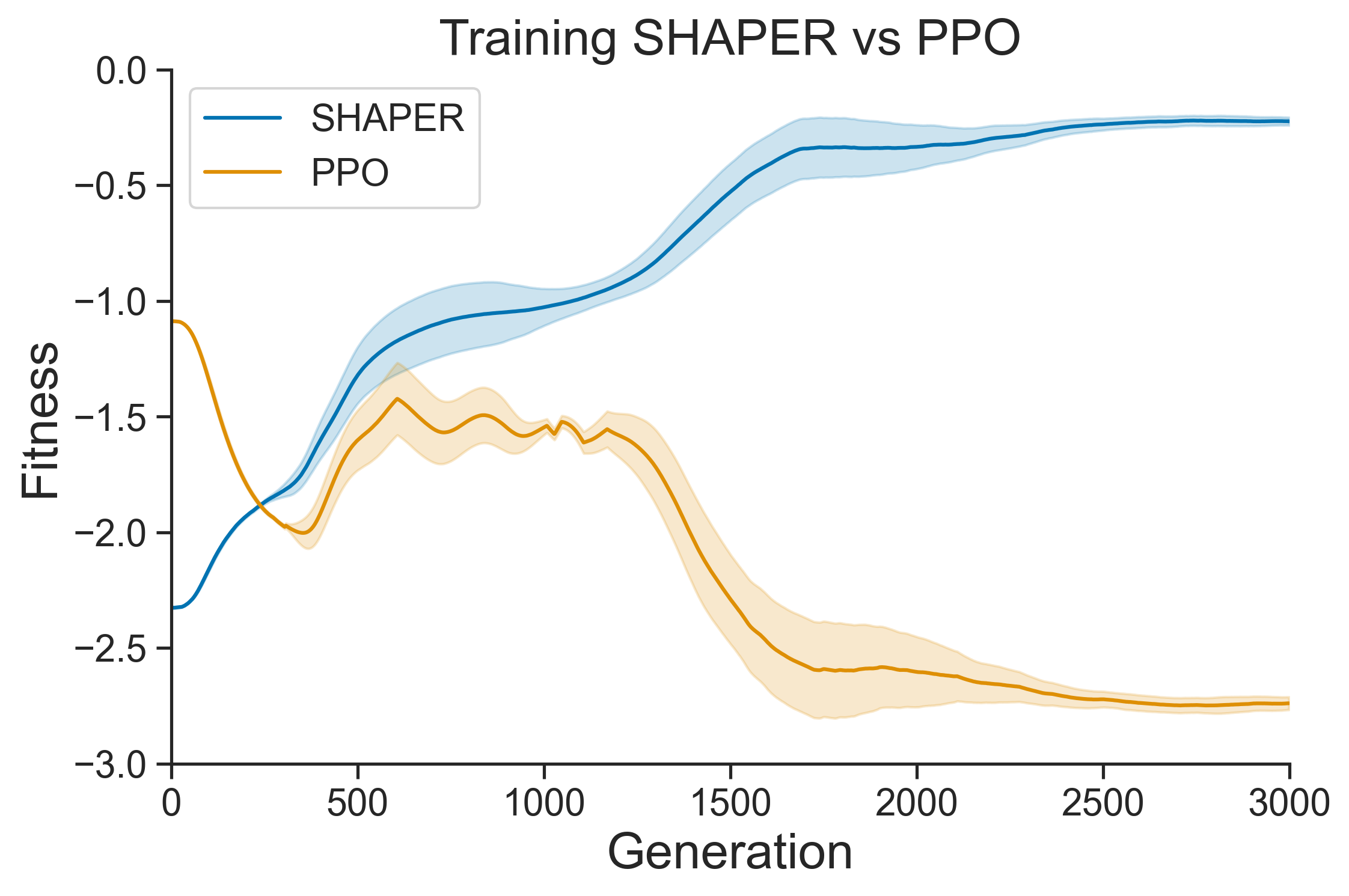}
      \caption{}
      \label{fig:earl_ipd_train_reward_appendix}
    \end{subfigure}
    \begin{subfigure}[b]{0.32\textwidth}
      \centering
      \includegraphics[width=\textwidth]{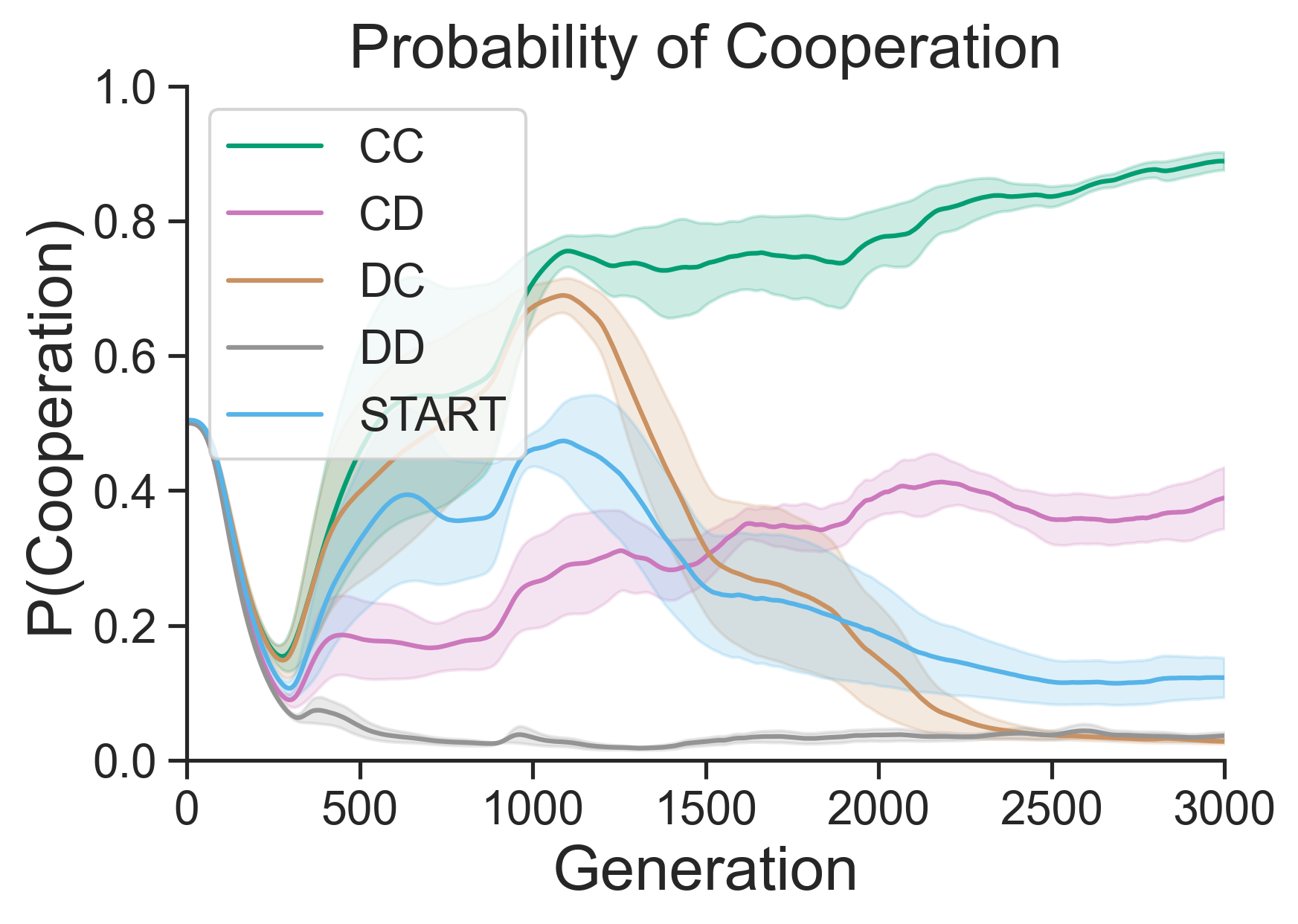}
      \caption{}
      \label{fig:earl_ipd_train_prob_appendix}
    \end{subfigure}
    \begin{subfigure}[b]{0.32\textwidth}
      \centering
      \includegraphics[width=\textwidth]{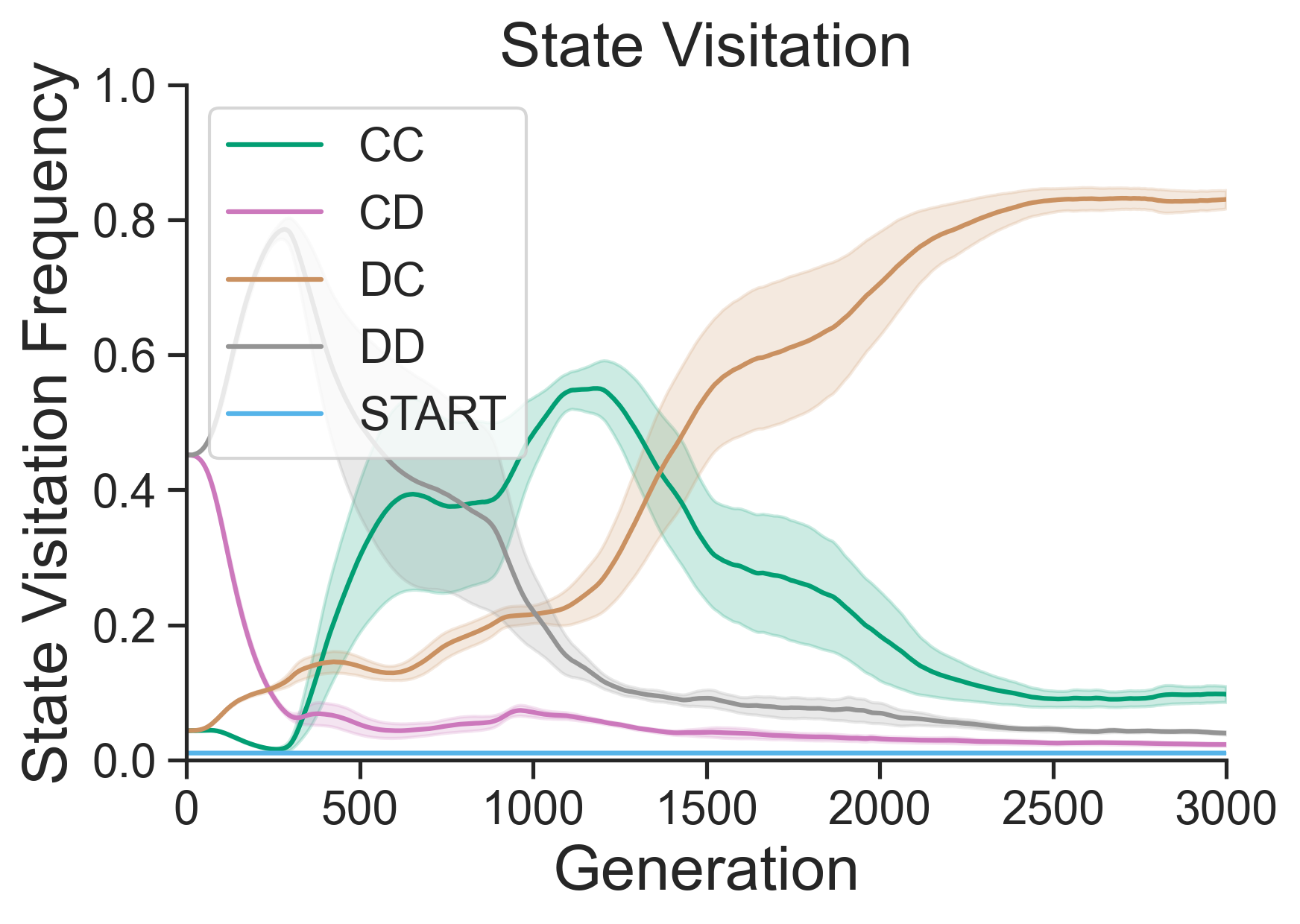}
      \caption{}
      \label{fig:earl_ipd_train_state_appendix}
    \end{subfigure}
    
    \centering
    \begin{subfigure}[b]{0.32\textwidth}
      \centering
      \includegraphics[width=\textwidth]{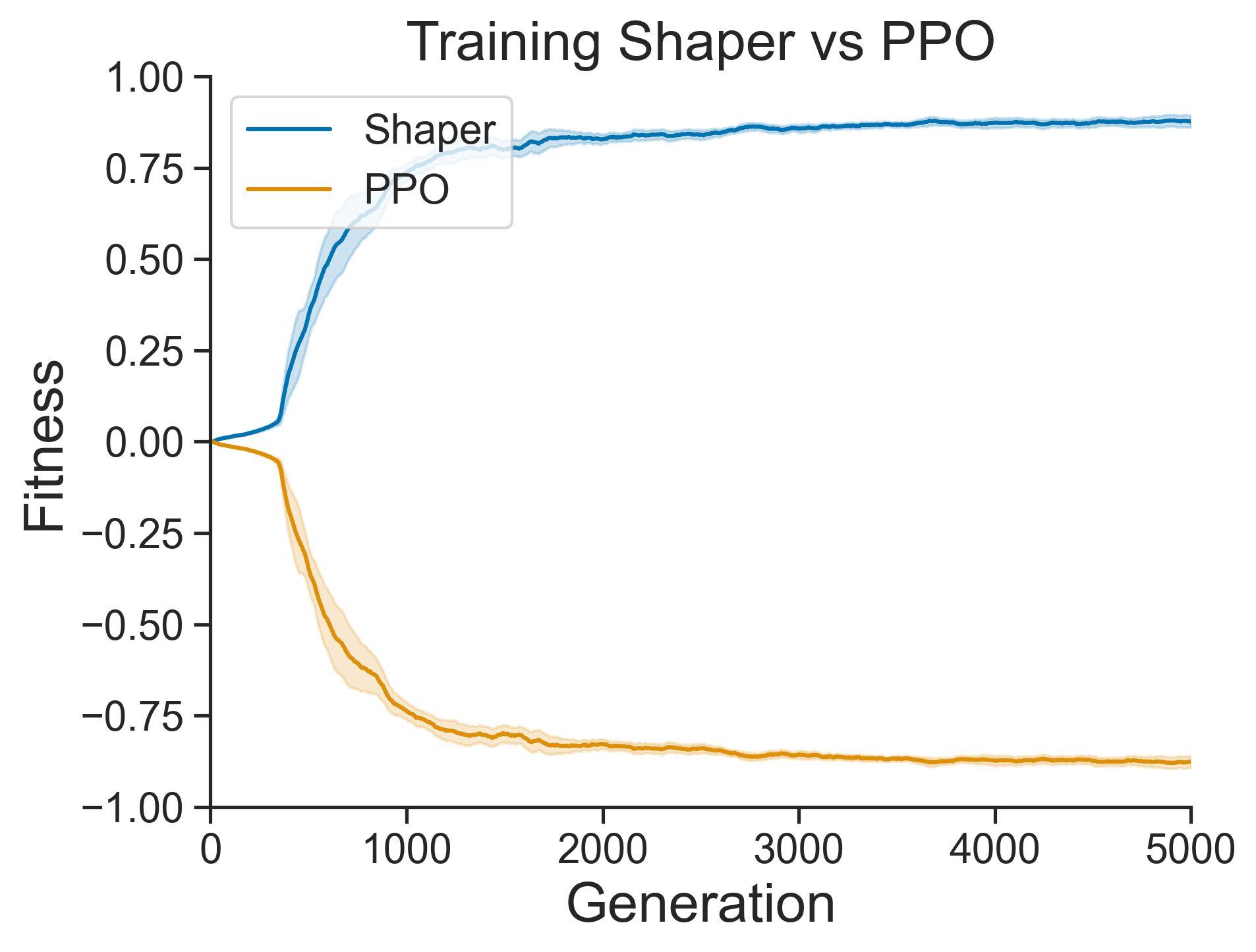}
      \caption{}
      \label{fig:earl_imp_train_reward}
    \end{subfigure}
    \begin{subfigure}[b]{0.32\textwidth}
      \centering
      \includegraphics[width=\textwidth]{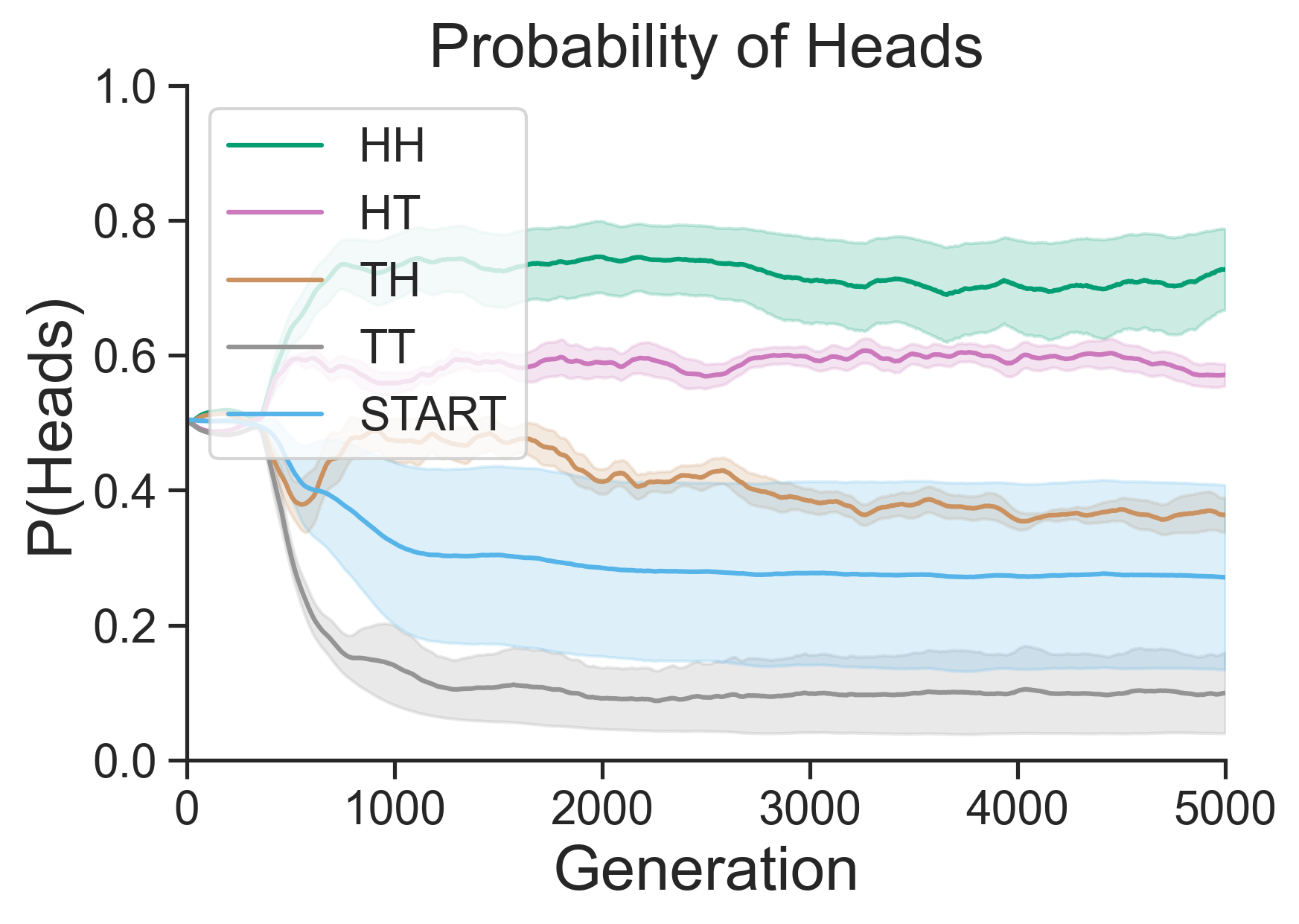}
      \caption{}
      \label{fig:earl_imp_train_prob}
    \end{subfigure}
    \begin{subfigure}[b]{0.32\textwidth}
      \centering
      \includegraphics[width=\textwidth]{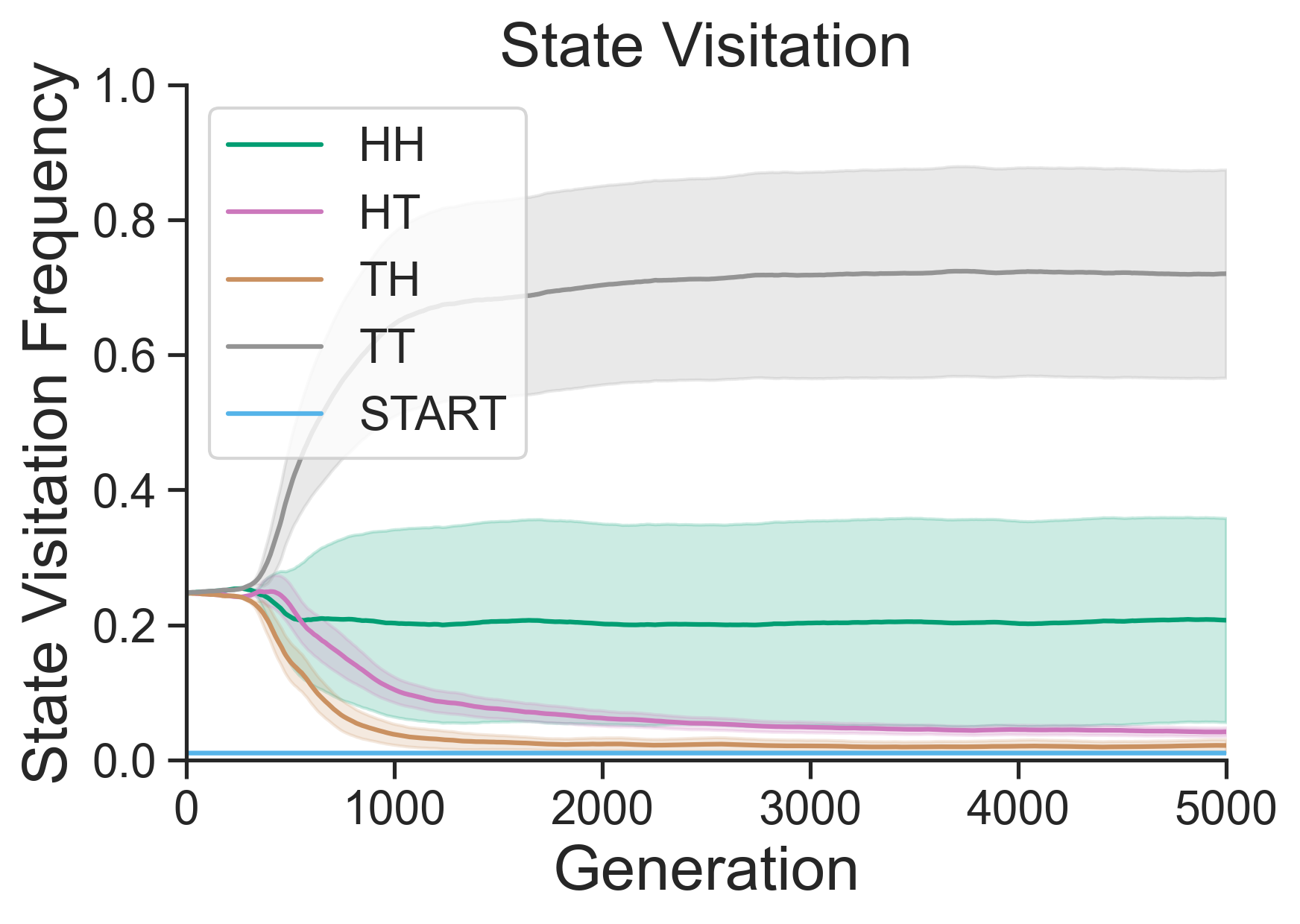}
      \caption{}
      \label{fig:earl_imp_train_state}
    \end{subfigure}
    
    \caption{Training results in the finite IPD over 5 seeds for \method{}. Here we display the (a) fitness, (b) conditional probability of cooperation, and (c) state visitation.
    Training results in the IMP over 5 seeds for \method{}. (d) Fitness (e) Empirical probability of Cooperative action conditioned by state and (f) state visitation.} 
\end{figure}

\subsection{Evaluation}
\begin{figure*}[ht]
    \centering
    \begin{subfigure}[b]{0.23\textwidth}
      \centering
      \includegraphics[width=\textwidth]{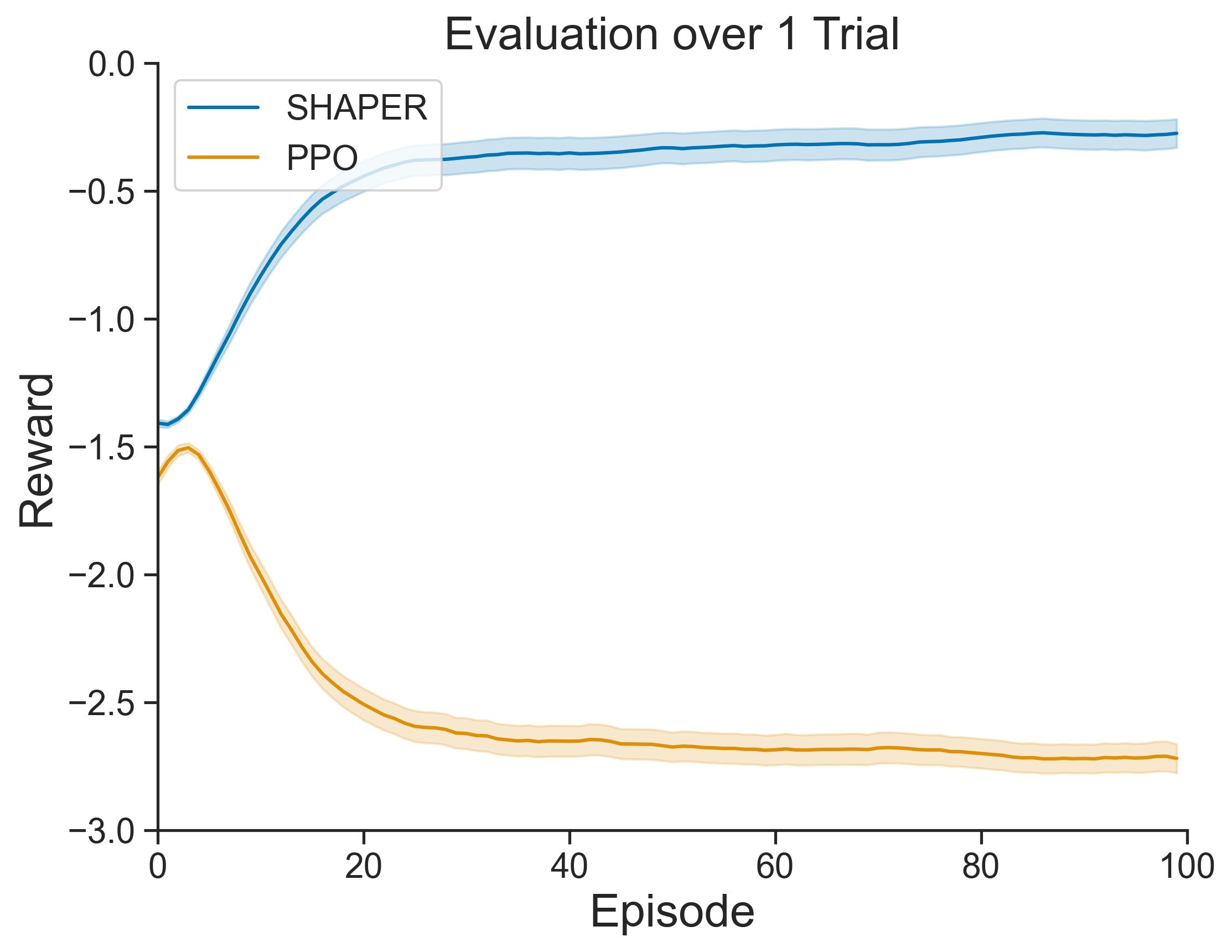}
      \caption{}
      \label{fig:earl_ipd_eval_reward}
    \end{subfigure}%
    \hfill
    \begin{subfigure}[b]{0.23\textwidth}
      \centering
      \includegraphics[width=\textwidth]{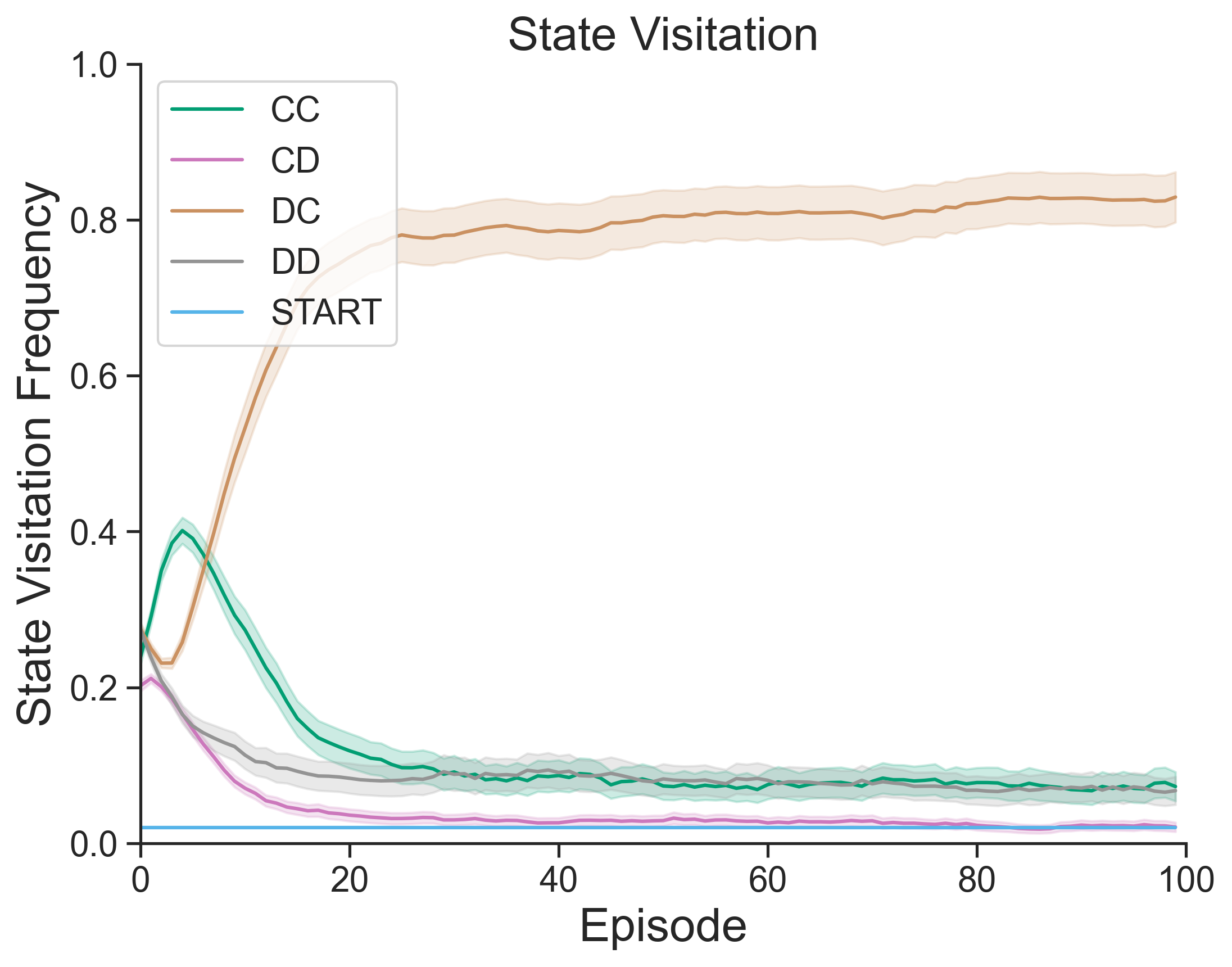}
      \caption{}
      \label{fig:earl_ipd_eval_state}
    \end{subfigure}%
    \hfill
    \begin{subfigure}[b]{0.23\textwidth}
      \centering
      \includegraphics[width=\textwidth]{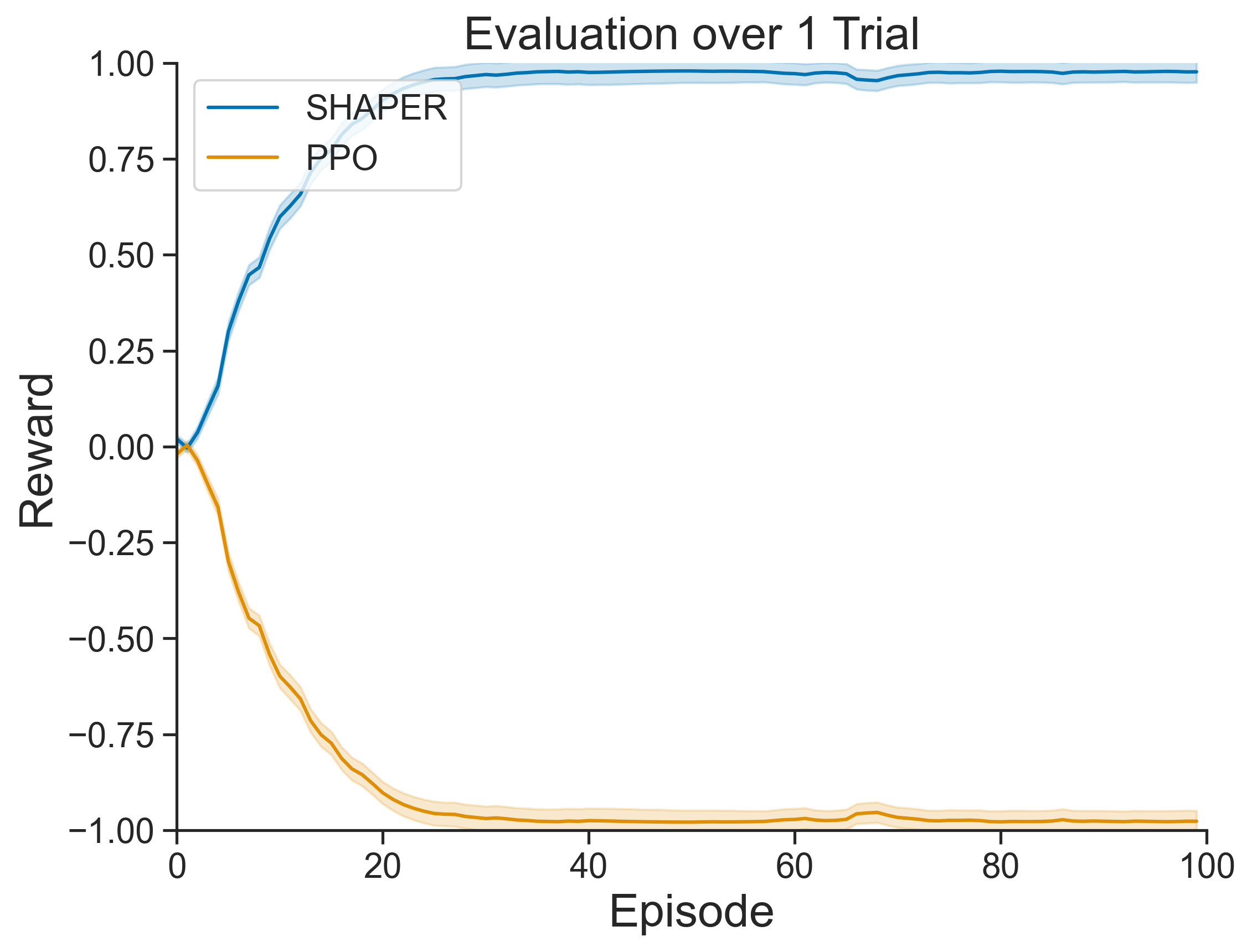}
      \caption{}
      \label{fig:earl_imp_eval_reward}
    \end{subfigure}%
    \hfill
    \begin{subfigure}[b]{0.23\textwidth}
      \centering
      \includegraphics[width=\textwidth]{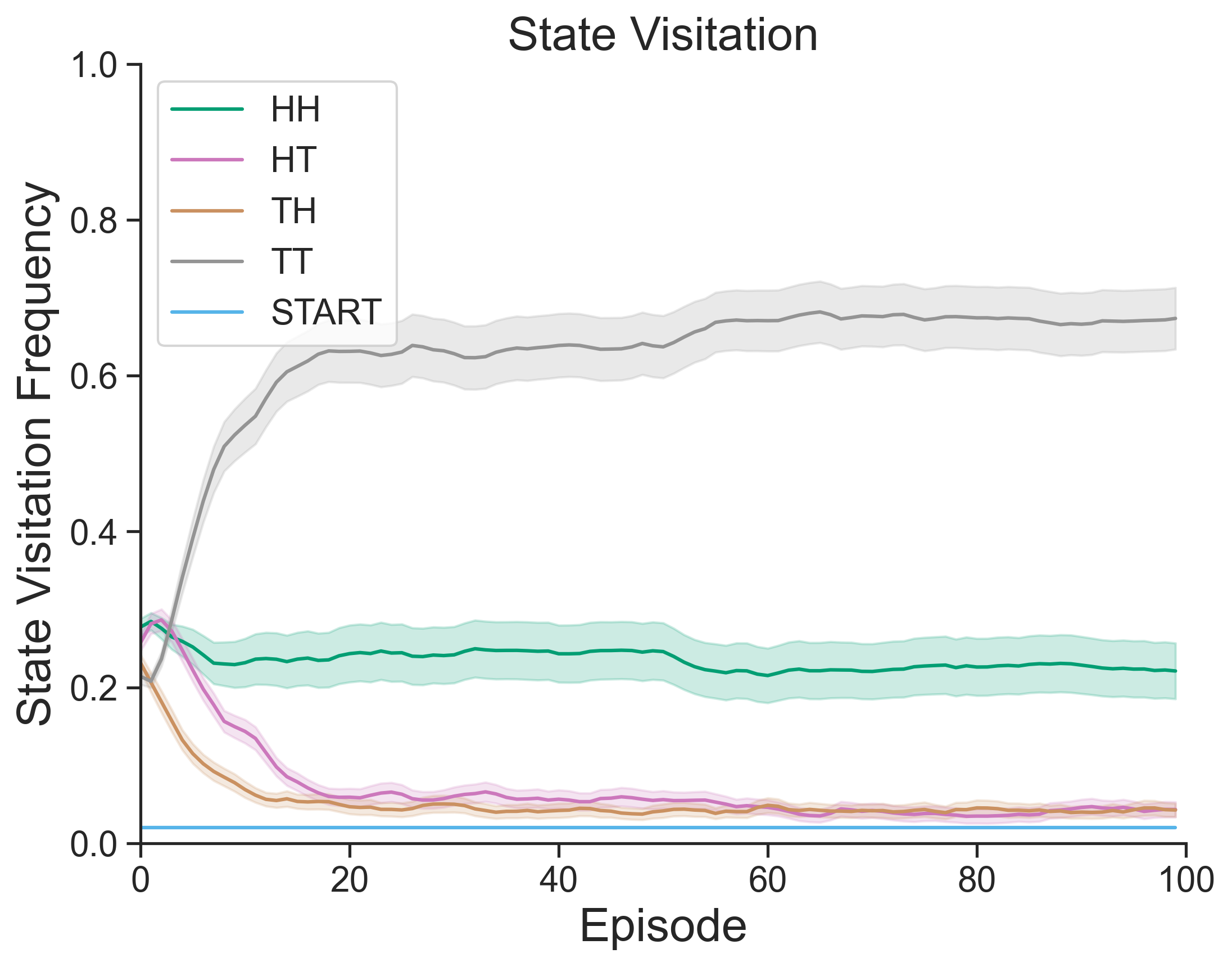}
      \caption{}
      \label{fig:earl_imp_eval_state}
    \end{subfigure}%
    \hfill
    \caption{
    \method{} finds extortion-like strategies in finite matrix games - evaluation trials composed of 100 episodes and over 20 seeds. In the IPD, we show reward (a) and state visitation (b) to demonstrate shaping by the high proportion of DC states. In the IMP, we evaluate (c) reward and (d) state visitation to demonstrate shaping by the high proportion of matching states HH and TT. 
    }
    \label{fig:earl_ipd_viz}
\end{figure*}
\pagebreak
\newpage

\section{Matrix Game Results}
\begin{table}[ht]
\caption{
Converged reward per step (meta-agent, co-player) for agents against Naive Learners in finite matrix games. \method{} can shape co-players to exploitative equilibria. We report mean and standard deviation over 20 randomised co-players.
\label{tab:matrix_results}}
    \centering
    \setlength{\tabcolsep}{2.5pt}
    \begin{tabular}{l cc}
     &  IPD  & IMP \\ 
    \midrule
     \method{} & \textbf{-0.1 $\pm$ 0.02, -2.8 $\pm$ 0.05}  & \textbf{0.9 $\pm$ 0.02, -0.9 $\pm$ 0.02}\\
     M-FOS & -0.6 $\pm$ 0.14, -2.3 $\pm$ 0.14
     & \textbf{0.8 $\pm$ 0.09, -0.8 $\pm$ 0.09} \\
     GS & -1.0 $\pm$ 0.03, -1.3 $\pm$ 0.10 & 
     0.0 $\pm$ 0.01, 0.0 $\pm$ 0.01 \\
     CT-NL & 
     -2.0 $\pm$ 0.00, -2.0 $\pm$ 0.00 & 0.0 $\pm$ 0.00, 0.0 $\pm$ 0.00 \\
    \bottomrule
    \end{tabular}
\end{table}
\section{Generalisability over long time period}

We also present the results for allowing the co-player to adapt longer to the meta-agent. This is in aims to understand what an RL agents best response to a meta-agent looks like. We report the scores below

\begin{table*}[ht]
\caption{Converged Reward (meta-agent, co-player) for agents trained with Naive Learners on the CoinGame, IPDitM and IMPitM. The reward is averaged per episode,  mean is reported across 100 seeds with standard deviations. Trial continued until episodic reward converged (for CoinGame=5000 episodes, for * in the Matrix = 1000).
\label{tab:result_gridworlds_1000}}
    \centering
    \begin{tabular}{l c c c}
     & CoinGame &IPD in the Matrix & IMP in the Matrix\\ 
    \midrule
     \method{} 
     & $\mathbf{5.21 \pm 0.66}$,$\mathbf{-4.84 \pm0.81}$
     & $\mathbf{21.94 \pm 1.12}$,$\mathbf{22.40 \pm 0.92}$
     & $\mathbf{0.14 \pm 0.06}$,$\mathbf{-0.14 \pm 0.06}$\\
     \textsc{M-FOS} (ES) & $3.12 \pm 0.42$,$ 2.27 \pm 0.40$
      & $14.69 \pm 1.28$,$ 24.93 \pm 1.14$
      & $\mathbf{0.11 \pm 0.70}$,$\mathbf{ -0.11 \pm 0.07}$
      \\
     \textsc{M-FOS} (RL) 
     & $0.85 \pm 0.60$,$ -0.23 \pm 0.44$
     & $7.58 \pm 0.21$,$ 7.26 \pm 0.17 $
     & $0.04 \pm 0.00$,$ -0.04 \pm 0.00$ \\
     GS 
     & $\mathbf{5.38 \pm 0.42}$, $\mathbf{-3.31 \pm 0.59}$
      & $15.43 \pm 1.39$,$ 25.43 \pm 1.00$ 
      & $0.00 \pm 0.00$,$ 0.00 \pm 0.00$ \\
     PT-NL 
     & $0.56 \pm 0.59$,$ 0.26 \pm 0.48$
     & $6.33 \pm 0.33$,$ 6.96 \pm 0.31 $
     & $-0.17 \pm 0.10$,$ 0.17 \pm 0.10 $\\ 
     CT-NL 
     & $0.44 \pm 0.65$,$ 0.31 \pm 0.61$
     & $5.56 \pm 0.02$,$ 5.56 \pm 0.02$
     & $-0.10 \pm 0.06$,$ 0.10 \pm 0.06$ \\ 
    \bottomrule
    \end{tabular}
\end{table*}

\section{CoinGame Details}
\label{appendix:cg_sanity}
\subsection{Evaluating Player's Competency}
In the CoinGame, agents struggle to learn (via RL) when trained against a pre-trained opponent. On inspection of trajectories, we found that competent agents removed a sufficient amount of coins to restrict reinforcement learners ability to capture signal from the game.

To address this, we show that adjusting the observations such that an agent receives to an \textit{egocentric} viewpoint (i.e. an agent always observes that it is in the centre of the grid) leads to competency against a competent opponent. In this case, we measured competency as an agent's ability to pick up coins. \textit{Competent} agents were those who picked up a similar number of coins to those trained against a stationary agent. Throughout the rest of the paper, we refer to the original version of CoinGame as \textit{non-egocentric} CoinGame and the modified observation version as \textit{egocentric} CoinGame. In addition, we deviate from the original 5 by 5 version of CoinGame to a 3 by 3 version, following the setting used in \citep{lu2022model}. 

\begin{figure}[H]
    \centering
    \begin{subfigure}[b]{0.32\textwidth}
      \centering
      \includegraphics[width=\textwidth]{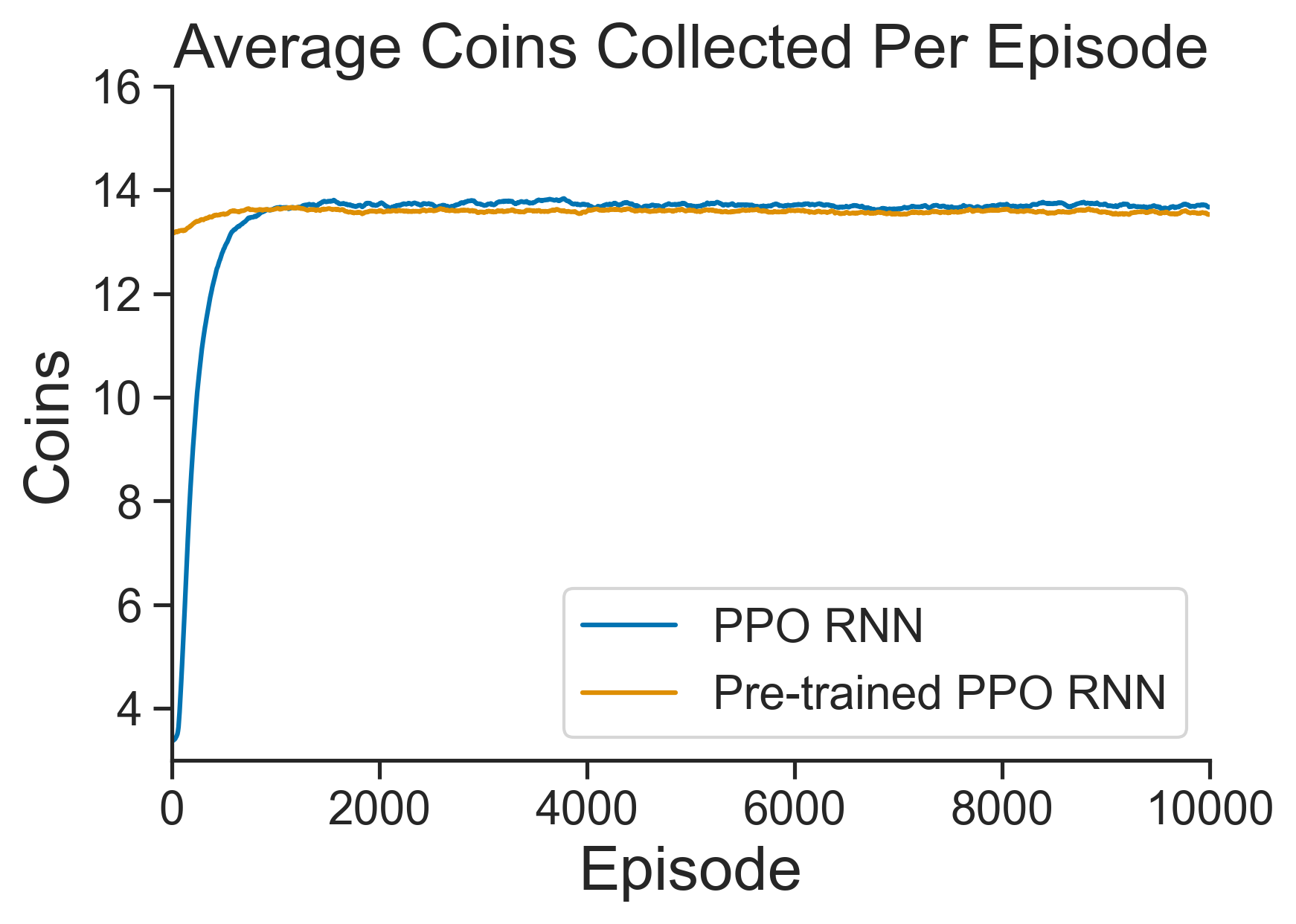}
      \caption{}
      \label{fig:ppo_rnn_sanity_cg_ppo_rnn}
    \end{subfigure}
    \begin{subfigure}[b]{0.32\textwidth}
      \centering
      \includegraphics[width=\textwidth]{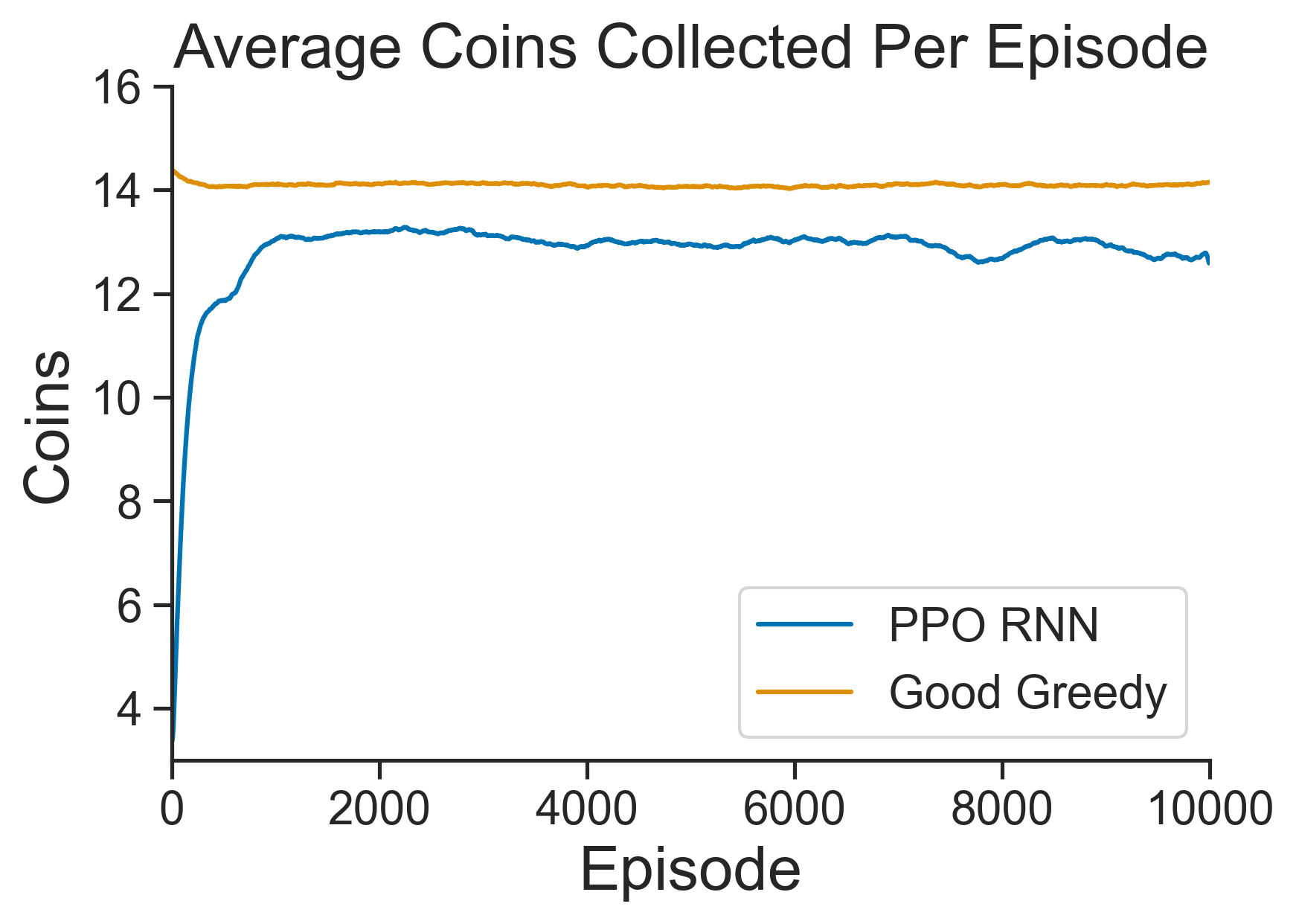}
      \caption{}
      \label{fig:ppo_rnn_sanity_cg_good_greedy}
    \end{subfigure}
    \begin{subfigure}[b]{0.32\textwidth}
      \centering
      \includegraphics[width=\textwidth]{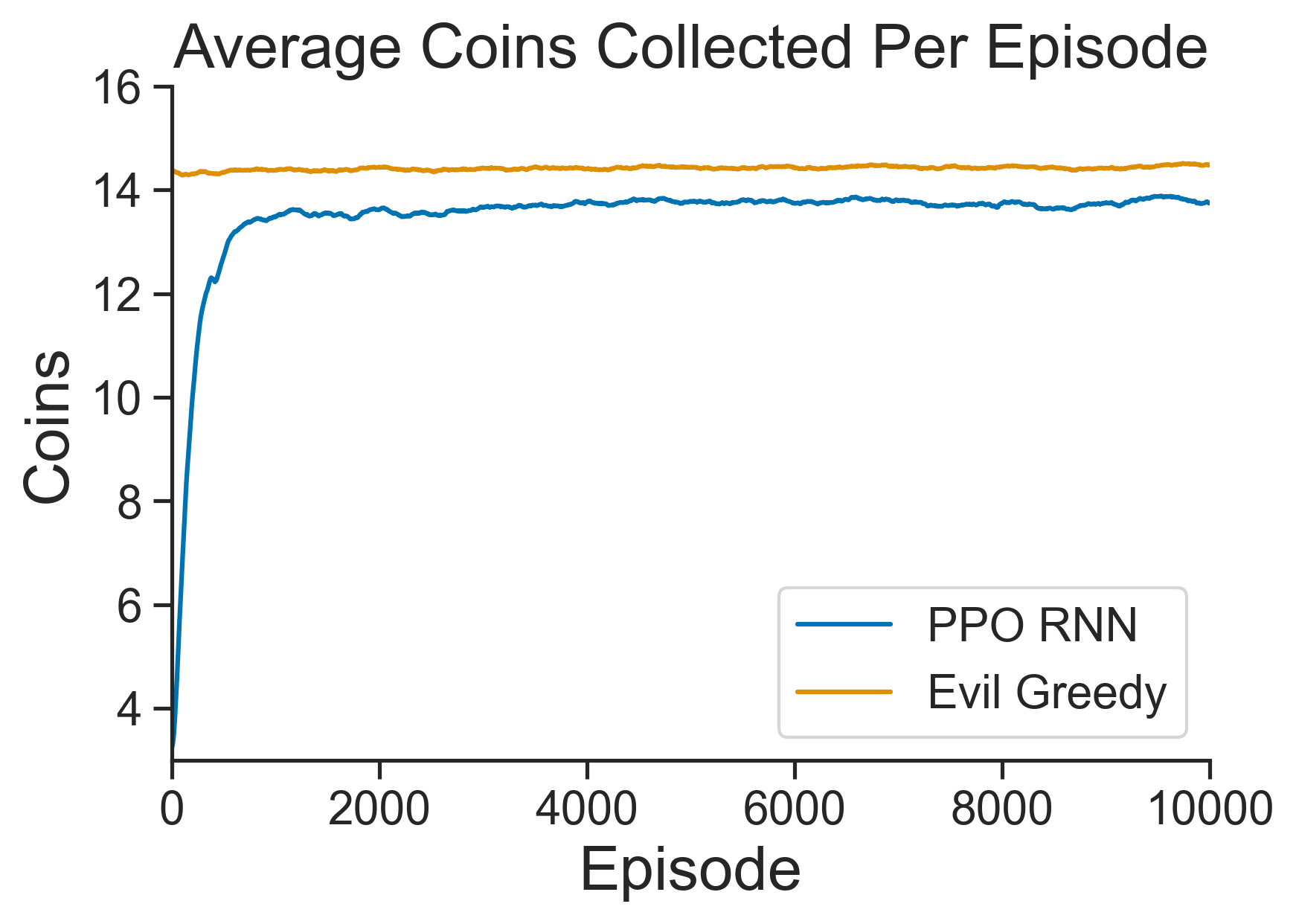}
      \caption{}
      \label{fig:ppo_rnn_sanity_cg_evil_greedy}
    \end{subfigure}

    \caption{Results of the experimental protocol verifying that PPO RNN learns to play the egocentric CoinGame against (a) pre-trained PPO RNN (b) Good Greedy (c) Evil Greedy. Notice that the agent learns to pick up roughly the same number of coins per episode as its competent opponents. 
    }
    \label{fig:coin_game_sanity_viz}
\end{figure}
\subsection{Training Details}
\begin{figure}[ht]
    \centering
    \begin{subfigure}[b]{0.24\textwidth}
      \centering
      \includegraphics[width=\textwidth]{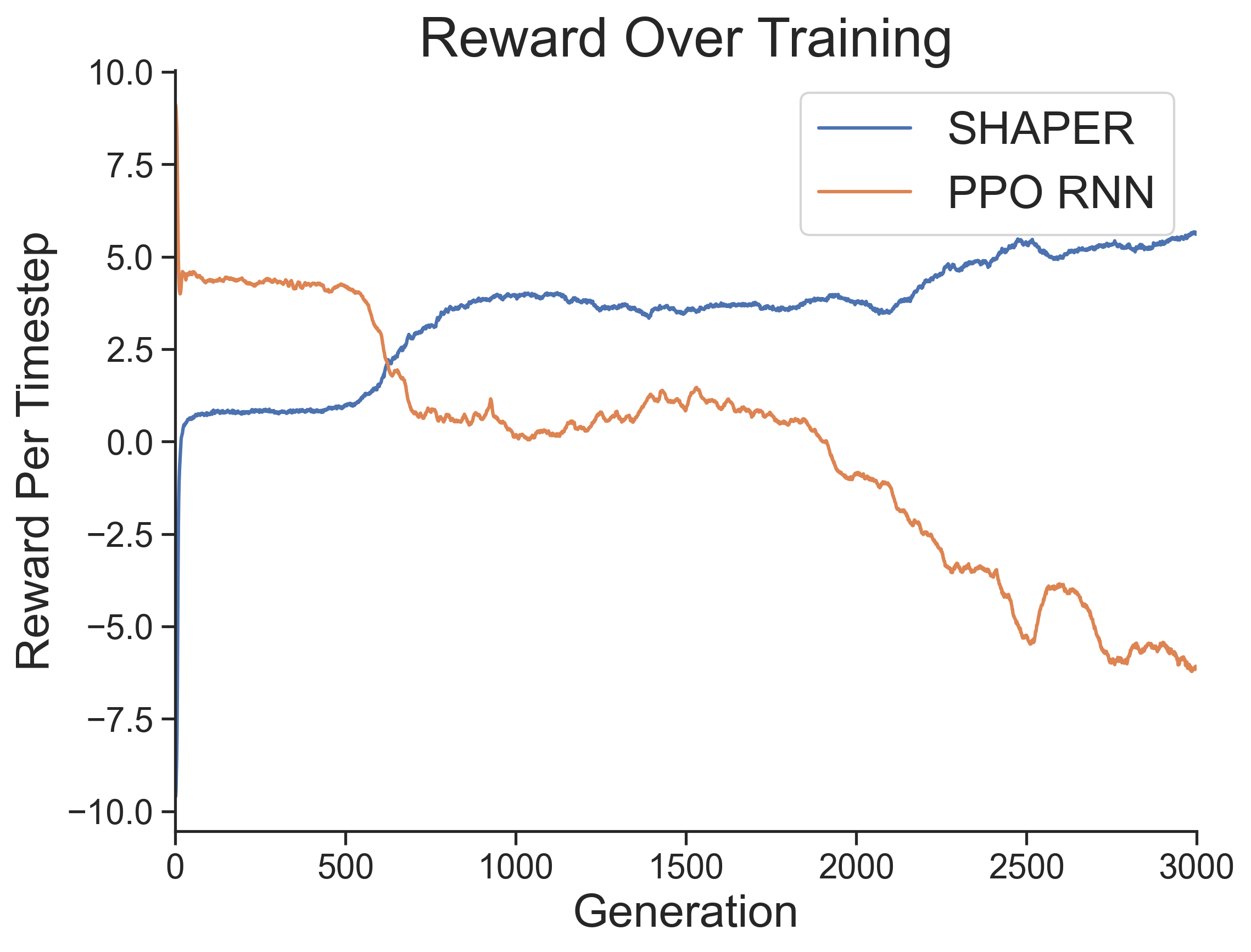}
      \caption{}
      \label{fig:earl_cg_train_reward}
    \end{subfigure}
    \begin{subfigure}[b]{0.24\textwidth}
      \centering
      \includegraphics[width=\textwidth]{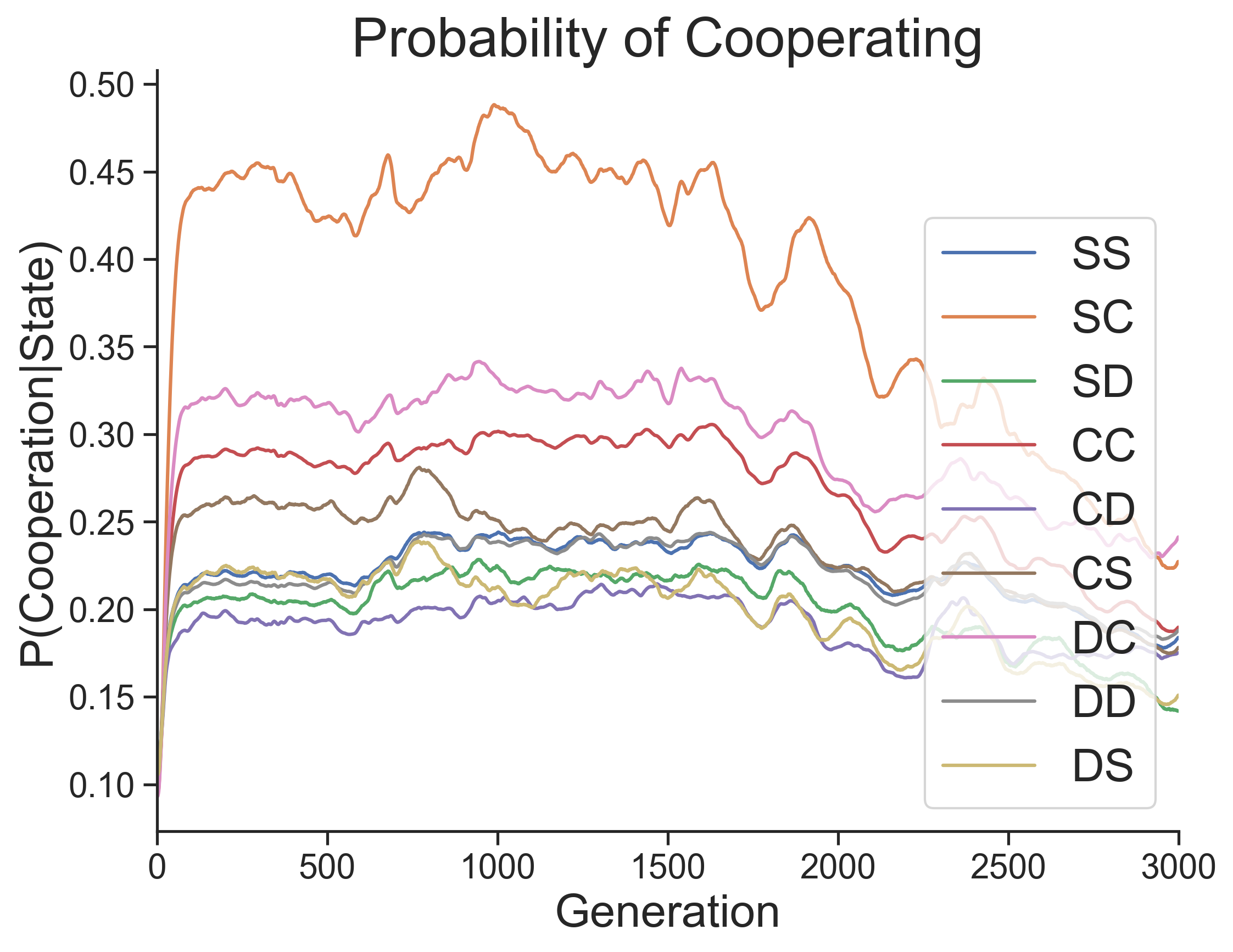}
      \caption{}
      \label{fig:earl_cg_fingerprint}
    \end{subfigure}
    \begin{subfigure}[b]{0.24\textwidth}
      \centering
      \includegraphics[width=\textwidth]{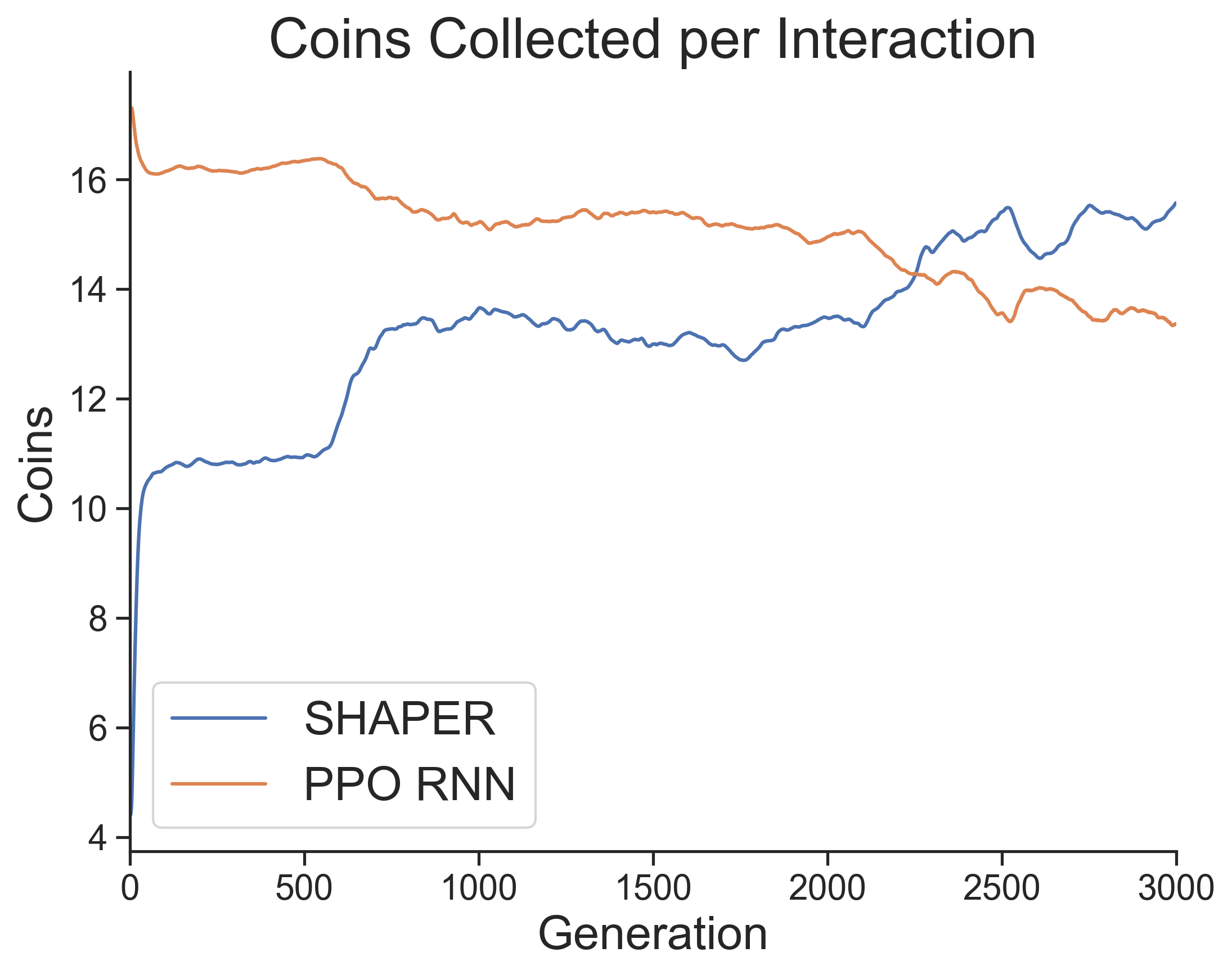}
      \caption{}
      \label{fig:earl_cg_train_coins}
    \end{subfigure}
    \begin{subfigure}[b]{0.24\textwidth}
      \centering
      \includegraphics[width=\textwidth]{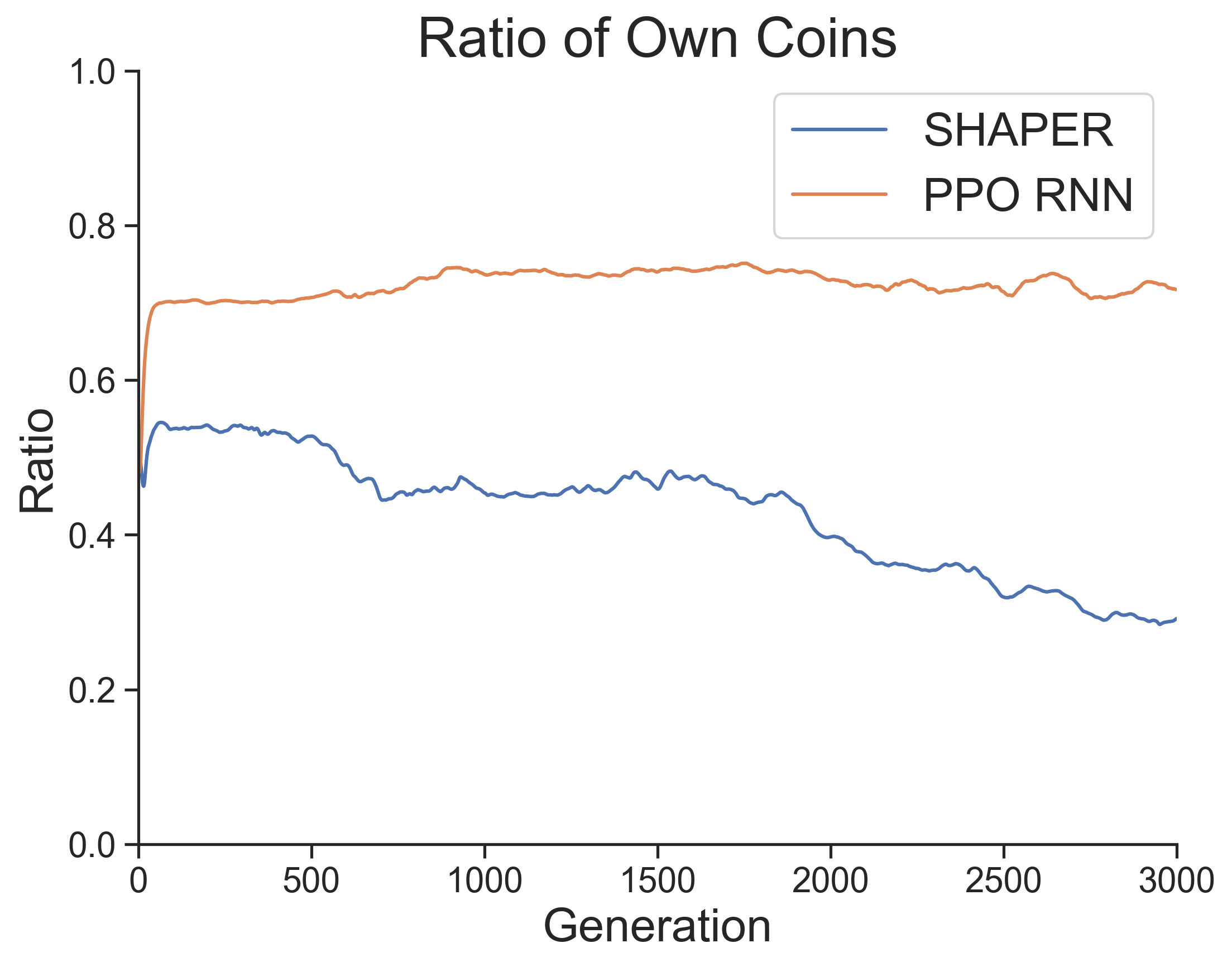}
      \caption{}
      \label{fig:earl_cg_train_prob}
    \end{subfigure}
    \caption{Training results of \method{} vs. PPO RNN in the egocentric CoinGame. (a) Fitness, (b) the meta-agent's frequency of picking up its own colour coin depending on existing convention, (c) the number of coins picked up per episode, (d) both agent's frequency of picking up its own colour coin over a full episode.}
\end{figure}

\clearpage
\section{Ablation Studies Results \& Details}

\begin{table}[ht]
\caption{
Ablations highlighting the importance of context and history for Shaping. We report converged reward per step (meta-agent, co-player) for agents against Naive Learners.
\label{tab:ablation}}
    \centering
    \begin{tabular}{l c }
    \toprule
     \multicolumn{2}{c}{Context Challenge:  IPD}   \\  
     \method{} & -0.8, -2.0 \\ 
     \method{} w/o Context & -1.25, -1.75\\ 
    \midrule
    \midrule
    \multicolumn{2}{c}{History Challenge: IMP (Length=2)} \\ 
     \method{} & 0.5, -0.5  \\ 
    \method{} w/o History & 
     0.0, 0.0 \\ 
    \midrule
    \midrule
    \multicolumn{2}{c}{History Challenge: IMP (Length=100)} \\ 
     \method{} & 0.5, -0.5 \\ 
    \method{} w/o History & 
     0.5, -0.5 \\ 
    \bottomrule
    \end{tabular}
\end{table}

We also include additional state visitation for the hardstop challenge. This helps validate that \method{} reacts to agents becoming exploitable after a hardstop has occurred.
\begin{figure}[h]
    \begin{subfigure}[]{0.45\textwidth}
      \centering
      \includegraphics[width=\textwidth]{images/hardstop/chaos_ipd_hardstop_coop.png}
    \end{subfigure}
    \begin{subfigure}[]{0.45\textwidth}
      \centering
      \includegraphics[width=\textwidth]{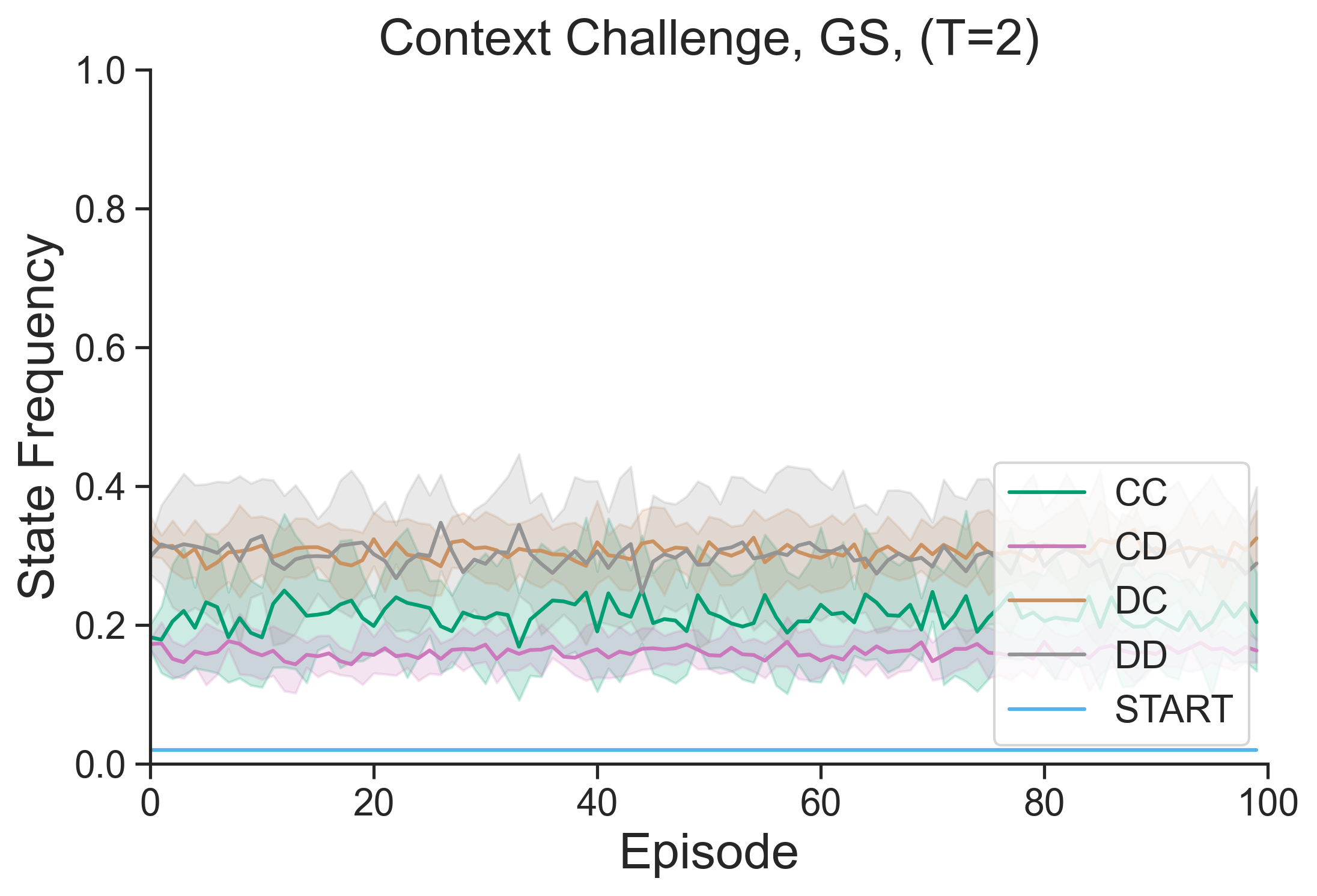}
      \label{fig:gs_coop_hardstop}
    \end{subfigure}
\caption{State visitation for the Hardstop Challenge with meta-agents (a) \method{} and (b) GS. Here we see  \method{} responds to co-players frozen stationary policy by moving into either DD (the best response to a defective agent) or DC (the best response to a fully cooperative agent), whilst GS is unable to adjust its approach after the co-player stop learning, continuing to plays the sub-optimal shaping strategy)}
\label{fig:hardstop_state_vis}
\end{figure}

\begin{figure}[h]
\begin{center}
    \begin{subfigure}[]{0.3\textwidth}
      \centering
      \includegraphics[width=\textwidth]{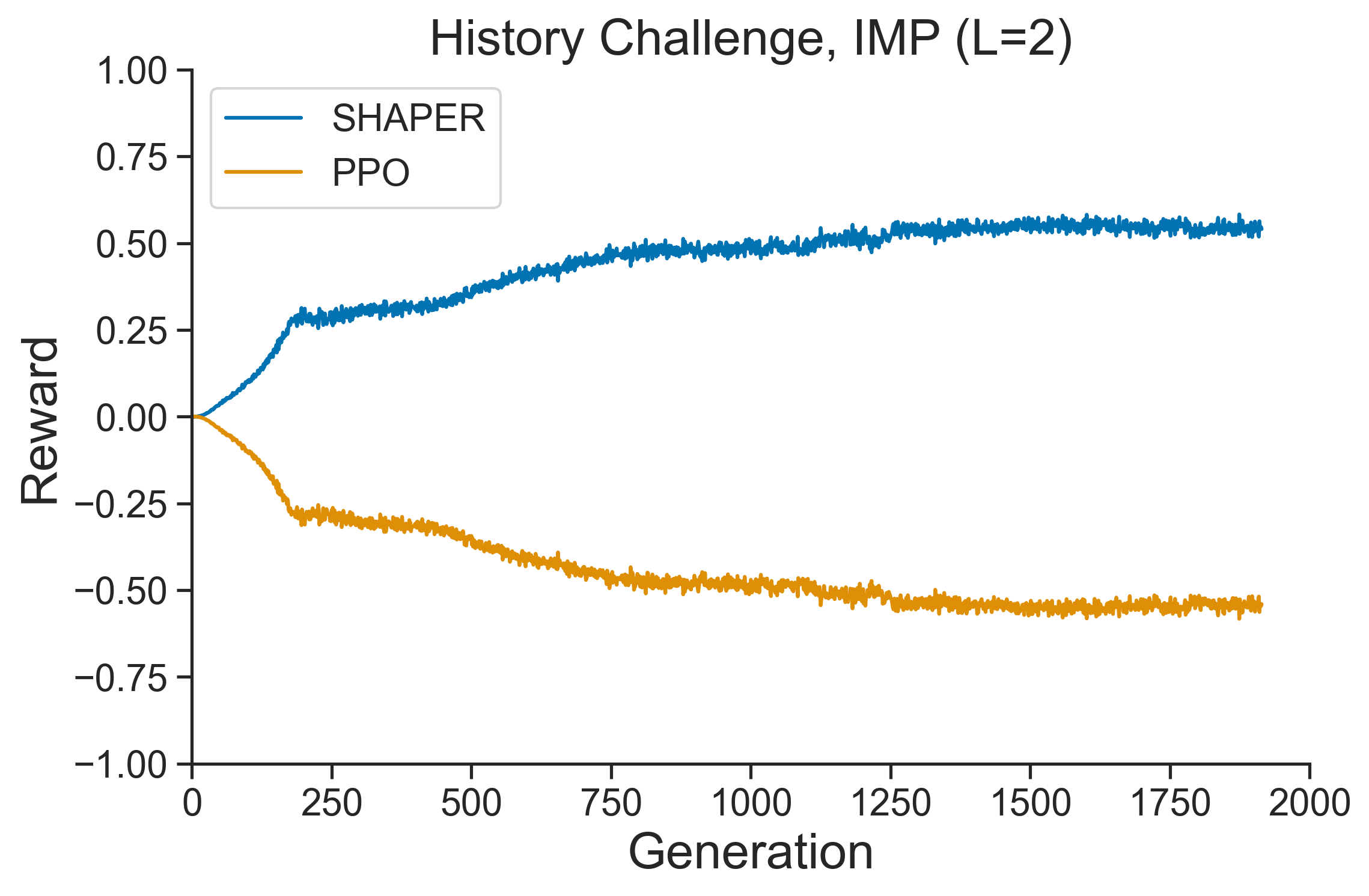}
        \caption{}
        \label{fig:earl_two_step}
    \end{subfigure}
    \begin{subfigure}[]{0.3\textwidth}
      \centering
      \includegraphics[width=\textwidth]{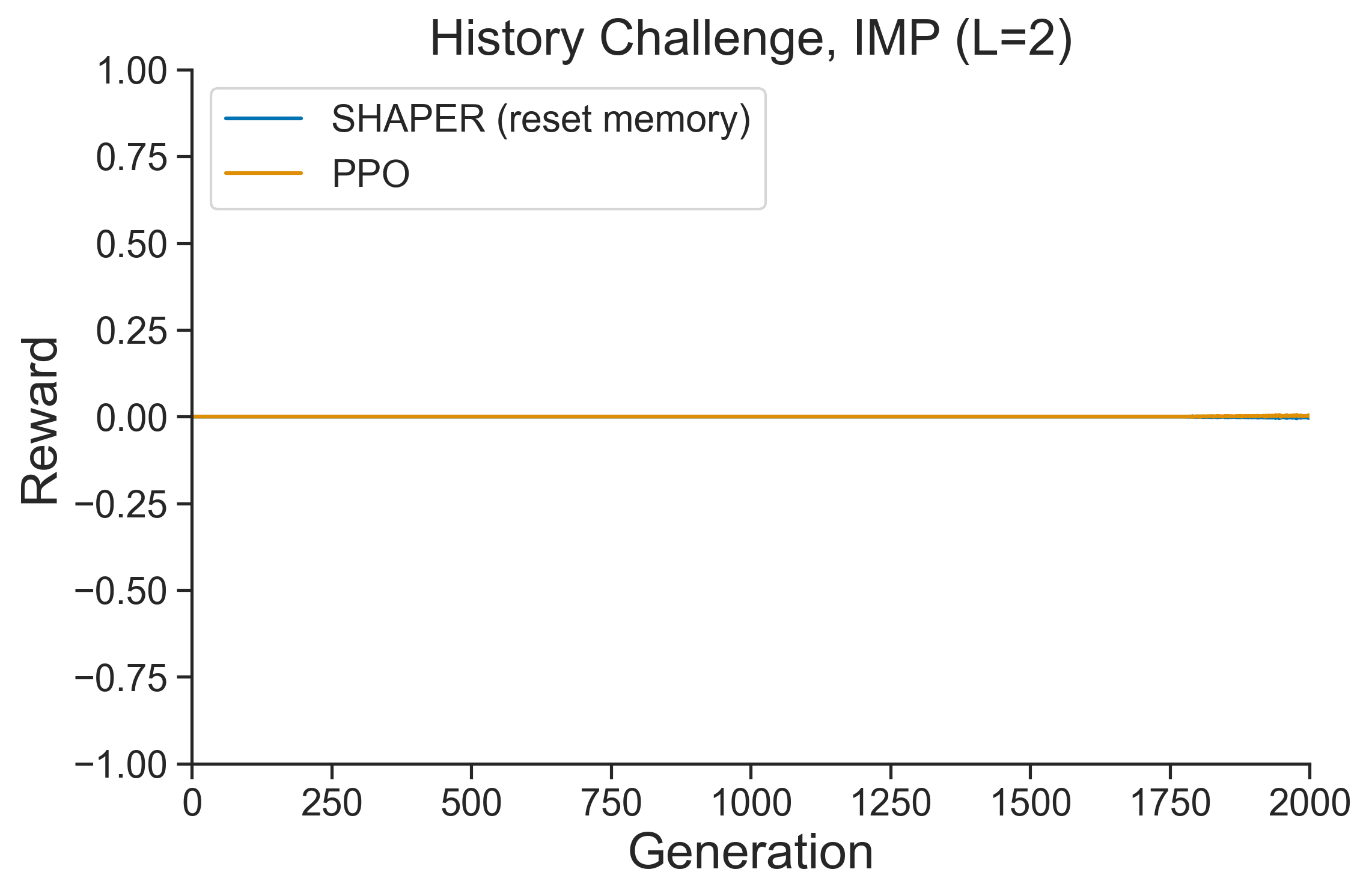}
    \caption{}
      \label{fig:gs_two_step}
    \end{subfigure}
    \begin{subfigure}[]{0.3\textwidth}
      \centering
      \includegraphics[width=\textwidth]{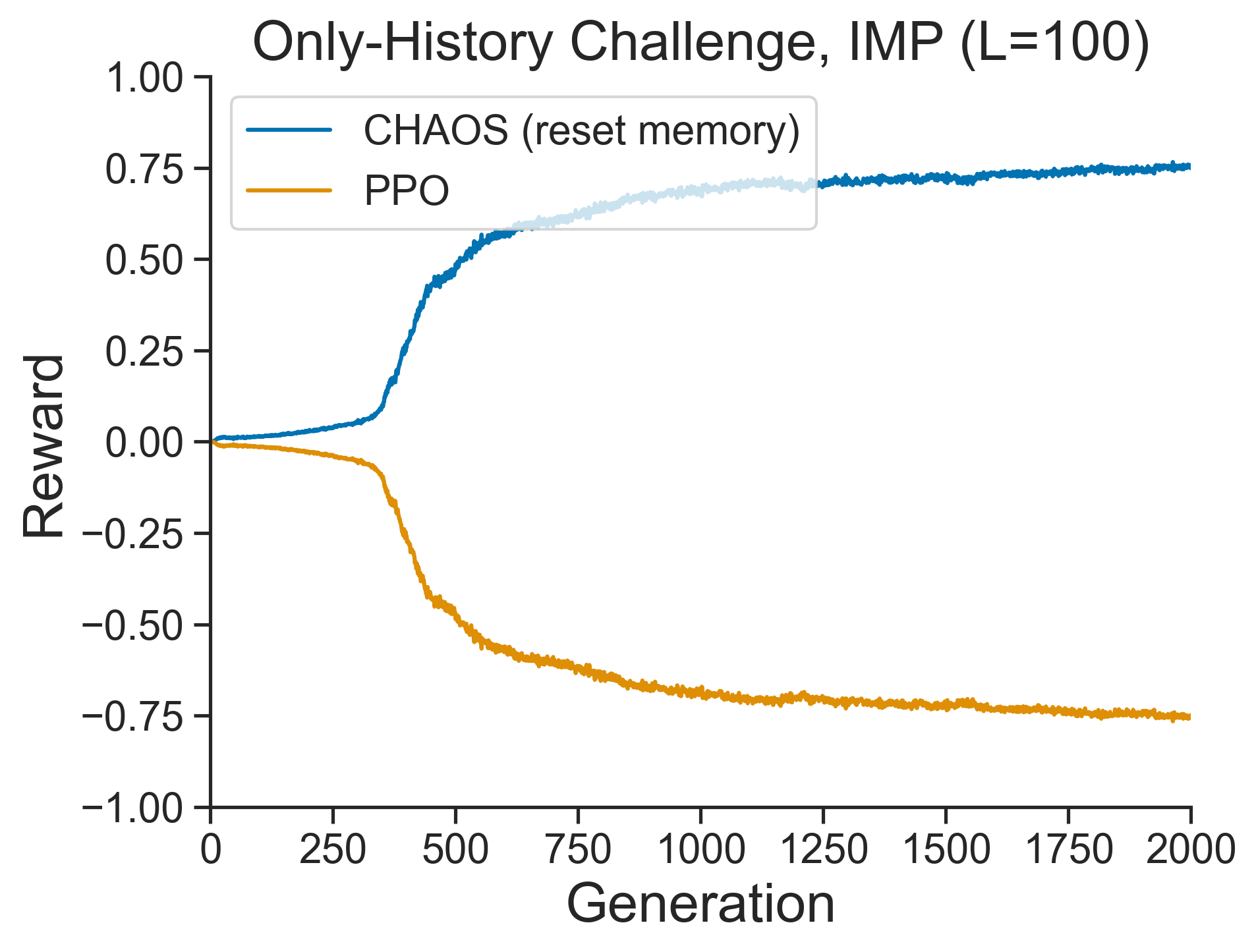}
    \caption{}
      \label{fig:gs_two_step_length_100}
    \end{subfigure}
\end{center}
\caption{The Only-History Challenge: Training curves in the IMP with episode length = 2  for (a) \method{} and (b) Shaper without context. Note that in short time-spans, history cannot enable shaping. Additionally (c) \method{} without context in IMP with episode length = 100 shows with sufficient timespan, history can enable shaping. 
\label{fig: only_history}} 
\end{figure}


\newpage
\pagebreak

\clearpage
\section{Gridworld Details}
\begin{figure}[H]
    \centering
    \begin{subfigure}[b]{0.23\textwidth}
      \includegraphics[width=\textwidth]{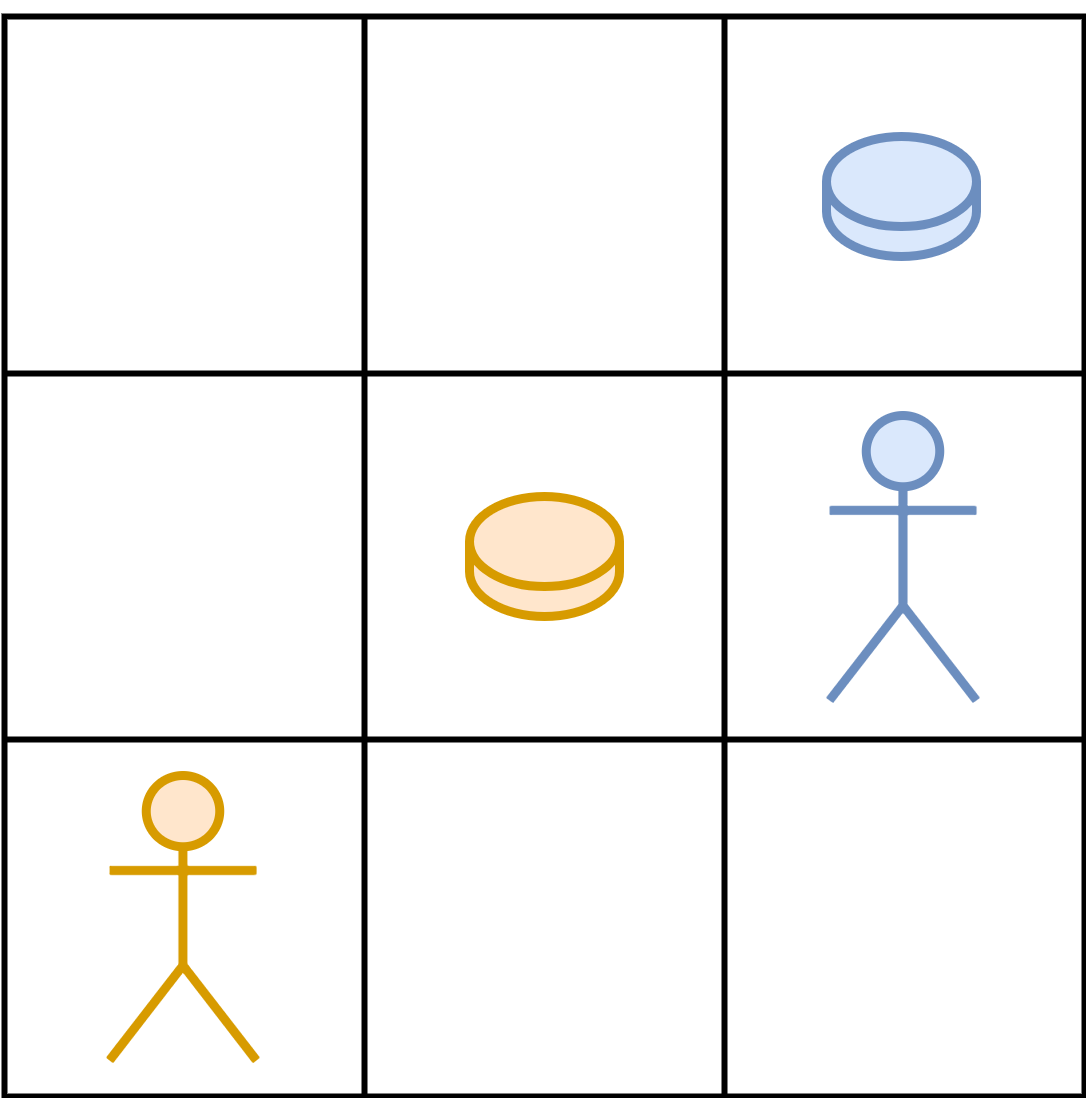}
      \caption{}
      \label{fig:coin_game_a}
    \end{subfigure}
    ~~~
    \begin{subfigure}[b]{0.23\textwidth}
      \includegraphics[width=\textwidth]{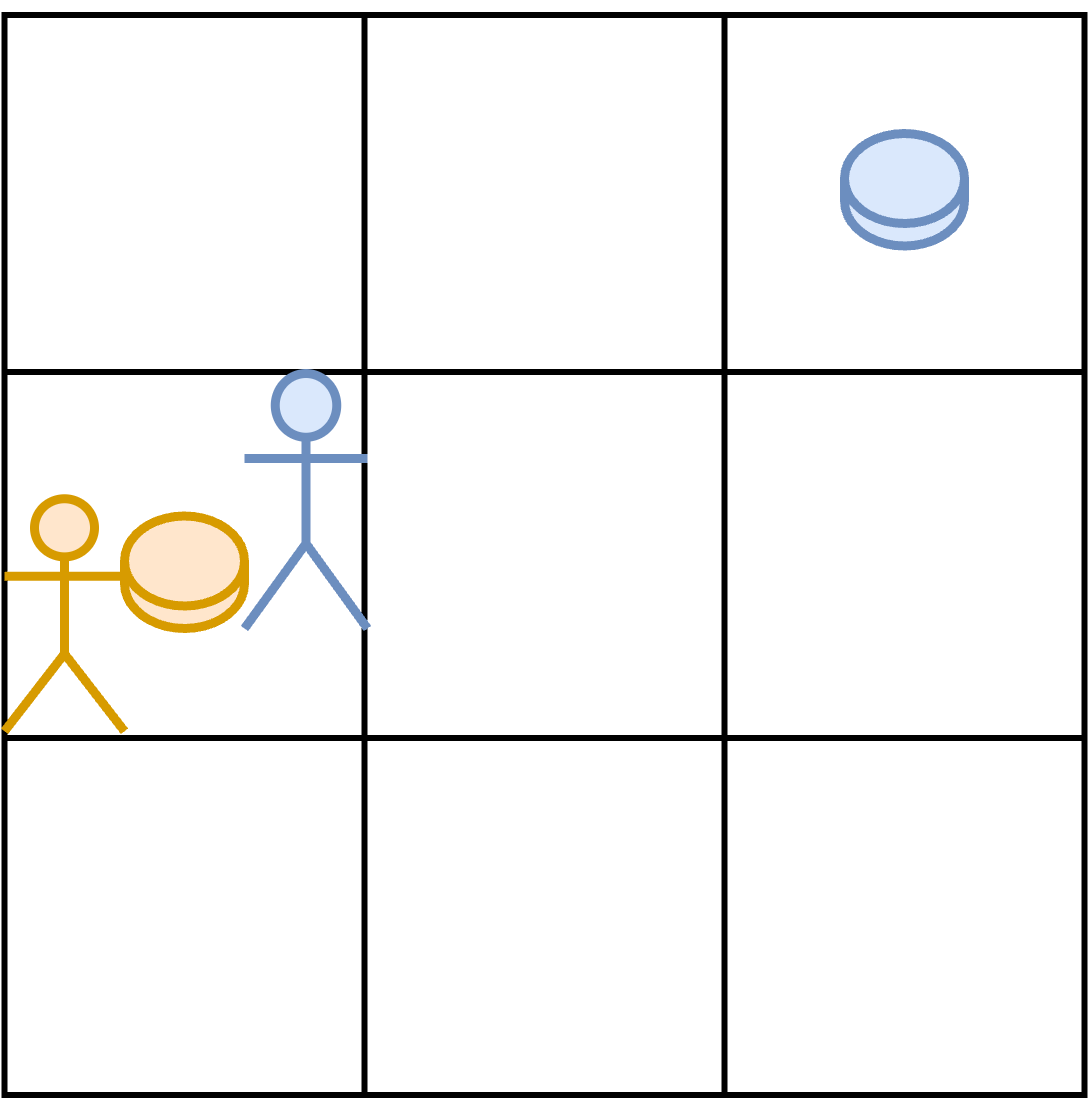}
      \caption{}
      \label{fig:coin_game_convention_break}
    \end{subfigure}
    \begin{subfigure}[b]{0.23\textwidth}
      \includegraphics[width=\textwidth]{images/ipditm/render/ipditm_start_state.png}
      \caption{}
      \label{fig:ipditm_a}
    \end{subfigure}
    \begin{subfigure}[b]{0.23\textwidth}
      \includegraphics[width=\textwidth]{images/ipditm/render/ipditm_state.png}
      \caption{}
      \label{fig:ipditm_b}
    \end{subfigure}
    \caption{Illustration of the CoinGame, a multi-step general-sum game. (a) shows a typical state, where agents gain $+1$ for collecting any coin and, if the coin is not theirs, inflict a penalty of $-2$ to the co-player. (b) demonstrates a degenerate state of the game, where memory-less agents infer co-player behaviour (in this case, the blue agent defects). (c-d) Render of the IPDitM games. Agents with restricted visibility and orientation traverse a grid picking up either \textit{Defect} or \textit{Cooperate} coins. (c) shows an initial state of the game before either agent has a coin. Once agents pick up a coin, their appearance changes, and they can interact. (d) shows agent one having collected a coin and agent 2 firing the interact beam.
    \vspace{-1.0\baselineskip}
    }
    \label{fig:coin_game_viz}
\end{figure}
\clearpage

\section{IPD in the Matrix Details}

\subsection{Environment}
\label{app:environment}

We first present a typical rollout from the environments below.

\begin{figure}[h]
  \centering
  \begin{subfigure}[b]{0.1\textwidth}
    \centering
    \includegraphics[width=\textwidth]{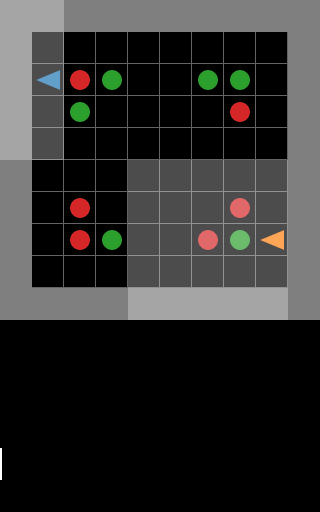}
  \end{subfigure}
  \begin{subfigure}[b]{0.1\textwidth}
    \centering
    \includegraphics[width=\textwidth]{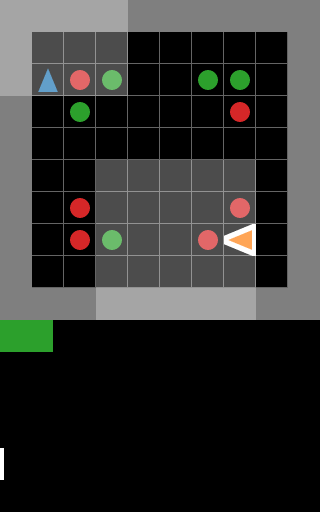}
  \end{subfigure}
  \begin{subfigure}[b]{0.1\textwidth}
    \centering
    \includegraphics[width=\textwidth]{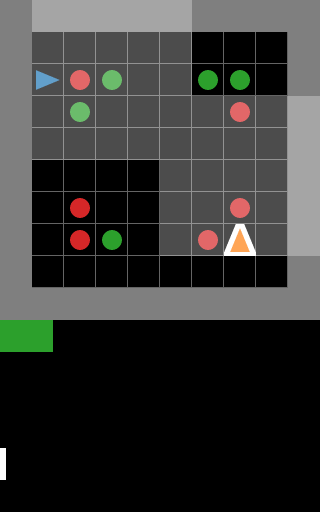}
  \end{subfigure}
  \begin{subfigure}[b]{0.1\textwidth}
    \centering
    \includegraphics[width=\textwidth]{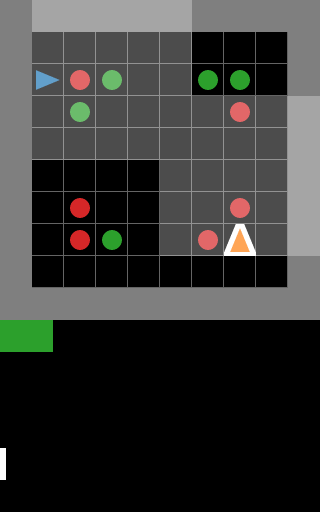}
  \end{subfigure}
  \begin{subfigure}[b]{0.1\textwidth}
    \centering
    \includegraphics[width=\textwidth]{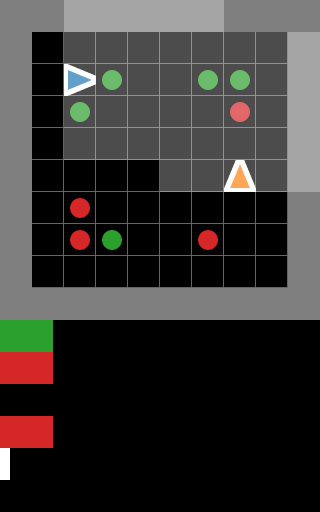}
  \end{subfigure}
  \begin{subfigure}[b]{0.1\textwidth}
    \centering
    \includegraphics[width=\textwidth]{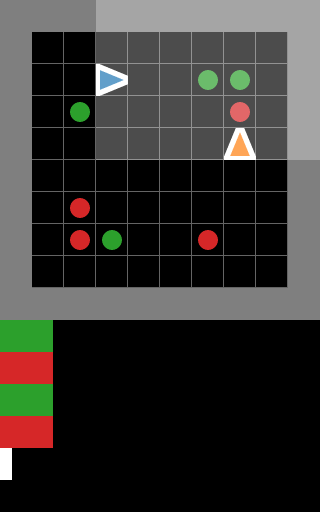}
  \end{subfigure}
  \begin{subfigure}[b]{0.1\textwidth}
    \centering
    \includegraphics[width=\textwidth]{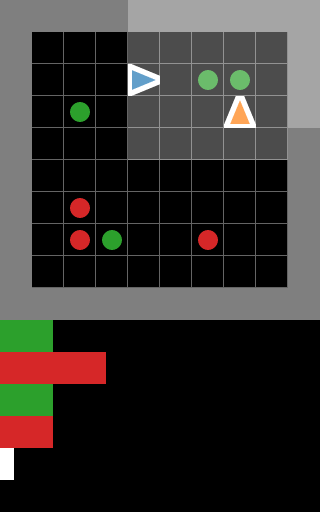}
  \end{subfigure}
  \begin{subfigure}[b]{0.1\textwidth}
    \centering
    \includegraphics[width=\textwidth]{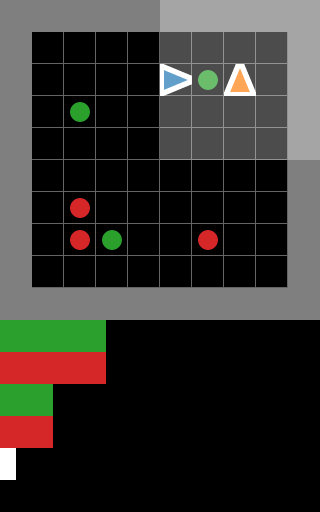}
  \end{subfigure}
  \begin{subfigure}[b]{0.1\textwidth}
    \centering
    \includegraphics[width=\textwidth]{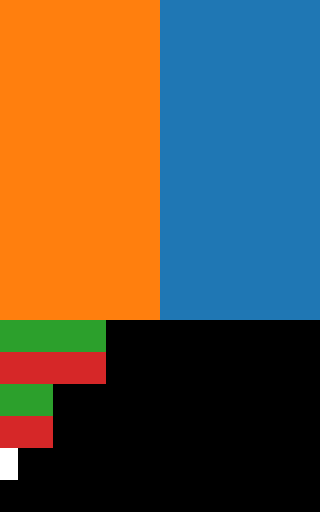}
  \end{subfigure}
  \caption{Rollout of a typical episode. Player 1 collects a green coin first then two reds before a final green coin; Player 2 collects a red then green coin. Both agents collect an equal ratio of coins and when interacting both play a mixed strategy of $(0.5, 0.5)$ resulting in them receiving equal reward when interacting. This state is presented for 5 frames (whilst frozen) after which coins are restored (in same positions) and agents respawn (in new positions).}
  \label{fig:8subfigures}
\end{figure}

\begin{figure}[h]
\centering
\includegraphics[width=\textwidth]{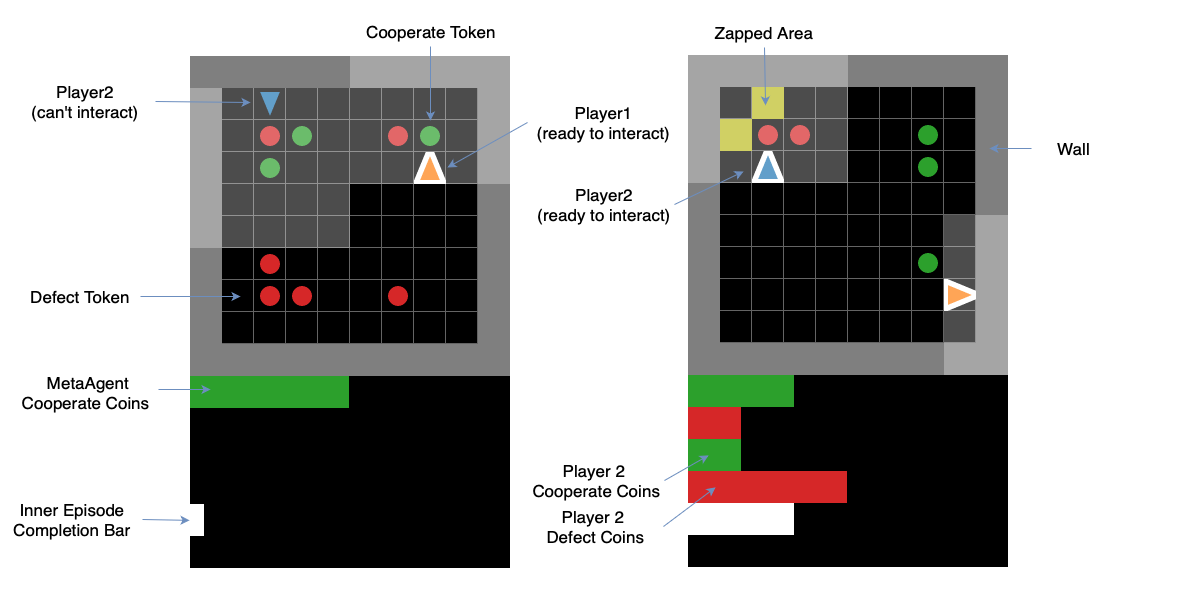}
\caption{Annotated Image of IPDiTM renders, demonstrating the objects within the game}
\label{fig:ipditm_annotated}
\end{figure}

\textbf{* in the Matrix} extends matrix games to gridworld environments~\citep{vezhnevets2020options}, where * is any normal-form game. For visual descriptions, see Figures \ref{fig:coin_game_viz}(c,d), \ref{fig:8subfigures}, and\ref{fig:ipditm_annotated}. Agents collect two types of resources into their inventory: \textit{Cooperate} and \textit{Defect} coins. Once an agent has collected any coin, the agent's colour changes, representing that the agent is ``ready'' for interaction. Agents can fire an `interact' beam to an area in front of them. If an agent's interact beam catches a ``ready'' agent, both receive rewards equivalent to playing a matrix game *, where their inventory represents their policy. For example, when agent 1's inventory is 1 \textit{Cooperate} coin and 3 \textit{Defect} coins, agent 1's probability to cooperate is 25\%. Upon a successful interaction, agents are frozen for five steps whilst their inventories are displayed to one another. After the freeze, agents respawn in new locations, and the current coins are reset. Agents have orientation, limited directed visibility and can step forward. Agents cannot occupy the same spot, with collisions giving preference to the stationary player; the agent who moves is randomised in the case both moved. Agents only observe their own inventory. On the completion of a full episode, coin locations are randomised.

\textbf{Difference to Melting Pot:} The meltingpot environment is 23x15, whereas our grid is 8x8. A size we chose to be as large as possible whilst still being able to optimise the methods given compute limitations. Meltingpot has walls placed within the environment, whereas ours does not. While our environment is smaller, we add additional stochasticity by randomising the coin positions, which are fixed in Meltingpot. Finally, unlike the meltingpot environment, agents spawned with no coins (as opposed to one of each), this made it easier for agents to choose pure cooperate or defect strategies. We found that randomising coin positions important as even large environments representing POMDPs with insufficient stochasticity, such as the The Starcraft Multi-Agent Challenge, can be solved without memory\citep{samvelyan19smac,ellis22022smacv2}. We found similar evidence in the IPDitM. In IMPitM, the same differences hold.

\subsection{Training Details}
\label{appendix:ipditm_training}
Below we present training curves for the three major shaping baselines.

\begin{figure}[h]
    \centering
    \begin{subfigure}[b]{0.24\textwidth}
      \centering
      \includegraphics[width=\textwidth]{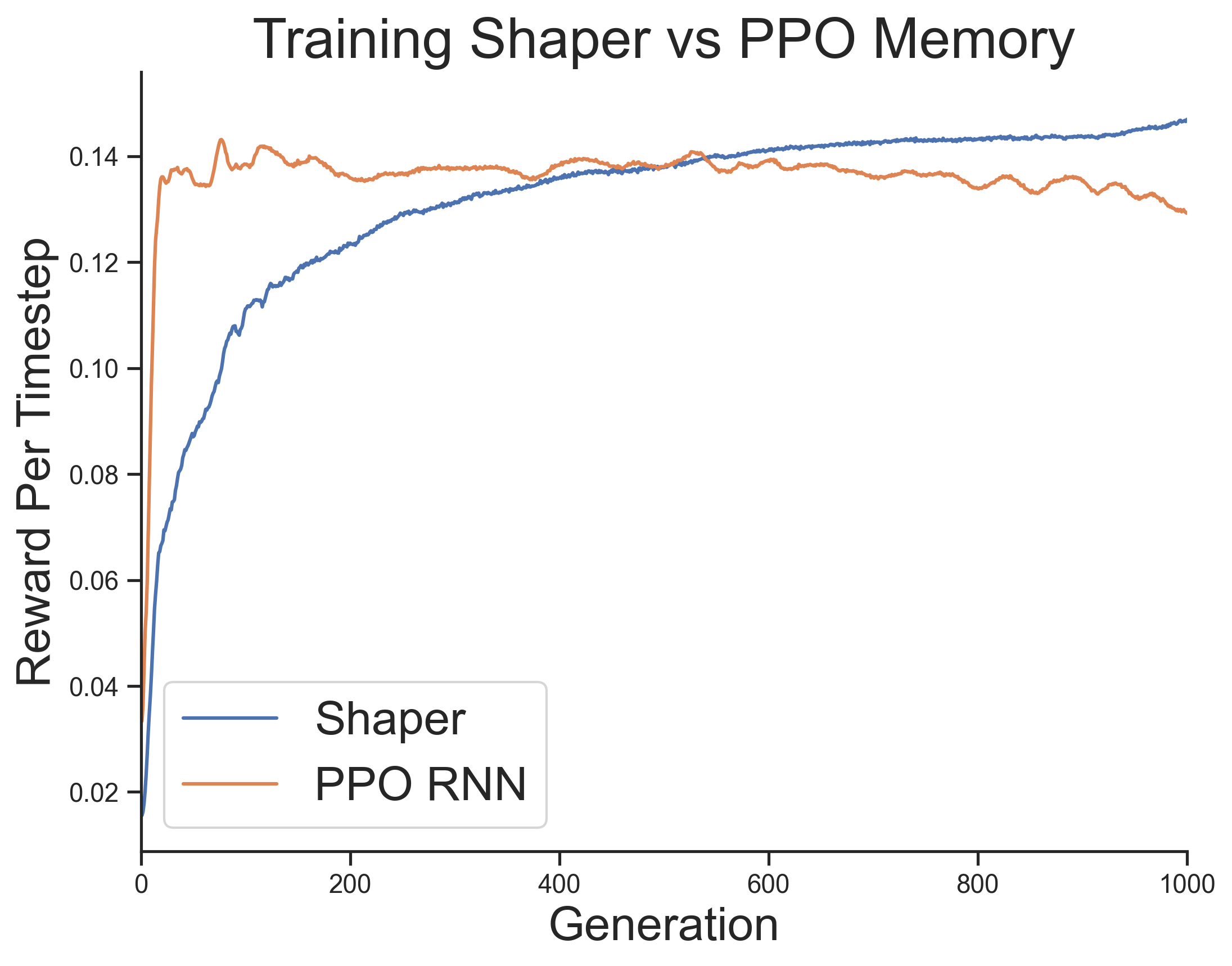}
      \label{fig:chaos_ipditm_train}
    \end{subfigure}
    \begin{subfigure}[b]{0.24\textwidth}
      \centering
      \includegraphics[width=\textwidth]{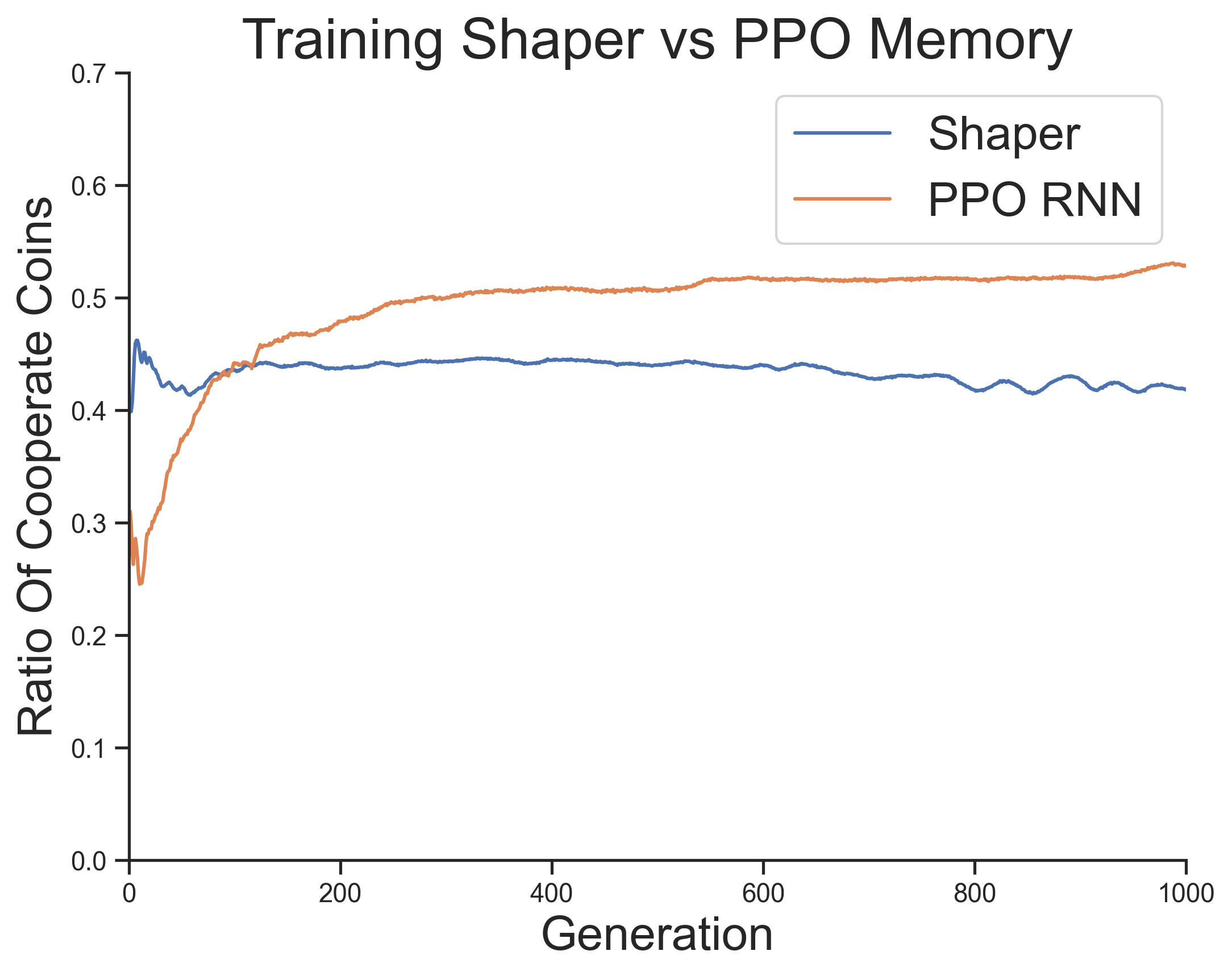}
      \label{fig:chaos_ipditm_ratio}
    \end{subfigure}
    \begin{subfigure}[b]{0.24\textwidth}
      \centering
      \includegraphics[width=\textwidth]{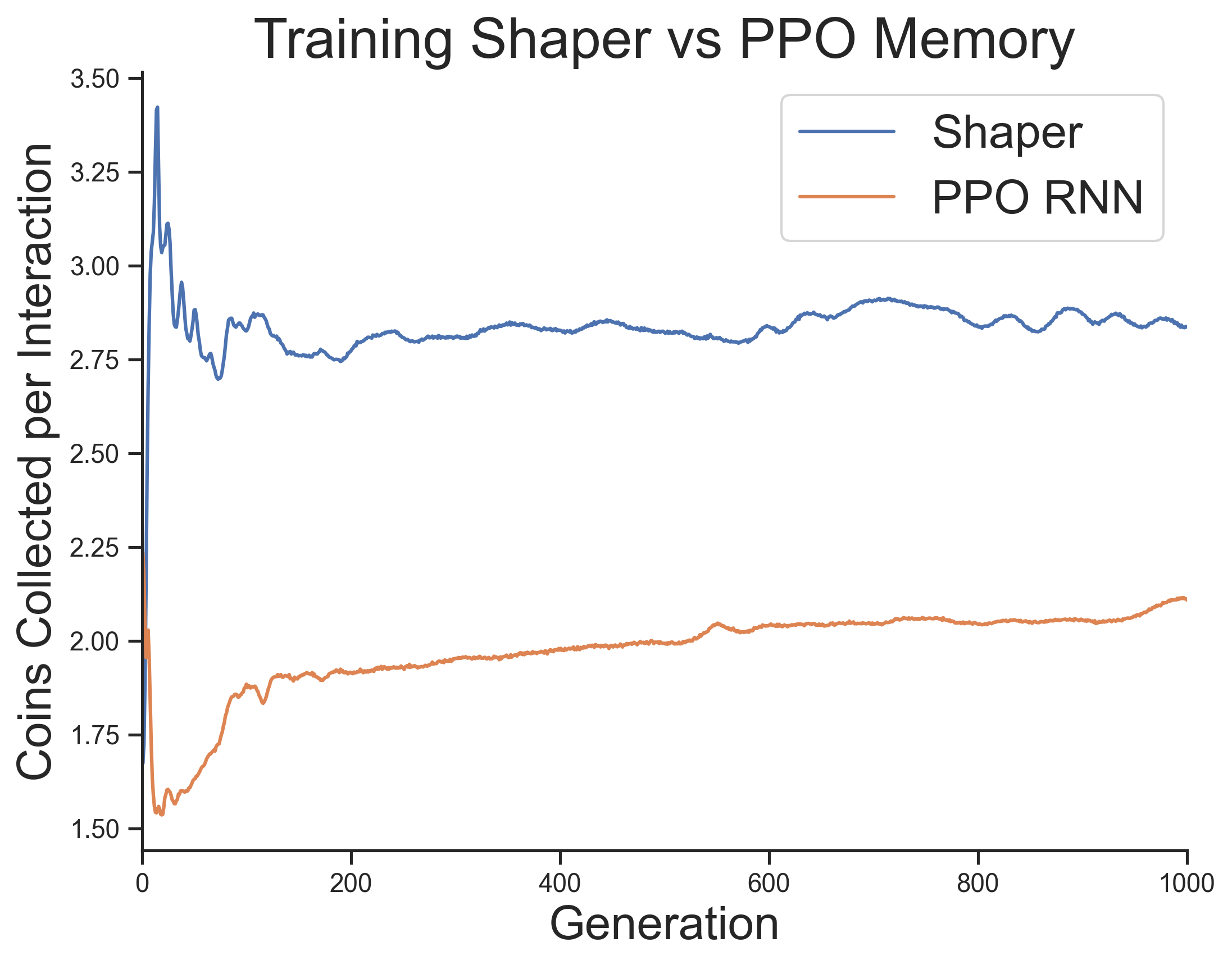}
      \label{fig:chaos_ipditm_coins}
    \end{subfigure}
    \begin{subfigure}[b]{0.24\textwidth}
      \centering
      \includegraphics[width=\textwidth]{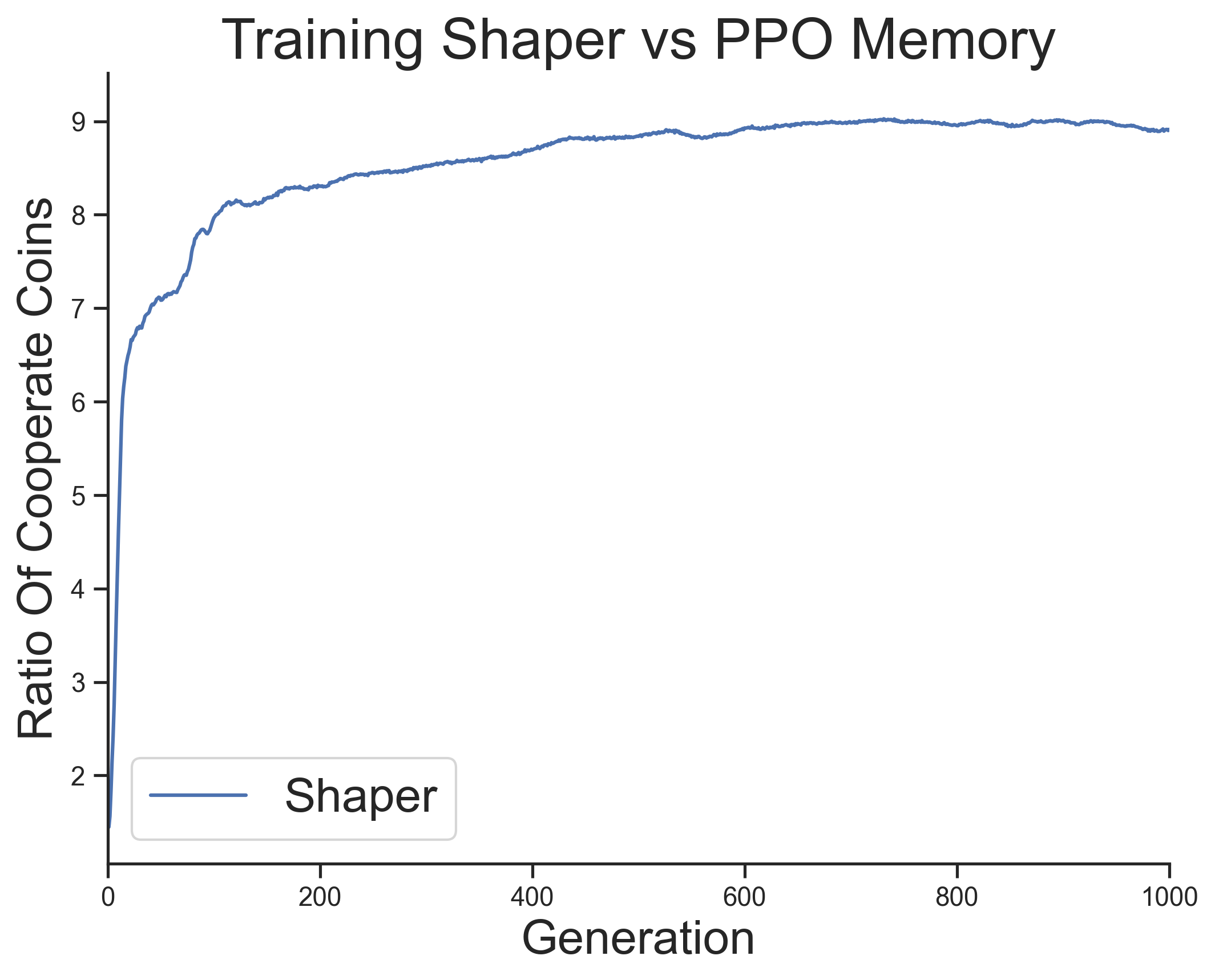}
      \label{fig:chaos_ipditm_reset}
    \end{subfigure}
    \begin{subfigure}[b]{0.24\textwidth}
      \centering
      \includegraphics[width=\textwidth]{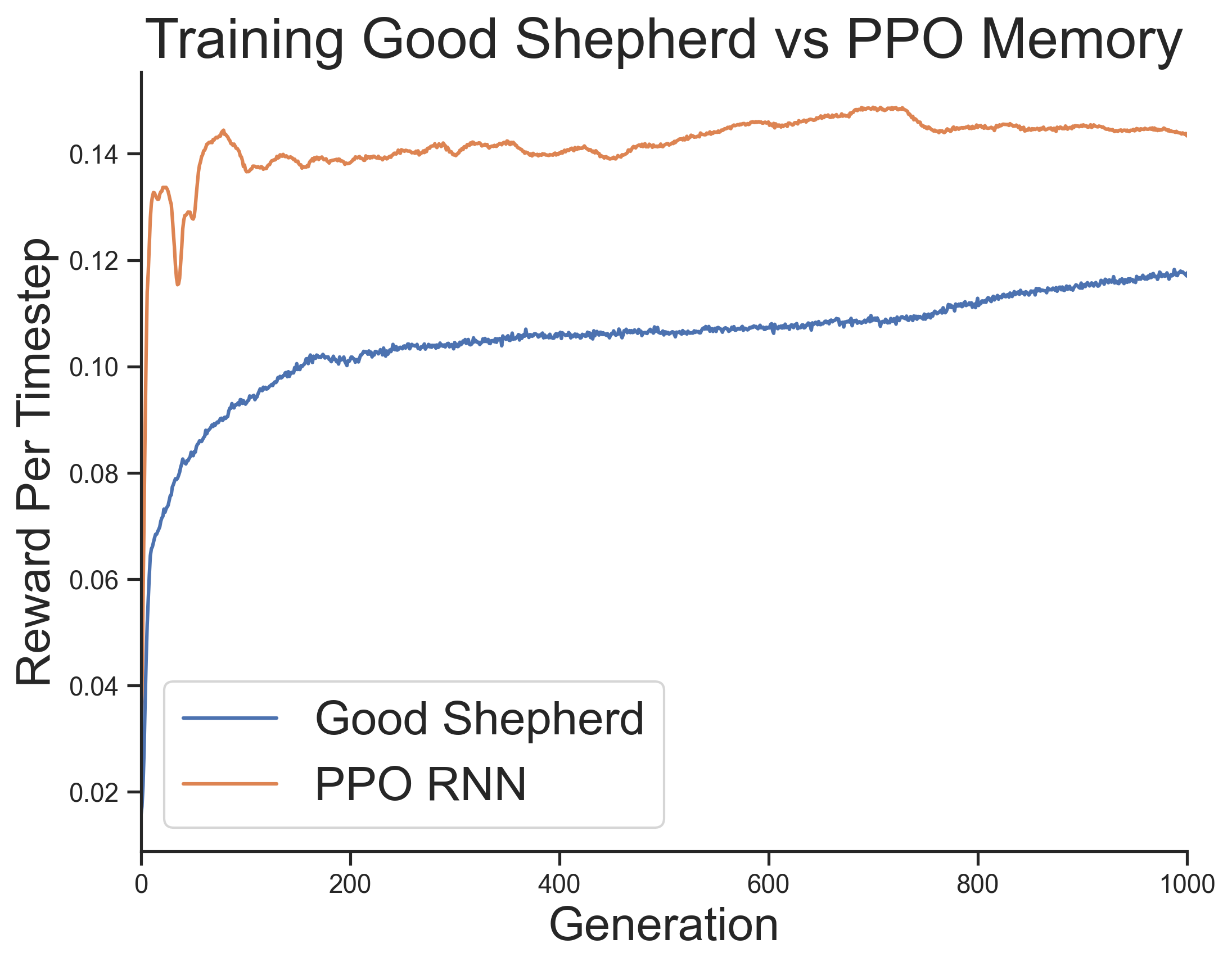}
      \label{fig:gs_ipditm_train}
    \end{subfigure}
    \begin{subfigure}[b]{0.24\textwidth}
      \centering
      \includegraphics[width=\textwidth]{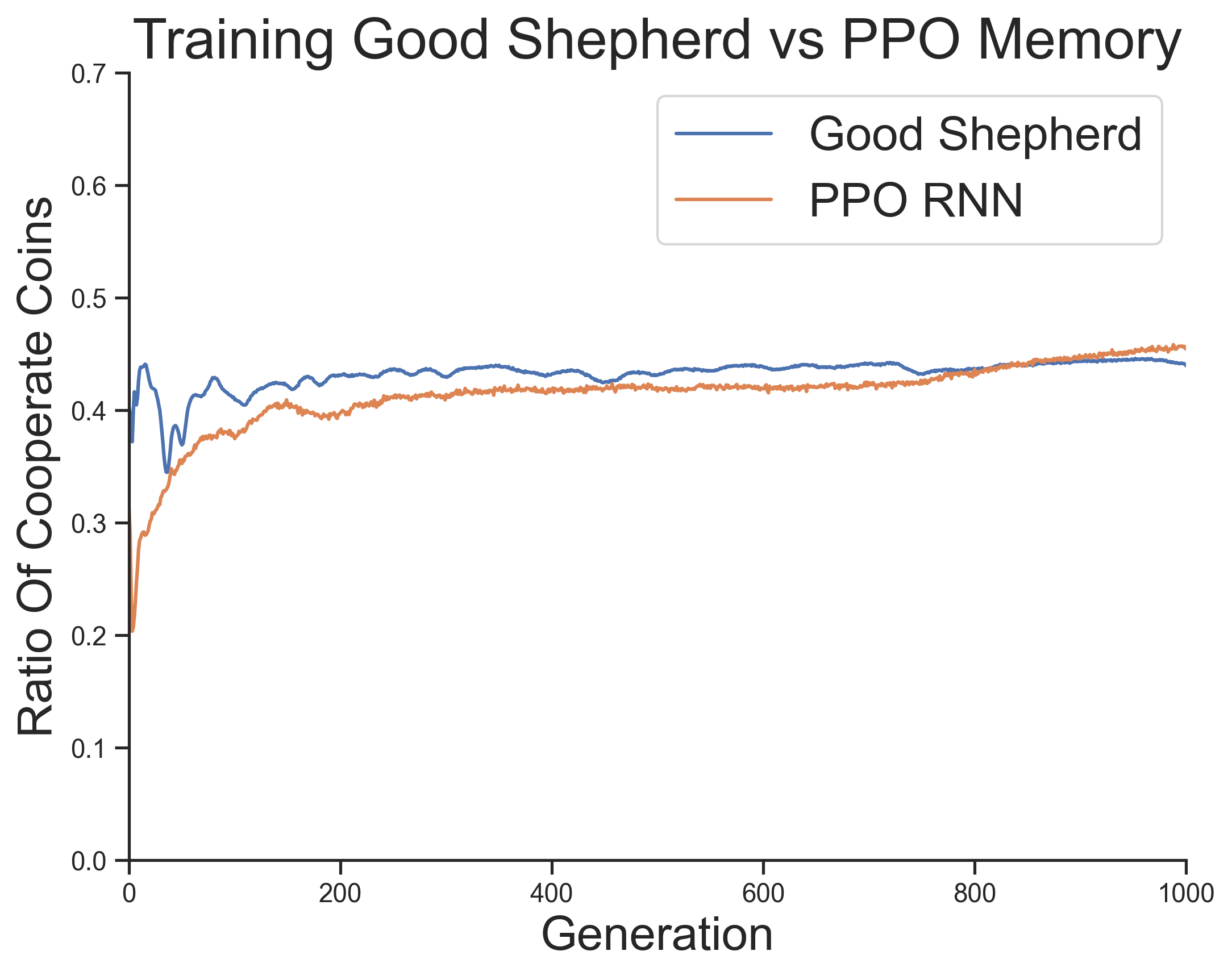}
      \label{fig:gs_ipditm_ratio}
    \end{subfigure}
    \begin{subfigure}[b]{0.24\textwidth}
      \centering
      \includegraphics[width=\textwidth]{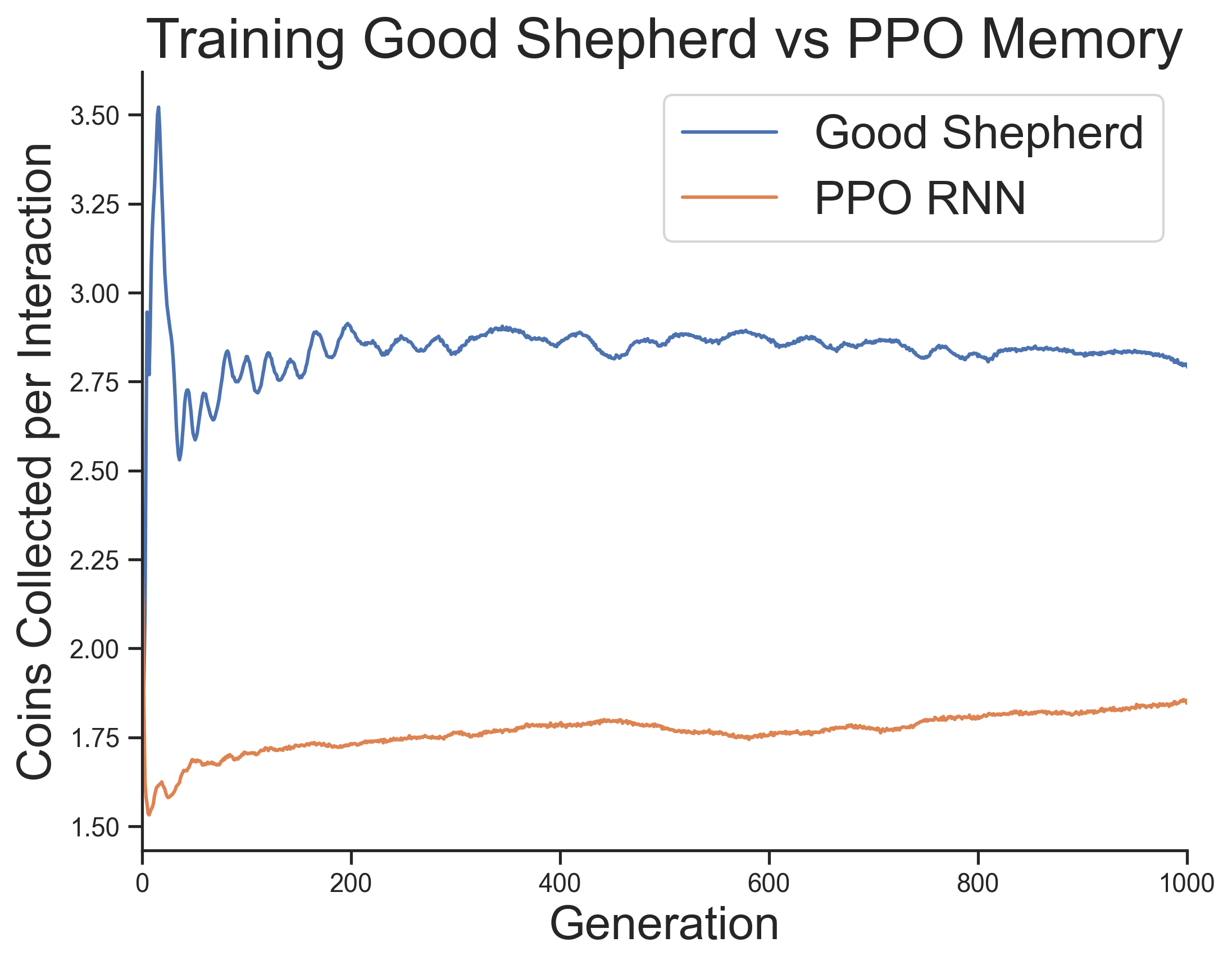}
      \label{fig:gs_ipditm_coins}
    \end{subfigure}
    \begin{subfigure}[b]{0.24\textwidth}
      \centering
      \includegraphics[width=\textwidth]{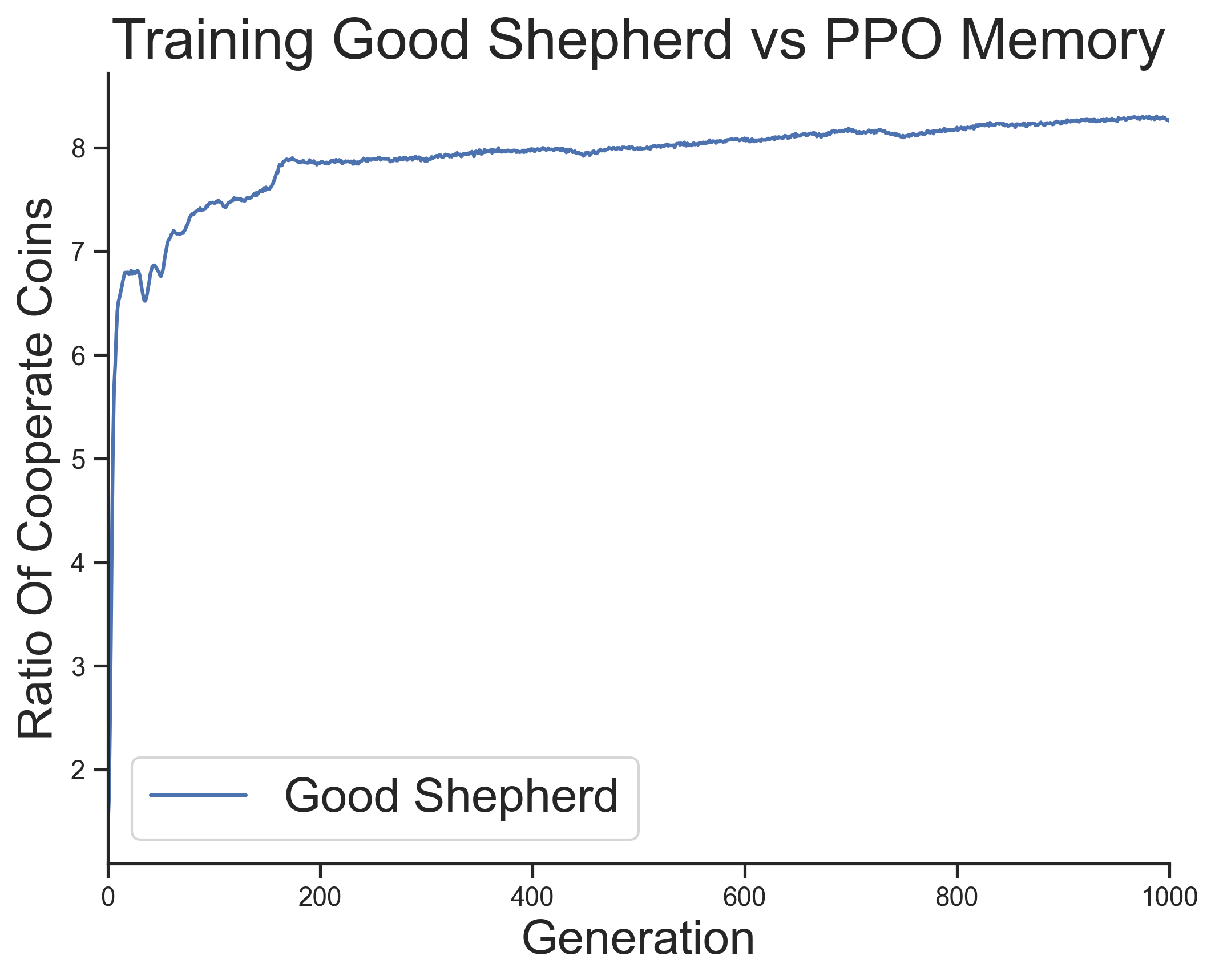}
      \label{fig:gs_ipditm_reset}
    \end{subfigure}
    \begin{subfigure}[b]{0.24\textwidth}
      \centering
      \includegraphics[width=\textwidth]{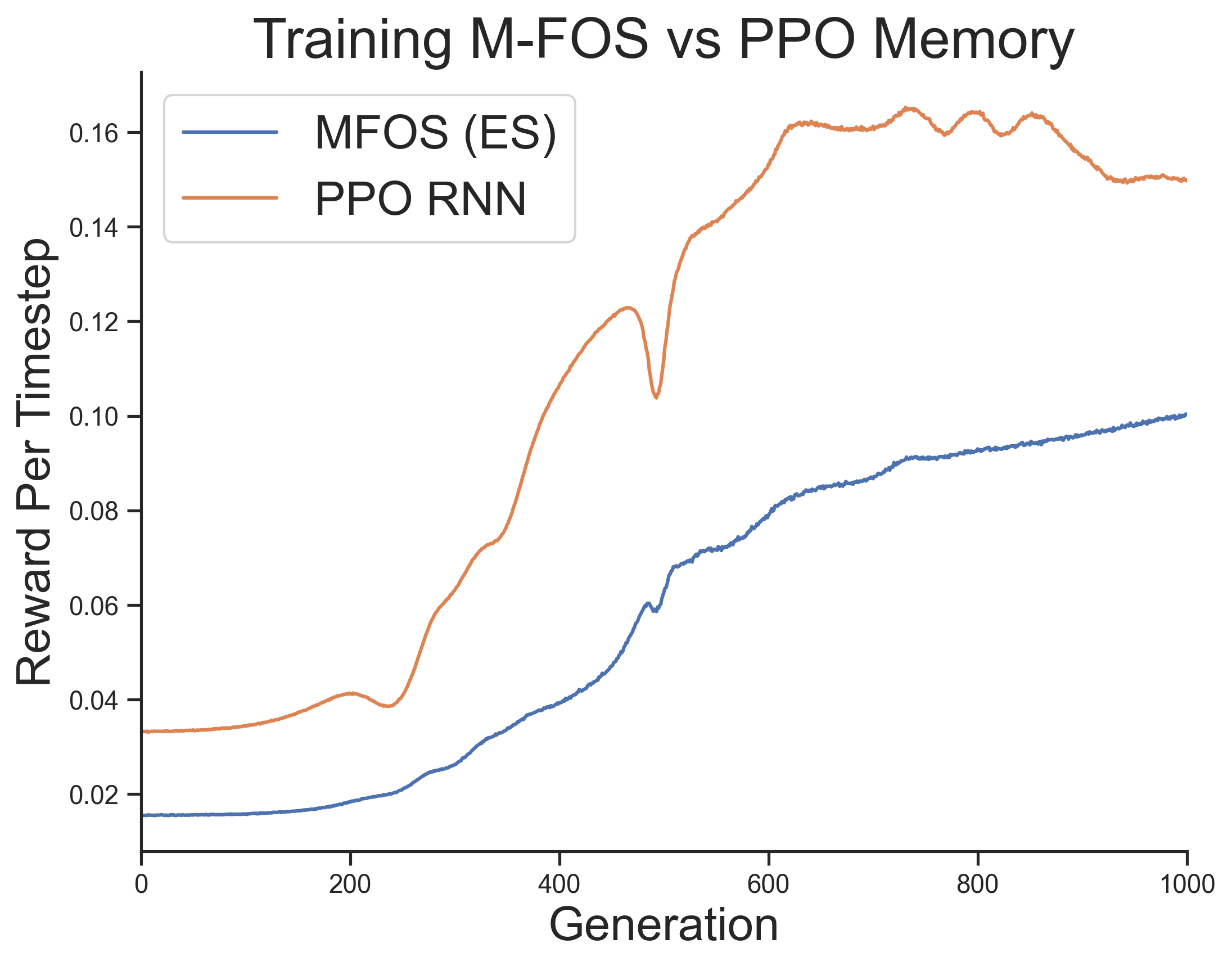}
      \label{fig:mfos_es_ipditm_train}
    \end{subfigure}
    \begin{subfigure}[b]{0.24\textwidth}
      \centering
      \includegraphics[width=\textwidth]{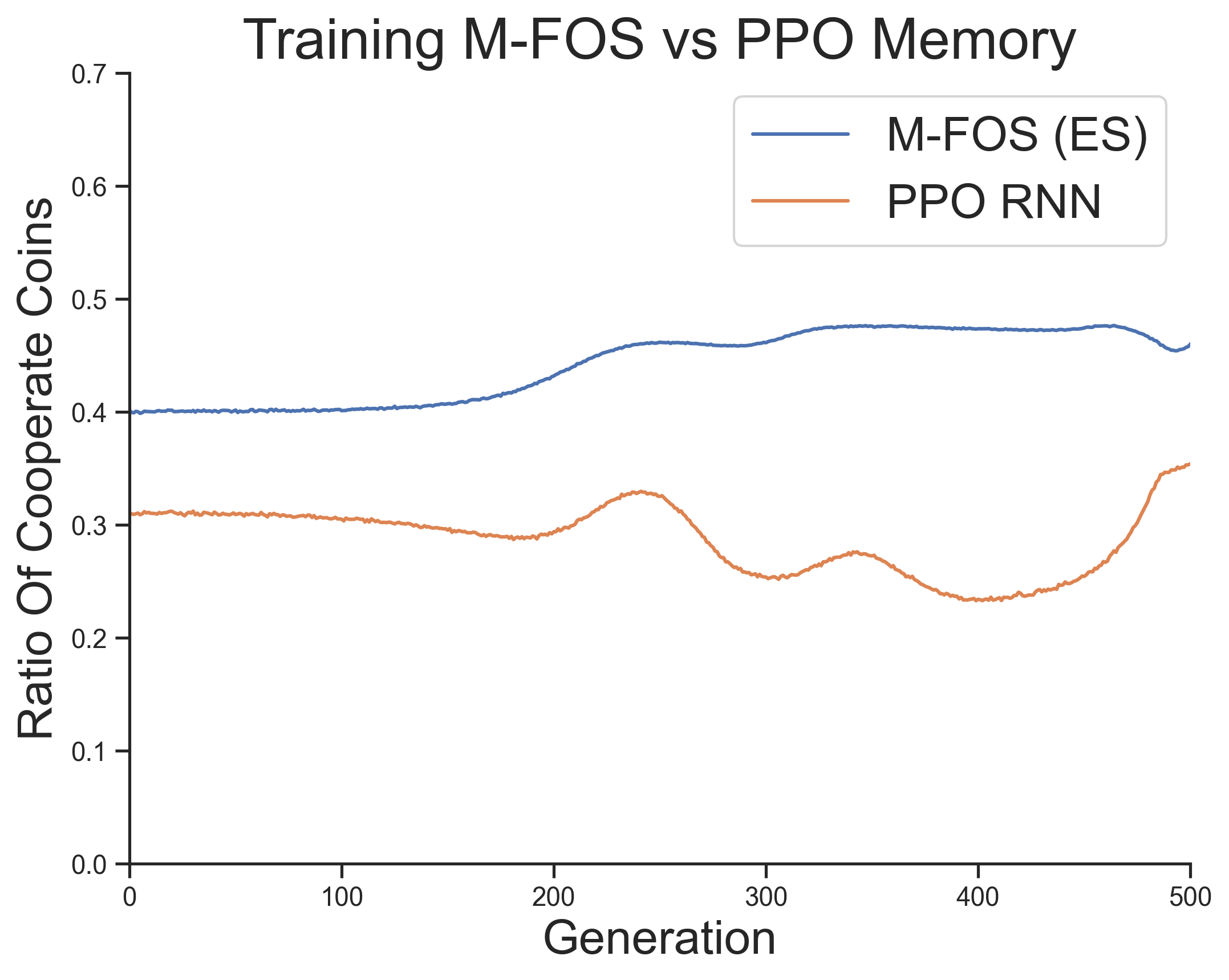}
      \label{fig:mfos_ipditm_ratio}
    \end{subfigure}
    \begin{subfigure}[b]{0.24\textwidth}
      \centering
      \includegraphics[width=\textwidth]{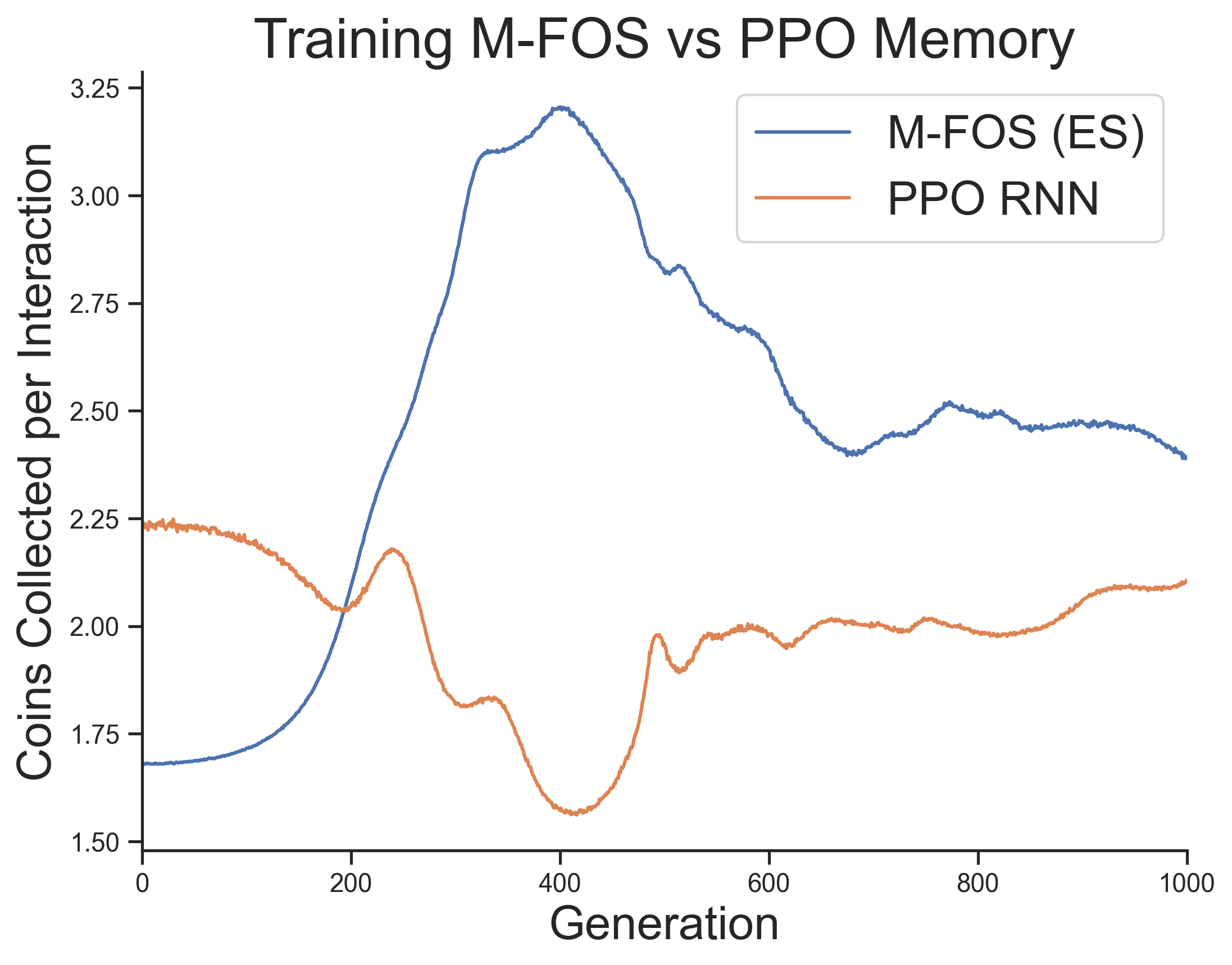}
      \label{fig:mfos_ipditm_coins}
    \end{subfigure}
    \begin{subfigure}[b]{0.24\textwidth}
      \centering
      \includegraphics[width=\textwidth]{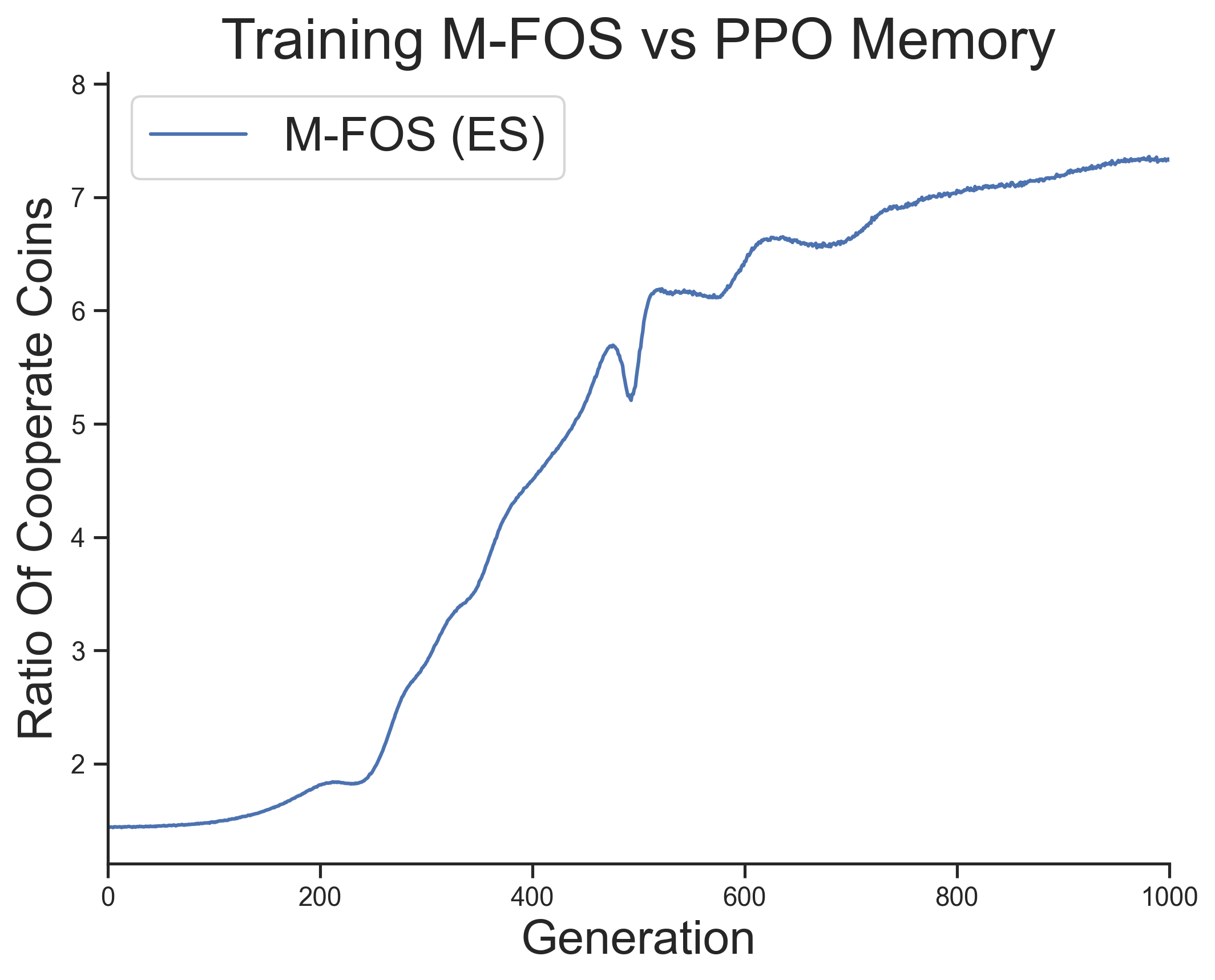}
      \label{fig:mfos_ipditm_reset}
    \end{subfigure}
    \caption{Training results for OS methods vs. PPO RNN in the IPDitM. (Column 1) Reward Per Timestep, (Column 2) the meta-agents's frequency of picking up its own colour coin depending on existing convention, (Column 3) the number of coins picked up per episode, (Column 4) the number of soft resets / successful interactions.}
\end{figure}

\begin{figure}[h]
    \centering
    \begin{subfigure}[b]{0.33\textwidth}
      \centering
      \includegraphics[width=\textwidth]{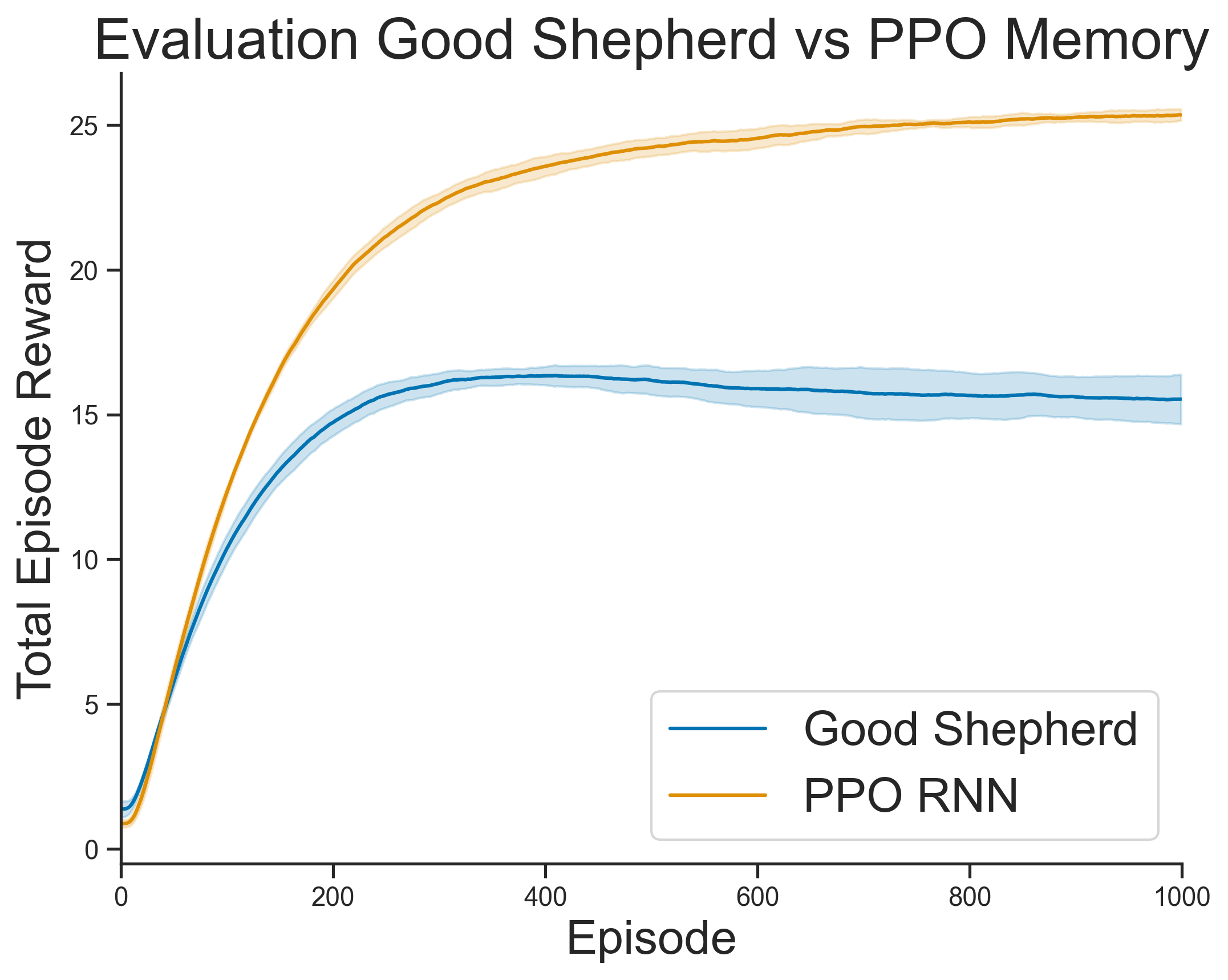}
      \caption{}
      \label{fig:gs_ipditm_eval_reward}
    \end{subfigure}
    \begin{subfigure}[b]{0.33\textwidth}
      \centering
      \includegraphics[width=\textwidth]{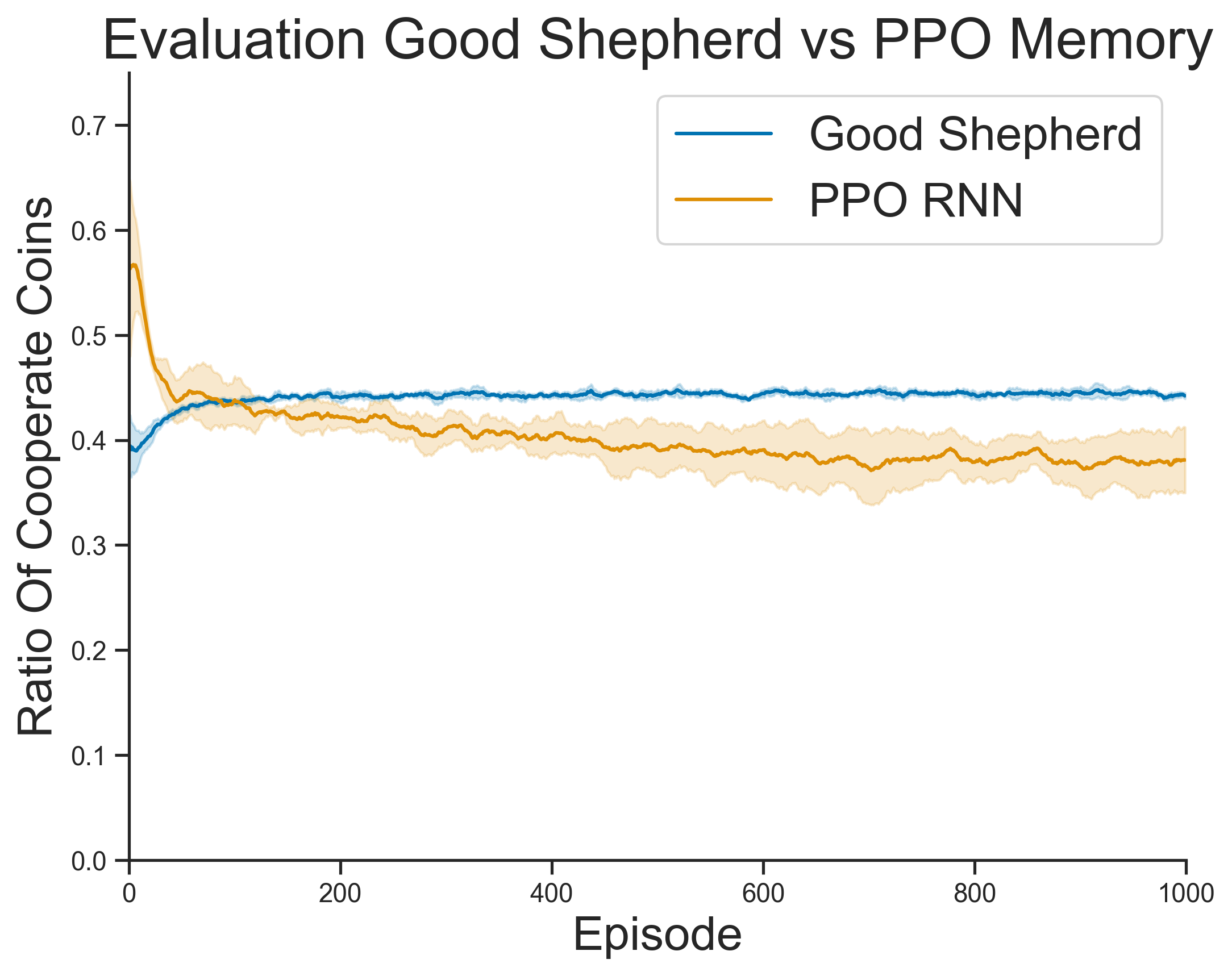}
      \caption{}
      \label{fig:gs_ipditm_eval_ratio}
    \end{subfigure}
    \begin{subfigure}[b]{0.33\textwidth}
      \centering
      \includegraphics[width=\textwidth]{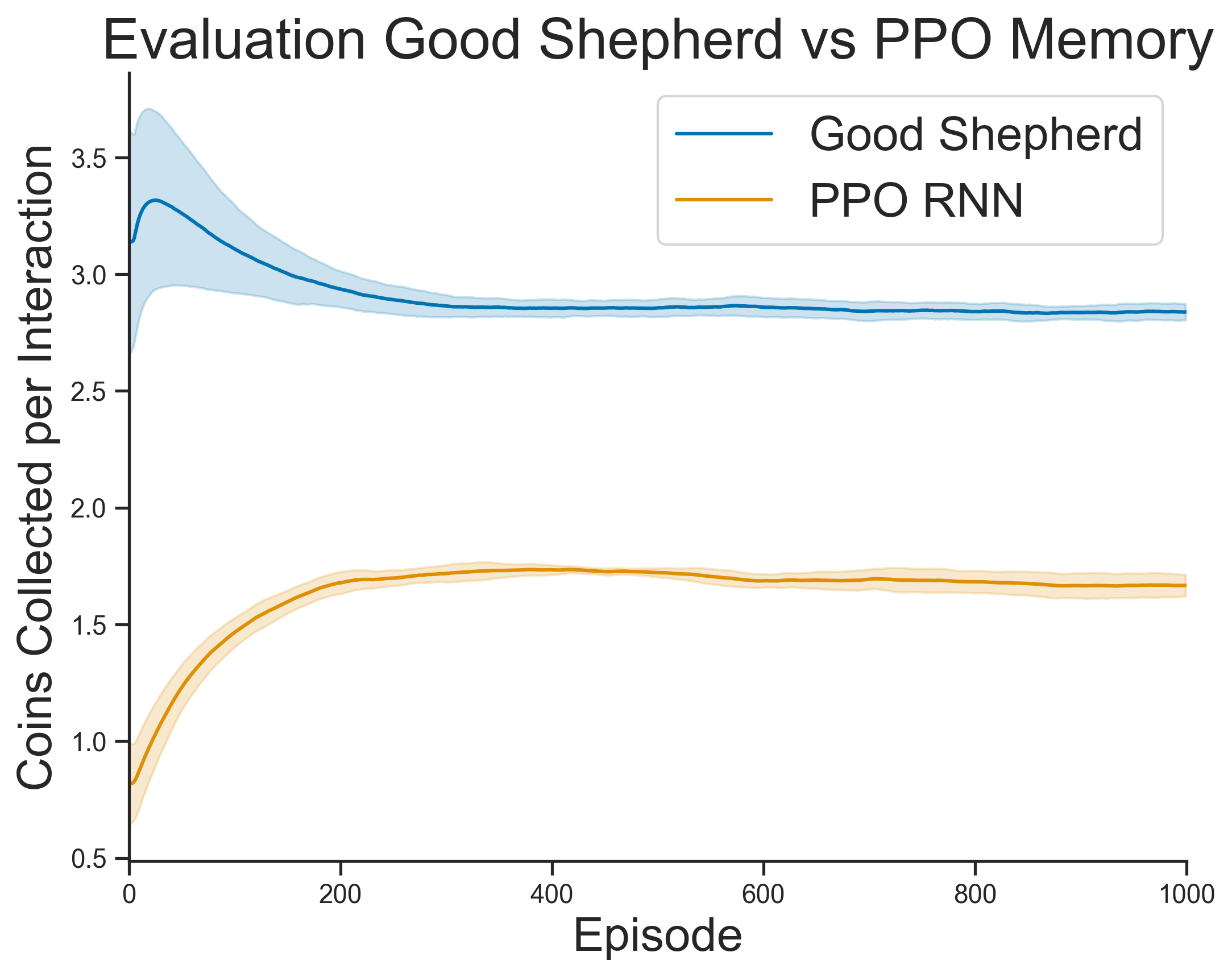}
      \caption{}
      \label{fig:gs_ipditm_eval_coins}
    \end{subfigure}
    \caption{Evaluation results over a single trial (with new co-player) compromising over five seeds for the IPDitM. (a) Mean reward per timestep, (b) mean ratio of picking up cooperate coins per soft-reset, (c) total number of coins picked up per soft-reset.
    }
    \label{fig:gs_ego_ipditm_viz}
\end{figure}

\begin{figure}[h]
    \centering
    \begin{subfigure}[b]{0.3\textwidth}
      \centering
      \includegraphics[width=\textwidth]{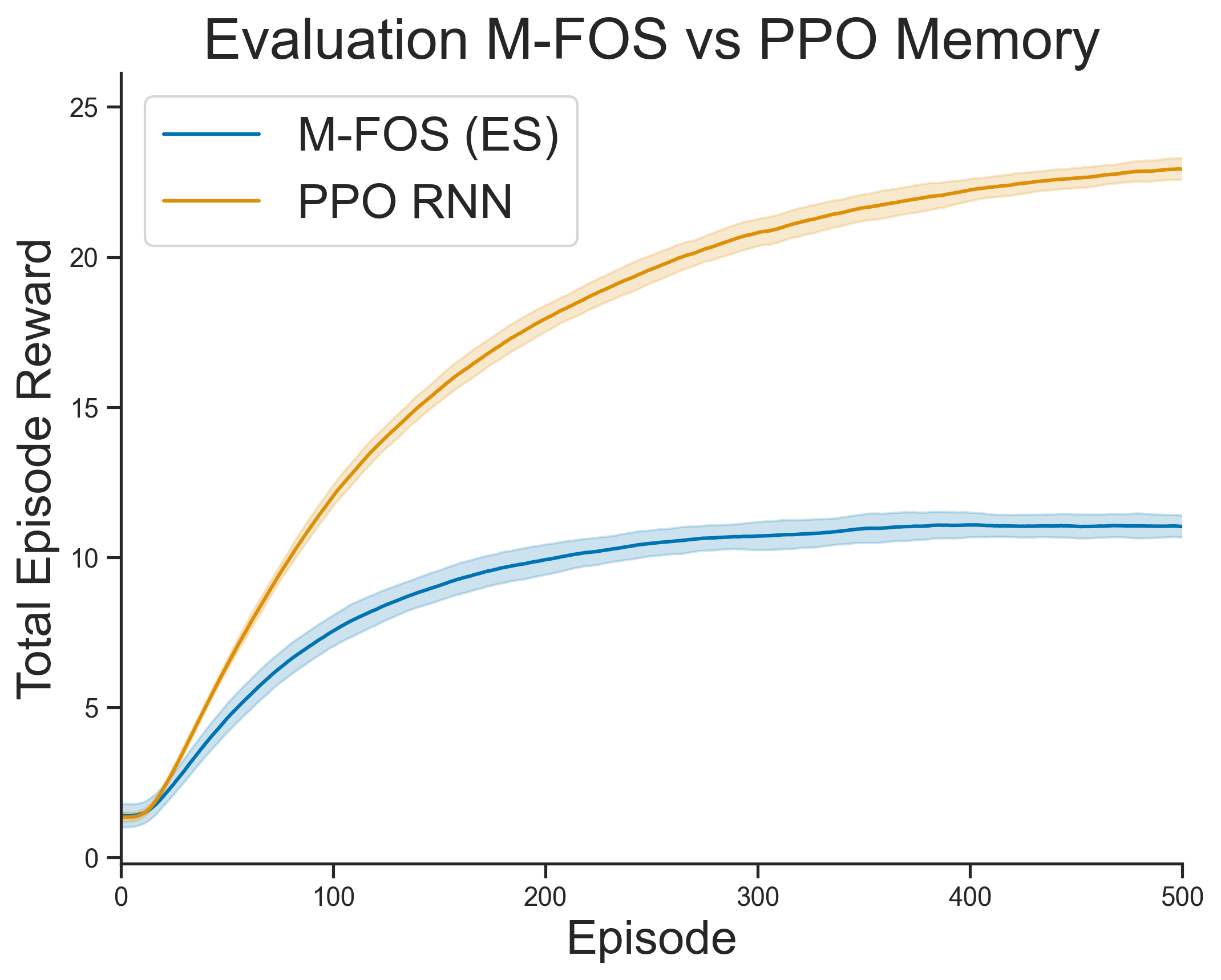}
      \caption{}
      \label{fig:mfos_ipditm_eval_reward}
    \end{subfigure}
    \begin{subfigure}[b]{0.3\textwidth}
      \centering
      \includegraphics[width=\textwidth]{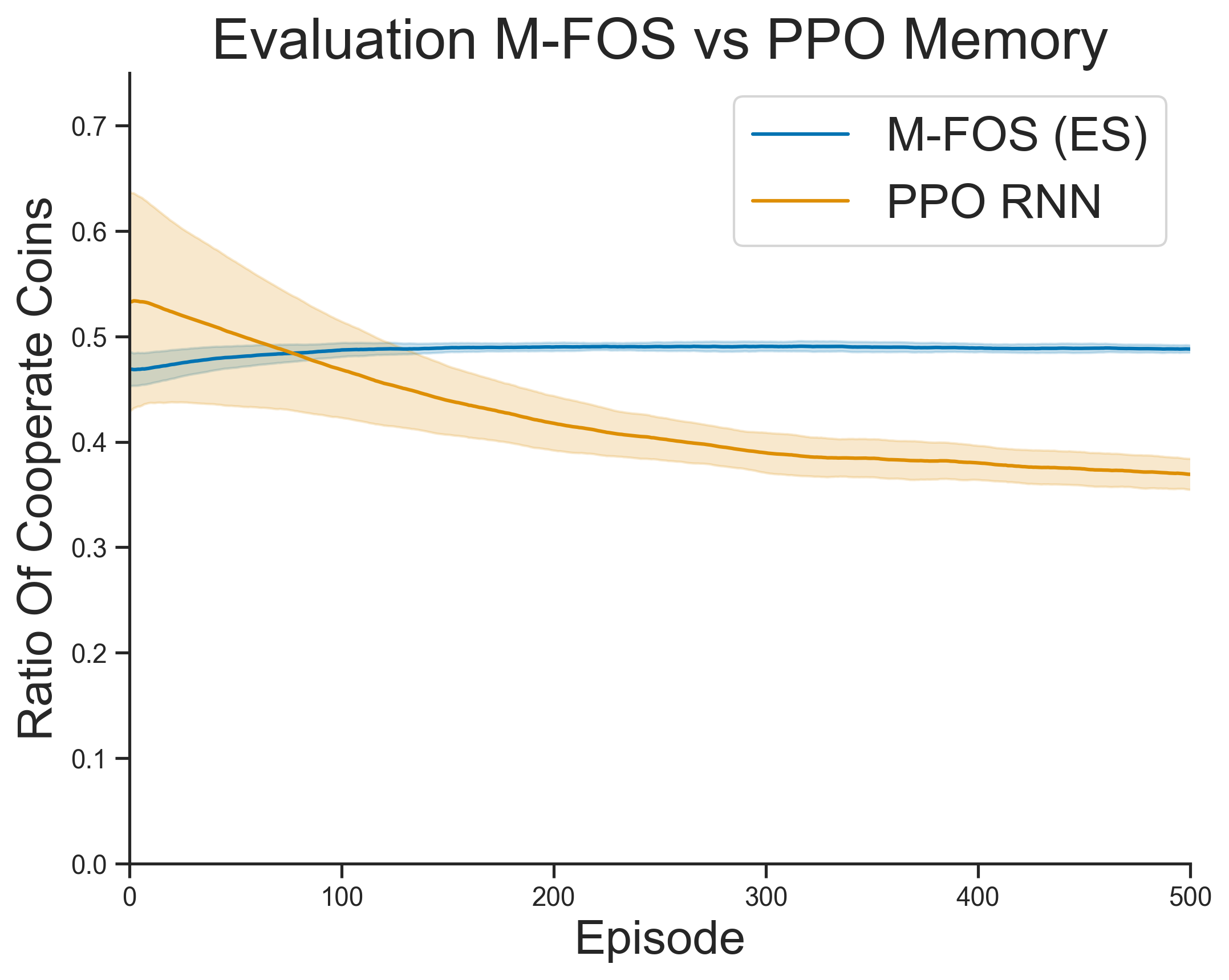}
      \caption{}
      \label{fig:mfos_ipditm_eval_ratio}
    \end{subfigure}
    \begin{subfigure}[b]{0.3\textwidth}
      \centering
      \includegraphics[width=\textwidth]{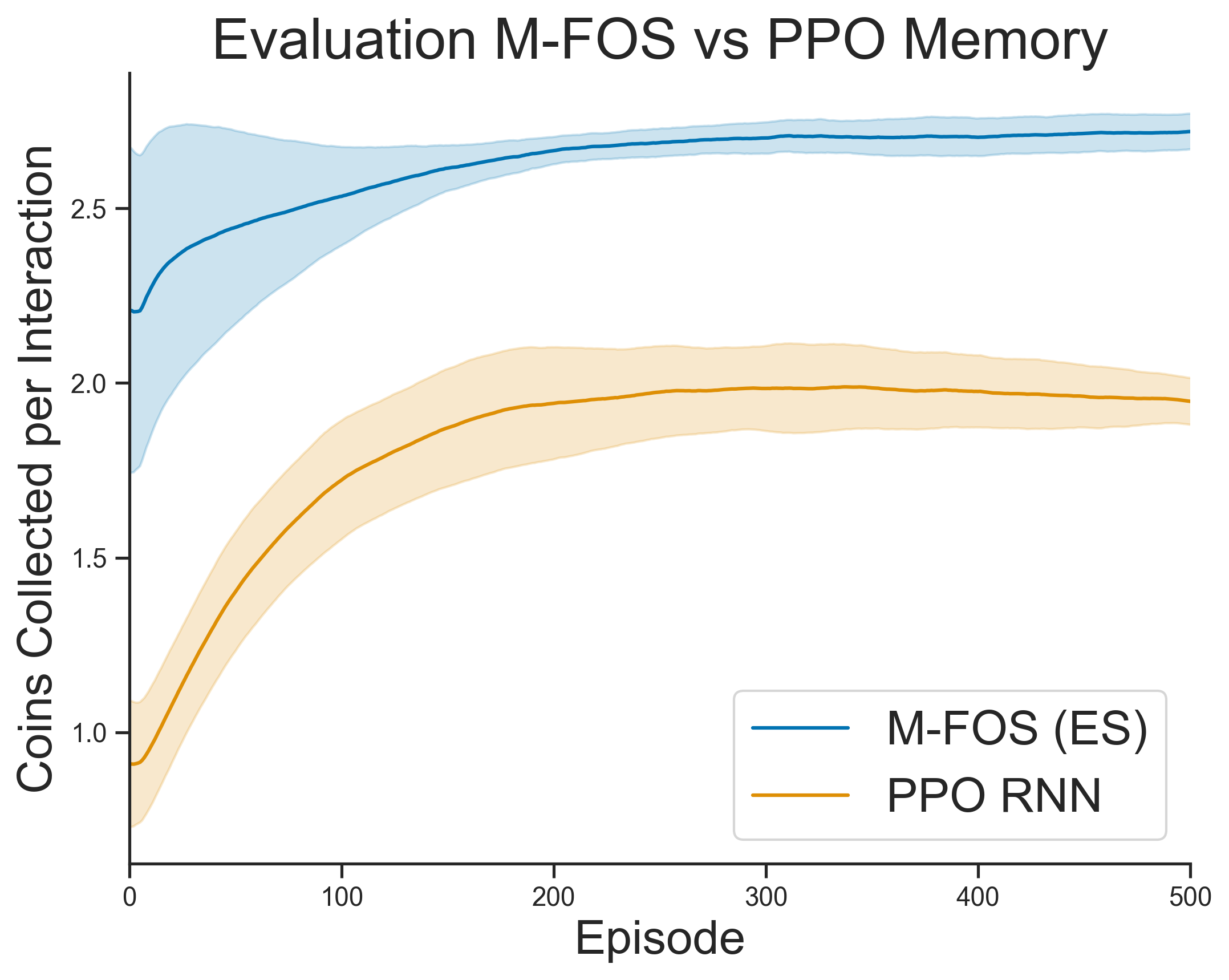}
      \caption{}
      \label{fig:mfos_es_ipditm_eval_coins}
    \end{subfigure}
    \caption{Evaluation results over a single trial (with new co-player) compromising over five seeds for the IPDitM. (a) Mean reward per timestep, (b) mean ratio of picking up cooperate coins per soft-reset, (c) total number of coins picked up per soft-reset.
    }
    \label{fig:mfos_ego_ipditm_viz}
\end{figure}

\begin{figure}[h]
    \centering
    \begin{subfigure}[b]{0.3\textwidth}
      \centering
      \includegraphics[width=\textwidth]{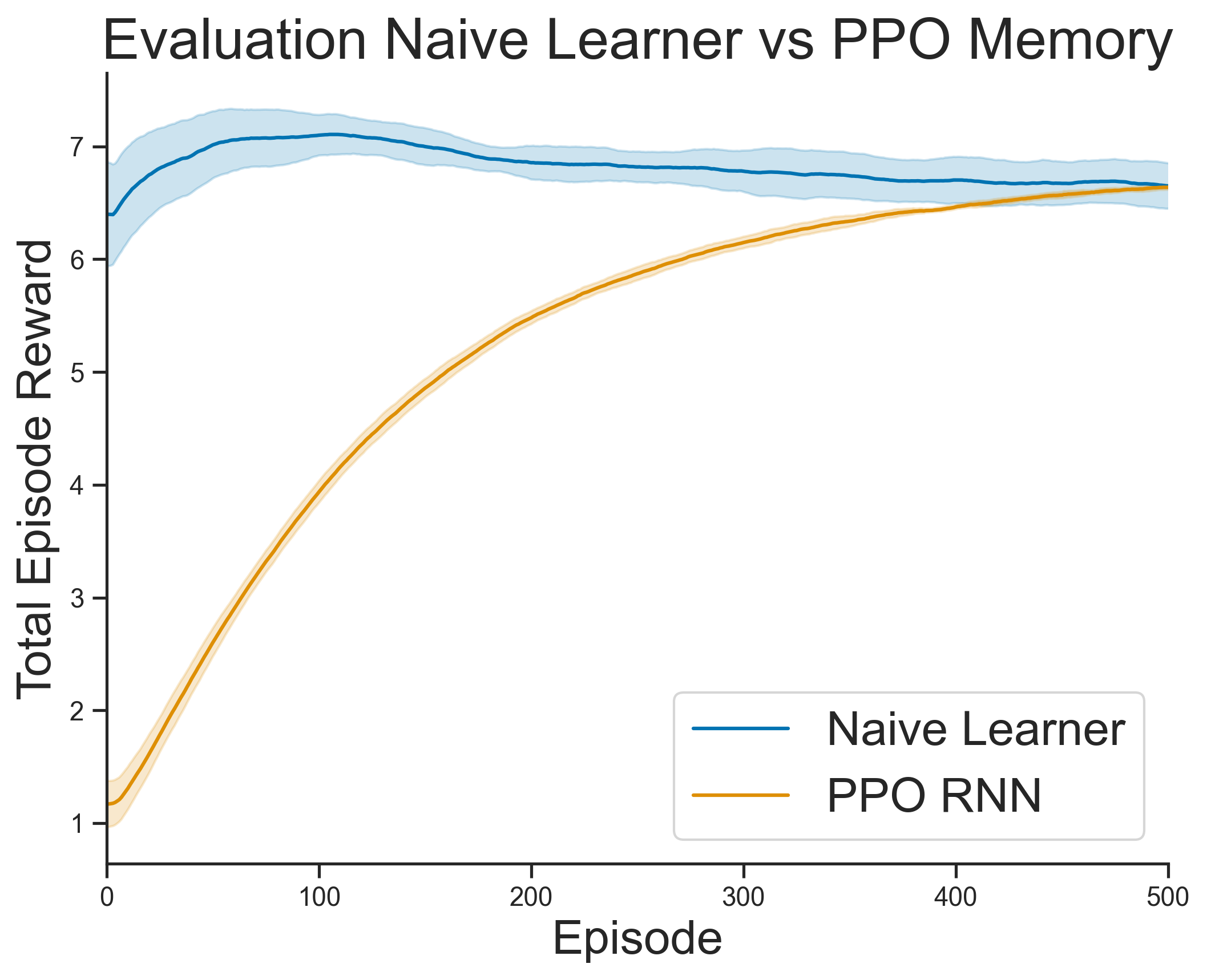}
      \caption{}
      \label{fig:nl_ipditm_eval_reward}
    \end{subfigure}
    \begin{subfigure}[b]{0.3\textwidth}
      \centering
      \includegraphics[width=\textwidth]{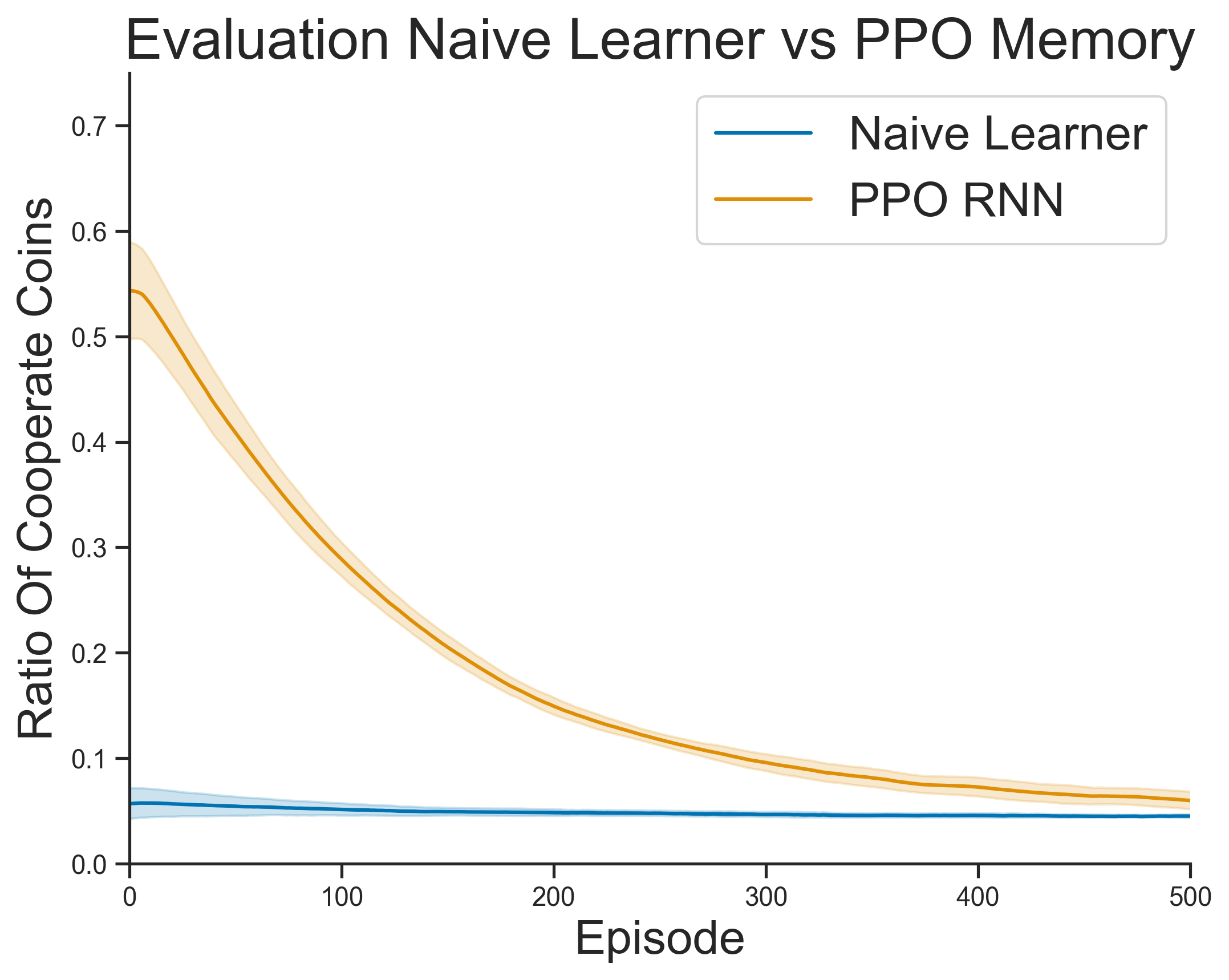}
      \caption{}
      \label{fig:nl_ipditm_eval_ratio}
    \end{subfigure}
    \begin{subfigure}[b]{0.3\textwidth}
      \centering
      \includegraphics[width=\textwidth]{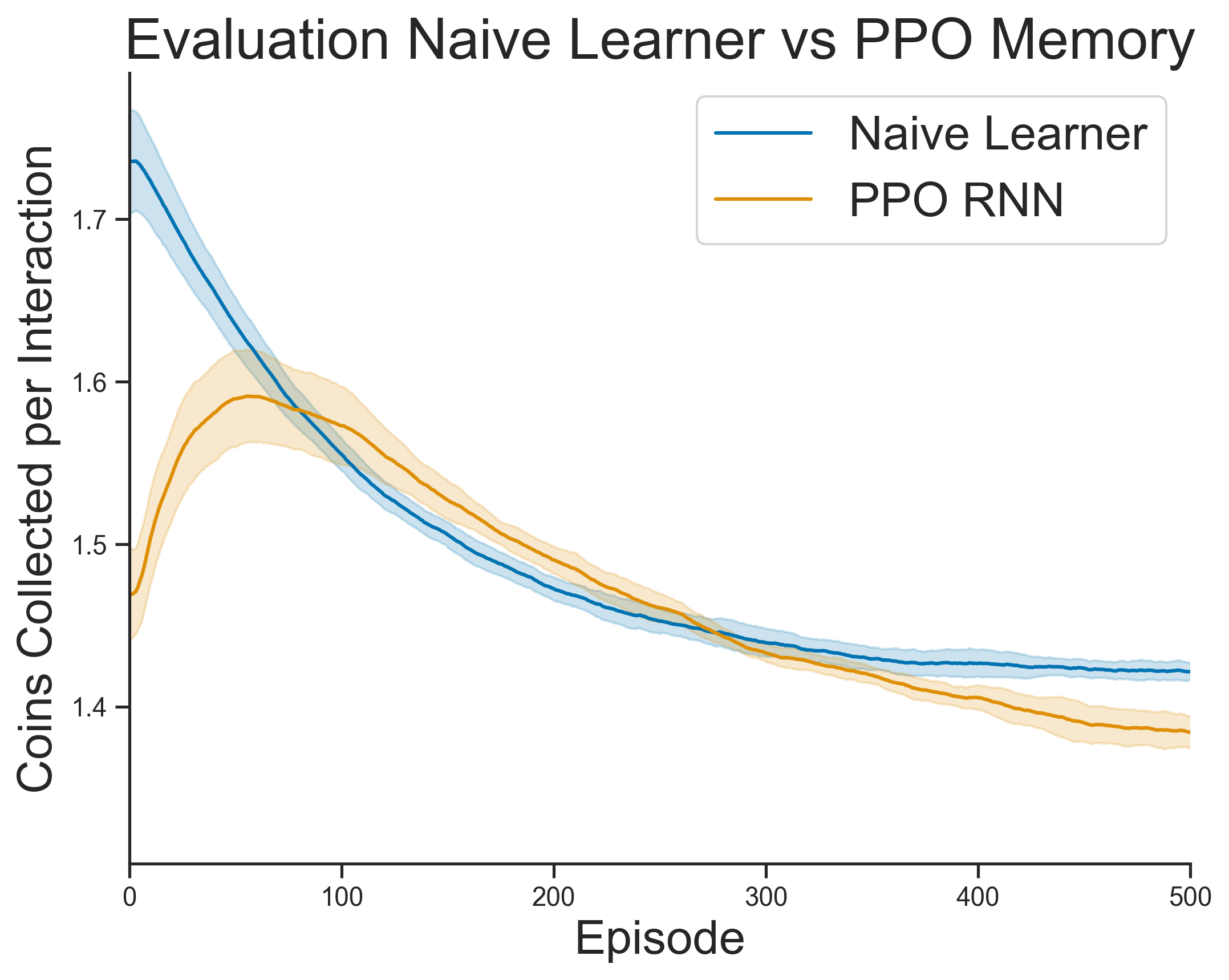}
      \caption{}
      \label{fig:nl_ipditm_eval_coins}
    \end{subfigure}
    \caption{Evaluation results over a single trial (with new co-player) compromising over five seeds for the IPDitM. (a) Mean reward per timestep, (b) mean ratio of picking up cooperate coins per soft-reset, (c) total number of coins picked up per soft-reset.
    }
    \label{fig:nl_ego_ipditm_viz}
\end{figure}
\clearpage
\section{IMP in the Matrix Details}

\begin{figure}[h]
    \centering
    \begin{subfigure}[b]{0.24\textwidth}
      \centering
      \includegraphics[width=\textwidth]{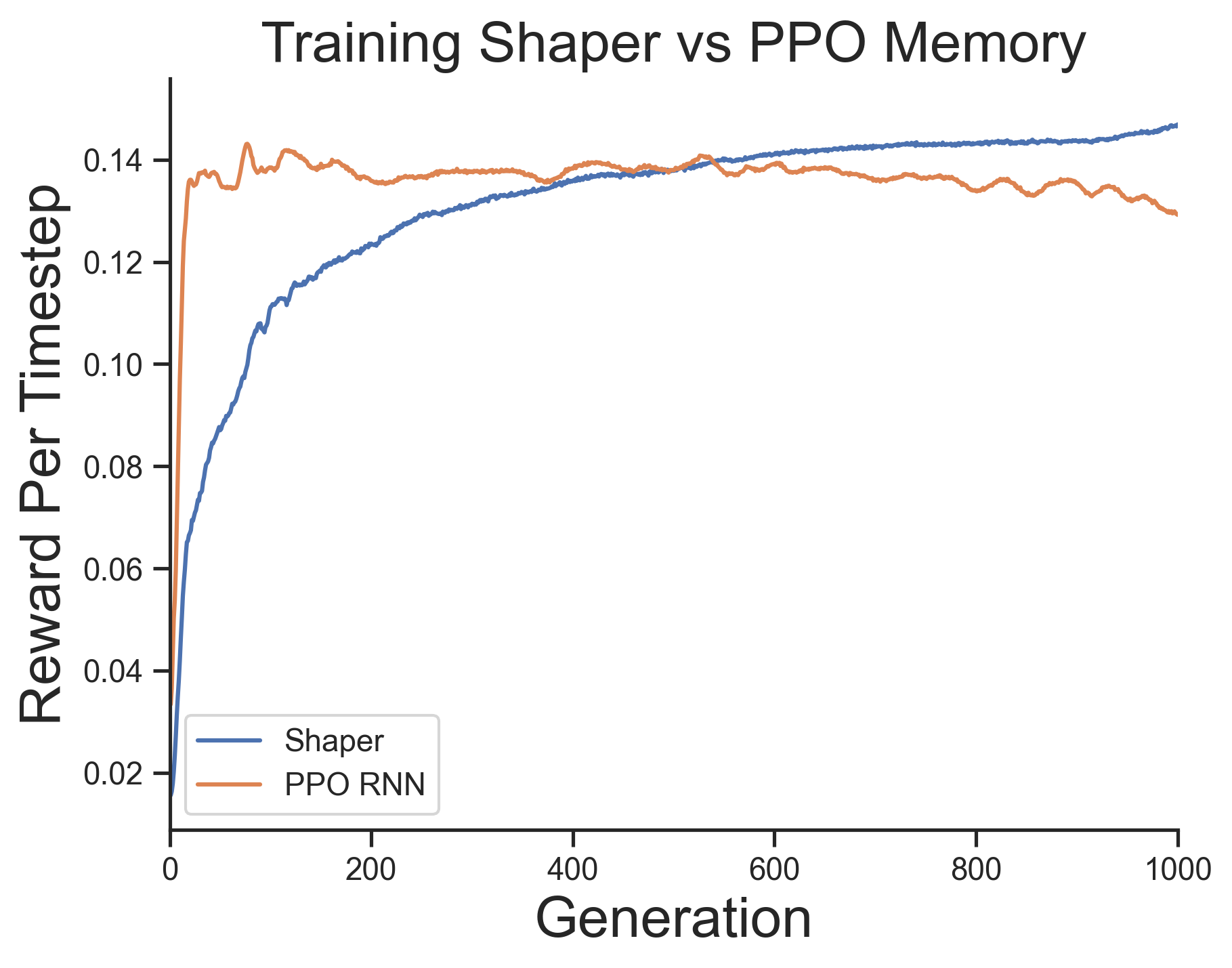}
      \label{fig:chaos_impitm_train}
    \end{subfigure}
    \begin{subfigure}[b]{0.24\textwidth}
      \centering
      \includegraphics[width=\textwidth]{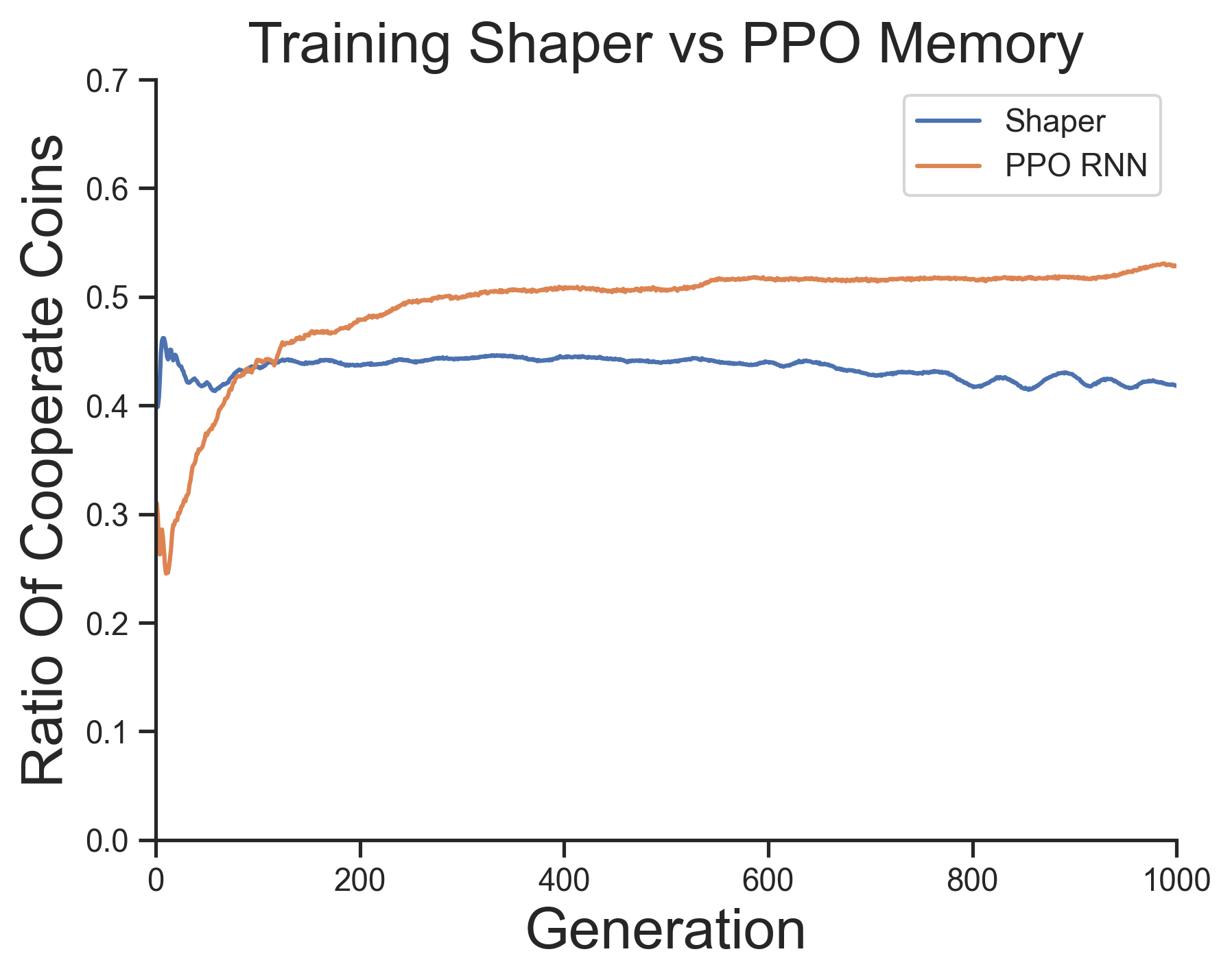}
      \label{fig:chaos_impitm_ratio}
    \end{subfigure}
    \begin{subfigure}[b]{0.24\textwidth}
      \centering
      \includegraphics[width=\textwidth]{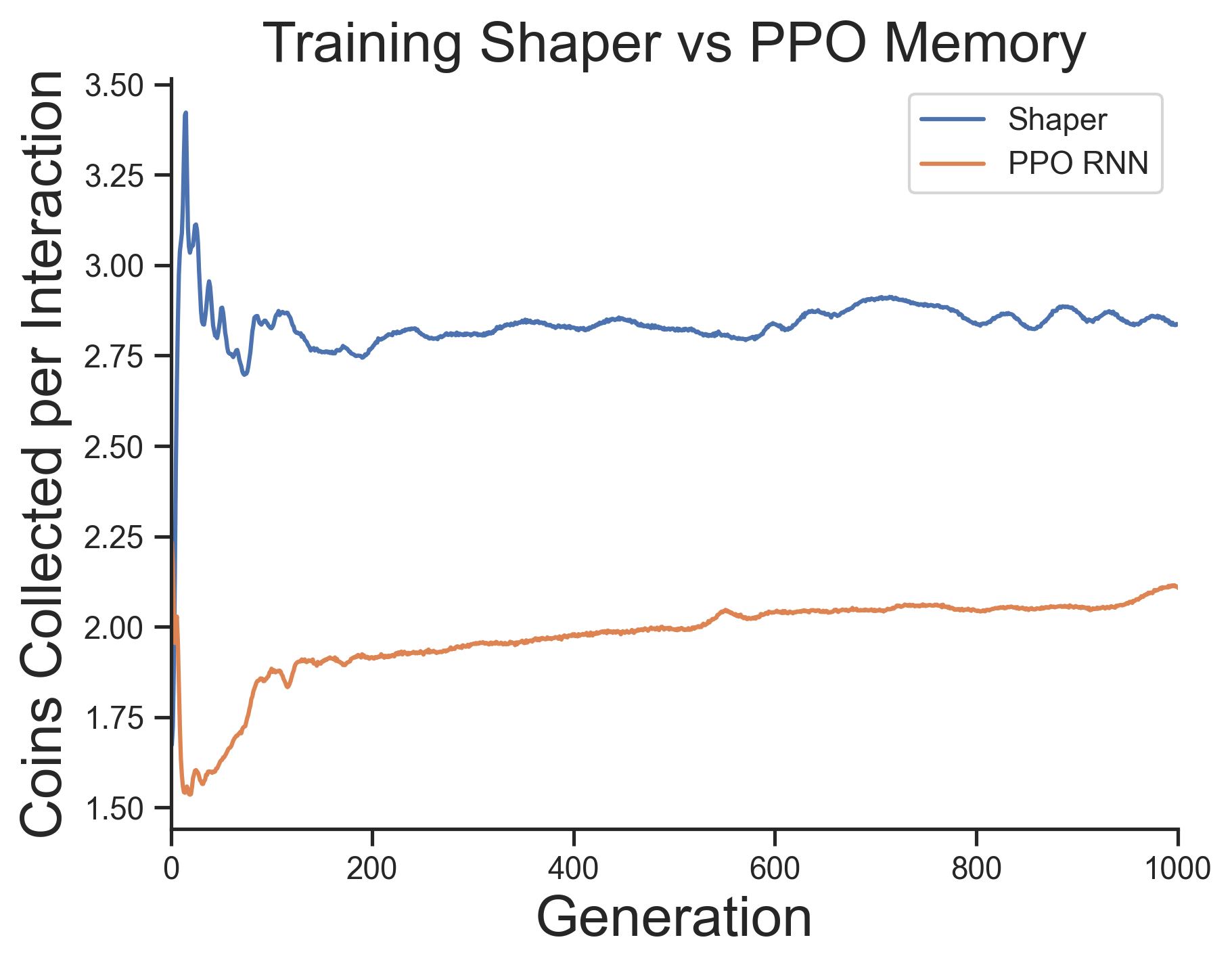}
      \label{fig:chaos_impitm_coins}
    \end{subfigure}
    \begin{subfigure}[b]{0.24\textwidth}
      \centering
      \includegraphics[width=\textwidth]{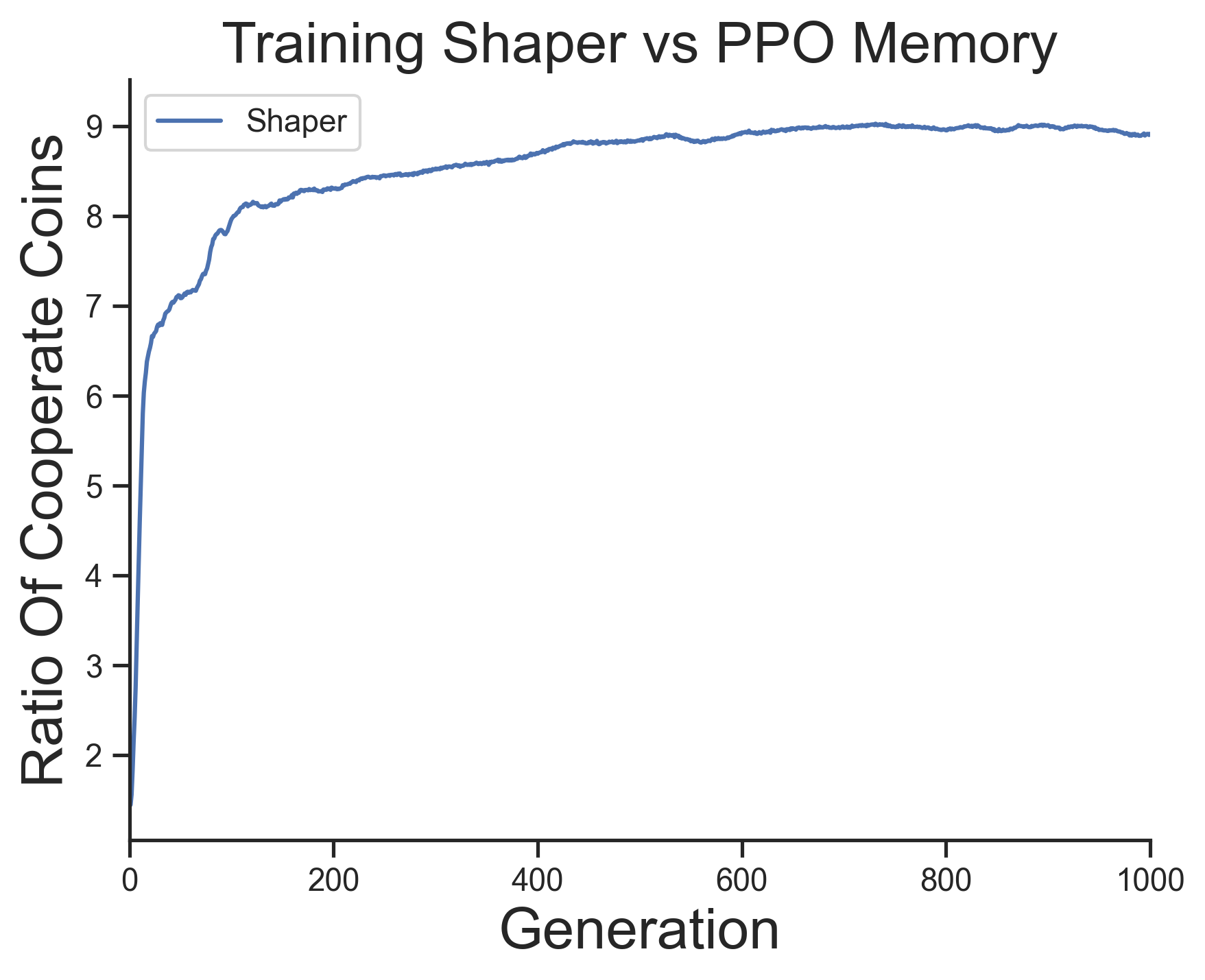}
      \label{fig:chaos_impitm_reset}
    \end{subfigure}
    \begin{subfigure}[b]{0.24\textwidth}
      \centering
      \includegraphics[width=\textwidth]{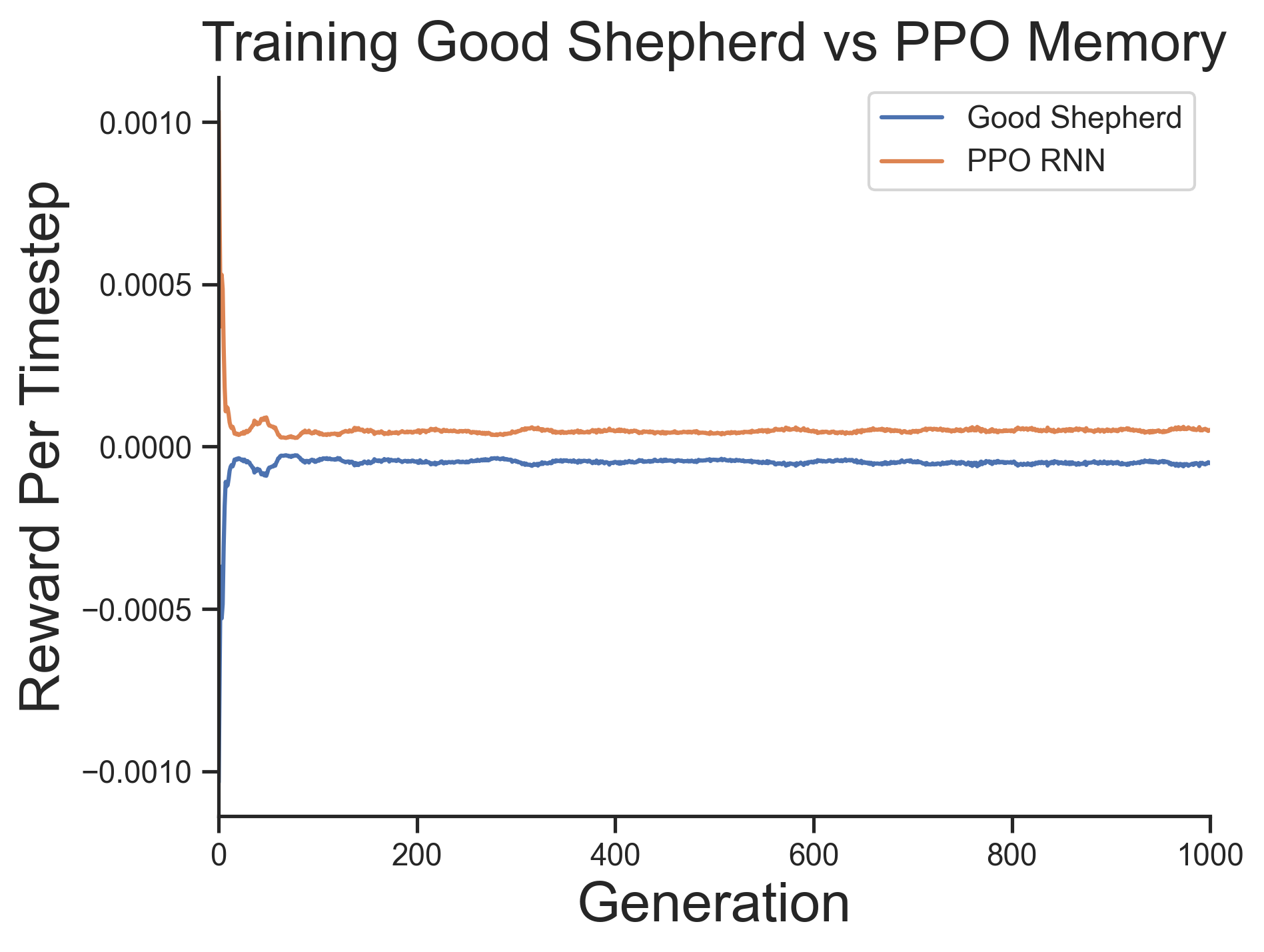}
      \label{fig:gs_impitm_train}
    \end{subfigure}
    \begin{subfigure}[b]{0.24\textwidth}
      \centering
      \includegraphics[width=\textwidth]{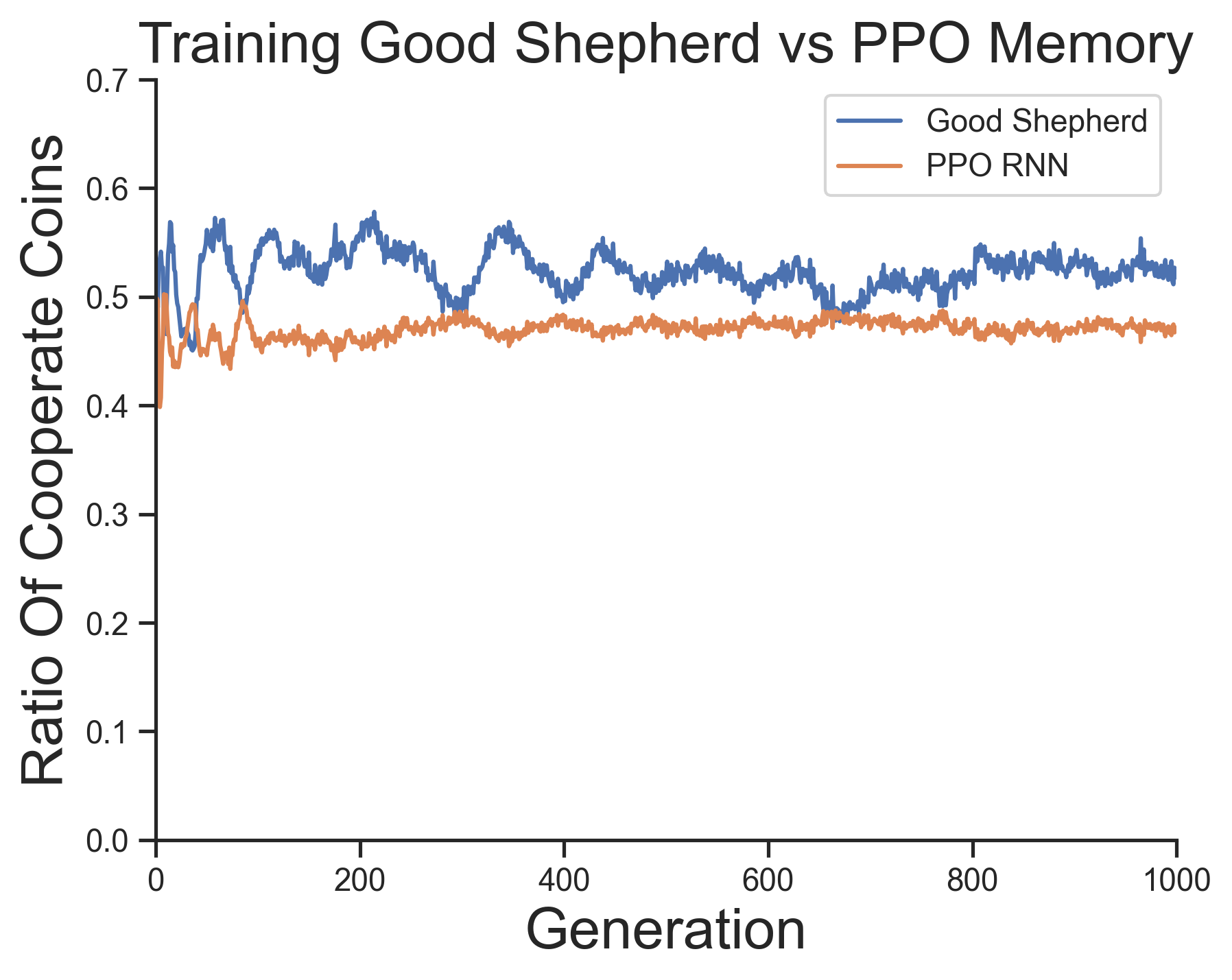}
      \label{fig:gs_impitm_ratio}
    \end{subfigure}
    \begin{subfigure}[b]{0.24\textwidth}
      \centering
      \includegraphics[width=\textwidth]{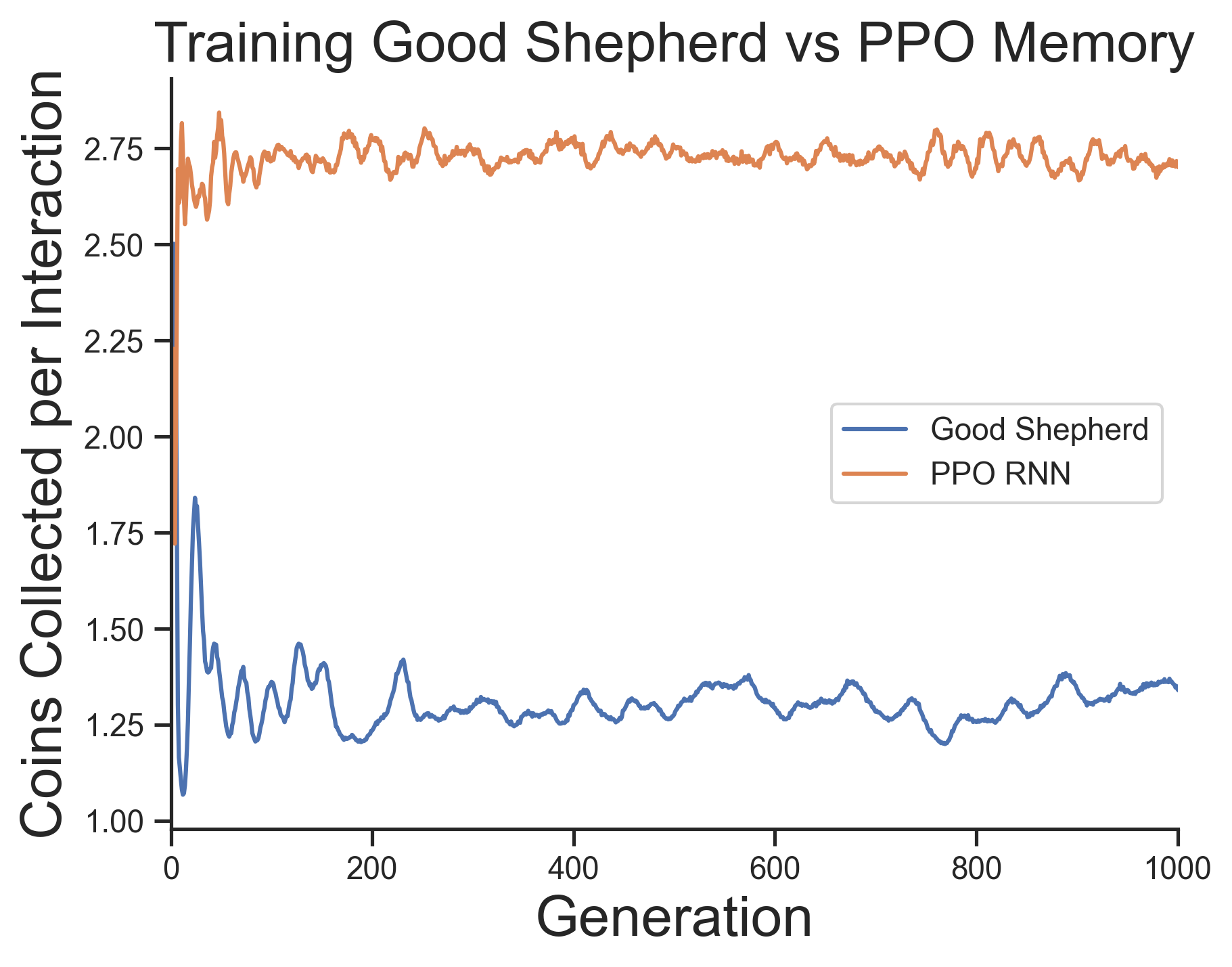}
      \label{fig:gs_impitm_coins}
    \end{subfigure}
    \begin{subfigure}[b]{0.24\textwidth}
      \centering
      \includegraphics[width=\textwidth]{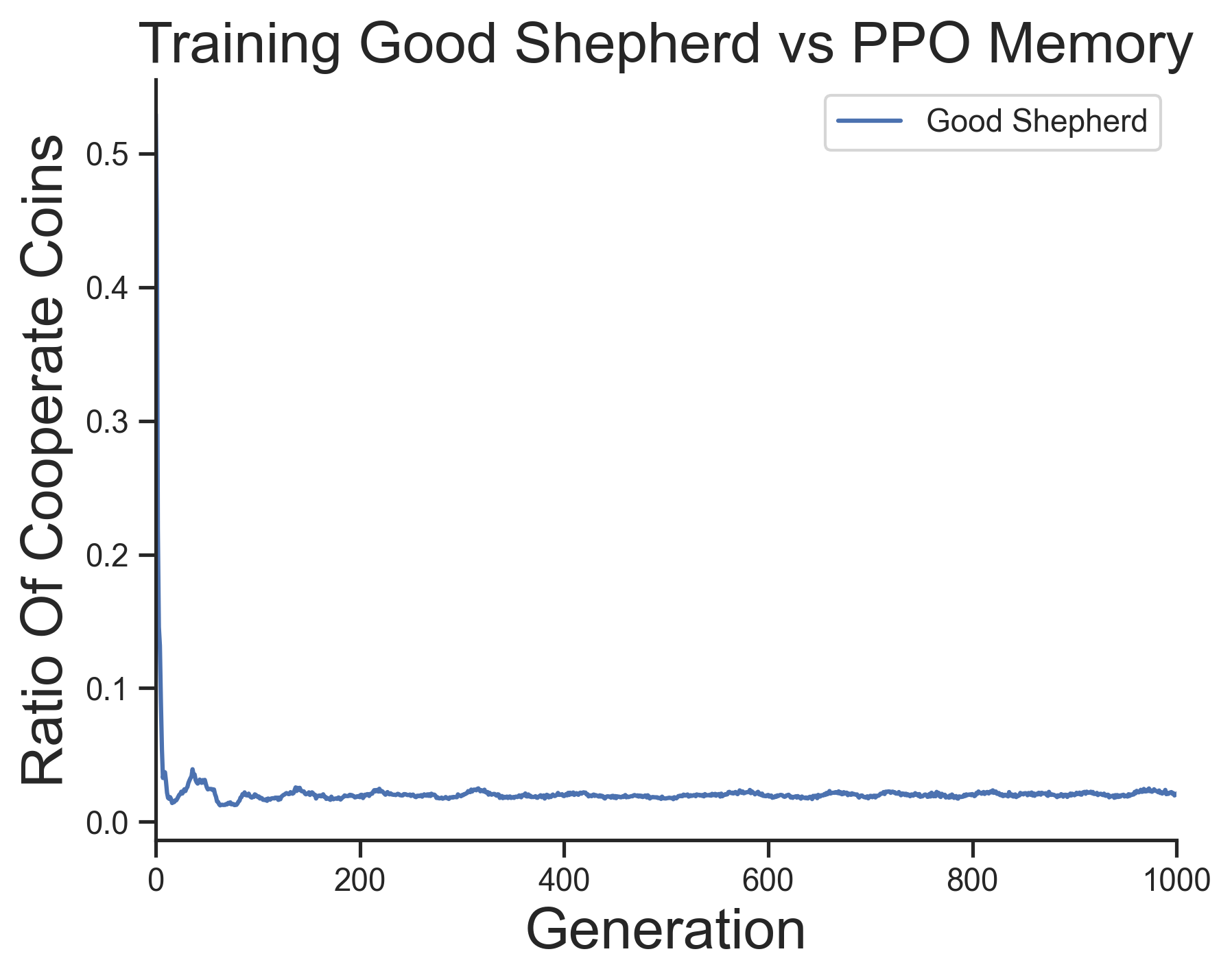}
      \label{fig:gs_impitm_reset}
    \end{subfigure}
    \begin{subfigure}[b]{0.24\textwidth}
      \centering
      \includegraphics[width=\textwidth]{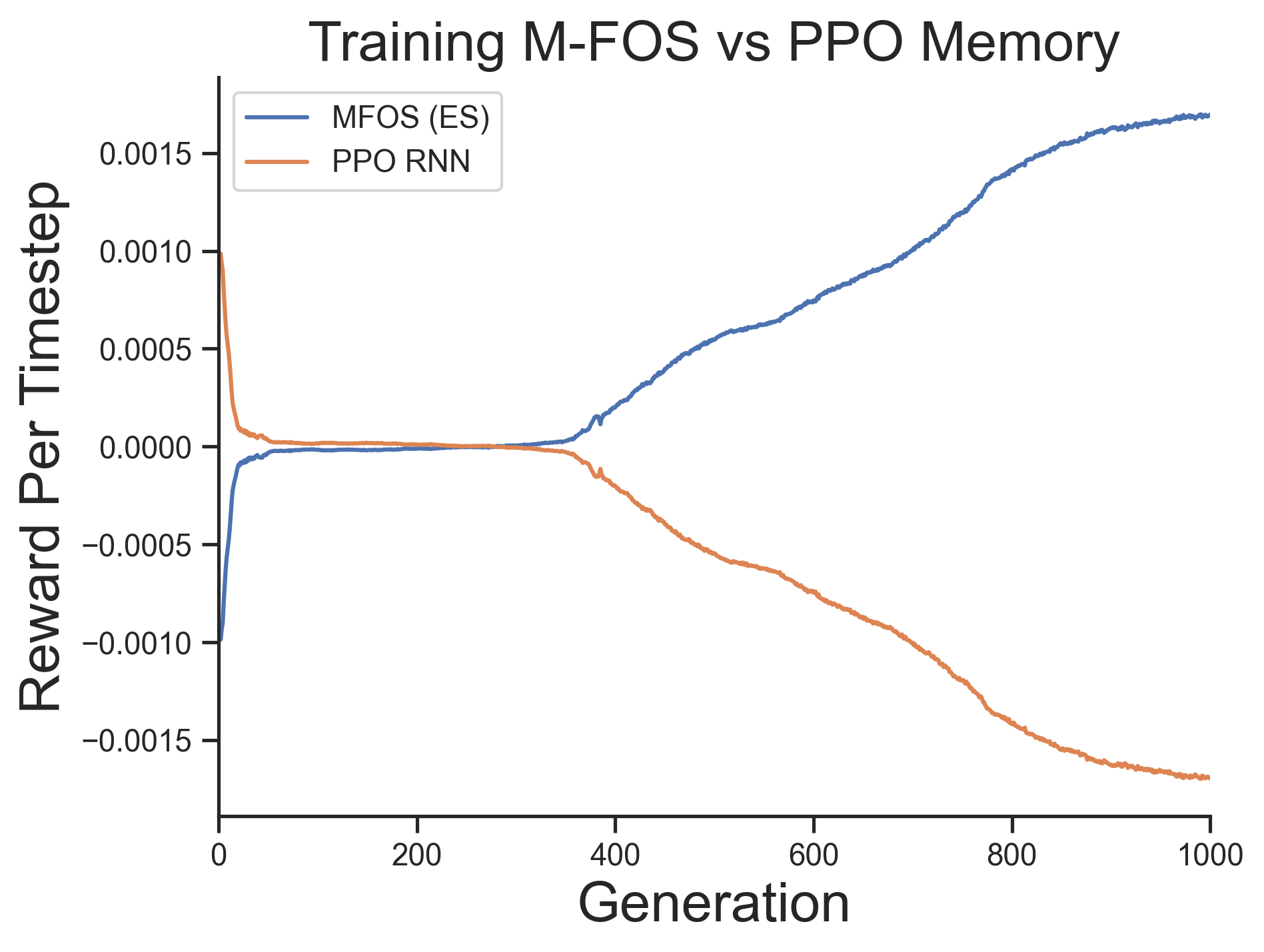}
      \label{fig:mfos_es_impitm_train}
    \end{subfigure}
    \begin{subfigure}[b]{0.24\textwidth}
      \centering
      \includegraphics[width=\textwidth]{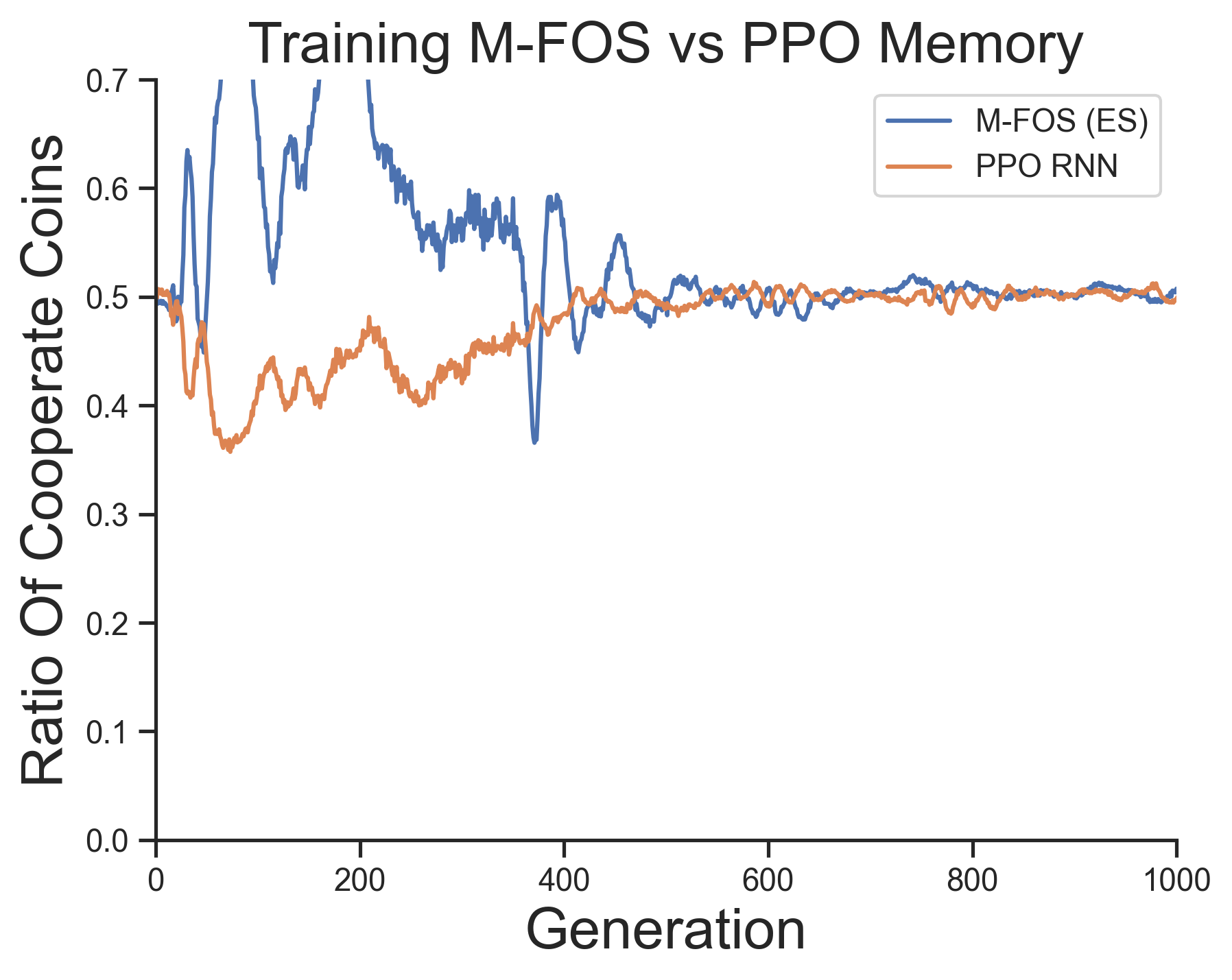}
      \label{fig:mfos_impitm_ratio}
    \end{subfigure}
    \begin{subfigure}[b]{0.24\textwidth}
      \centering
      \includegraphics[width=\textwidth]{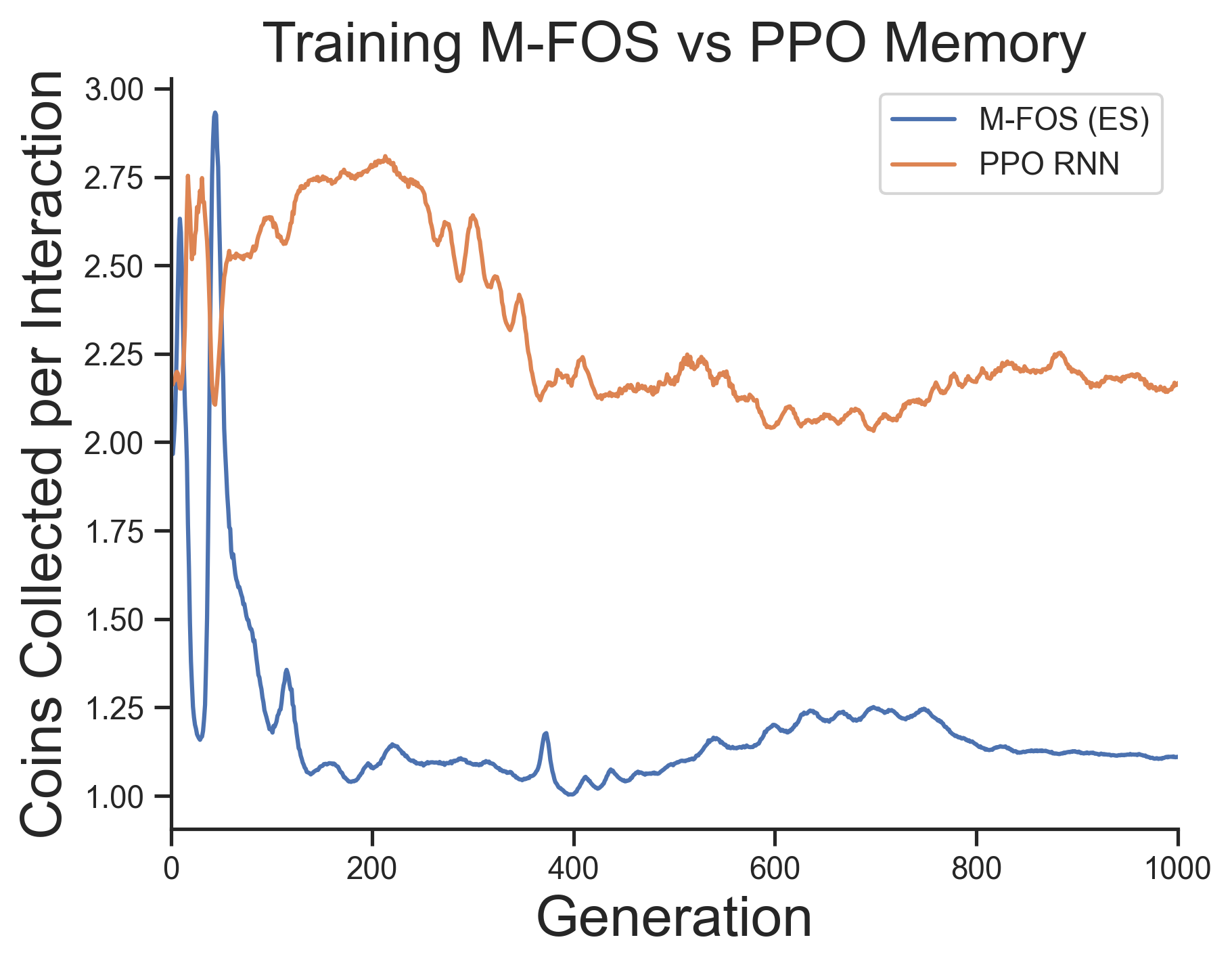}
      \label{fig:mfos_impitm_coins}
    \end{subfigure}
    \begin{subfigure}[b]{0.24\textwidth}
      \centering
      \includegraphics[width=\textwidth]{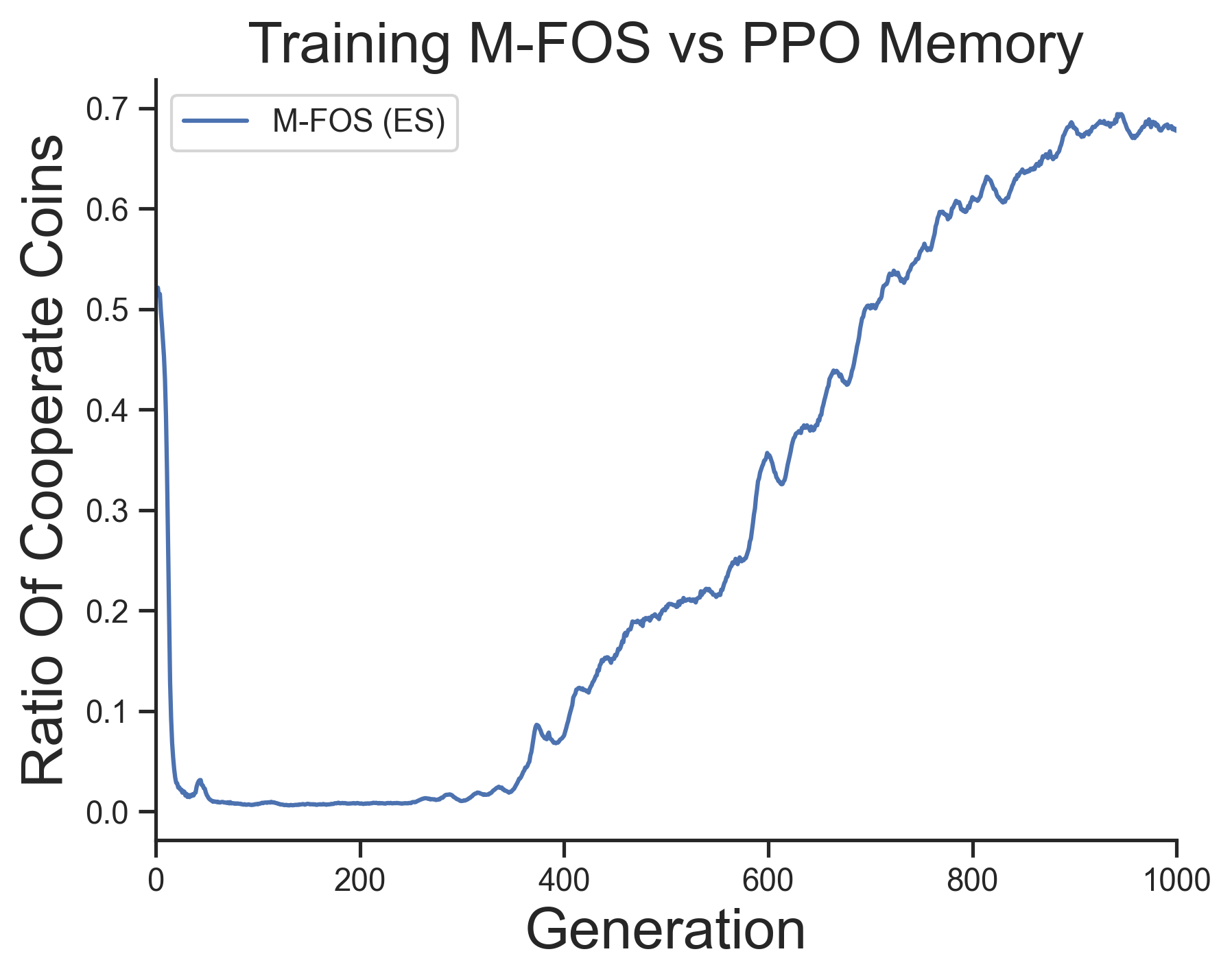}
      \label{fig:mfos_impitm_reset}
    \end{subfigure}
    \caption{Training results for Shapers vs. PPO RNN in the IMPitM. (Column 1) Reward Per Timestep, (Column 2) the meta-agent's frequency of picking up its own colour coin depending on existing convention, (Column 3) the number of coins picked up per episode, (Column 4) the number of soft resets / successful interactions.}
\end{figure}

\begin{figure}
    \centering
    \begin{subfigure}[b]{0.32\textwidth}
      \centering
      \includegraphics[width=\textwidth]{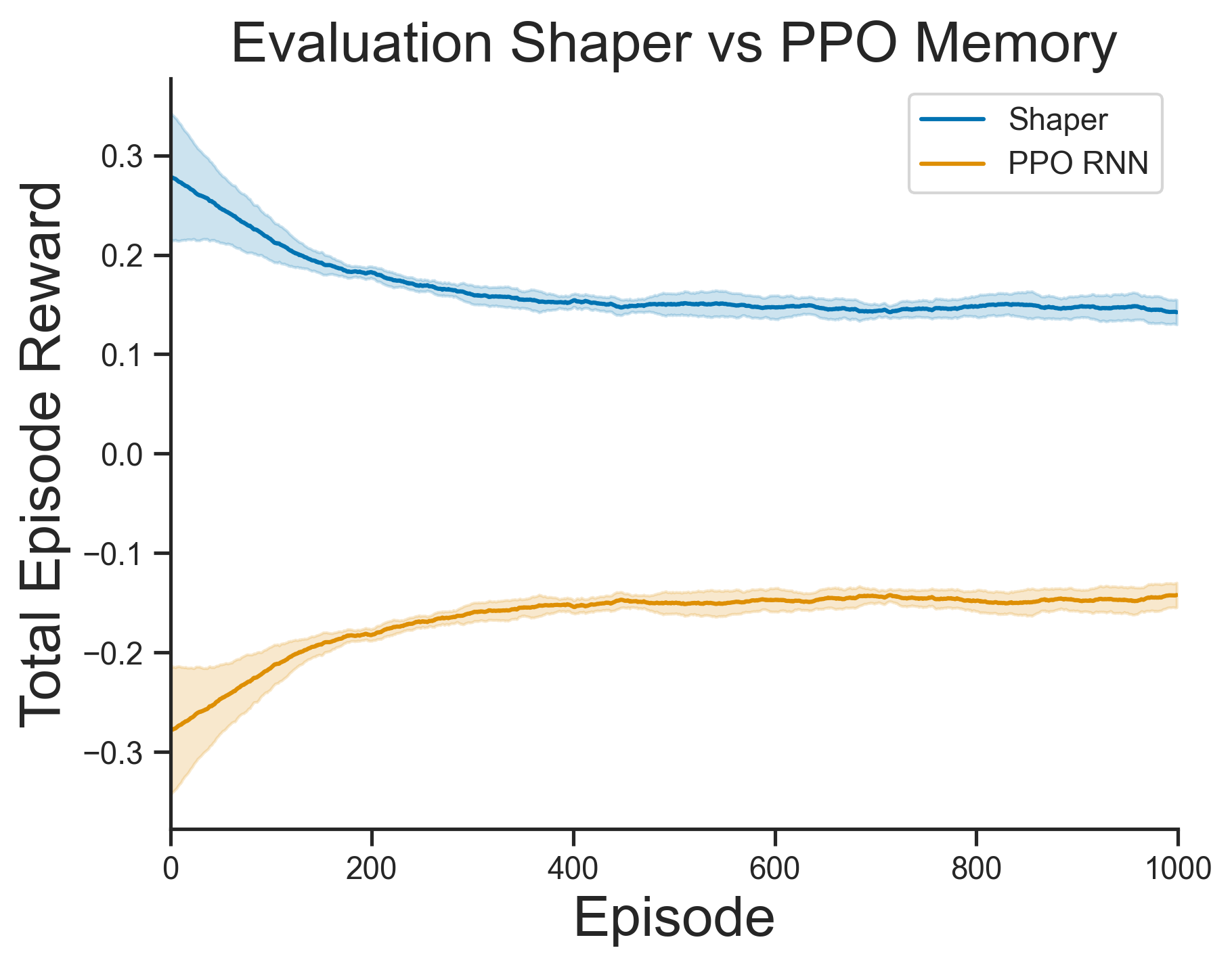}
      \caption{}
      \label{fig:shaper_impitm_eval_reward}
    \end{subfigure}
    \begin{subfigure}[b]{0.32\textwidth}
      \centering
      \includegraphics[width=\textwidth]{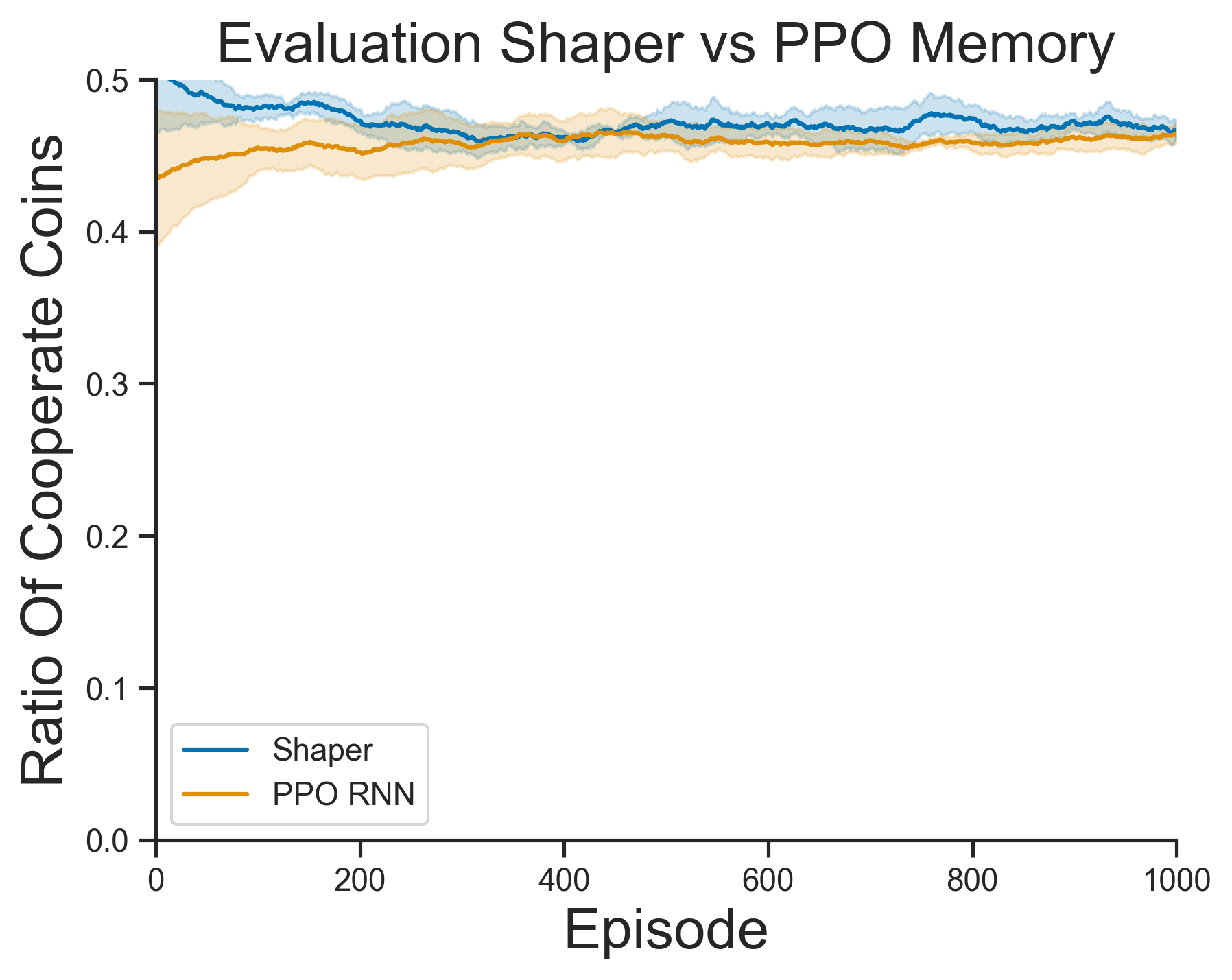}
      \caption{}
      \label{fig:shaper_impitm_eval_ratio}
    \end{subfigure}
    \begin{subfigure}[b]{0.32\textwidth}
      \centering
      \includegraphics[width=\textwidth]{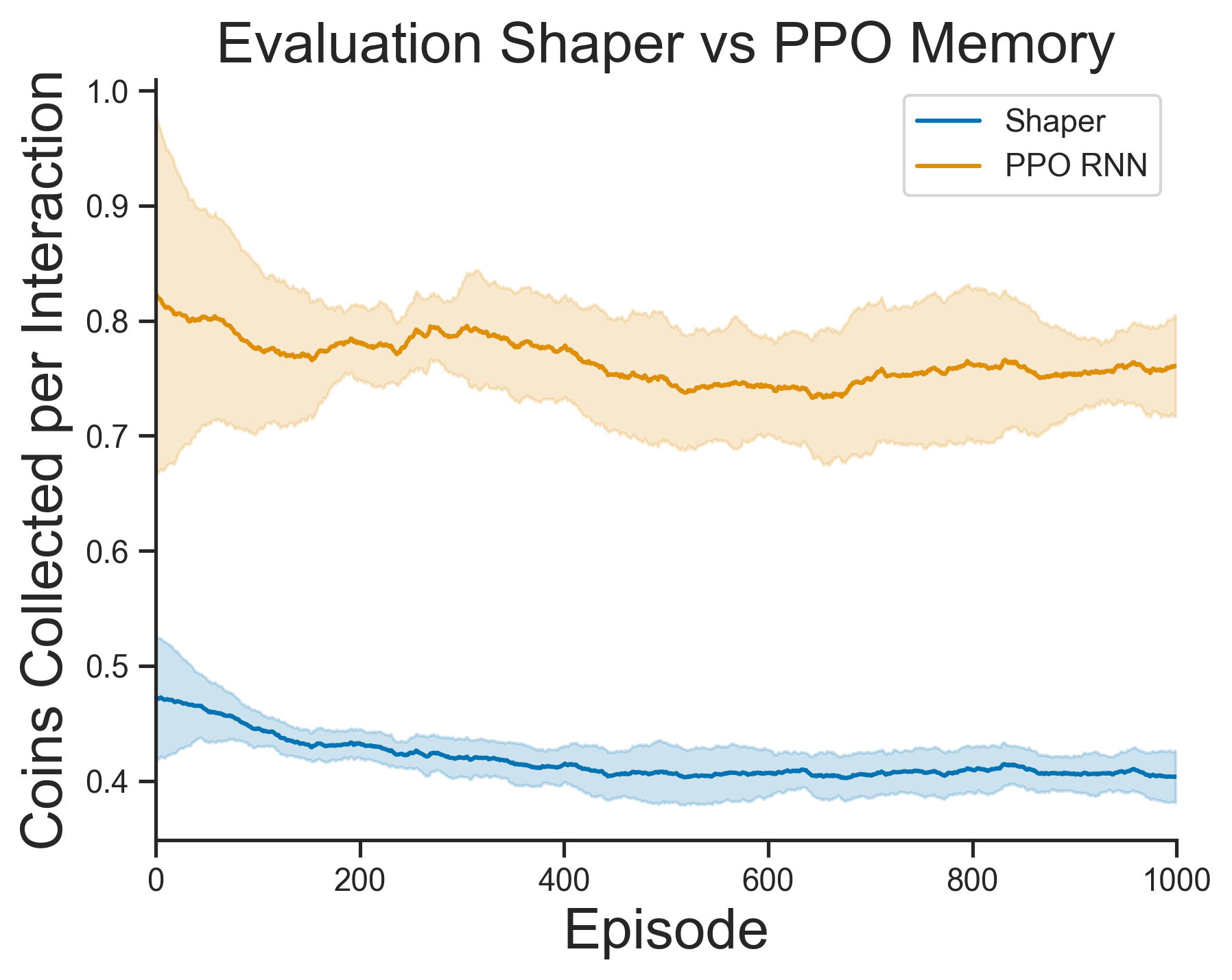}
      \caption{}
      \label{fig:shaper_impitm_eval_coins}
    \end{subfigure}
    \caption{Evaluation results over a single trial (with new co-player) compromising over five seeds for the IPDitM. (a) Mean reward per timestep, (b) mean ratio of picking up cooperate coins per soft-reset, (c) total number of coins picked up per soft-reset.
    }
    \label{fig:Shaper_ego_impitm_viz}
\end{figure}

\begin{figure}[h]
    \centering
    \begin{subfigure}[b]{0.33\textwidth}
      \centering
      \includegraphics[width=\textwidth]{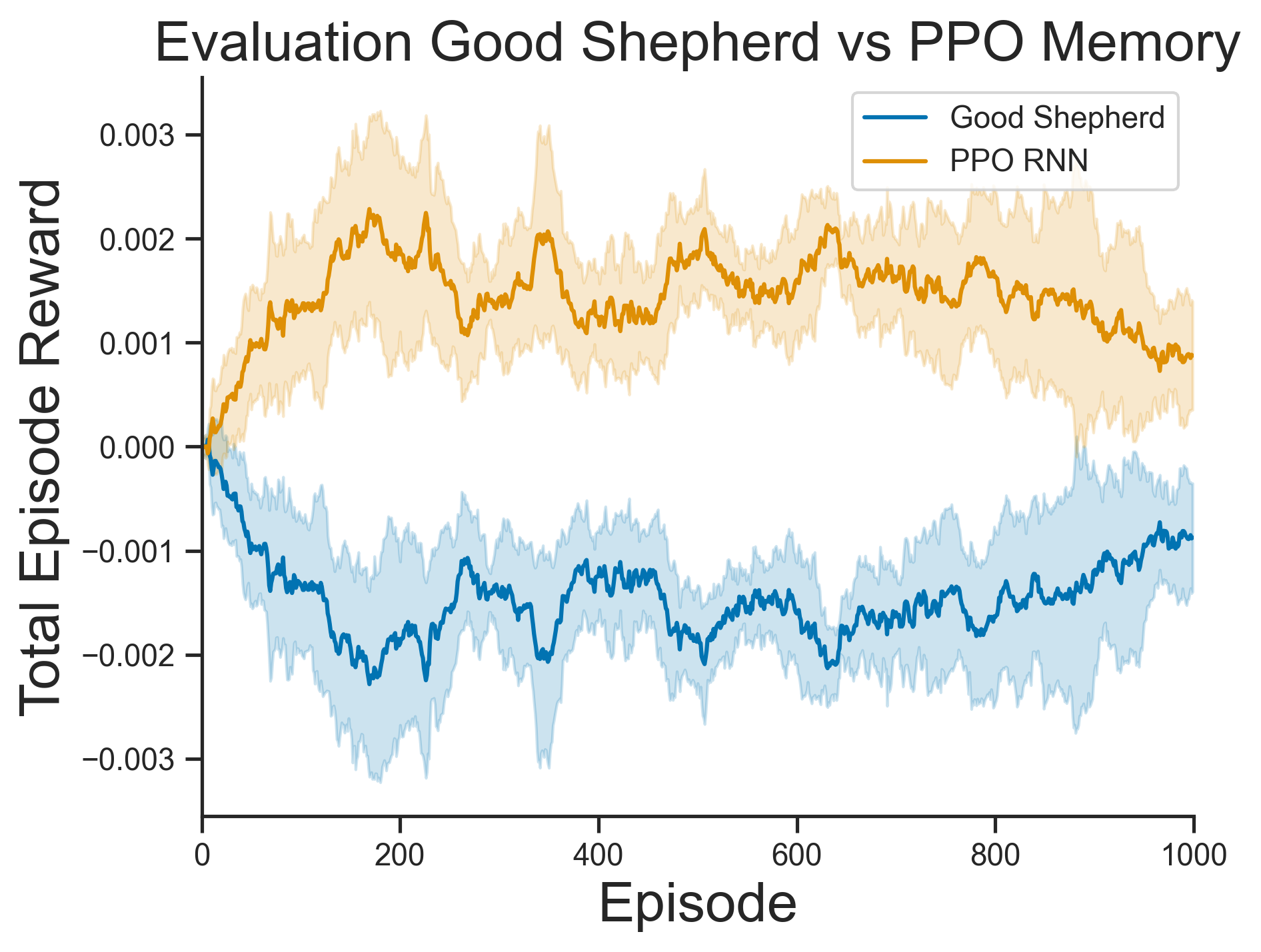}
      \caption{}
      \label{fig:gs_impitm_eval_reward}
    \end{subfigure}
    \begin{subfigure}[b]{0.33\textwidth}
      \centering
      \includegraphics[width=\textwidth]{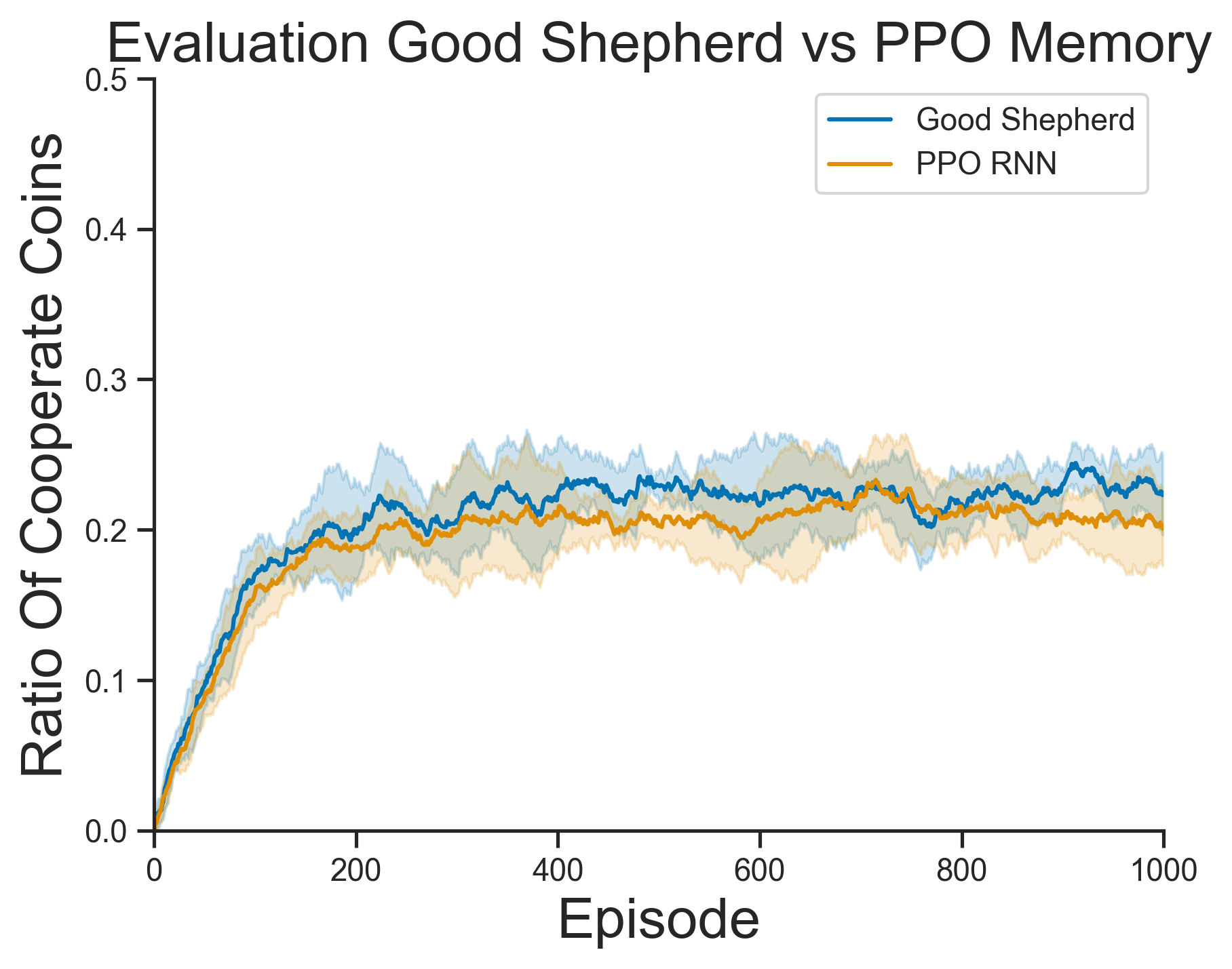}
      \caption{}
      \label{fig:gs_impitm_eval_ratio}
    \end{subfigure}
    \begin{subfigure}[b]{0.33\textwidth}
      \centering
      \includegraphics[width=\textwidth]{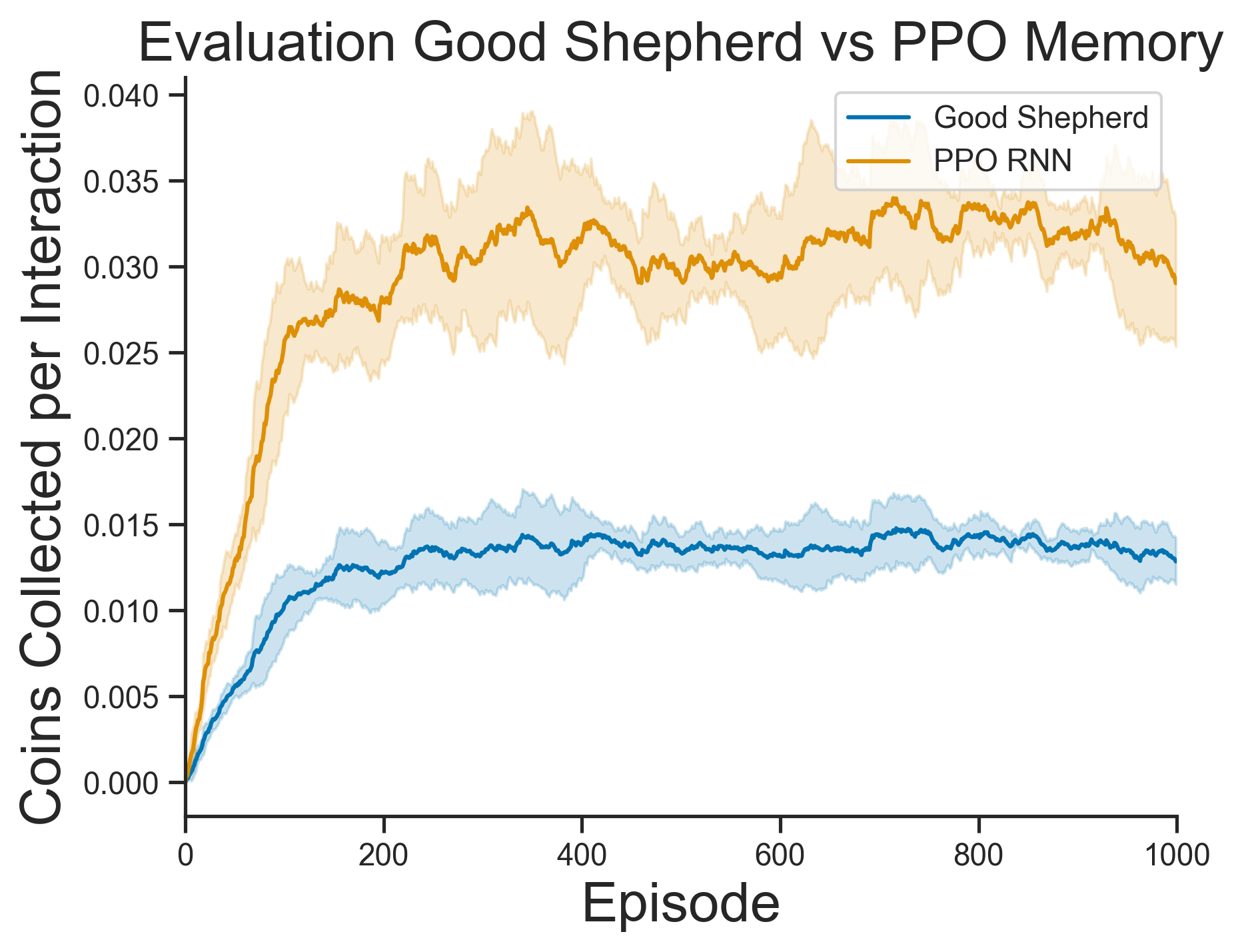}
      \caption{}
      \label{fig:gs_impitm_eval_coins}
    \end{subfigure}
    \caption{Evaluation results over a single trial (with new co-player) compromising over five seeds for the IPDitM. (a) Mean reward per timestep, (b) mean ratio of picking up cooperate coins per soft-reset, (c) total number of coins picked up per soft-reset. 
    }
    \label{fig:gs_ego_impitm_viz}
\end{figure}

\begin{figure}[h]
    \centering
    \begin{subfigure}[b]{0.32\textwidth}
      \centering
      \includegraphics[width=\textwidth]{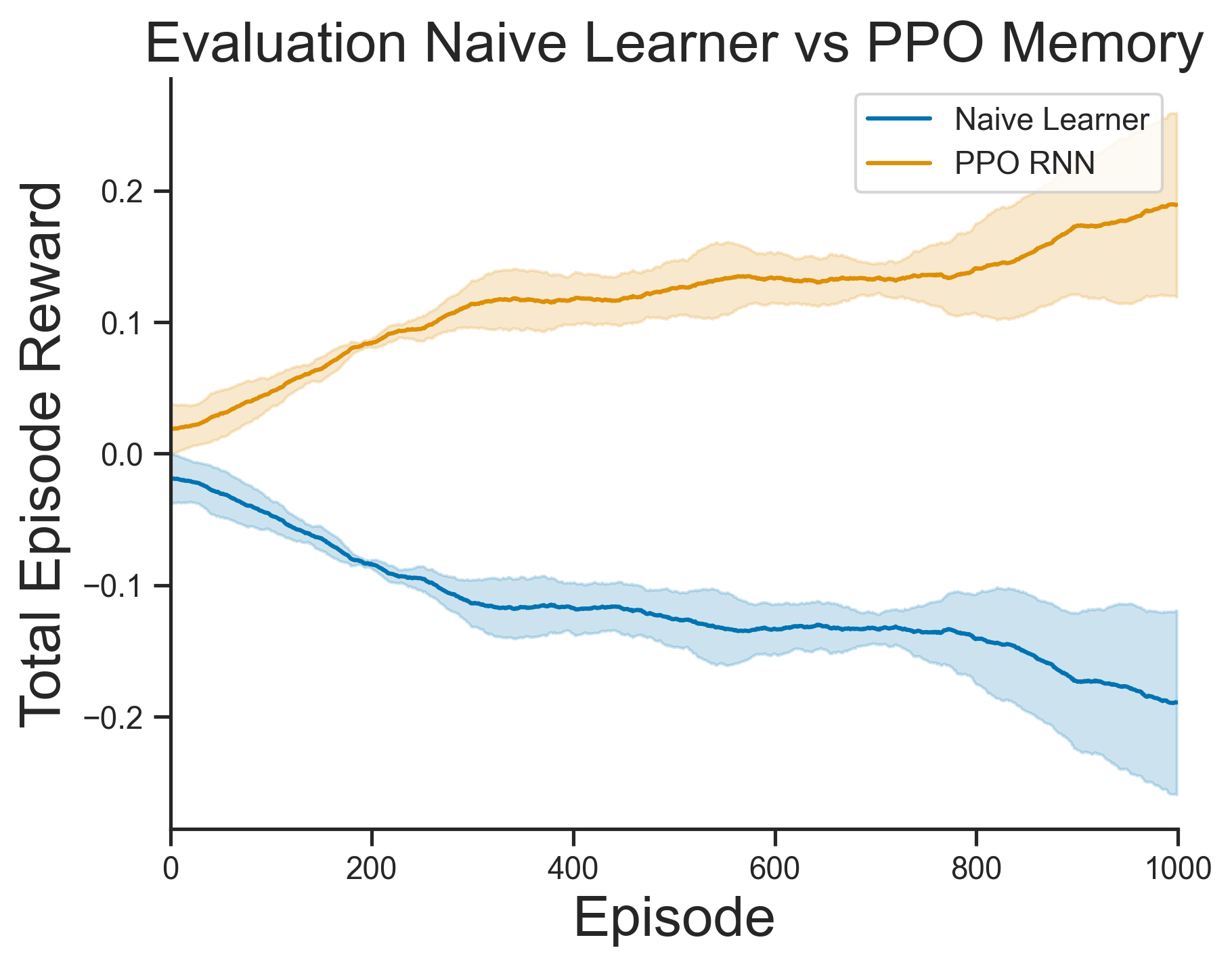}
      \caption{}
      \label{fig:nl_impitm_eval_reward}
    \end{subfigure}
    \begin{subfigure}[b]{0.32\textwidth}
      \centering
      \includegraphics[width=\textwidth]{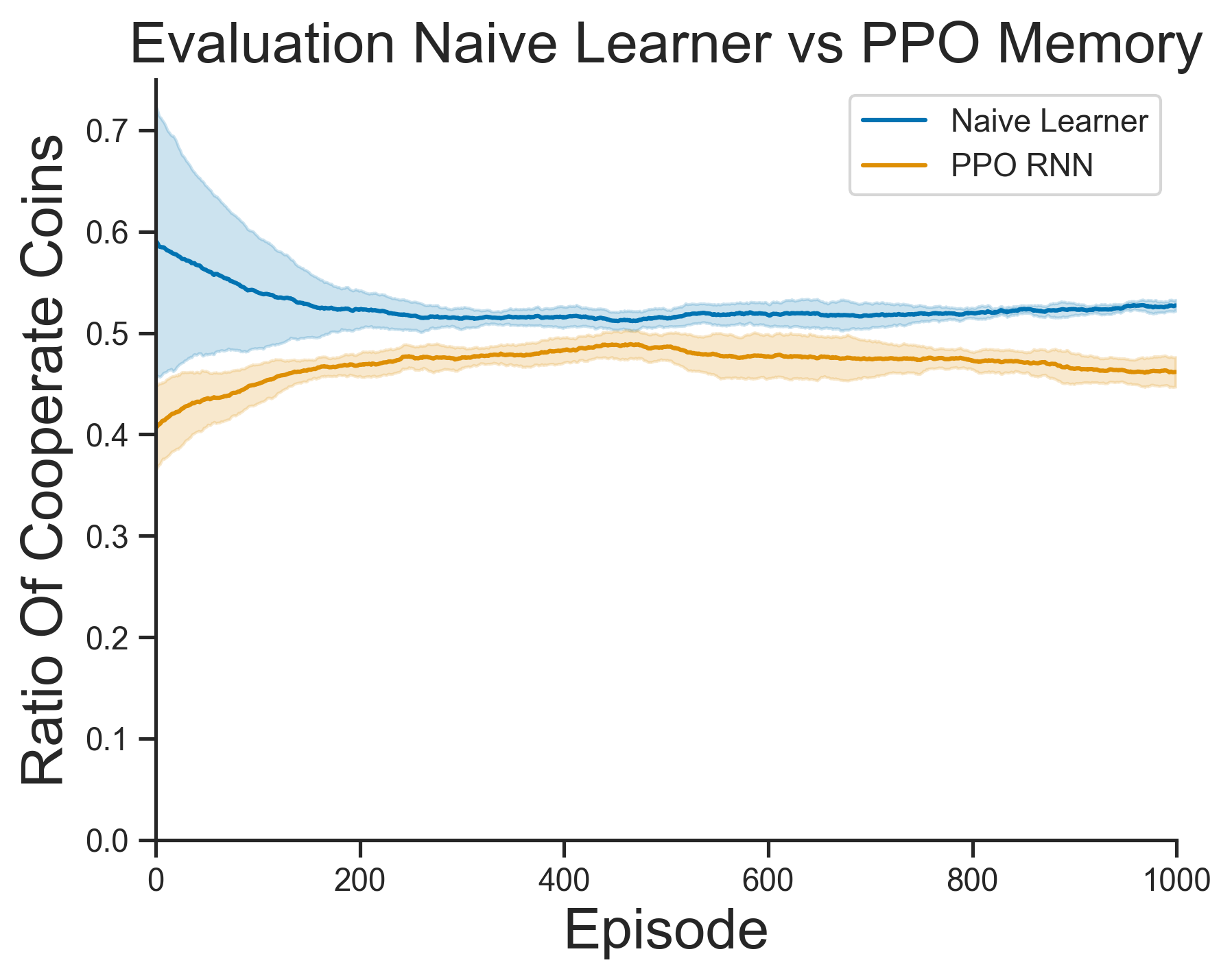}
      \caption{}
      \label{fig:nl_impitm_eval_ratio}
    \end{subfigure}
    \begin{subfigure}[b]{0.32\textwidth}
      \centering
      \includegraphics[width=\textwidth]{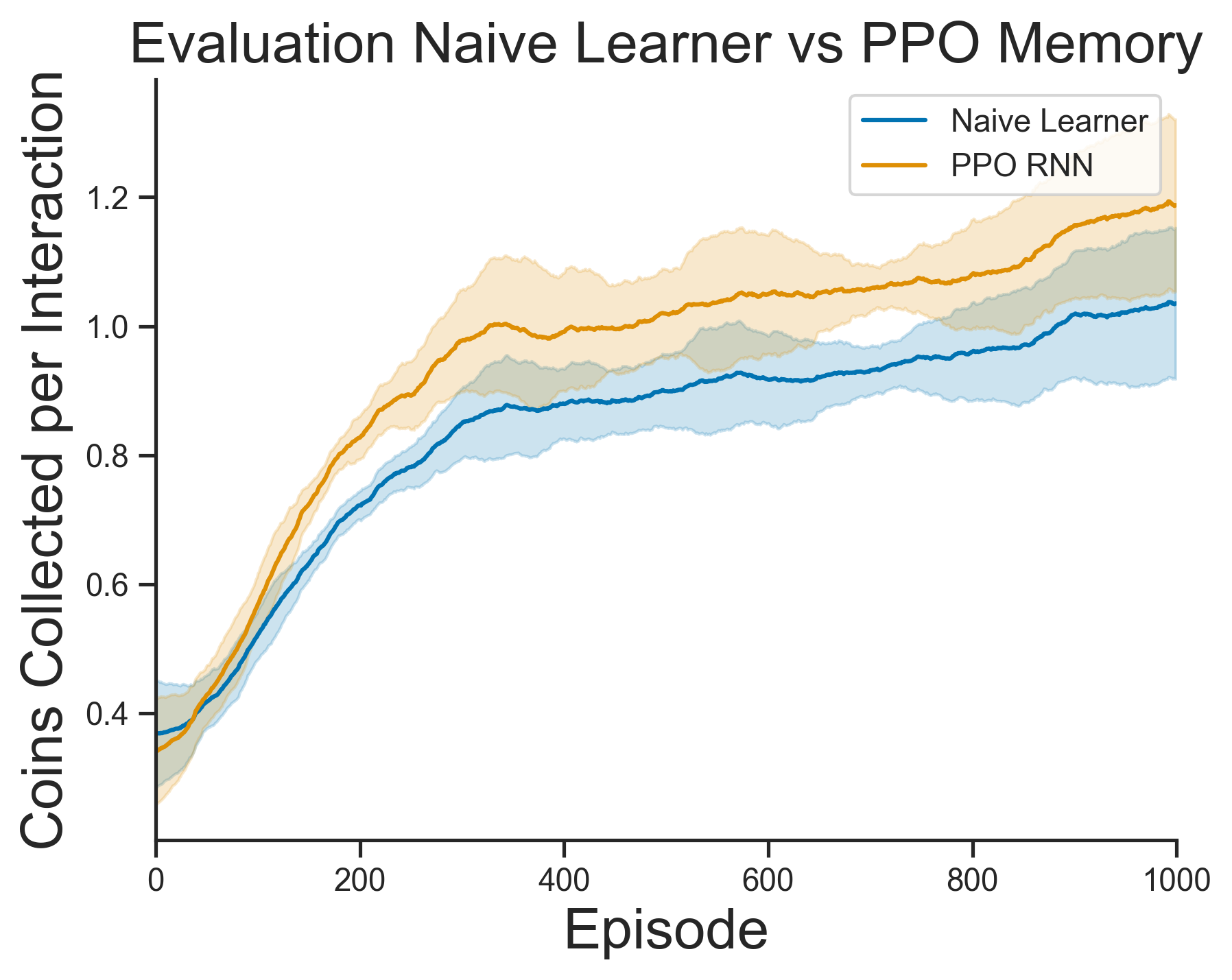}
      \caption{}
      \label{fig:nl_impitm_eval_coins}
    \end{subfigure}
    \caption{Evaluation results over a single trial (with new co-player) compromising over five seeds for the IPDitM. (a) Mean reward per timestep, (b) mean ratio of picking up cooperate coins per soft-reset, (c) total number of coins picked up per soft-reset.
    }
    \label{fig:nl_ego_impitm_viz}
\end{figure}
\clearpage

\section{Cross-Play Results}
We also present cross-play for shaping algorithms against each other on the IPD in the Matrix game. 

\begin{table}[ht]
    \centering
    \caption{Episode reward for a single evaluation trial against different OS shaping methods. Neither agent takes gradient updates, but those with memory \textsc{Shaper} and \textsc{M-FOS} are able to use memory to change their policy during the trial. We report mean and std over 5 seeds.}
    \label{tab:x_play_results}

    \begin{tabular}{@{}lccc@{}}
        \toprule
        & \textsc{Shaper} & \textsc{GS} & \textsc{M-FOS} \\
        \midrule
    \textsc{shaper} & $16.48 \pm 0.88$ & $28.61 \pm 1.82$ & $7.32 \pm 0.34$ \\
    \textsc{gs} & $20.23 \pm 1.27$ & $0 \pm 0$ & $1.91 \pm 0.27$ \\
    \textsc{m-fos} & $5.08 \pm 0.36$ & $1.35 \pm 0.28$ & $16.25 \pm 0.95$ \\
    \bottomrule
			\bottomrule
		\end{tabular}
\end{table}

\section{Variance over seeds}
Here we also report the scores for each game by median.

\begin{table*}[ht]
\caption{Converged episode reward per episode (meta-agent, co-player) for agents trained with Naive Learners on the CoinGame, IPDitM and IMPitM. The median is reported across 100 seeds with standard error of mean.
\label{tab:result_mean_gridworlds}}
    \centering
    \begin{tabular}{l rl rl rl}  
     & \multicolumn{2}{c}{CoinGame} & \multicolumn{2}{c}{IPD in the Matrix} & \multicolumn{2}{c}{IMP in the Matrix} \\ 
    \midrule
     \method{} 
     & $\mathbf{3.46 \pm 0.66}$,&$ \mathbf{-1.73 \pm 0.09}$
     & $\mathbf{22.29 \pm 0.11}$, & $\mathbf{21.99 \pm 0.11}$
     & $\mathbf{0.09\pm 0.03}$, & $\mathbf{-0.09 \pm 0.03}$\\
     \textsc{M-FOS} (ES) & $3.19 \pm 0.09$,&$ 3.28 \pm 0.11$
      & $10.47 \pm 0.35$,&$ 25.50 \pm 0.33$
      & $\mathbf{0.11 \pm 0.02}$,&$\mathbf{-0.11 \pm 0.02}$
      \\
     \textsc{M-FOS} (RL) 
     & $0.24 \pm 0.14$,&$ 0.819 \pm 0.08$
     & $7.39 \pm 0.08$,&$ 7.29 \pm 0.05$
     & $0.07\pm 0.02$,&$ -0.07 \pm 0.02$ \\
     GS 
     & $\mathbf{4.48 \pm 0.14}$, & $\mathbf{-2.63 \pm 0.12}$
      & $15.24 \pm 0.19$,&$ 6.84 \pm 0.11$ 
      & $0.00 \pm 0.00$,&$ 0.00 \pm 0.00$ \\
     PT-NL 
     & $0.07 \pm 0.15$,&$ 1.48 \pm 0.26$
     & $6.41 \pm 0.12$,&$ 6.89 \pm 0.13 $
     & $-0.13 \pm 0.07$,&$ 0.13 \pm 0.07 $\\ 
     CT-NL 
     & $0.34 \pm 0.66$,&$ 0.09 \pm 0.93$
     & $6.03 \pm 0.02$,&$ 5.07 \pm 0.18$
     & $-0.11 \pm 0.02$,&$ 0.11 \pm 0.02$ \\ 
    \bottomrule
    \end{tabular}
\end{table*}
\clearpage  

\section{Hyper-parameters}
\label{appendix:ipd_hyper}
\label{appendix:cg_hyper}

We used the Jax library \citep{jax2018github} with the Haiku framework \citep{haiku2020github} to implement our neural networks. For the Evolution strategies, we relied on the Evosax library \citep{evosax2022github}. Our experiments were performed on NVIDIA A100, A40 and V100 GPUs.

\subsection{Implementation Details} 
We performed hyperparameter optimisation over GS, \textsc{M-FOS} and \method{} - evaluating network sizes ($8, 16, 32$), co-player learning rates $(2.5e^{-3}, 2e^{-2}, 1^{e-1}, 1)$, co-player discount ($0.96, 0.98, 0.99$), and population size ($128, 256, 512, 1000$). We report best parameters in the Appendix \ref{appendix:ipd_hyper}.

\begin{table}[h]
\centering
\begin{tabular}{l|l}
Hyperparameter & Value \\ \hline
Number of Actor Hidden Layers & 1 \\
Number of Critic Hidden Layers & 1 \\
Torso GRU Size & [25] \\
Length of Meta-Episode  & 100 \\
Length of Inner Episode & 100 \\
Number of Generations & 5000 \\
Batch Size & 100 \\
Population Size  & 1000\\
OpenES sigma init& 0.04    \\
OpenES sigma decay & 0.999 \\
OpenES sigma limit& 0.01  \\
OpenES init min& 0.0  \\
OpenES init max& 0.0 \\ 
OpenES clip min& -1e10  \\  
OpenES clip max& 1e10     \\ 
OpenES lrate init& 0.01    \\
OpenES lrate decay& 0.9999 \\
OpenES lrate limit& 0.001 \\
OpenES beta 1& 0.99       \\
OpenES beta 2& 0.999       \\
OpenES eps& 1e-8          \\
\end{tabular}
\caption{Hyperparameters for \method{} in Iterated Prisoner's Dilemma}
\end{table}

\begin{table}[h]
\centering
\begin{tabular}{l|l}
Hyperparameter & Value \\ \hline
Number of Minibatches & 4\\
Number of Epochs  & 2\\
Gamma & 0.96\\
GAE Lambda & 0.95\\
PPP clipping epsilon& 0.2\\
Value Coefficient & 0.5\\
Clip Value& True \\
Max Gradient Norm& 0.5\\
Entropy Coefficient Start & 0.02\\
Entropy Coefficient Horizon& 2000000\\
Entropy Coefficient End& 0.001\\
Learning rate& 1\\
ADAM epsilon& 1e-5\\
\end{tabular}
\caption{Hyperparameters for Tabular-PPO in Iterated Prisoner's Dilemma}
\end{table}
\begin{table}
\centering
\begin{tabular}{l|l}
\small
Hyperparameter & Value \\ \hline
Number of Actor Hidden Layers & 2 \\
Number of Critic Hidden Layers & 2 \\
Network Hidden Size & [16, 16] \\
Length of Meta-Episode  & 100 \\
Length of Inner Episode & 100 \\
Number of Generations & 5000 \\
Batch Size & 100 \\
Population Size  & 1000\\
OpenES sigma init& 0.04    \\
OpenES sigma decay & 0.999 \\
OpenES sigma limit& 0.01  \\
OpenES init min& 0.0  \\
OpenES init max& 0.0 \\ 
OpenES clip min& -1e10  \\  
OpenES clip max& 1e10     \\ 
OpenES lrate init& 0.01    \\
OpenES lrate decay& 0.9999 \\
OpenES lrate limit& 0.001 \\
OpenES beta 1& 0.99       \\
OpenES beta 2& 0.999       \\
OpenES eps& 1e-8          \\
\end{tabular}
\caption{Hyperparameters for GS in Iterated Prisoner's Dilemma}
\end{table}

\begin{table}[h]
\centering
\begin{tabular}{l|l}
Hyperparameter & Value \\ \hline
Number of Actor Hidden Layers & 1 \\
Number of Critic Hidden Layers & 1 \\
Actor GRU Hidden Size & 16 \\
Critic GRU Hidden Size & 16 \\
Meta-Agent Gru Hidden Size & 16 \\
Hidden Layer Size & 16 \\
Length of Meta-Episode  & 100 \\
Length of Inner Episode & 100 \\
Number of Generations & 5000 \\
Batch Size & 100 \\
Population Size  & 1000\\
OpenES sigma init& 0.04    \\
OpenES sigma decay & 0.999 \\
OpenES sigma limit& 0.01  \\
OpenES init min& 0.0  \\
OpenES init max& 0.0 \\ 
OpenES clip min& -1e10  \\  
OpenES clip max& 1e10     \\ 
OpenES lrate init& 0.01    \\
OpenES lrate decay& 0.9999 \\
OpenES lrate limit& 0.001 \\
OpenES beta 1& 0.99       \\
OpenES beta 2& 0.999       \\
OpenES eps& 1e-8          \\
\end{tabular}
\caption{Hyperparameters for \textsc{M-FOS} in Iterated Prisoner's Dilemma}
\end{table}

\begin{table}[]
\centering
\begin{tabular}{l|l}
Hyperparameter & Value \\ \hline
Number of Actor Hidden Layers & 1 \\
Number of Critic Hidden Layers & 1 \\
Torso Gru Size & [16] \\
Length of Meta-Episode  & 600 \\
Length of Inner Episode & 16 \\
Number of Generations & 3000 \\
Batch Size & 100 \\
Population Size  & 4000\\
OpenES sigma init& 0.04    \\
OpenES sigma decay & 0.999 \\
OpenES sigma limit& 0.01  \\
OpenES init min& 0.0  \\
OpenES init max& 0.0 \\ 
OpenES clip min& -1e10  \\  
OpenES clip max& 1e10     \\ 
OpenES lrate init& 0.01    \\
OpenES lrate decay& 0.9999 \\
OpenES lrate limit& 0.001 \\
OpenES beta 1& 0.99       \\
OpenES beta 2& 0.999       \\
OpenES eps& 1e-8          \\
\end{tabular}
\caption{Hyperparameters for \textsc{Shaper} in Iterated Matching Pennies}
\end{table}

\begin{table}[]
\centering
\begin{tabular}{l|l}
Hyperparameter & Value \\ \hline
Number of Minibatches & 8\\
Number of Epochs  & 2\\
Gamma & 0.96\\
GAE Lambda & 0.95\\
PPO clipping epsilon& 0.2\\
Value Coefficient& 0.5\\
Clip Value& True \\
Max Gradient Norm& 0.5\\
Anneal Entropy & False\\
Entropy Coefficient Start& 0.02\\
Entropy Coefficient Horizon& 2000000\\
Entropy Coefficient End& 0.001\\
LR Scheduling& False\\
Learning Rate& 0.005\\
ADAM Epsilon& 1e-5\\
With CNN& False\\
\end{tabular}
\caption{Hyperparameters for PPO Memory and Tabular in the CoinGame}
\end{table}

\begin{table}[]
\centering
\begin{tabular}{l|l}
Hyperparameter & Value \\ \hline
Number of Actor Hidden Layers & 1 \\
Number of Critic Hidden Layers & 1 \\
Hidden Size & [16] \\
Length of Meta-Episode  & 600 \\
Length of Inner Episode & 16 \\
Number of Generations & 3000 \\
Batch Size & 100 \\
Population Size  & 4000\\
OpenES sigma init& 0.04    \\
OpenES sigma decay & 0.999 \\
OpenES sigma limit& 0.01  \\
OpenES init min& 0.0  \\
OpenES init max& 0.0 \\ 
OpenES clip min& -1e10  \\  
OpenES clip max& 1e10     \\ 
OpenES lrate init& 0.01    \\
OpenES lrate decay& 0.9999 \\
OpenES lrate limit& 0.001 \\
OpenES beta 1& 0.99       \\
OpenES beta 2& 0.999       \\
OpenES eps& 1e-8          \\
\end{tabular}
\end{table}

\begin{table}[]
\centering
\begin{tabular}{l|l}
Hyperparameter & Value \\ \hline
Number of Actor Hidden Layers & 1 \\
Number of Critic Hidden Layers & 1 \\
Actor GRU Hidden Size & 16 \\
Critic GRU Hidden Size & 16 \\
Meta-Agent Gru Hidden Size & 16 \\
Hidden Layer Size & 16 \\
Length of Meta-Episode  & 100 \\
Length of Inner Episode & 100 \\
Number of Generations & 5000 \\
Batch Size & 100 \\
Population Size  & 1000\\
OpenES sigma init& 0.04    \\
OpenES sigma decay & 0.999 \\
OpenES sigma limit& 0.01  \\
OpenES init min& 0.0  \\
OpenES init max& 0.0 \\ 
OpenES clip min& -1e10  \\  
OpenES clip max& 1e10     \\ 
OpenES lrate init& 0.01    \\
OpenES lrate decay& 0.9999 \\
OpenES lrate limit& 0.001 \\
OpenES beta 1& 0.99       \\
OpenES beta 2& 0.999       \\
OpenES eps& 1e-8          \\
\end{tabular}
\caption{Hyperparameters for \textsc{M-FOS} in CoinGame}
\end{table}

\begin{table}[]
\centering
\begin{tabular}{l|l}
Hyperparameter & Value \\ \hline
Number of Actor Hidden Layers & 1 \\
Number of Critic Hidden Layers & 1 \\
GRU Hidden Size & 32 \\
Kernel Shape & [3,3] \\ 
Hidden Layer Size & 16 \\
Length of Meta-Episode  & 500 \\
Length of Inner Episode & 128 \\
Number of Generations & 1000 \\
Batch Size & 50 \\
Population Size  & 1000\\
OpenES sigma init& 0.075    \\
OpenES sigma decay & 0.999 \\
OpenES sigma limit& 0.01  \\
OpenES init min& 0.0  \\
OpenES init max& 0.0 \\ 
OpenES clip min& -1e10  \\  
OpenES clip max& 1e10     \\ 
OpenES lrate init& 0.05    \\
OpenES lrate decay& 0.9999 \\
OpenES lrate limit& 0.001 \\
OpenES beta 1& 0.99       \\
OpenES beta 2& 0.999       \\
OpenES eps& 1e-8          \\
\end{tabular}
\caption{Hyperparameters for \textsc{Shaper} in * in the Matrix games }
\end{table}

\begin{table}[h]
\centering
\begin{tabular}{l|l}
Hyperparameter & Value \\ \hline
Number of Minibatches & 8\\
Number of Epochs  & 2\\
Gamma & 0.96\\
GAE Lambda & 0.95\\
PPO clipping epsilon& 0.2\\
Value Coefficient& 0.5\\
Clip Value& True \\
Max Gradient Norm& 0.5\\
Anneal Entropy & False\\
Entropy Coefficient Start& 0.02\\
Entropy Coefficient Horizon& 2000000\\
Entropy Coefficient End& 0.001\\
LR Scheduling& False\\
Learning Rate& 0.005\\
ADAM Epsilon& 1e-5\\
With CNN& True\\
Output Channels & 16 \\
Kernel Shape & [3,3] \\
\end{tabular}
\caption{Hyperparameters for PPO Memory in * in the Matrix games}
\end{table}
\clearpage

\section{FAQ}
\label{app:faq}
\begin{faq}
  \item[Why didn’t you use meltingpot directly?]
    We rely heavily on using Jax and its computational efficiencies. Therefore, we need our environments vectorised, which is not the case for meltingpot.
  \item[What are implementation differences between your implementation and meltingpot?]
  Generally, meltingpot, though a gridworld, has pixel-based observations whereas our environment provides access to the grid directly. 
  
  In IPDitM, the meltingpot environment is 23x15, whereas our grid is 8x8. A size we chose to be as large as possible whilst still being able to optimise the methods given compute limitations. Meltingpot has walls placed within the environment, whereas ours does not. While our environment is smaller, we add additional stochasticity by randomising the coin positions, which are fixed in Meltingpot. Finally, unlike the meltingpot environment, agents spawned with no coins (as opposed to one of each), this made it easier for agents to choose pure cooperate or defect strategies. We found that randomising coin positions important as even large environments representing POMDPs with insufficient stochasticity, such as the The Starcraft Multi-Agent Challenge, can be solved without memory\citep{samvelyan19smac,ellis22022smacv2}. We found similar evidence in the IPDitM. In IMPitM, the same differences hold.
  \item[What compute resources did you use?]
  We had access to 32 A40s, 32 A100s, distributed over 8-GPU machines. An experiment is distributed over 8 GPUs. Training \textsc{GS}, \textsc{M-FOS} or \method{} on IPDitM takes approximately 5 days respectively.
  \item[What frameworks did you use?]
  We used the Jax library (Bradbury et al., 2018) with the Haiku framework (Hennigan et al., 2020) to implement our neural networks. We use the Evosax library \citep{evosax2022github} for the Evolution Strategies method and have adapted the interface of gymnax \citep{gymnax2022github} for our environment implementations.

  \item[Are there any other differences between Shaper and M-FOS?]
In the algorithm definition of \textsc{M-FOS}, there are two action spaces: the meta-action space $\bar{\mathcal{A}}$  and the underlying action space $\mathcal{A}$. 
The meta-action space consists of the policy parameters of the underlying agent or a conditional vector parameterising this, and the conventional action space is the action space of the game.
In \method{}, the only action space is that of the underlying game, meaning \textit{there is only one agent in \method{}, whereas there are two agents in \textsc{M-FOS}.} Consequently, \method{} is a special case of \textsc{M-FOS}. 

Moreover, \method{}'s architecture is different from the architectures proposed in M-FOS.
 M-FOS proposes \textit{two} significantly different architectures \citep{lu2022model}.
 For the CoinGame, in which a simple table cannot represent policies, M-FOS proposes an architecture akin to Hierarchical RL, consisting of the meta-agent and agent. Both the meta-agent and the underlying agent are recurrent neural networks. The underlying agent resets their hidden state after each episode, whereas the meta-agent does not. In this architecture, the meta-agent does not output the full parameterisation of the underlying agent but instead outputs a conditioning variable, which the underlying agent uses as input. M-FOS assumes that outputting a conditioning variable is equivalent to outputting a policy parameterisation. The conditioning variable is fixed during an episode. In contrast, \method{} only uses one recurrent neural network that does not reset its hidden state after an episode. \textit{Both M-FOS and \method{} capture context and history; however, \method{} only needs one agent. Originally, M-FOS only conditions on the past meta-episode. To ensure the fairest comparison, our M-FOS conditions on all past meta-episodes.}

\item[How do the full network sizes of these methods compare, and how does their performance compare when you control for it?]
We control for the network sizes in our experiments as much as possible. Note, however, that we performed hyperoptimization over our hidden layer size for M-FOS and we chose the best performing hyperparameters. More specifically, M-FOS and Shaper use the same network architectures. M-FOS has two networks with a hidden size of 16, respectively. Shaper has one network with a hidden size of 32. We report total number of parameters below

\begin{table}
  \centering
  \caption{Number of Parameters for each environment}
  \label{tab:matchups}
  \begin{tabular}{@{}lccc@{}}
    \toprule
     & \textbf{Matrix Games} & \textbf{Coin Game} & \textbf{*in The Matrix} \\
    \midrule
    \textsc{shaper} & 1104 & 2272 & 21896 \\
    \textsc{gs} & 416 & 688 & 3892 \\
    \textsc{m-fos} & 5360 & 6896 & 29024 \\
    \bottomrule
  \end{tabular}
\end{table}

\item[Are there any other differences between Shaper and GS?]
GS does not use a recurrent agent, and there is no discussion about notions of history or context. Thus GS cannot capture history or context. Even though it is not discussed, the framework is extensible to a recurrent meta-agent. However, that meta-agent only captures history as the hidden state is reset after each episode. By not capturing context, GS fails to shape in zero-sum games, such as the Matching Pennies, which we show in Section \ref{sec:experiments}.

\end{faq}

\end{document}